# Mining the contribution of intensive care clinical course to outcome after traumatic brain injury

**Shubhayu Bhattacharyay[1,2,3,*], Pier Francesco Caruso[1,4], Cecilia Åkerlund[5], Lindsay Wilson[6], Robert D Stevens[3,7], David K Menon[1], Ewout W Steyerberg[8], David W Nelson[5], Ari Ercole[1,9], and the CENTER-TBI investigators and participants[†]**

[1]Division of Anaesthesia, University of Cambridge, Cambridge, United Kingdom.
[2]Department of Clinical Neurosciences, University of Cambridge, Cambridge, United Kingdom.
[3]Laboratory of Computational Intensive Care Medicine, Johns Hopkins University, Baltimore, MD, USA.
[4]Department of Biomedical Sciences, Humanitas University, Milan, Italy
[5]Department of Physiology and Pharmacology, Section for Perioperative Medicine and Intensive Care, Karolinska Institutet, Stockholm, Sweden.
[6]Division of Psychology, University of Stirling, Stirling, United Kingdom.
[7]Department of Anesthesiology and Critical Care Medicine, Johns Hopkins University, Baltimore, MD, USA.
[8]Department of Biomedical Data Sciences, Leiden University Medical Center, Leiden, The Netherlands.
[9]Cambridge Centre for Artificial Intelligence in Medicine, Cambridge, United Kingdom.

*Corresponding author: sb2406@cam.ac.uk (SB)
[†]A full list of the CENTER-TBI investigators and participants can be found in Supplementary Note 4.

# ABSTRACT

Existing methods to characterise the evolving condition of traumatic brain injury (TBI) patients in the intensive care unit (ICU) do not capture the context necessary for individualising treatment. Here, we integrate all heterogenous data stored in medical records (1,166 pre-ICU and ICU variables) to model the individualised contribution of clinical course to six-month functional outcome on the Glasgow Outcome Scale – Extended (GOSE). On a prospective cohort (*n*=1,550, 65 centres) of TBI patients, we train recurrent neural network models to map a token-embedded time series representation of all variables (including missing values) to an ordinal GOSE prognosis every two hours. The full range of variables explains up to 52% (95% CI: 50%–54%) of the ordinal variance in functional outcome. Up to 91% (95% CI: 90%–91%) of this explanation is derived from pre-ICU and admission information (i.e., static variables). Information collected in the ICU (i.e., dynamic variables) increases explanation (by up to 5% [95% CI: 4%–6%]), though not enough to counter poorer overall performance in longer-stay (>5.75 days) patients. Highest-contributing variables include physician-based prognoses, CT features, and markers of neurological function. Whilst static information currently accounts for the majority of functional outcome explanation after TBI, data-driven analysis highlights investigative avenues to improve dynamic characterisation of longer-stay patients. Moreover, our modelling strategy proves useful for converting large patient records into interpretable time series with missing data integration and minimal processing.

# MAIN TEXT

## Introduction

Traumatic brain injury (TBI) is the most frequently occurring neurological disorder and imposes a substantial public health burden.[1,2] Whilst TBI is increasingly appreciated as a progressive condition rather than a single event, the disease course of TBI patients in the intensive care unit (ICU) has not been well characterised. As a result, existing ICU treatments are based on limited evidence and do not target the heterogeneous mechanisms of an individual's TBI.[3] Answering the call for patient-tailored treatments (i.e., precision medicine), issued by *The Lancet Neurology* Commissions on TBI,[1,2] must start with an evidence-based understanding of individual patient trajectories in the ICU.

The main instrument for characterising TBI severity in the ICU is the Glasgow Coma Scale (GCS), for which a patient's best motor, verbal, and eye responses are assessed.[4] The GCS, however, is not sufficient for precision medicine as it does not capture a patient's pathophysiological profile and is confounded by external factors (e.g., drug use, medications, and tracheal intubation).[5] An alternative approach is to characterise severity through functional outcome prognosis. Functional outcome after TBI is typically evaluated on the ordinal, eight-point Glasgow Outcome Scale – Extended (GOSE),[6] and currently, the best validated prognostic tools for moderate-to-severe TBI (GCS≤12) are the International Mission for Prognosis and Analysis of Clinical Trials in TBI (IMPACT) models.[7] The IMPACT extended model estimates the probability of either survival (GOSE>1) or functional independence (GOSE>4) at six months post-injury from ten static predictors collected from the first 24 hours of ICU stay and explains approximately 35% of the pseudo-variance in dichotomised

GOSE.[7] Considering a patient's full, dynamic clinical course and increasing model output granularity (i.e., ordinal GOSE prognosis[8]) would further enable clinical characterisation through prognosis whilst uncovering the outcome contribution of ICU events and treatments.

In this work, we take a full-context, data-driven approach to assess the limits of dynamic ICU characterisation after TBI. The Collaborative European NeuroTrauma Effectiveness Research in TBI (CENTER-TBI) project represents the most comprehensive set of pre-ICU and ICU data for TBI patients across Europe.[5,9] Mining clinical trajectories from this data – which comprises a complex combination of modalities with varying structure, sampling, and missingness – is not straightforward. We therefore develop a regularised disease course modelling strategy which integrates all this heterogenous information and returns an interpretable, detailed proxy for severity over each patient's ICU stay.

Upon developing our TBI modelling strategy, our central aims were to: (1) evaluate the additive prognostic significance of incorporating the most complete description of ICU stay available and (2) uncover clinical events most strongly associated with transitions in an individual's trajectory. We also assess the reliability (i.e., calibration) and information content of our explanatory modelling approach to validate its application in deriving insight from medical data.

# Results

## *Study population*

Of the 2,138 CENTER-TBI patients available for analysis in the ICU stratum of the core study, 1,550 met the additional inclusion criteria of this work (Supplementary Figure 1). Since the regularity of bihourly assessments collected for CENTER-TBI decreased after a week (Supplementary Figure 2), and since over half the population remained at this point (Supplementary Figure 3), we focused our analysis on the first week after ICU admission and the last week before ICU discharge. Summary characteristics of our study population are detailed in Table 1. Additional characteristics (e.g., race, comorbidities, and vitals) of our study population have been previously published,[5] and distributions of all study variables are available online (https://www.center-tbi.eu/data/dictionary).

## *Disease course modelling*

We developed a modelling strategy to map all 738 static (i.e., fixed at ICU admission) and 428 dynamic (i.e., collected during ICU stay) variables in CENTER-TBI (Supplementary Note 1) to a multidimensional, evolving prognostic trajectory over each patient's ICU stay. Through supervised learning, our optimised models were trained with three main components: (1) a token-embedding encoder to integrate all variable types and missing values (Figure 1a), (2) a recurrent neural network (RNN), and (3) an ordinal outcome decoder (Figure 1b). Since model performance was independent of time window length (Supplementary Figure 4), we focused on models with two-hour time windows to offer the greatest possible trajectory resolution. With both calibration slope (averaged across the GOSE thresholds, Figure 1c) and smoothed calibration curves (Figure 1d), we observed that our modelling strategy required eight hours of information to achieve sufficient calibration for analysis (Figure 1c). However, after three days post-admission, the calibration slope of both the dynamic and baseline comparison models began decreasing, indicating a slight overfitting for TBI patients with longer ICU stays.

## *Explanation of functional outcome*

At best, the entire set of 1,166 core CENTER-TBI variables combined with our modelling strategy explained 51.3% (95% CI: 49.1%–53.3%) of the ordinal variance in six-month GOSE at 32 hours post-admission and 52.2% (95% CI: 50.2%–54.3%) at discharge (Figure 2a). Whilst overall explanation performance consistently decreased after approximately three days post-admission, the added explanation over both baseline comparison models increased over time (Figure 2b). The additional explanation of the full CENTER-TBI set over the ten IMPACT variables (Supplementary Table 1) increased from 7.0% (95% CI: 5.2%–8.7%) at 24 hours to 11.5% (95% CI: 8.5%–14.5%) at one week, and the additional explanation of ICU information increased from 2.1% (95% CI: 1.6%–2.5%) at admission to 5.2% (95% CI: 4.2%–6.2%) at one week. Therefore, at one week after admission, the ten IMPACT variables accounted for 73.9% of the explanation of functional outcome achieved by all 1,166 CENTER-TBI variables and 82.9% of that achieved by the 738 static variables.

Additionally, functional outcome explanation was, on average, 11.4% (95% CI: 6.6%–16.9%) greater in patients who stayed in the ICU for 5.75 days or less (*n*=619) than in those who stayed longer (*n*=931) (Figure 2c). Explanation performance was significantly better in shorter-stay patients from ICU admission, but the difference was not significant closer to discharge (Figure 2d). Longer-stay patients were more likely to have presented with severe TBI, received more intense treatment, and remained alive but severely disabled at six months post-injury (Supplementary Figure 5). Patients who died in the ICU were significantly more likely to have shorter stays (Supplementary Figure 6).

## *Contributions of clinical events to transitions in outcome*

In the calibrated analysis region (ten hours to one week after admission), we found a median of one (IQR: zero–three) high-magnitude transition (as defined in the Methods and Supplementary Table 2) per patient's ICU stay. The majority of identified transitions occurred within two days of ICU admission, but clinical worsening transitions occurred earlier than improvement transitions, on average (Figure 2e).

According to the TimeSHAP values[10] associated with high-magnitude transitions across the population (Figure 3a), physician-based prognostic estimates at emergency room (ER) discharge had the highest contribution to model trajectories. However, when we retrained the dynamic models without physician-based impressions (Supplementary Note 2), we found no statistically significant drop (95% confidence) in explanation percentage until one week after admission: -4.7% (95% CI: -0.2% – -9.2%) (Supplementary Figure 7). Of the remaining static variables, certain demographic (i.e., employment, age, education, and living situation), CT (i.e., subarachnoid haemorrhage, intraventricular haemorrhage, and epidural haematoma), and clinical presentation (i.e., loss of consciousness and amnesia) variables ranked highest in terms of contribution to model output. For dynamic variables, markers of raised intracranial pressure, neurological function (i.e., pupillary, motor, and verbal reactivity), and administered medication contributed the most. The highest-contributing variables were largely consistent across the six-month GOSE thresholds (Supplementary Figure 8), but TimeSHAP amplitudes generally decreased at higher thresholds, and the incidence of mechanical ventilation had a strong negative association with achieving full functional recovery (GOSE > 7). Observing the TimeSHAP values of top-contributing variables per category (Supplementary Figure 9), we found that whilst certain ER lab measurements (e.g., glucose) had significant contributions, the same lab measurements taken in the ICU did not. When comparing patients receiving withdrawal of life-sustaining treatment (WLST, $n$=203) with those who did not, TimeSHAP values for models trained with and without physician-based impressions did not reveal a significant difference in clinical events associated with high-magnitude transitions (Supplementary Figure 10). The TimeSHAP values of missing variables (Supplementary Figure 11) demonstrated that missingness of a variable could have a significant negative (e.g., missing level of education) or positive (e.g., missing heart rate value at ER admission) effect on model output. Absolute TimeSHAP values of timepoints leading up to a high-magnitude transition (Figure 3b) suggested that only ICU events that occurred within ten hours before a high-magnitude transition offered considerable contribution to the change in model output.

## *Individualised trajectories*

In Figure 4, we show the prognostic trajectories for a typical individual in our population. The patient, a c. 50-year-old male, was admitted to the ICU after a moderate TBI (GCS 10) caused by a traffic collision. The patient survived for at least six months after the injury but became severely disabled and completely functionally dependent (GOSE 2 or 3). The models correctly returned low probabilities for all six-month GOSE thresholds but GOSE > 1, for which the prognostic trajectory oscillates above and below 50% with high-magnitude transitions. In the highlighted positive high-magnitude transition (centre of Figure 4), we found improvements from the last day's GCS and the start of pharmacological thromboprophylaxis to be most strongly associated with the improvement in the patient's condition. The penultimate time window before the transition contributed the most towards the model output. In Supplementary Figure 12, we also show similarly dynamic individual trajectories for patient cases at each remaining six-month GOSE score.

# Discussion

In this work, we develop a dynamic, data-driven approach to exploit the full clinical context available in the CENTER-TBI dataset and produce individual trajectories of TBI disease course. Notably, our modelling strategy required minimal data processing for a large set of variables and imposed no constraints on the number or type of variables per patients (Figure 1a).[11] Moreover, by including missing value tokens, models discovered meaningful patterns of missingness (Supplementary Figure 11).[12] Finally, our approach detailed clinical events in terms of prognostic transitions on ordinal levels of functional recovery (Figure 1b), which is an improvement in statistical power and clinical information over using a dichotomised outcome (e.g., mortality).[8] Our modelling strategy can potentially be valuable in other heterogenous-data-intensive domains in medicine to: (1) qualify information in a dataset, (2) explore high-magnitude transitions in individuals, or (3) automate state-space characterisation for applications in reinforcement learning[13] and individualised treatment effect estimation.[14]

Our principal finding is that the full range of 1,166 core CENTER-TBI variables explained up to 52.2% (95% CI: 50.2%–54.3%) of the variance in ordinal, six-month functional outcome (Figure 2a). Up to 90.9% (95% CI: 90.3%–91.3%) of this explanation was derived from static (i.e., pre-ICU or admission) information, which constituted approximately 80% of the variables in the average patient time window (Supplementary Figure 2). Over time, the dynamic (i.e., collected during ICU stay) information increased explanation over the baseline (Figure 2b), though not enough to compensate for poorer overall performance in longer-stay (>5.75 days in ICU) patients. These patients more likely experienced severe TBI and received intense treatment without early discharge (Supplementary Figures 5–6 and as described previously[15]), and our results suggest greater, unexplained variation in recovery for this subpopulation. Remarkably, the ten IMPACT variables covered 86.3% of the explanation achieved by all 1,166 CENTER-TBI variables at 24 hours post-admission.

The greater outcome influence of static over dynamic information has several plausible explanations which could guide future TBI research and clinical management. Amongst high-resource ICUs, the variation in TBI treatment strategies has previously been shown not to result in a commensurate variation in functional outcome.[15] Since ICU treatments for TBI are mostly effective in mitigating secondary insults,[3] primary injury severity and pre-ICU circumstances may account for greater outcome differences amongst ICU patients. These explanations support both the continued use of baseline prognosis to guide treatment planning in existing practice[8] and a paradigm shift in future practice towards targeted treatment of primary injury mechanisms.[2] Moreover, the information currently collected in the ICU may not sufficiently capture pathophysiological changes that take place from the acute stage of TBI.[1] This motivates the development of more precise ICU metrics inspired by scientific discoveries of longitudinal TBI effects. Furthermore, six-month GOSE may not reflect the full contribution of ICU clinical course towards functional recovery. Upper levels of GOSE have been shown not to discriminate cognitive function well,[16] and GOSE may require repeated measurements to account adequately for day-to-day variation in questionnaire responses during recovery.[6] A multidimensional measure of TBI outcome, which integrates assessments of mental, cognitive, and physical health over time, may be preferable to six-month GOSE in revealing specific contributions of ICU events.[17] Finally, a fraction of functional outcome is likely to be explained by variations in rehabilitative care and longer-term sequalae of TBI.[18] Therefore, extending data collection past ICU discharge may reveal outcome contributions of dynamic information overlooked in our study.

The data-driven results also highlight several avenues to help account for the remaining half of functional outcome explanation. Amongst potential static variables for inclusion, genetic factors (not available for this study) are the most promising as they have been shown to explain up to 26% of the variation in dichotomised GOSE.[19] Whilst the exclusion of physician-based impressions did not significantly worsen GOSE explanation until one week post-admission (Supplementary Figure 7), the relatively high contribution of these impressions (Figure 3a) merits an investigation into the extent to which they affect self-fulfilling prophecies[20] or simply summarise other clinical variables.[21] For dynamic variables, the logical first step would be to test the inclusion of high-resolution time-series – both routinely collected (e.g., cerebral perfusion pressure[22]) and experimental (e.g., accelerometry[23]) – that have been shown to correlate with pathophysiological changes after TBI. The fact that clinical assessments of pupillary reactivity and the GCS rank amongst the highest-contributing dynamic variables (Figure 3a) may encourage the development of methods that more precisely characterise neurological mechanisms underpinning reactivity. At the same time, the relatively high information coverage of the ten IMPACT variables may suggest that the existing CENTER-TBI set could be made more concise for prognosis-based characterisation.

There are two important considerations when understanding the results of this study. First, TimeSHAP values on observational data are merely associative and cannot be interpreted for causal inference. For example, the consistently positive contribution of pharmacological thromboprophylaxis (Figure 3a) is likely explained more by the clinical selection of patients with reduced risk of intracranial bleeding for such treatment than by the effect of the treatment itself.[24] Moreover, TimeSHAP is not sufficient for developing clinical rules as it does not reveal important variable interactions.[25] We used TimeSHAP in this work not to derive claims but rather to highlight potential areas of investigation from a wider, data-driven approach, even if many of these associations may be confounded. For instance, employment status had a strong model contribution in this (Figure 3a) and prior work;[8] whether that is due to an indication of frailty (shown to be associated with lower GOSE, regardless of age[26]) or a spurious association may be worth exploring. Second, we strongly advise against using our models for clinical outcome prediction. Our explanatory modelling strategy was designed for mining patient trajectories from observational datasets and is not deployable for real-time prediction due to concerns of self-fulfilling prophecies, generalisability, and variable robustness. We refer readers interested in dynamic TBI prediction model development to the following studies.[22,27,28]

We recognise several additional limitations in this study. Our modelling strategy discretised both numerical variables into binned tokens and time into windows, which caused some loss of information. To bypass the discretisation of time, neural differential equations[29] may be a suitable alternative to RNNs but still require greater validation in medical problems. Additionally, our definition of high-magnitude transitions based on a percentile cut-off of model outputs was ultimately arbitrary. We encourage investigators either to try other percentiles or assess TimeSHAP values at known clinical events and transitions. Finally, our results may encode recruitment, collection, and clinical biases native to our European patient set and may not generalise to other populations.[30] We encourage investigators to apply our modelling strategy to other longitudinal, granular datasets of critically ill TBI patients – particularly in low- and middle-income countries where the burden of TBI is disproportionately higher[31] – and compare their results.

# Methods

## *Study design and participants*

CENTER-TBI is a longitudinal, observational cohort study (NCT02210221) involving 65 medical centres across 19 European countries.[5,9] TBI patients were prospectively recruited between 19 December 2014 and 17 December 2017 if they met the following criteria: (1) presentation within 24 hours of a TBI, (2) clinical indication for a CT scan, and (3) no severe pre-existing neurological disorder. In accordance with relevant laws of the European Union and the local country, ethical approval was obtained for each site, and written informed consent by the patient or legal representative was documented electronically. The study sites, ethical committees, approval numbers, and approval dates are listed in Supplementary Table 3. The project objectives and design of CENTER-TBI have been described in detail previously.[5,9]

In this work, we apply the following inclusion criteria in addition to those of CENTER-TBI: (1) primary admission to the ICU for at least 24 hours, (2) at least 16 years old, and (3) availability of functional outcome assessment at six months post-injury.

## *Variables and functional outcome*

We extracted all variables collected before and during ICU stays for the CENTER-TBI core study[9] (v3.0, ICU stratum) using Opal database software.[32] These variables were sourced from medical records and online test results and include structured (i.e., numerical, binary, or categorical), unstructured (i.e., free text), and missing values. We manually excluded variables which explicitly indicate death or withdrawal of life-sustaining treatment (Supplementary Note 3). In total, we included 1,166 variables (Supplementary Note 1): 738 static (i.e., fixed at ICU admission) variables and 428 dynamic variables (i.e., collected during ICU stay). We organised the variables into the nine categories listed in Table 2 and further indicated whether variables represented an ICU intervention or a physician-based impression. The highest resolution amongst regularly collected variables was once every two hours.

Additionally, we extracted the eight-point, ordinal GOSE functional outcome score at six months post-injury (heading of Table 1). Since CENTER-TBI does not distinguish vegetative patients (GOSE=2) into a separate category, GOSE scores 2 and 3 (lower severe disability) were combined to one category (GOSE$\in$\{2,3\}). For 12.8% of our study patients, six-month GOSE scores were previously imputed by CENTER-TBI using a Markov multi-state model based on the observed GOSE scores recorded at different timepoints between two weeks to one-year post-injury.[33]

## *Modelling strategy*

We created 100 partitions of our patient population for repeated *k*-fold cross-validation (20 repeats, 5 folds), stratified by six-month GOSE, with validation sets nested within training sets.

Our explanatory modelling strategy is outlined in Figure 1 and builds upon our previous work.[8,11] We started by partitioning ICU stays into non-overlapping time windows of either 2, 8, 12, or 24 hours. Static variables were carried forward across all windows (Figure 1a). All variables were tokenised through one of the following methods: (1) for categorical variables, appending the value to the variable name, (2) for numerical variables, learning the training set distribution and discretising into either 3, 4, 5, 7, 10, or 20 quantile bins, (3) for text-based entries, removing all special characters, spaces, and capitalisation from the text and appending to the variable name, and (4) for missing values, creating a separate token to designate missingness (Figure 1a). By labelling missing values with separate tokens instead of imputing them, the models could learn potentially significant patterns of missingness and integrate a diverse range of missing data without needing to validate the assumptions of imputation methods on each variable. During training, the models learned a low dimensional vector (of either 16, 32, 64, or 128 units) and a “relevance” weight for each token in the training set. Therefore, models would take the unique tokens from each time window of a patient, replace them with the corresponding vectors, and average the vectors – weighted by relevance – into a single vector per time window (Figure 1a).

Each patient’s sequence of low-dimensional vectors then fed into a RNN – either long short-term memory (LSTM) or gated recurrent unit (GRU) – to output another vector per time window. In this manner, the models learned temporal patterns of variable interactions from training set ICU records and updated outputs with each new time window of data. Finally, each RNN output vector was decoded – either with a multinomial (i.e., softmax) or ordinal (i.e., constrained sigmoid) output layer – to return a probability at each threshold of six-month GOSE over time (Figure 1b).

The combinations of hyperparameters – in addition to those already mentioned (time window length, quantile bin count, embedding vector dimension, RNN type, and output layer) – and their optimisation results are reported in the Supplementary Methods.

## Model and information evaluation

All metrics, curves, and associated confidence intervals (CIs) were calculated on the testing sets using the repeated Bootstrap Bias Corrected Cross-Validation (BBC-CV) method.[34] We calculated metrics and CIs at each timepoint after ICU admission as well as at each timepoint leading up to ICU discharge.

The reliability of model-generated trajectories was assessed through the calibration of output probabilities at each threshold of six-month GOSE. Using the logistic recalibration framework,[35] we first measured calibration slope. Calibration slope less(/greater) than one indicates overfitting(/underfitting).[35] Additionally, we examined smoothed probability calibration curves to detect miscalibrations that might have been overlooked by the logistic recalibration framework.[35]

We also assessed the information quality achieved by the combination of our modelling strategy and the CENTER-TBI variables by calculating Somers' $D_{xy}$.[36] In our context, Somers' $D_{xy}$ is interpreted as the proportion of ordinal variation in six-month GOSE that is explained by the variation in model output.[37] The calculation of Somers' $D_{xy}$ is detailed in the Supplementary Methods.

We compared the performance of our modelling strategy with that of two baseline models on the same remaining patients over time. The first was a multinomial logistic regression model trained on the ten, static IMPACT extended variables (Supplementary Table 1) from the first 24 hours of ICU stay (i.e., the validated standard for static prognosis).[8] The second was our developed modelling strategy but trained only on the 738 static variables in CENTER-TBI to measure added explanation by ICU information.

## *High-magnitude transition identification and explanation*

Within the calibrated region of testing set model outputs, we found high-magnitude transitions, both negative (i.e., worsening) and positive (i.e., improvement), of model-generated probabilities at each threshold of six-month GOSE. High-magnitude transitions were arbitrarily defined by a consecutive time-window difference in probability that:

- for negative transitions, was less than or equal to the 1$^{st}$ percentile of negative differences for a given GOSE threshold across the population,
- for positive transitions, was greater than or equal to the 99$^{th}$ percentile of positive differences for a given GOSE threshold across the population.

The cut-offs for high-magnitude transitions are listed in Supplementary Table 2.

To uncover the variables associated with high-magnitude transitions, we applied the TimeSHAP algorithm.[10] TimeSHAP estimates the relative contribution of both tokens and time windows towards an individual's model output by perturbing the clinical events leading up to a high-magnitude transition. TimeSHAP applies a temporal coalition pruning algorithm which groups low-contributing time windows in the distant past together as a single feature (otherwise, the calculation would be computationally intractable given the number of tokens).

At the timepoints of high-magnitude transition, we calculated TimeSHAP for contributions towards both the threshold probability of six-month GOSE and the expected six-month GOSE index (Supplementary Figure 13 and the Supplementary Methods). For the TimeSHAP baseline, we defined an "average patient" to be one with tokens that are in 50$^{+}$% of training set time windows. Therefore, TimeSHAP values were interpreted as associative contributions of tokens or timesteps towards the difference in a patient's model output from that of the average patient.

# Data availability

Individual participant data, including data dictionary, the study protocol, and analysis scripts are available online, conditional to approved study proposal, with no end date. Interested investigators must submit a study proposal to the management committee at https://www.center-tbi.eu/data. Signed confirmation of a data access agreement is required, and all access must comply with regulatory restrictions imposed on the original study.

# Code availability

All code used in this project can be found at the following online repository: https://github.com/sbhattacharyay/dynamic_GOSE_model (doi:10.5281/zenodo.7668551).

# Acknowledgments

This research was supported by the National Institute for Health Research (NIHR) Brain Injury MedTech Co-operative. CENTER-TBI was supported by the European Union 7th Framework programme (EC grant 602150). Additional funding was obtained from the Hannelore Kohl Stiftung (Germany), from OneMind (USA), and from Integra LifeSciences Corporation (USA). CENTER-TBI also acknowledges interactions and support from the International Initiative for TBI Research (InTBIR) investigators. SB is funded by a Gates Cambridge fellowship. The funders had no role in study design, data collection and analysis, decision to publish, or preparation of the manuscript.

We are grateful to the patients and families of our study for making our efforts to improve TBI care possible. S.B. would like to thank Kathleen Mitchell-Fox (Princeton University) and Andrew Maas (Antwerp University Hospital) for offering comments on the manuscript.

# Author contributions

S.B. co-conceptualised the aims, developed the methodology and design, analysed and visualised the data, acquired funding, and wrote the manuscript. P.F.C. aided in model design and reviewed the manuscript. C.A. and L.W. curated data and reviewed the manuscript. R.D.S. reviewed the manuscript. D.K.M. and E.W.S. curated data, acquired funding, advised statistical analysis, and reviewed the manuscript. D.W.N. curated data, aided in methodology development, and reviewed the manuscript. A.E. served as principal investigator, curated data, co-conceptualized the aims, and reviewed the manuscript. All authors read and approved the final manuscript.

# Competing interests

All authors declare no financial or non-financial competing interests.

# Figure legends

**Figure 1. Illustration and reliability of disease course modelling strategy.** Unless otherwise specified, all shaded regions surrounding curves are 95% confidence intervals derived using bias-corrected bootstrapping (1,000 resamples) to represent the variation across 20 repeated five-fold cross-validation partitions. (**a**) Tokenisation and embedding of the variables in a sample patient's ICU stay into a single, low dimensional vector ($x_t$) per time window. The patient's ICU stay (sample timeline) was first discretised into non-overlapping, two-hour time windows. From each time window, values for up to 428 dynamic variables were combined with values for up to 738 static variables to form the variable set (Supplementary Note 1). The variable values were converted to tokens by discretising numerical values (e.g., intracranial pressure [ICP] and neurofilament light chain [NF-L]) into 20-quantile bins from the training set and removing special formatting from text-based entries. Through an embedding layer, a vector was learned for each token encountered in the training set, and tokens were replaced with these vectors. Finally, a positive relevance weight, also learned for each token, was used to weight-average the vectors of a time window into a single, low-dimensional vector. The patient stock image is accredited to iStock.com/SiberianArt and was purchased under Standard License. (**b**) The sequence of low-dimensional vectors ($x_t$) representing a patient's ICU stay were fed into a recurrent neural network (RNN) with either long short-term memory (LSTM) or gated recurrent unit (GRU) cells. The RNN outputs were then decoded at each time window into an ordinal prognosis of six-month functional outcome. The level of recovery associated with each threshold of six-month GOSE is decoded in the heading of Table 1 (e.g., GOSE > 1 represents survival at six months post-injury). (**c**) Probability calibration slope, averaged across the six functional

outcome thresholds, in the first (*top*) and last (*bottom*) week of ICU stay for models trained on the full variable set (*blue*) and on the static IMPACT extended set from the first 24 hours of ICU stay (*red*). The ideal calibration slope of one is marked with a horizontal orange line. (**d**) Ordinal probability calibration curves at four different timepoints after ICU admission. The diagonal dashed line represents the line of perfect calibration. The values in each panel correspond to the mean absolute error (95% confidence interval) between the curve and the perfect calibration line.

**Figure 2. Explanation of outcome by modelling strategy and distribution of high-magnitude transitions.** Unless otherwise specified, all shaded regions surrounding curves are 95% confidence intervals derived using bias-corrected bootstrapping (1,000 resamples) to represent the variation across 20 repeated five-fold cross-validation partitions. (**a**) Explanation of ordinal six-month functional outcome – measured by Somers' $D_{xy}$ – in the first (*top*) and last (*bottom*) week of ICU stay by models trained on the full variable set (*blue*) and on the static IMPACT extended set from the first 24 hours of ICU stay (*red*). (**b**) Added explanation of ordinal six-month functional outcome – measured by difference in Somers' $D_{xy}$ – in the first (*top*) and last (*bottom*) week of ICU stay achieved by the full variable model over baseline models trained on all static variables (*blue*) and on the static IMPACT extended set from the first 24 hours of ICU stay (*red*). (**c**) Mean difference in full variable model explanation – measured by difference in Somers' $D_{xy}$ – between subpopulation with ICU stay less than or equal to cut-off and subpopulation with ICU stay greater than the same cut-off. Positive values designate greater explanation in shorter-stay subpopulation, and the horizontal orange line designates no difference. (**d**) Explanation of ordinal six-month functional outcome – measured by Somers' $D_{xy}$

– in the first (*top*) and last (*bottom*) week of ICU stay by the full variable model on the subpopulation with ICU stay less than or equal to 5.75 days (*blue*) and on the subpopulation with ICU greater than 5.75 days (*red*). (**e**) Scaled histograms (bin width equals a two-hour time window) and density curves of high-magnitude transitions identified by model trajectories. A high-magnitude transition is defined as a change of output probability that is in the 99$^{th}$ percentile of changes for a specific threshold of six-month functional outcome. The blue histogram/density represent positive (i.e., improvement) transitions whilst the purple histogram/density represent negative (i.e., worsening) transitions. The dashed orange lines mark the median time since ICU admission for each type of transition.

**Figure 3. Population-level variable and time window contributions to expected six-month functional outcome output at high-magnitude transitions.** TimeSHAP values are interpreted as contributions of variables or time windows towards the difference in a patient's expected six-month functional outcome output from that of the average patient (Supplementary Figure 13). (**a**) TimeSHAP values of the 20 highest-contribution static (left) and 20 highest-contribution dynamic (right) variables. The variables were selected by first identifying the ten variables with non-missing value tokens with the most negative median TimeSHAP values across the population (above the ellipses) and then, amongst the remaining variables, selecting the ten with non-missing value tokens with the most positive median TimeSHAP values (below the ellipses). Each point represents the mean TimeSHAP value for a token across an individual patient's high-magnitude transitions. The colour of the point represents the relative ordered value of a token within a variable, and for unordered variables (e.g., employment status before injury), tokens were sorted

alphanumerically (the sort index per possible unordered variable token is provided in the Supplementary Note 1). Green points represent variable tokens that are not missing but explicitly encode an unknown value (i.e., GCS motor score untestable due to sedation). New variable abbreviations include deep vein thrombosis (DVT), fraction of inspired oxygen (FiO2), partial pressure of oxygen (PaO2), and unfavourable outcome (UO) as defined by GOSE $\leq$ 4 at six months post-injury. (**b**) TimeSHAP amplitude distributions of two-hour time windows leading up to high-magnitude transitions. The width of violin plots is scaled for each time window, but the width of the points inside them demonstrates relative frequency across the windows.

**Figure 4. Example of individual ICU disease course with explanations for high-magnitude transition.** TimeSHAP values are interpreted as contributions of variables or time windows towards the difference in this patient's expected six-month functional outcome output from that of the average patient (Supplementary Figure 13). The patient was a c. 50-year-old male, admitted to the ICU after a moderate traumatic brain injury (GCS 10), who became severely disabled (SD) with full functional dependency by six months post-injury (GOSE 2 or 3). The patient presented with a subarachnoid haemorrhage (SAH) and received emergency intracranial surgery (IC) and a decompressive craniectomy (DC). New variable abbreviations include deep vein thrombosis (DVT), eye component score of GCS (GCSe), left (L), and right (R).

# Figure 1

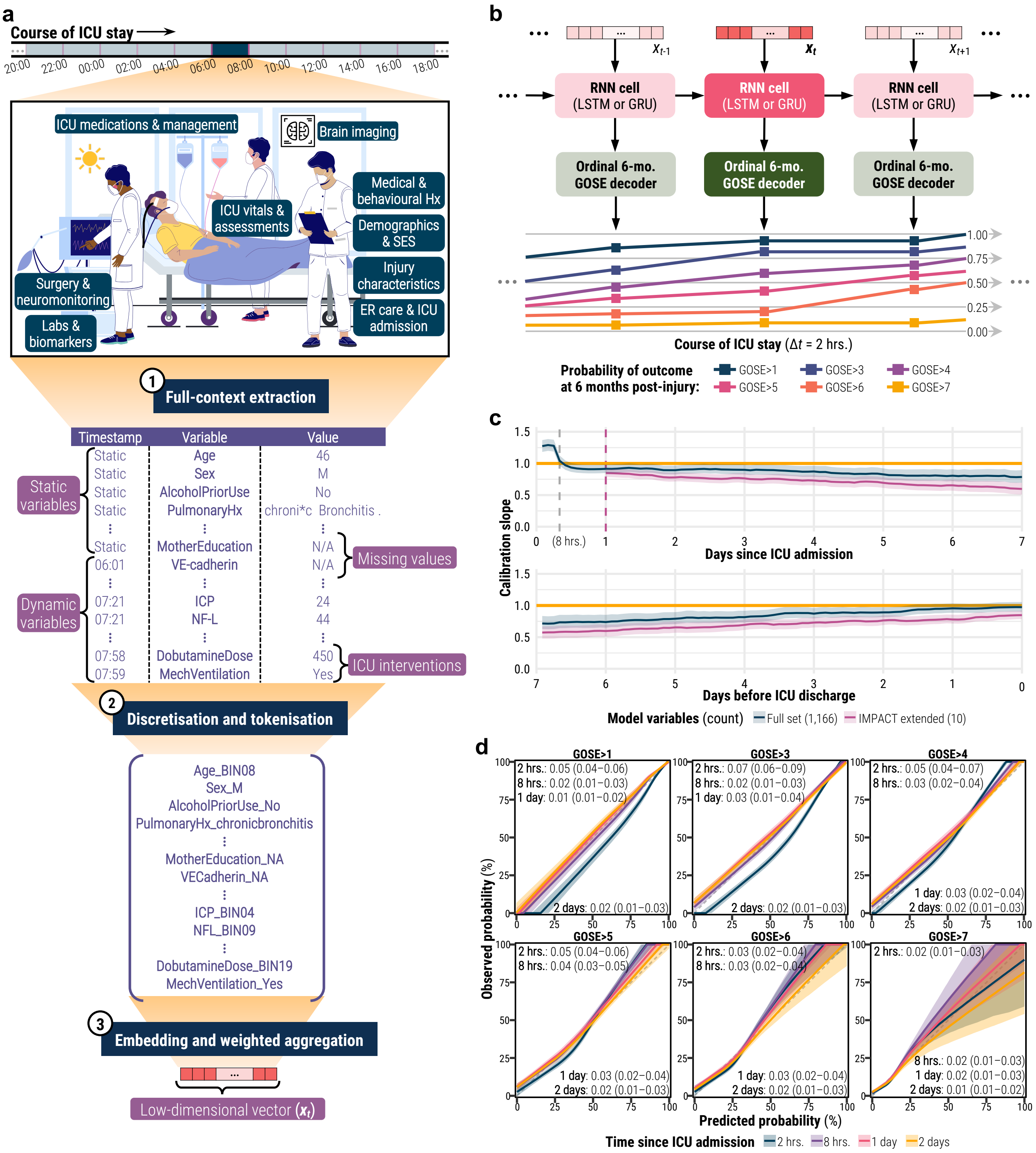

a
Course of ICU stay
20:00 22:00 00:00 02:00 04:00 06:00 08:00 10:00 12:00 14:00 16:00 18:00
ICU medications & management
Brain imaging
Medical & behavioural Hx
ICU vitals & assessments
Demographics & SES
Injury characteristics
Surgery & neuromonitoring
ER care & ICU admission
Labs & biomarkers
1 Full-context extraction
Timestamp Variable Value
Static Age 46
Static Sex M
Static AlcoholPriorUse No
Static PulmonaryHx chroni*c Bronchitis .
Static MotherEducation N/A
06:01 VE-cadherin N/A
07:21 ICP 24
07:21 NF-L 44
07:58 DobutamineDose 450
07:59 MechVentilation Yes
Static variables
Dynamic variables
Missing values
ICU interventions
2 Discretisation and tokenisation
Age_BIN08
Sex_M
AlcoholPriorUse_No
PulmonaryHx_chronicbronchitis
MotherEducation_NA
VECadherin_NA
ICP_BIN04
NFL_BIN09
DobutamineDose_BIN19
MechVentilation_Yes
3 Embedding and weighted aggregation
Low-dimensional vector ($x_t$)
b
$x_{t-1}$ $x_t$ $x_{t+1}$
RNN cell (LSTM or GRU)
Ordinal 6-mo. GOSE decoder
1.00 0.75 0.50 0.25 0.00
Course of ICU stay (Δt = 2 hrs.)
Probability of outcome at 6 months post-injury:
GOSE>1 GOSE>3 GOSE>4 GOSE>5 GOSE>6 GOSE>7
c
Calibration slope
Days since ICU admission
0 (8 hrs.) 1 2 3 4 5 6 7
Days before ICU discharge
7 6 5 4 3 2 1 0
Model variables (count) Full set (1,166) IMPACT extended (10)
d
GOSE>1
2 hrs.: 0.05 (0.04–0.06)
8 hrs.: 0.02 (0.01–0.03)
1 day: 0.01 (0.01–0.02)
2 days: 0.02 (0.01–0.03)
GOSE>3
2 hrs.: 0.07 (0.06–0.09)
8 hrs.: 0.02 (0.01–0.03)
1 day: 0.03 (0.01–0.04)
2 days: 0.02 (0.01–0.03)
GOSE>4
2 hrs.: 0.05 (0.04–0.07)
8 hrs.: 0.03 (0.02–0.04)
1 day: 0.03 (0.02–0.04)
2 days: 0.02 (0.01–0.03)
GOSE>5
2 hrs.: 0.05 (0.04–0.06)
8 hrs.: 0.04 (0.03–0.05)
1 day: 0.03 (0.02–0.04)
2 days: 0.02 (0.01–0.03)
GOSE>6
2 hrs.: 0.03 (0.02–0.04)
8 hrs.: 0.03 (0.02–0.04)
1 day: 0.03 (0.02–0.04)
2 days: 0.02 (0.01–0.03)
GOSE>7
2 hrs.: 0.02 (0.01–0.03)
8 hrs.: 0.02 (0.01–0.03)
1 day: 0.02 (0.01–0.03)
2 days: 0.01 (0.01–0.02)
Observed probability (%)
Predicted probability (%)
Time since ICU admission 2 hrs. 8 hrs. 1 day 2 days

## Figure 2

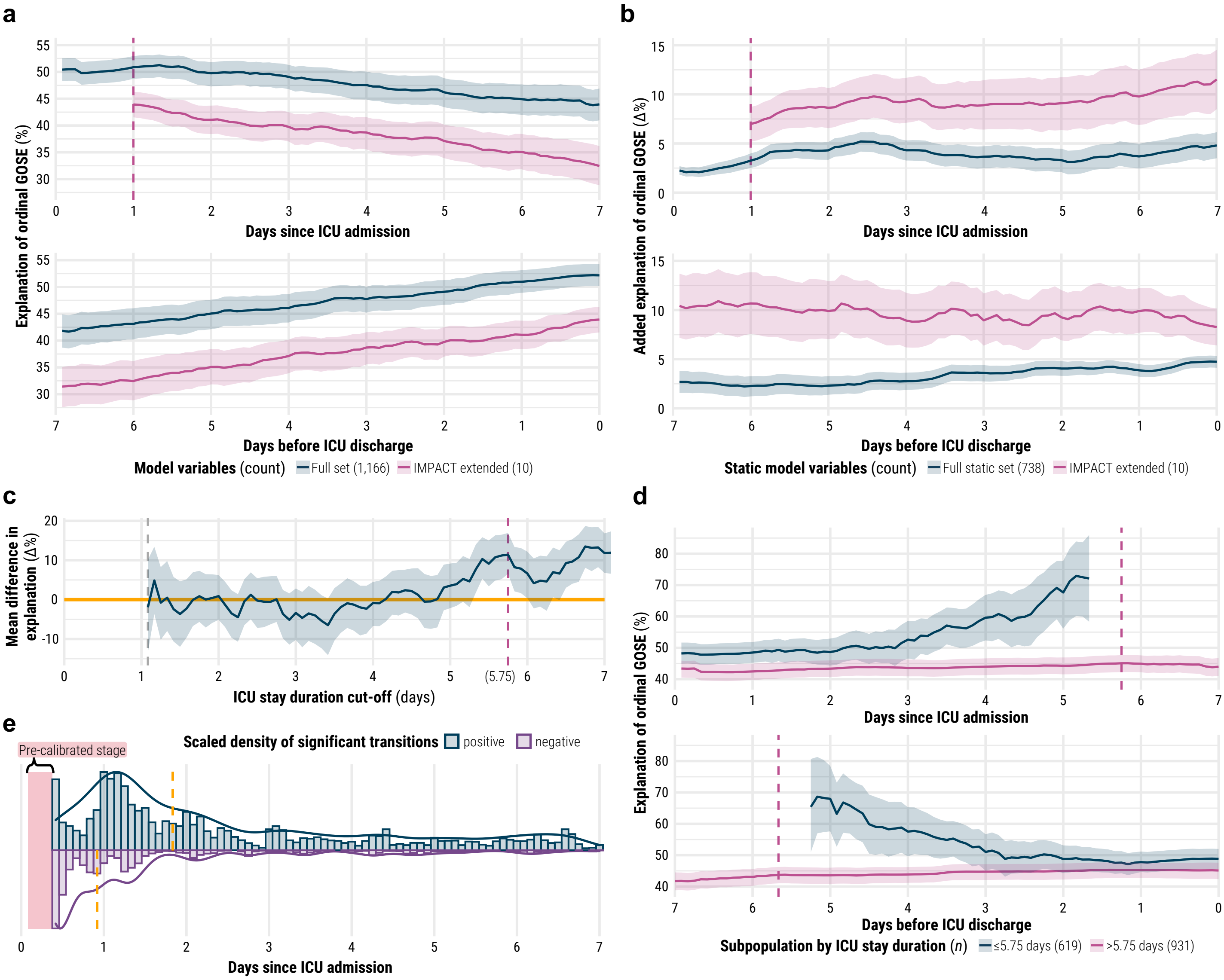

a
Explanation of ordinal GOSE (%)
Days since ICU admission
Days before ICU discharge
Model variables (count)
Full set (1,166)
IMPACT extended (10)
b
Added explanation of ordinal GOSE (Δ%)
Days since ICU admission
Days before ICU discharge
Static model variables (count)
Full static set (738)
IMPACT extended (10)
c
Mean difference in explanation (Δ%)
ICU stay duration cut-off (days)
(5.75)
d
Explanation of ordinal GOSE (%)
Days since ICU admission
Days before ICU discharge
Subpopulation by ICU stay duration (n)
≤5.75 days (619)
>5.75 days (931)
e
Pre-calibrated stage
Scaled density of significant transitions
positive
negative
Days since ICU admission

# Figure 3

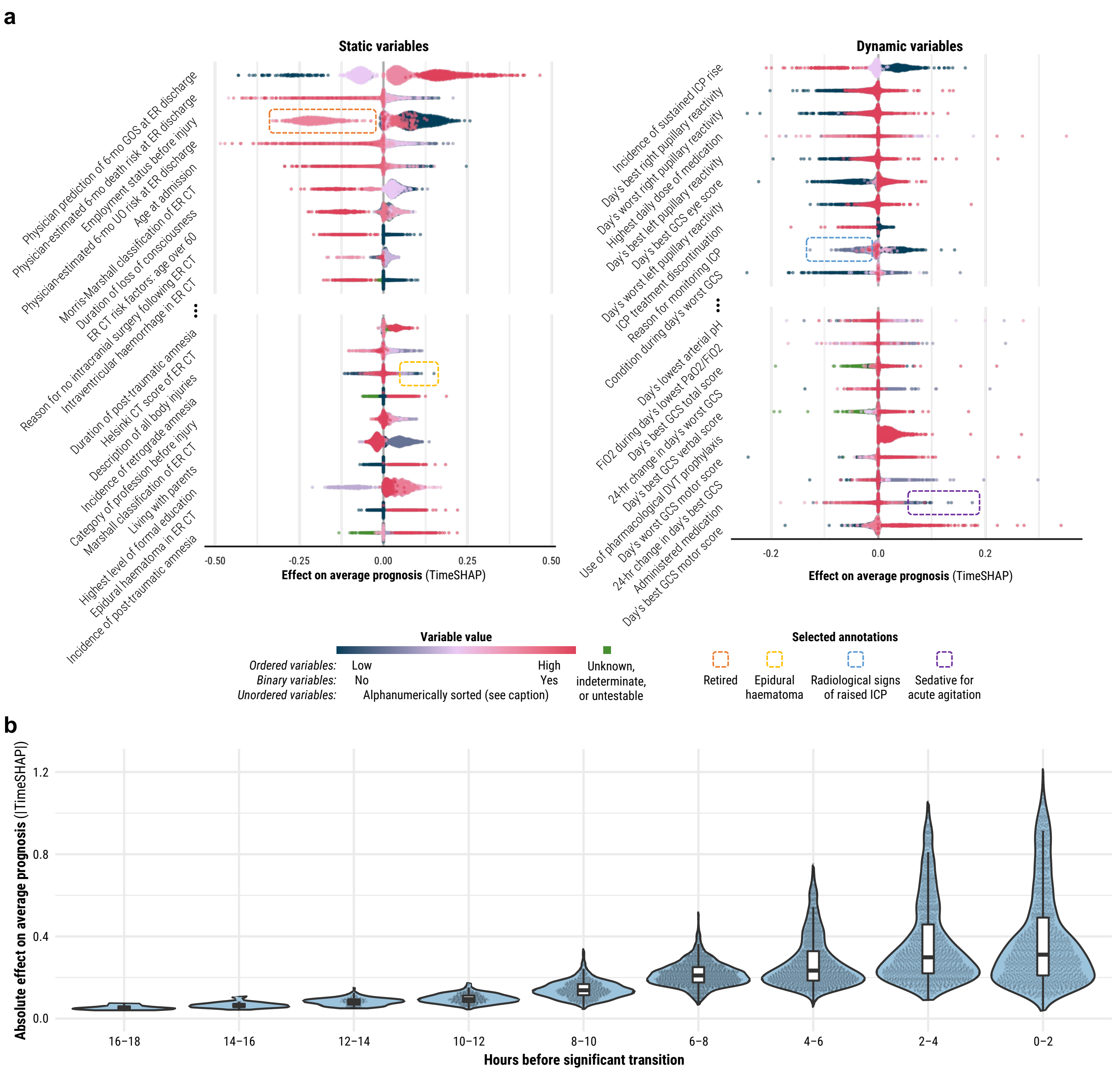
a
Static variables
Dynamic variables
Physician prediction of 6-mo GOS at ER discharge
Physician-estimated 6-mo death risk at ER discharge
Employment status before injury
Physician-estimated 6-mo UO risk at ER discharge
Age at admission
Morris-Marshall classification of ER CT
Duration of loss of consciousness
ER CT risk factors: age over 60
Reason for no intracranial surgery following ER CT
Intraventricular haemorrhage in ER CT
Duration of post-traumatic amnesia
Helsinki CT score of ER CT
Description of all body injuries
Incidence of retrograde amnesia
Category of profession before injury
Marshall classification of ER CT
Living with parents
Highest level of formal education
Epidural haematoma in ER CT
Incidence of post-traumatic amnesia
Incidence of sustained ICP rise
Day's best right pupillary reactivity
Day's worst right pupillary reactivity
Highest daily dose of medication
Day's best left pupillary reactivity
Day's best GCS eye score
Day's worst left pupillary reactivity
ICP treatment discontinuation
Reason for monitoring ICP
Condition during day's worst GCS
Day's lowest arterial pH
FiO2 during day's lowest PaO2/FiO2
Day's best GCS total score
24-hr change in day's worst GCS
Day's best GCS verbal score
Use of pharmacological DVT prophylaxis
Day's worst GCS motor score
24-hr change in day's best GCS
Administered medication
Day's best GCS motor score
-0.50
-0.25
0.00
0.25
0.50
Effect on average prognosis (TimeSHAP)
-0.2
0.0
0.2
Effect on average prognosis (TimeSHAP)
Variable value
Ordered variables: Low High
Binary variables: No Yes
Unordered variables: Alphanumerically sorted (see caption)
Unknown, indeterminate, or untestable
Selected annotations
Retired
Epidural haematoma
Radiological signs of raised ICP
Sedative for acute agitation
b
Absolute effect on average prognosis (|TimeSHAP|)
0.0
0.4
0.8
1.2
16–18
14–16
12–14
10–12
8–10
6–8
4–6
2–4
0–2
Hours before significant transition

# Figure 4

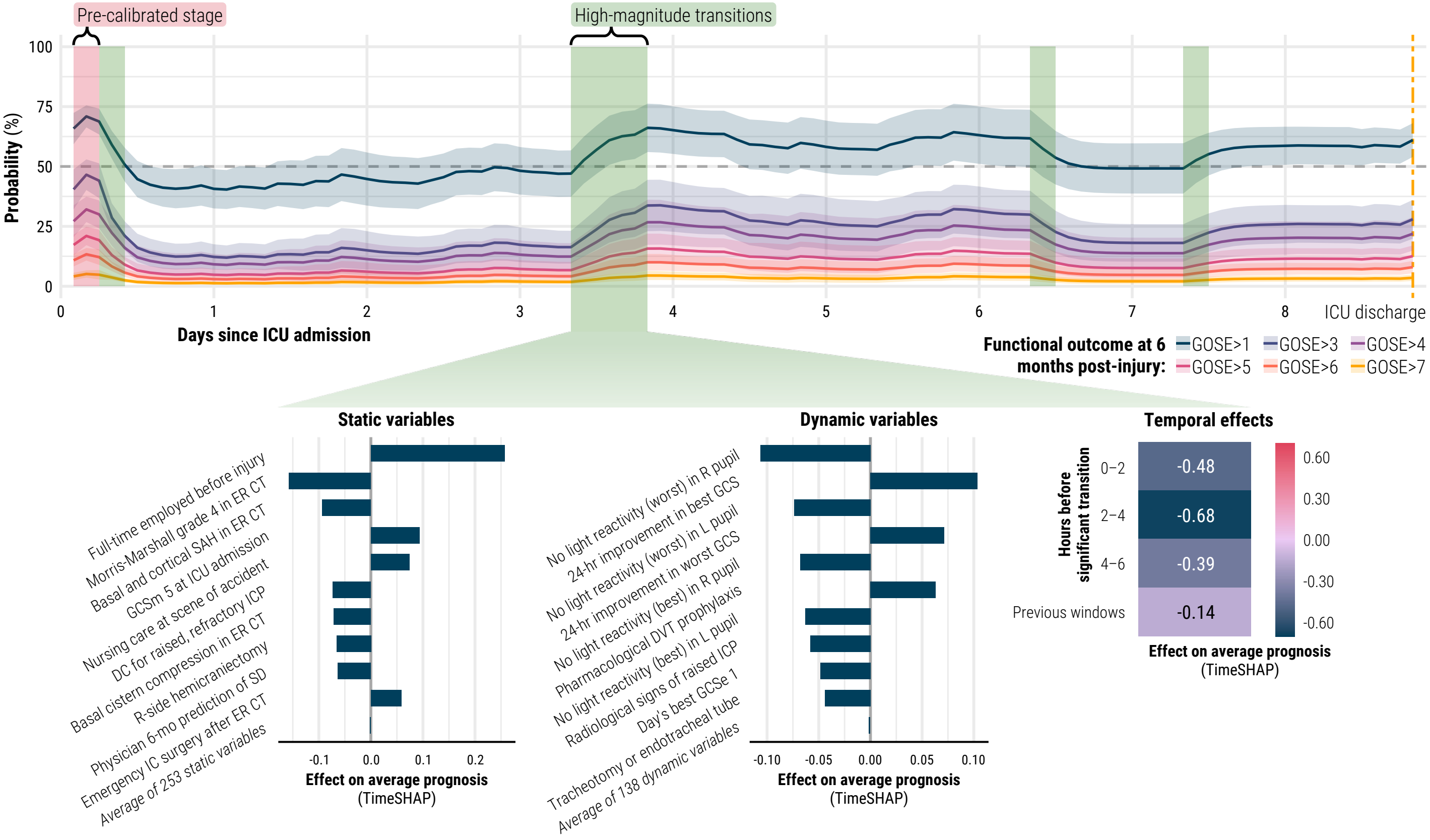
Pre-calibrated stage
High-magnitude transitions
Probability (%)
100
75
50
25
0
0
1
2
3
4
5
6
7
8
ICU discharge
Days since ICU admission
Functional outcome at 6 months post-injury:
GOSE>1
GOSE>3
GOSE>4
GOSE>5
GOSE>6
GOSE>7
Static variables
Full-time employed before injury
Morris-Marshall grade 4 in ER CT
Basal and cortical SAH in ER CT
GCSm 5 at ICU admission
Nursing care at scene of accident
DC for raised, refractory ICP
Basal cistern compression in ER CT
R-side hemicraniectomy
Physician 6-mo prediction of SD
Emergency IC surgery after ER CT
Average of 253 static variables
-0.1
0.0
0.1
0.2
Effect on average prognosis
(TimeSHAP)
Dynamic variables
No light reactivity (worst) in R pupil
24-hr improvement in best GCS
No light reactivity (worst) in L pupil
24-hr improvement in worst GCS
No light reactivity (best) in R pupil
Pharmacological DVT prophylaxis
No light reactivity (best) in L pupil
Radiological signs of raised ICP
Day's best GCSe 1
Tracheotomy or endotracheal tube
Average of 138 dynamic variables
-0.10
-0.05
0.00
0.05
0.10
Effect on average prognosis
(TimeSHAP)
Temporal effects
Hours before significant transition
0–2
2–4
4–6
Previous windows
-0.48
-0.68
-0.39
-0.14
0.60
0.30
0.00
-0.30
-0.60
Effect on average prognosis
(TimeSHAP)

**Table 1. Summary characteristics of the study population at ICU admission stratified by ordinal 6-month outcomes.**

| Summary characteristics | Overall | Glasgow Outcome Scale – Extended (GOSE) at 6 months post-injury | | | | | | | *p*-value |
|---|---|---|---|---|---|---|---|---|---|
| | | (1) Death | (2 or 3) Vegetative or lower severe disability | (4) Upper severe disability | (5) Lower moderate disability | (6) Upper moderate disability | (7) Lower good recovery | (8) Upper good recovery | |
| *n* | 1550 | 318 (20.5%) | 262 (16.9%) | 120 (7.7%) | 227 (14.6%) | 200 (12.9%) | 206 (13.3%) | 217 (14.0%) | |
| Age [years] | 51 (31–66) | 66 (50–76) | 55 (36–68) | 48 (29–61) | 44 (31–56) | 41 (27–53) | 48 (31–65) | 41 (24–61) | <0.0001 |
| Sex | | | | | | | | | 0.94 |
| Female | 409 (26.4%) | 78 (24.5%) | 71 (27.1%) | 43 (35.8%) | 64 (28.2%) | 49 (24.5%) | 59 (28.6%) | 45 (20.7%) | |
| Baseline GCS (*n* = 1465) | 8 (4–14) | 5 (3–10) | 6 (3–10) | 8 (4–13) | 8 (5–13) | 9 (6–14) | 13 (7–15) | 13 (8–15) | <0.0001 |
| Mild [13–15] | 390 (26.6%) | 30 (10.3%) | 38 (15.3%) | 26 (23.4%) | 42 (19.5%) | 66 (34.9%) | 91 (45.3%) | 97 (46.4%) | |
| Moderate [9–12] | 331 (22.6%) | 65 (22.3%) | 41 (16.5%) | 28 (25.2%) | 65 (30.2%) | 36 (19.0%) | 40 (19.9%) | 56 (26.8%) | |
| Severe [3–8] | 744 (50.8%) | 196 (67.4%) | 170 (68.3%) | 57 (51.4%) | 108 (50.2%) | 87 (46.0%) | 70 (34.8%) | 56 (26.8%) | |
| Marshall CT (*n* = 1255) | VI (II–VI) | III (II–VI) | II (II–VI) | II (II–VI) | II (II–II) | II (II–III) | II (II–II) | VI (II–VI) | 0.02 |
| No visible pathology (I) | 118 (9.4%) | 8 (3.3%) | 11 (5.3%) | 5 (5.2%) | 17 (8.7%) | 25 (15.2%) | 24 (13.6%) | 28 (16.5%) | |
| Diffuse injury II | 592 (47.2%) | 56 (22.8%) | 84 (40.6%) | 54 (56.2%) | 92 (47.2%) | 100 (60.6%) | 103 (58.5%) | 103 (60.6%) | |
| Diffuse injury III | 108 (8.6%) | 42 (17.1%) | 17 (8.2%) | 10 (10.4%) | 14 (7.2%) | 9 (5.5%) | 6 (3.4%) | 10 (5.9%) | |
| Diffuse injury IV | 16 (1.3%) | 7 (2.8%) | 1 (0.5%) | 1 (1.0%) | 4 (2.1%) | 1 (0.6%) | 1 (0.6%) | 1 (0.6%) | |
| Mass lesion (V & VI) | 421 (33.5%) | 133 (54.0%) | 94 (45.4%) | 26 (27.1%) | 68 (34.9%) | 30 (18.2%) | 42 (23.9%) | 28 (16.5%) | |
| tSAH (*n* = 1254) | 957 (76.3%) | 221 (90.2%) | 176 (84.2%) | 73 (76.0%) | 150 (76.9%) | 106 (63.9%) | 125 (71.4%) | 106 (63.1%) | 0.12 |
| EDH (*n* = 1257) | 244 (19.4%) | 31 (12.7%) | 32 (15.3%) | 21 (21.9%) | 46 (23.6%) | 32 (19.3%) | 42 (23.9%) | 40 (23.5%) | 0.01 |
| Retired (*n* = 1312) | 353 (26.9%) | 136 (61.3%) | 74 (33.6%) | 23 (22.1%) | 12 (5.9%) | 13 (7.3%) | 52 (28.1%) | 43 (21.8%) | 0.02 |
| Length of ICU stay [days] | 8.3 (3.0–16.9) | 7.0 (2.9–14.2) | 17.5 (10.1–24.9) | 13.1 (4.6–21.7) | 9.1 (3.7–16.3) | 6.9 (2.8–14.9) | 3.7 (2.1–9.7) | 3.9 (1.8–9.1) | <0.0001 |

Data are count (% of total study) for sample size (*n*), median (IQR) for continuous characteristics, and *n* (% of column group) for categorical characteristics. Units or numerical definitions of characteristics are provided in square brackets. If a characteristic had missing values for some patients in the population, the non-missing sample size was provided in parentheses – e.g., Marshall CT (*n* = 1255). Conventionally, TBI severity is categorically defined by baseline GCS scores as indicated in square brackets. Incidence of epidural haematoma (EDH) or traumatic subarachnoid haemorrhage (tSAH) was assessed from CT scan at ICU admission. *p*-values are determined from proportional odds logistic regression (POLR) coefficient analysis trained on all summary characteristics concurrently. For categorical variables with $k > 2$ categories (e.g., Baseline GCS), $p$-values were calculated with a likelihood ratio test (with $k$-1 degrees of freedom) on POLR.

**Table 2. Variable count per category and subtype.**

| Category | Example variable | Count by subtypes | | | | |
|---|---|---|---|---|---|---|
| | | All | Static | Dynamic | Interventions | Physician impressions |
| Demographics and socioeconomic status | Lives with parents | 22 | 22 | 0 | 0 | 0 |
| Medical and behavioural history | Takes beta blockers | 188 | 188 | 0 | 0 | 0 |
| Injury characteristics and severity | Helmet on during accident | 84 | 84 | 0 | 0 | 0 |
| Emergency care and ICU admission | Physician prognosis at ER discharge | 246 | 246 | 0 | 0 | 16 |
| Brain imaging reports | Midline shift | 186 | 108 | 78 | 0 | 25 |
| Laboratory measurements | Glial fibrillary acidic protein | 228 | 81 | 147 | 0 | 1 |
| ICU medications and management | Fluid loading | 108 | 3 | 105 | 75 | 17 |
| ICU vitals and assessments | Bihourly systolic blood pressure | 67 | 0 | 67 | 0 | 0 |
| Surgery and neuromonitoring | Decompressive craniectomy | 37 | 6 | 31 | 7 | 18 |
| Total | | 1166 | 738 | 428 | 82 | 77 |

Data represent the number of subtype (column) variables per category (row).

## Supplementary Information

TABLE OF CONTENTS

## SUPPLEMENTARY FIGURES

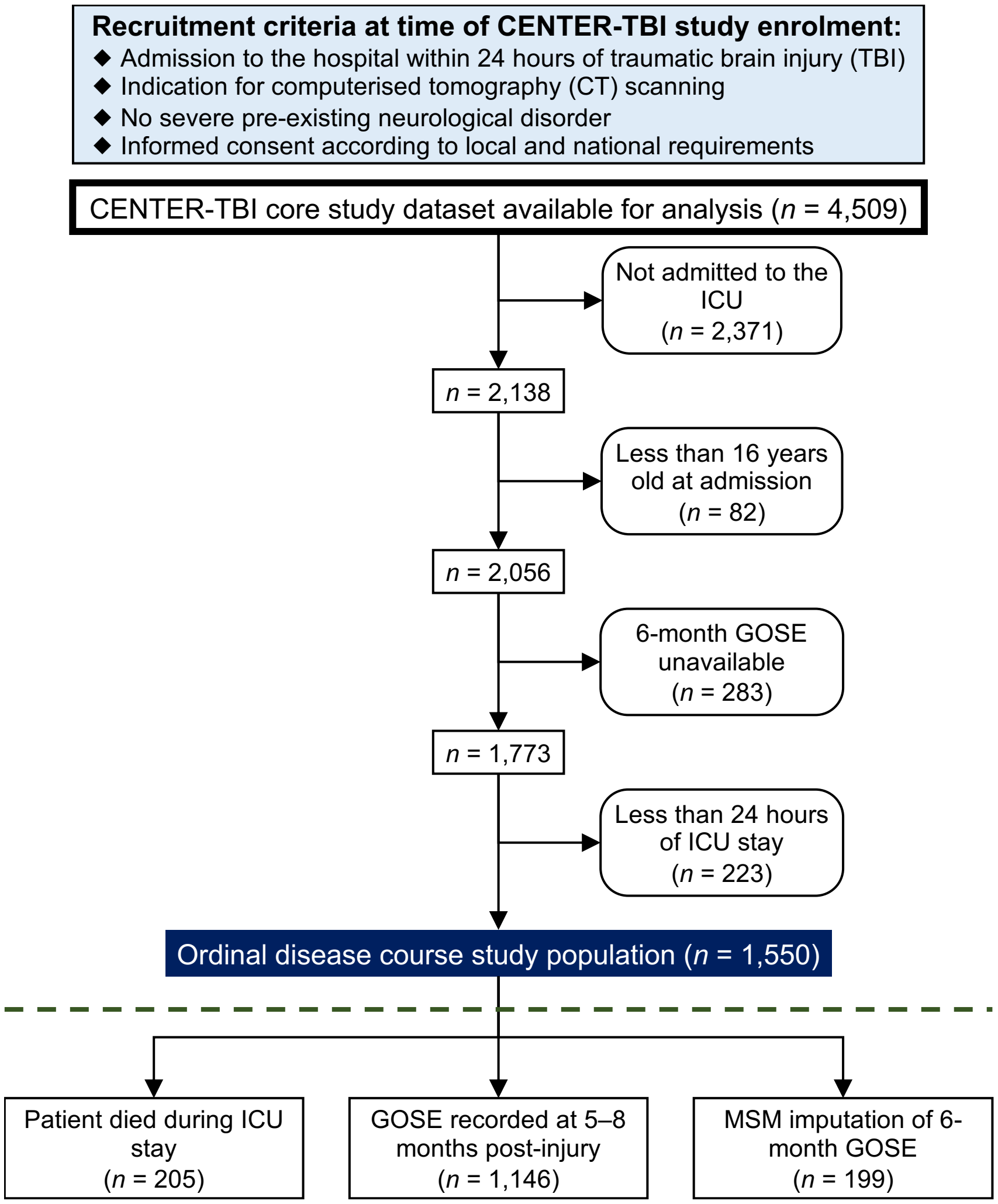

**Supplementary Figure 1. Flow diagram for patient enrolment and follow-up.** The dashed, olive-green line in the lower-middle of the diagram divides the enrolment flow diagram (above) and the follow-up breakdown (below). Imputation of missing six-month GOSE values with a Markov multi-state model (MSM) is described in the Methods.

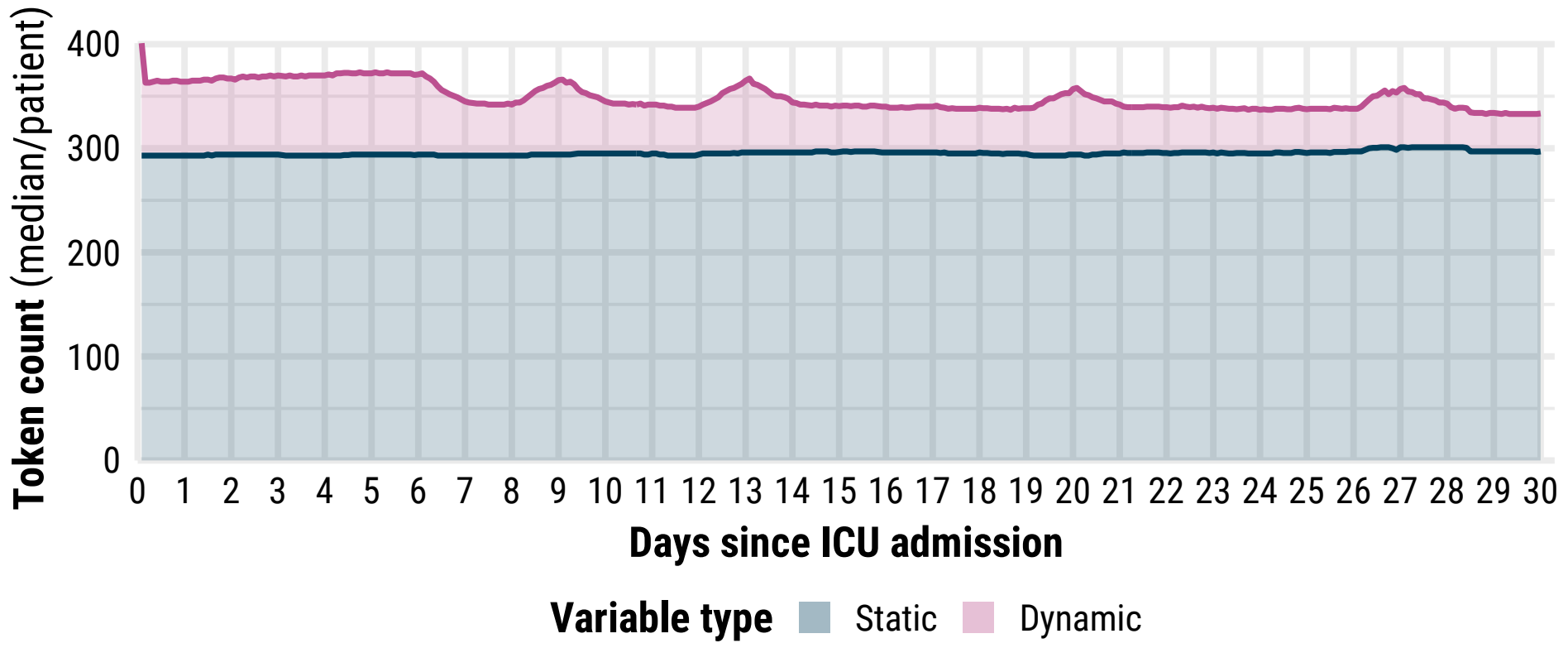

**Supplementary Figure 2. Per-patient token information over the first month of ICU stay.** The blue and purple lines represent the median number of non-missing value tokens of static and dynamic variables, respectively, per patient's time window over the first month of ICU stay.

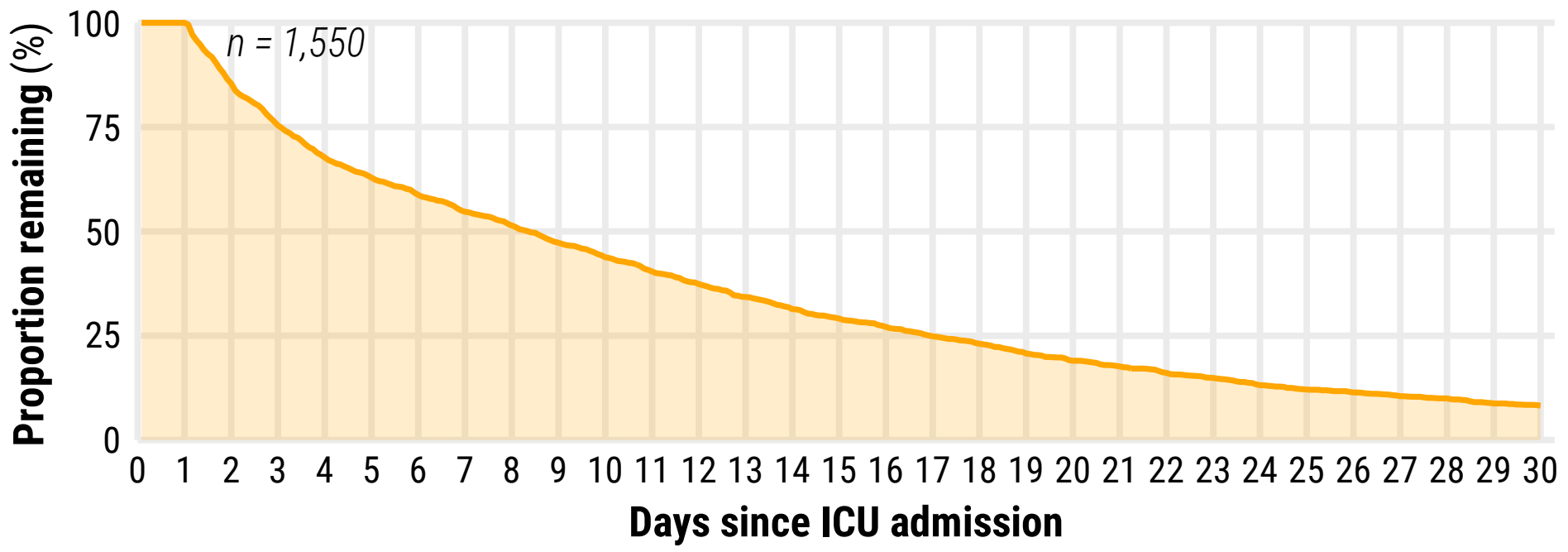

**Supplementary Figure 3. Proportion of study population remaining over the first month of ICU stay.**

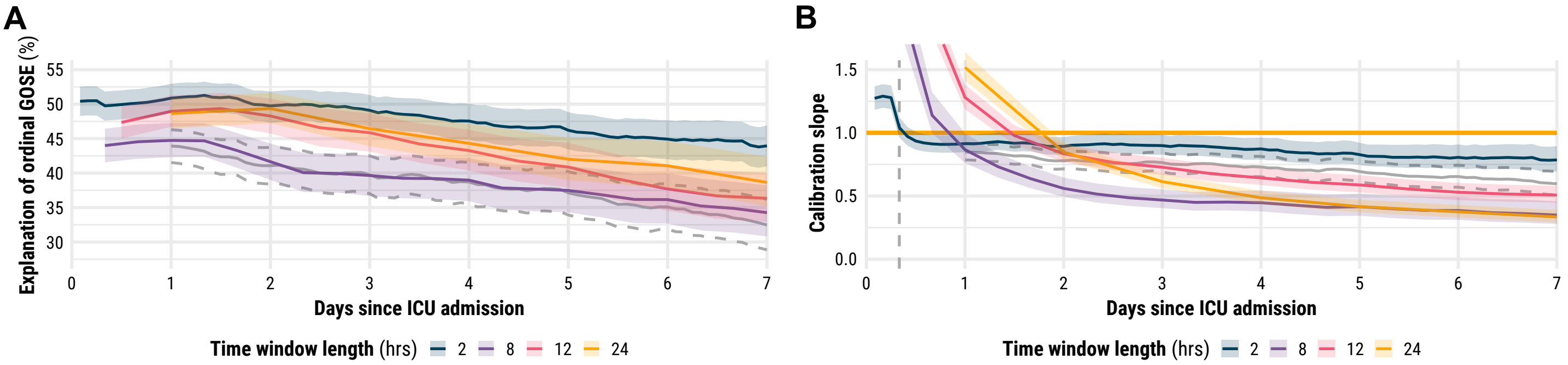

**Supplementary Figure 4. Effect of time-window length on explanation of ordinal six-month functional outcome.** The shaded regions surrounding curves are 95% confidence intervals derived using bias-corrected bootstrapping (1,000 resamples) to represent the variation across 20 repeated five-fold cross-validation partitions. (**A**) Model explanation of ordinal six-month functional outcome, measured by Somers' $D_{xy}$, in the first week of ICU stay. The grey line (with 95% confidence interval in dashed lines) is the explanation achieved by a static model trained on ten validated variables from the first 24 hours of ICU stay. (**B**) Model probability calibration slope, averaged across the six functional outcome thresholds, in the first week of ICU stay. The ideal calibration slope of one is marked with a horizontal orange line. The grey line (with 95% confidence interval in dashed lines) is the average calibration slope of a static model trained on ten validated predictors from the first 24 hours of ICU stay.

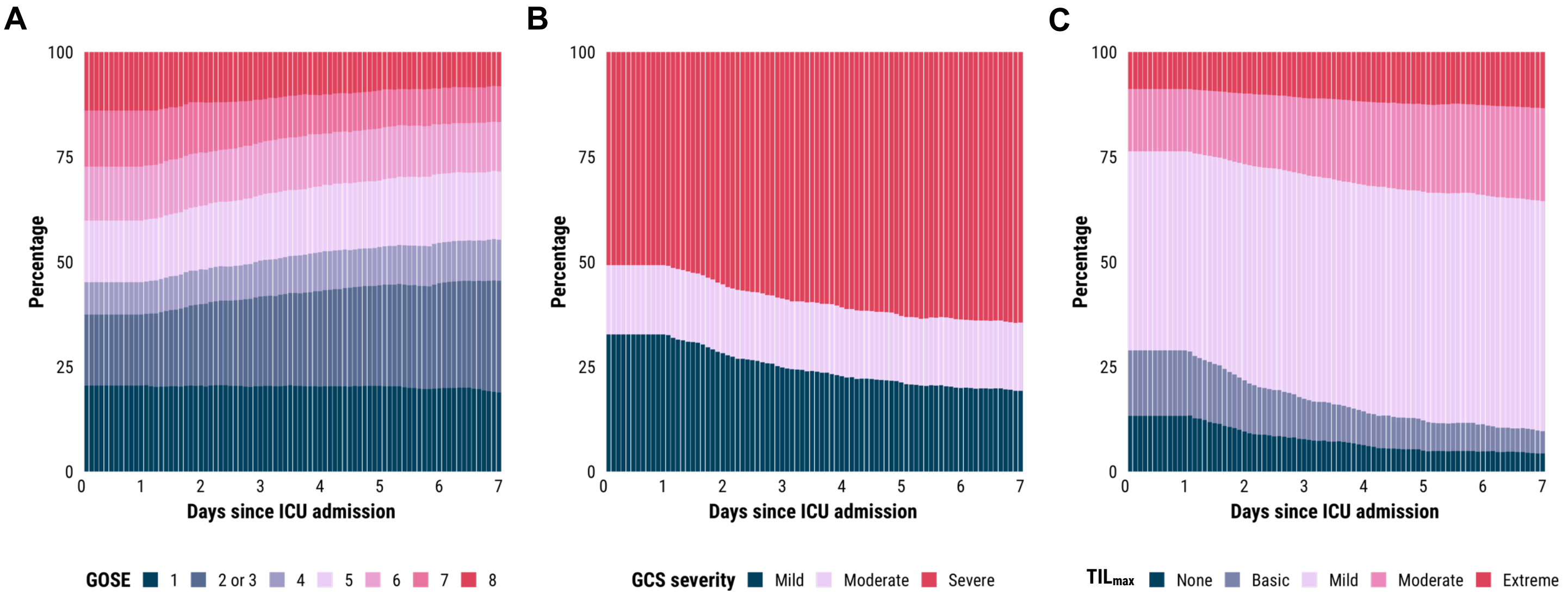

**Supplementary Figure 5. Characteristics of study population remaining over first week of ICU stay.** (**A**) The eight, ordinal scores of GOSE are decoded in the heading of Table 2. (**B**) GCS is categorised into severity as follows: Mild=13–15, Moderate=9–12, and Severe=3–8. (**C**) Maximum therapy intensity level over ICU stay ($TIL_{max}$) is categorised into severity as follows: None=0, Basic=1–2, Mild=3–11, Moderate=12–18, and Extreme=19–38.

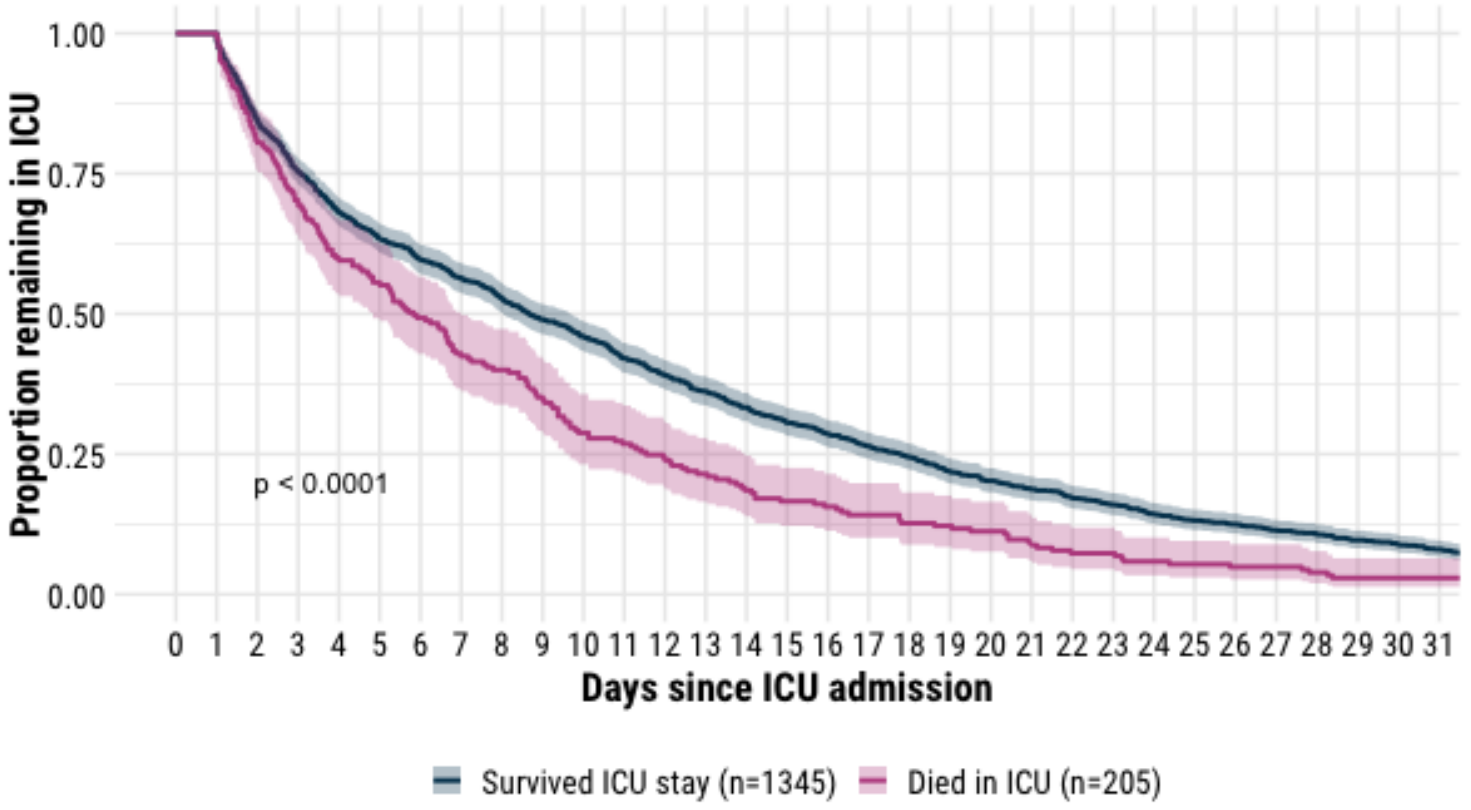

**Supplementary Figure 6. Proportion of study population, stratified by ICU survival, remaining over the first month of ICU stay.** The shaded regions surrounding curves are 95% confidence intervals and *p*-values are derived from a log-rank test.

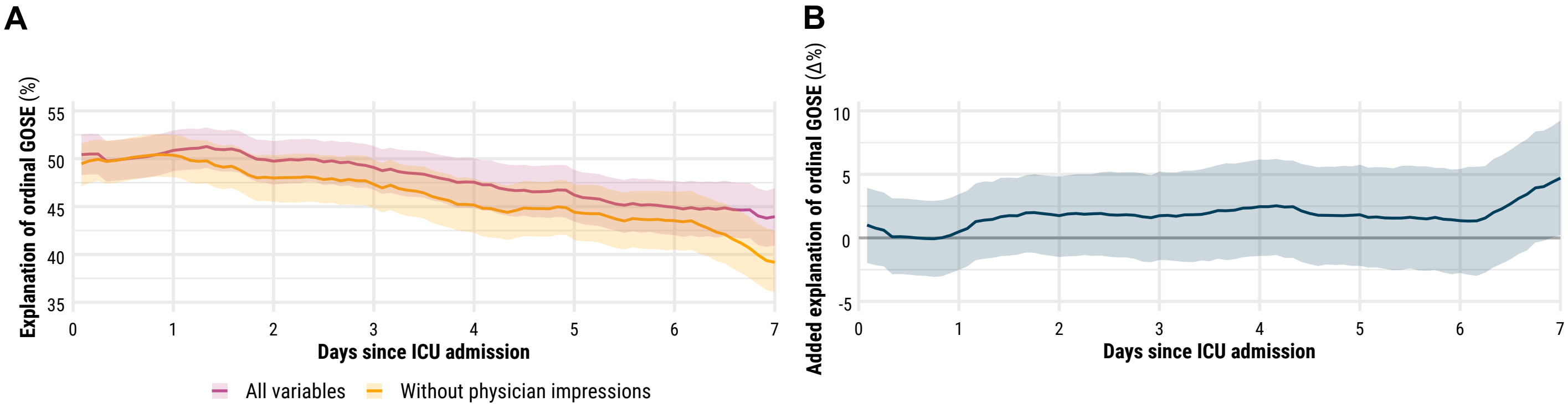

**Supplementary Figure 7. Effect of physician-based impressions on explanation of ordinal six-month functional outcome.** The shaded regions surrounding curves are 95% confidence intervals derived using bias-corrected bootstrapping (1,000 resamples) to represent the variation across 20 repeated five-fold cross-validation partitions. All physician-based impression variables are listed in Supplementary Note 2. (**A**) Model explanation of ordinal six-month functional outcome, measured by Somers' $D_{xy}$, in the first week of ICU stay. (**B**) Added model explanation of ordinal six-month functional outcome is measured by difference in Somers' $D_{xy}$ in the first week of ICU stay.

**Supplementary Figure 8. Population-level TimeSHAP values across levels of six-month functional recovery.** Legend provided at end of figure (pp 8).

A

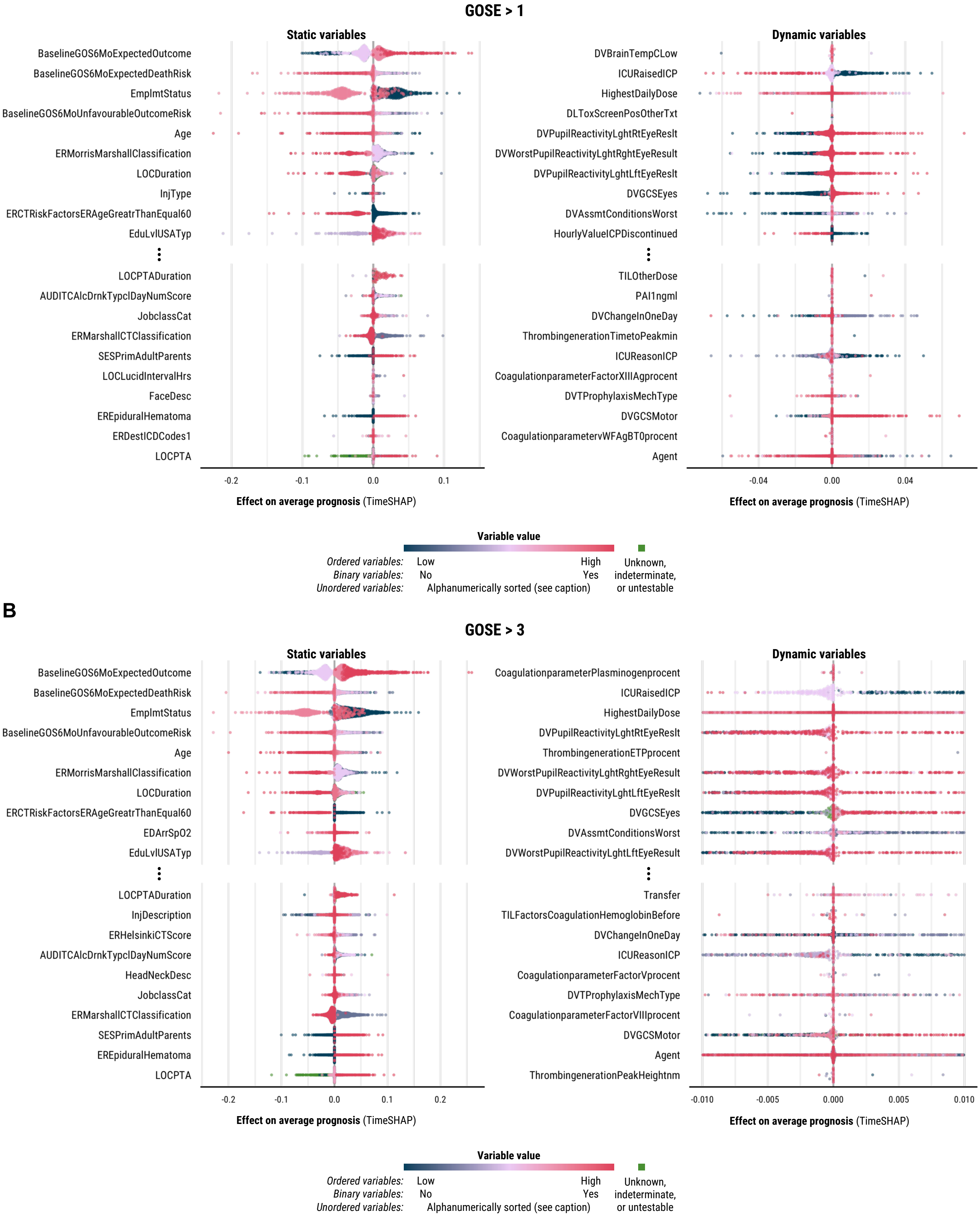

**Supplementary Figure 8** (*continued*).

**C**

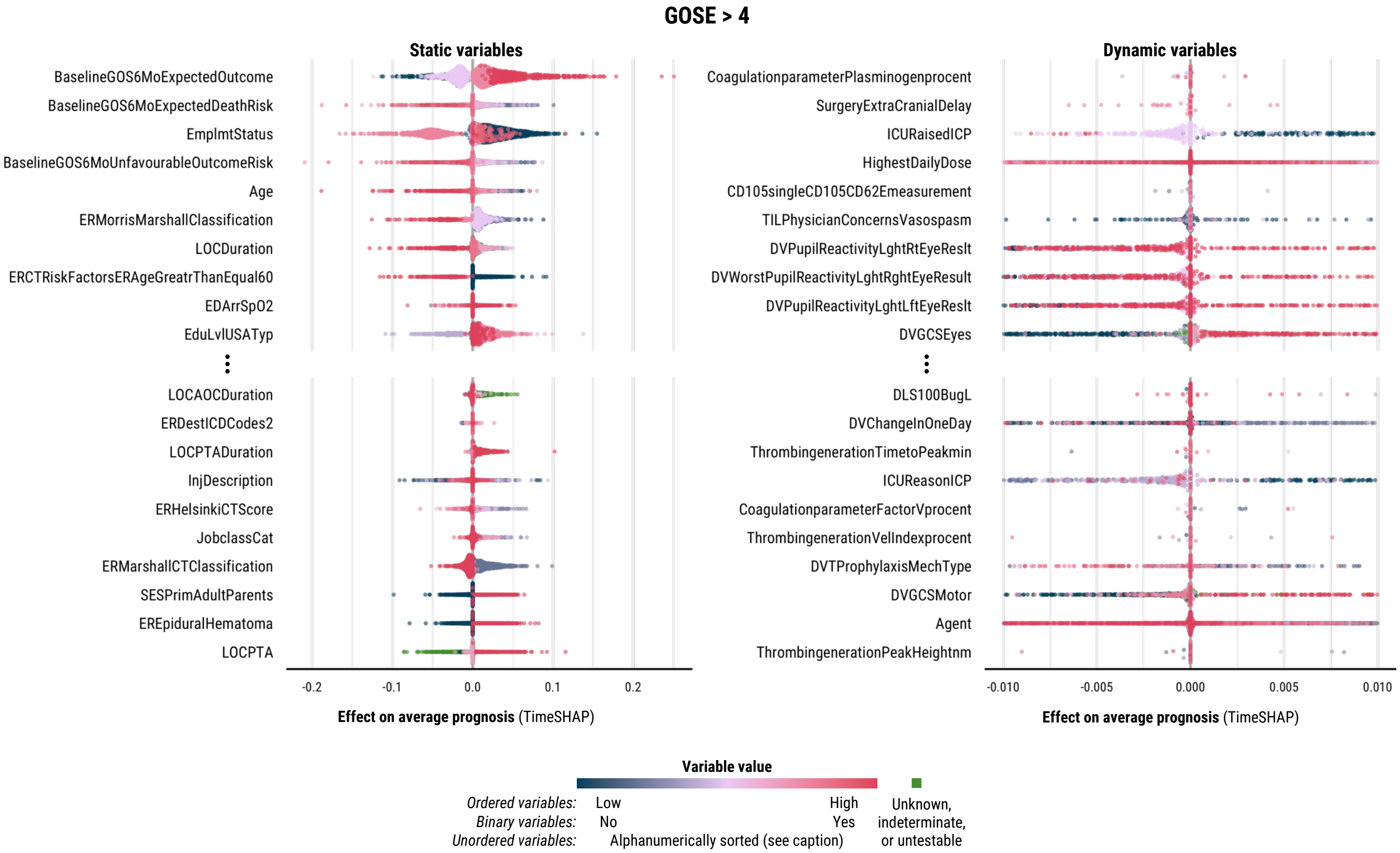

**D**

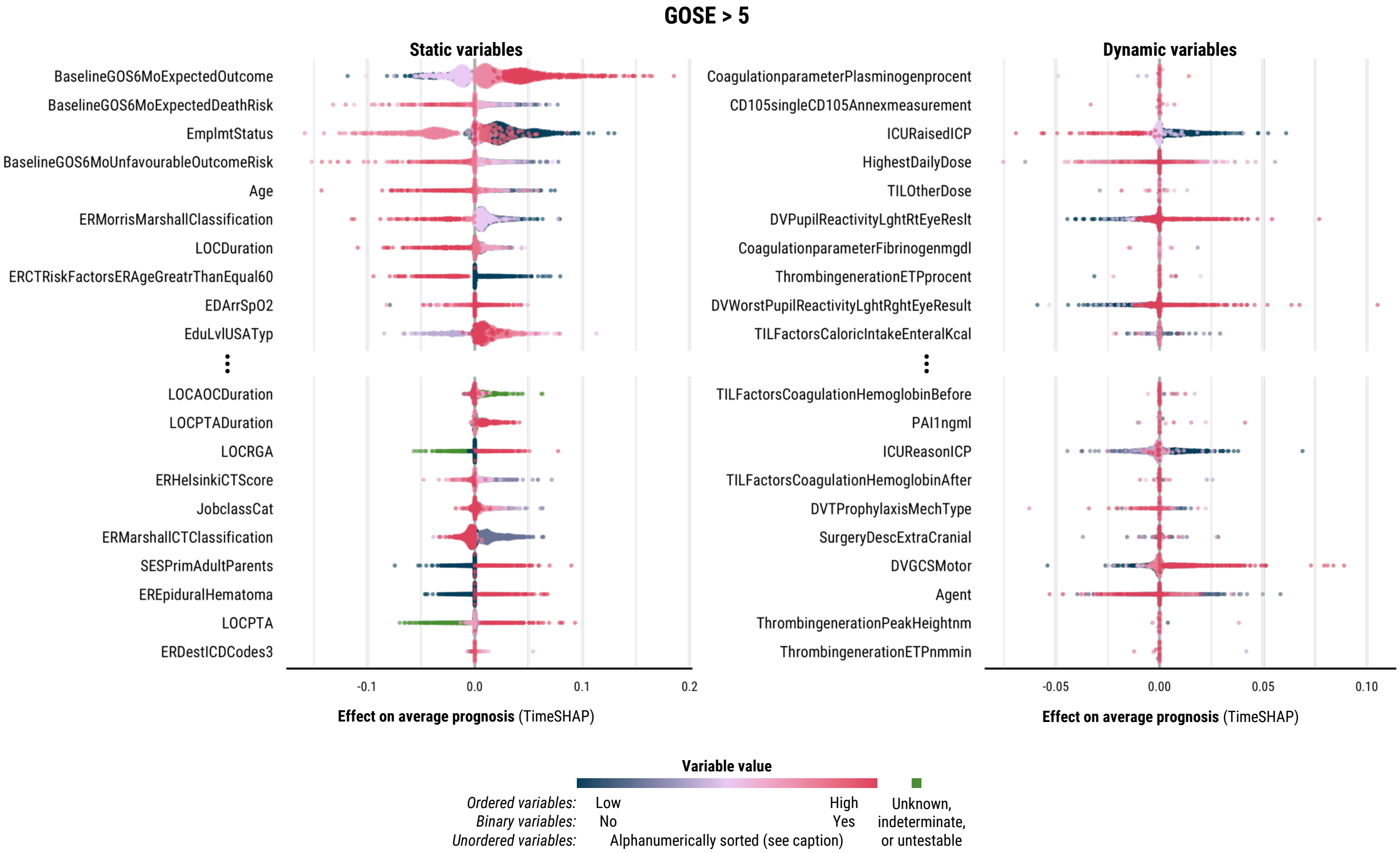

**Supplementary Figure 8** (*continued*).

**E**

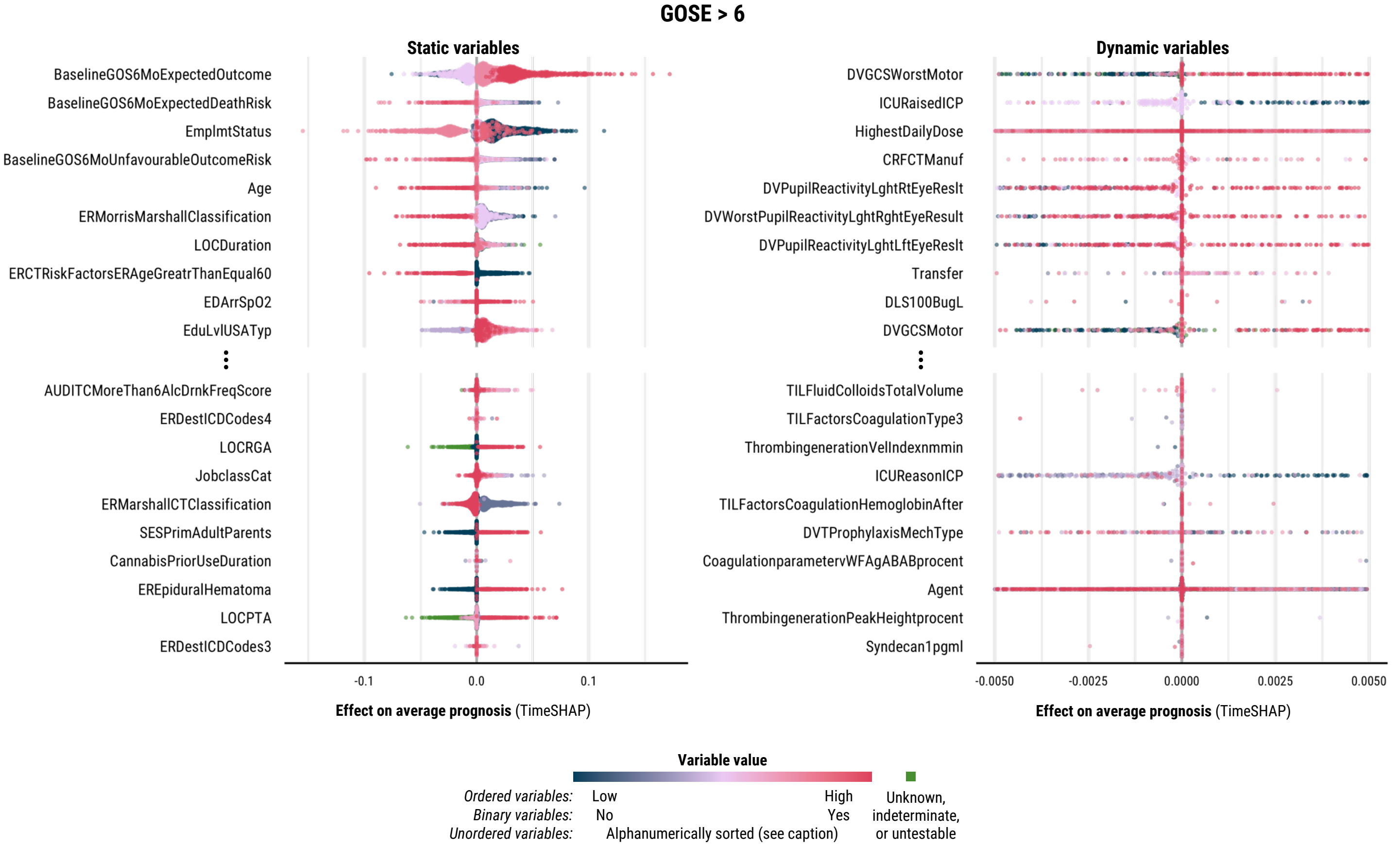

**F**

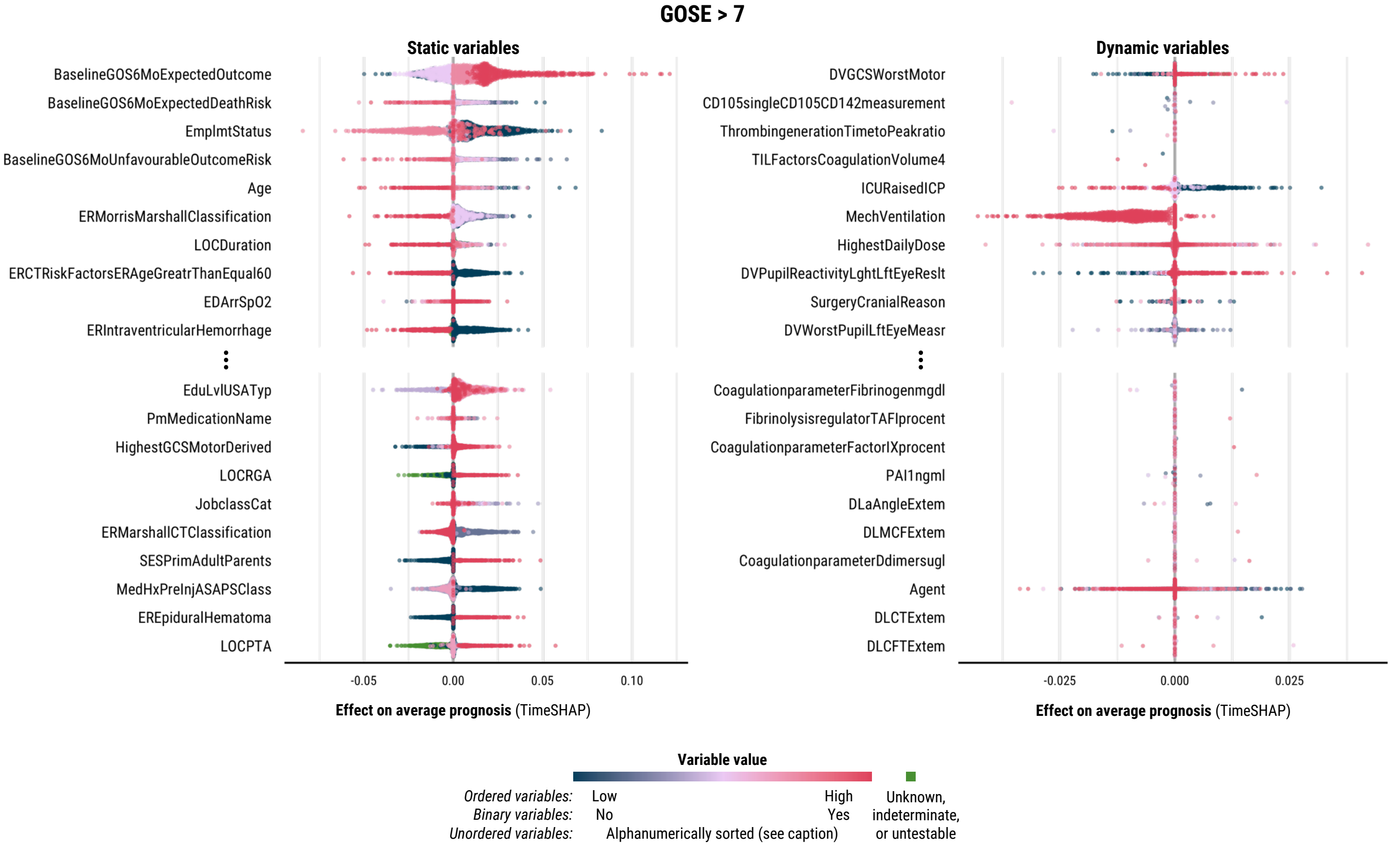

**Supplementary Figure 8. Population-level TimeSHAP values across levels of six-month functional recovery.** All abbreviated variable names are decoded in Supplementary Note 1. TimeSHAP values are interpreted as contributions of variables or time windows towards the difference in a patient's probability at each threshold of six-month GOSE from that of the average patient. The variables were selected – for each threshold of GOSE – by first identifying the ten variables with non-missing value tokens with the most negative median TimeSHAP values across the population (above the ellipses) and then, among the remaining variables, selecting the ten with non-missing value tokens with the most positive median TimeSHAP values (below the ellipses). Each point represents the mean TimeSHAP value for a token across an individual patient's high-magnitude transitions. The colour of the point represents the relative ordered value of a token within a variable, and for unordered variables (e.g., employment status before injury), tokens were sorted alphanumerically (the sort index per possible unordered variable token is provided in Supplementary Note 1). Green points represent variable tokens that are not missing but explicitly encode an unknown value. The TimeSHAP values are provided for each threshold of GOSE: (**A**) GOSE>1, i.e., survival, (**B**) GOSE>3, i.e., regaining consciousness, (**C**) GOSE>4, i.e., regaining independence, (**D**) GOSE>5, i.e., regaining ability to participate in one or more life roles, (**E**) GOSE>6, i.e., symptomatic return to normal life, and (**F**) GOSE>7, i.e., full return to normal life.

**Supplementary Figure 9. Population-level TimeSHAP values for each category of variables.** Legend provided at end of figure (pp 13).

A

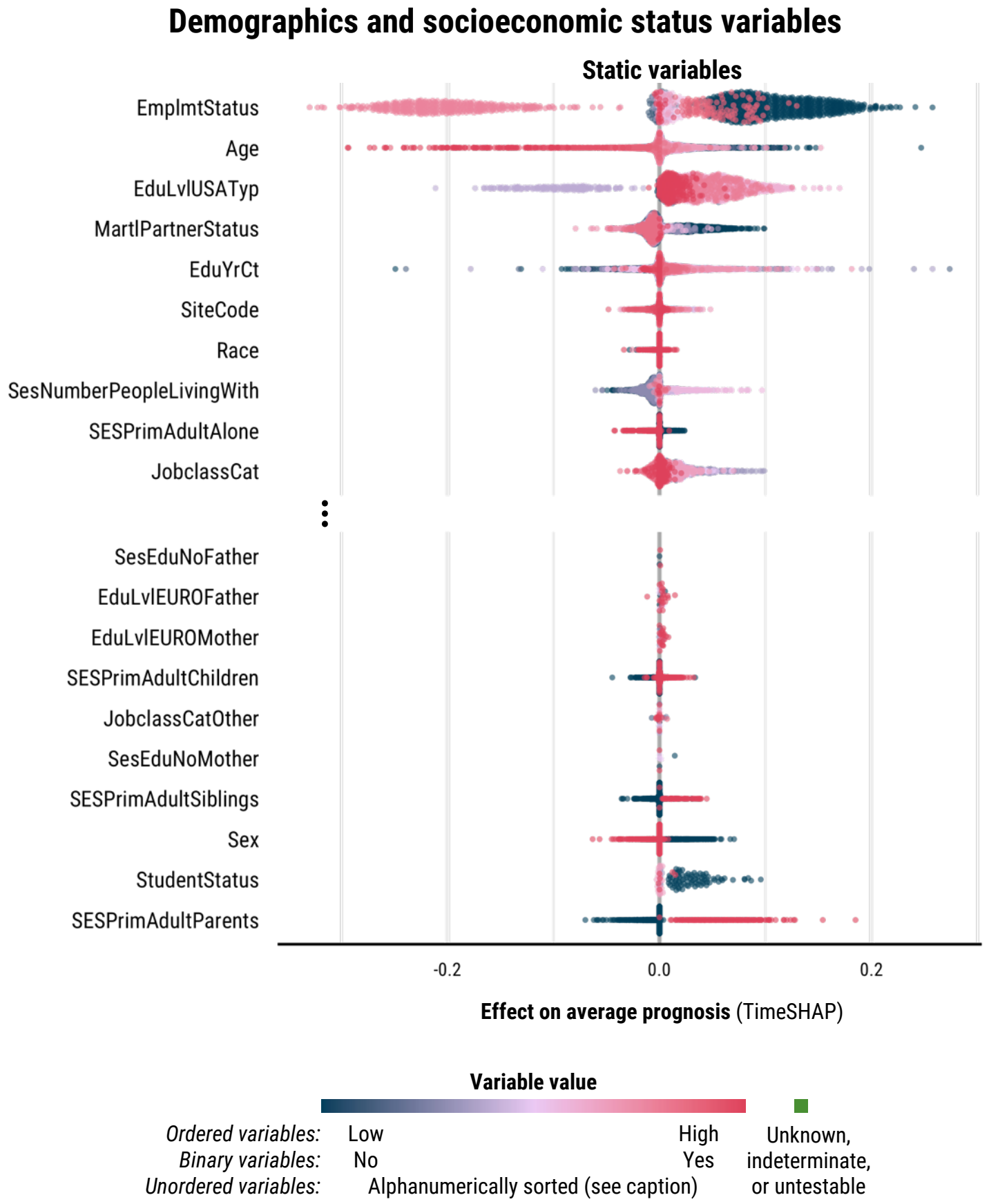

B

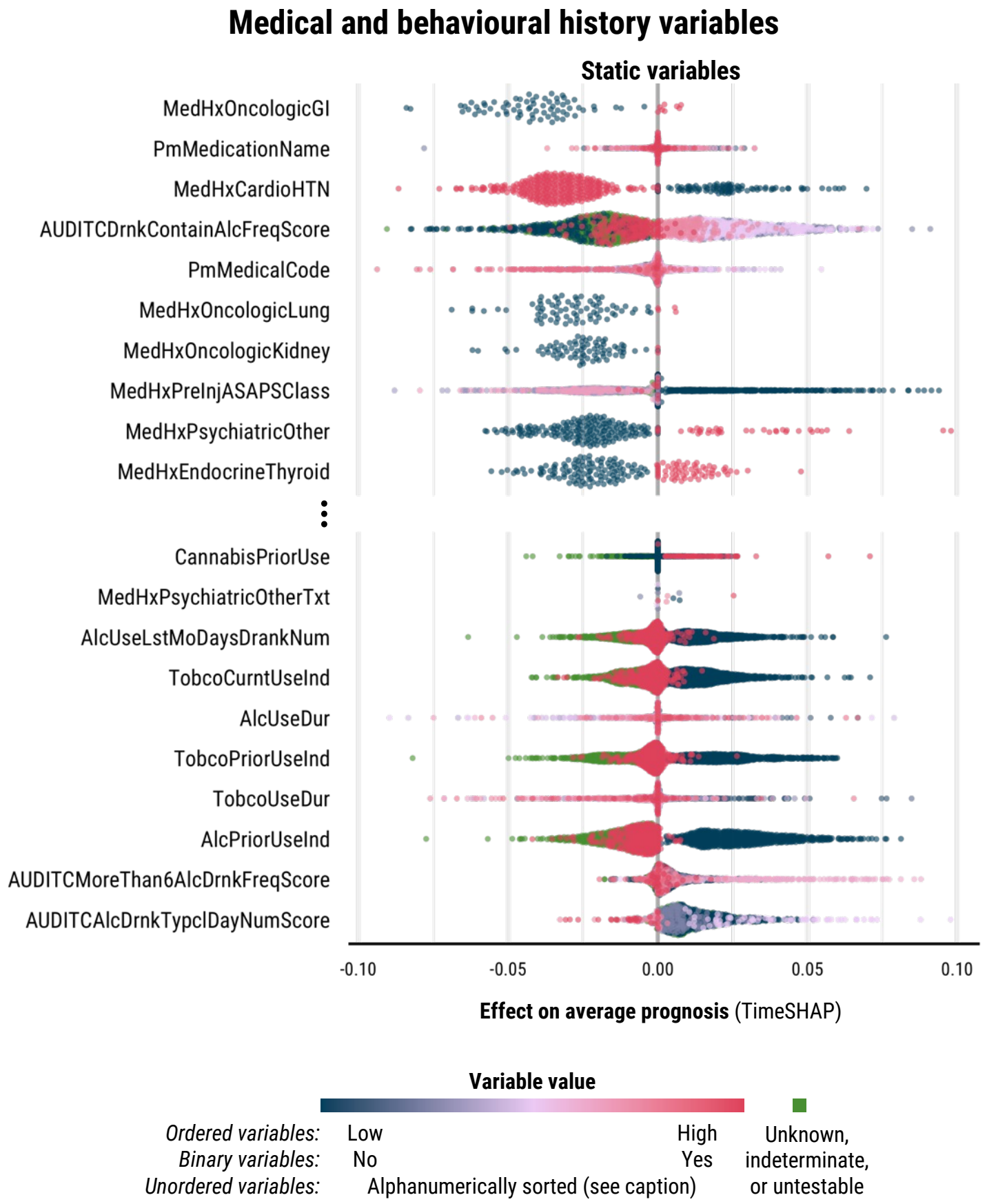

**Supplementary Figure 9** (*continued*).

**C**

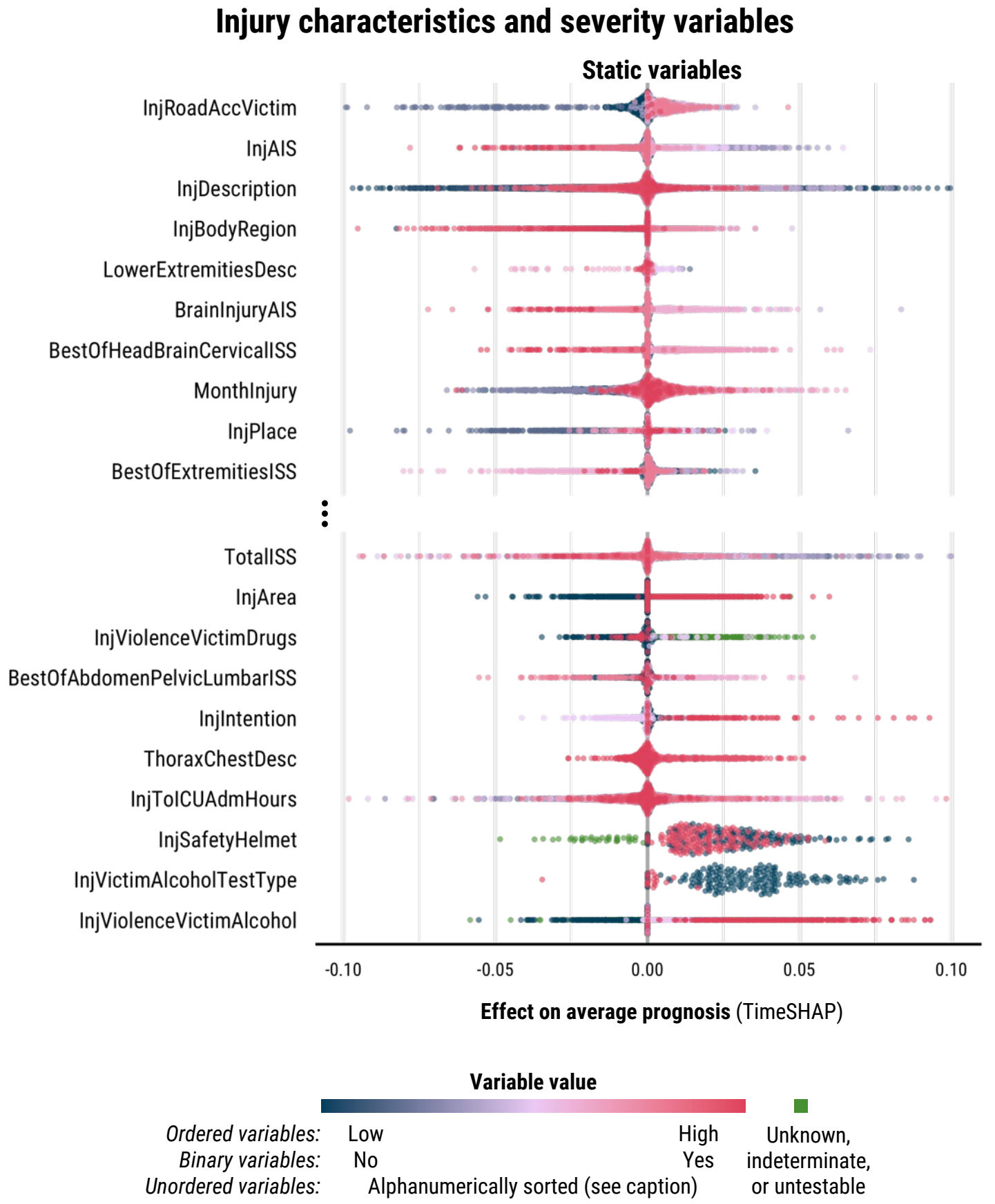

**D**

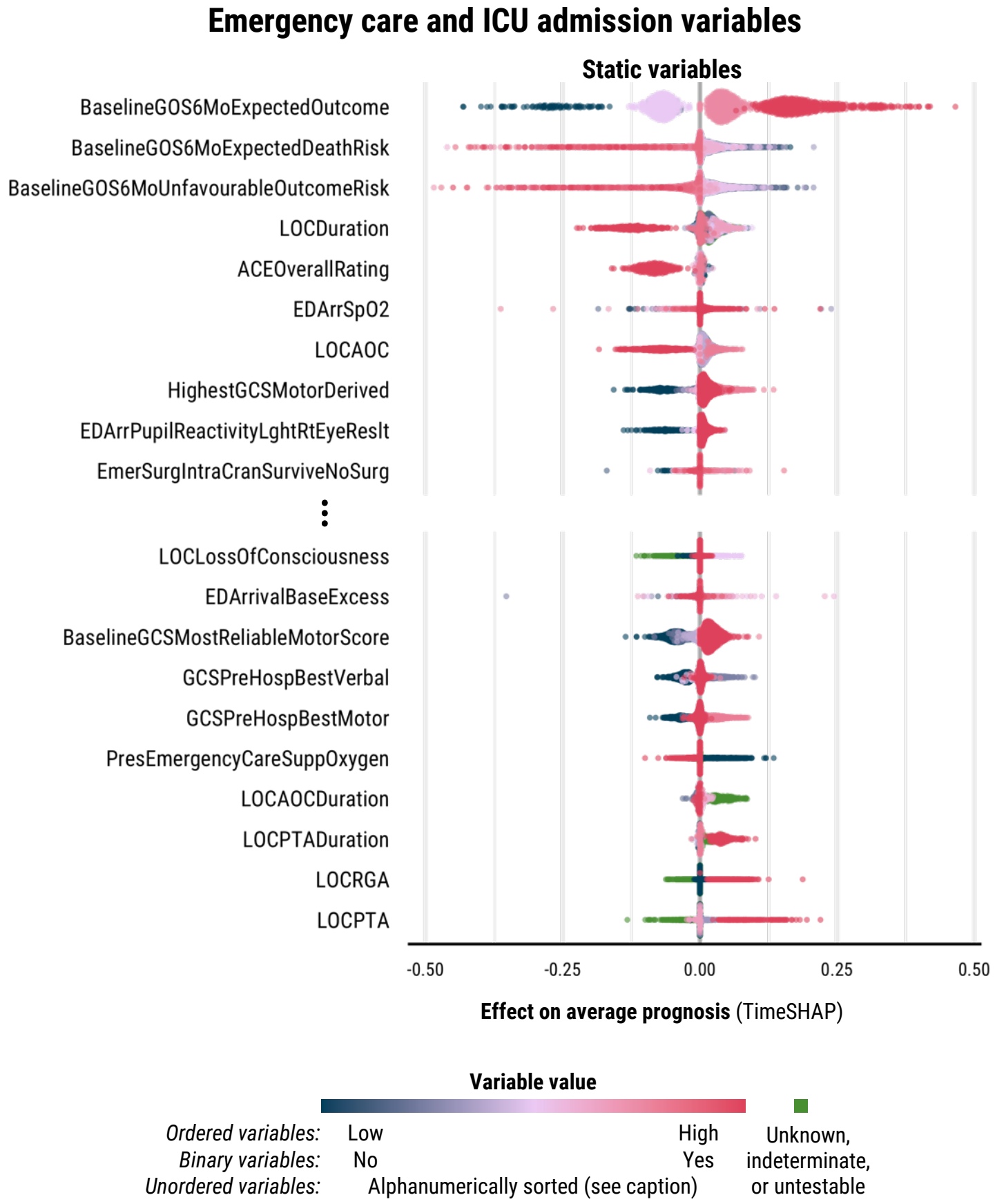

**Supplementary Figure 9** (*continued*).

E

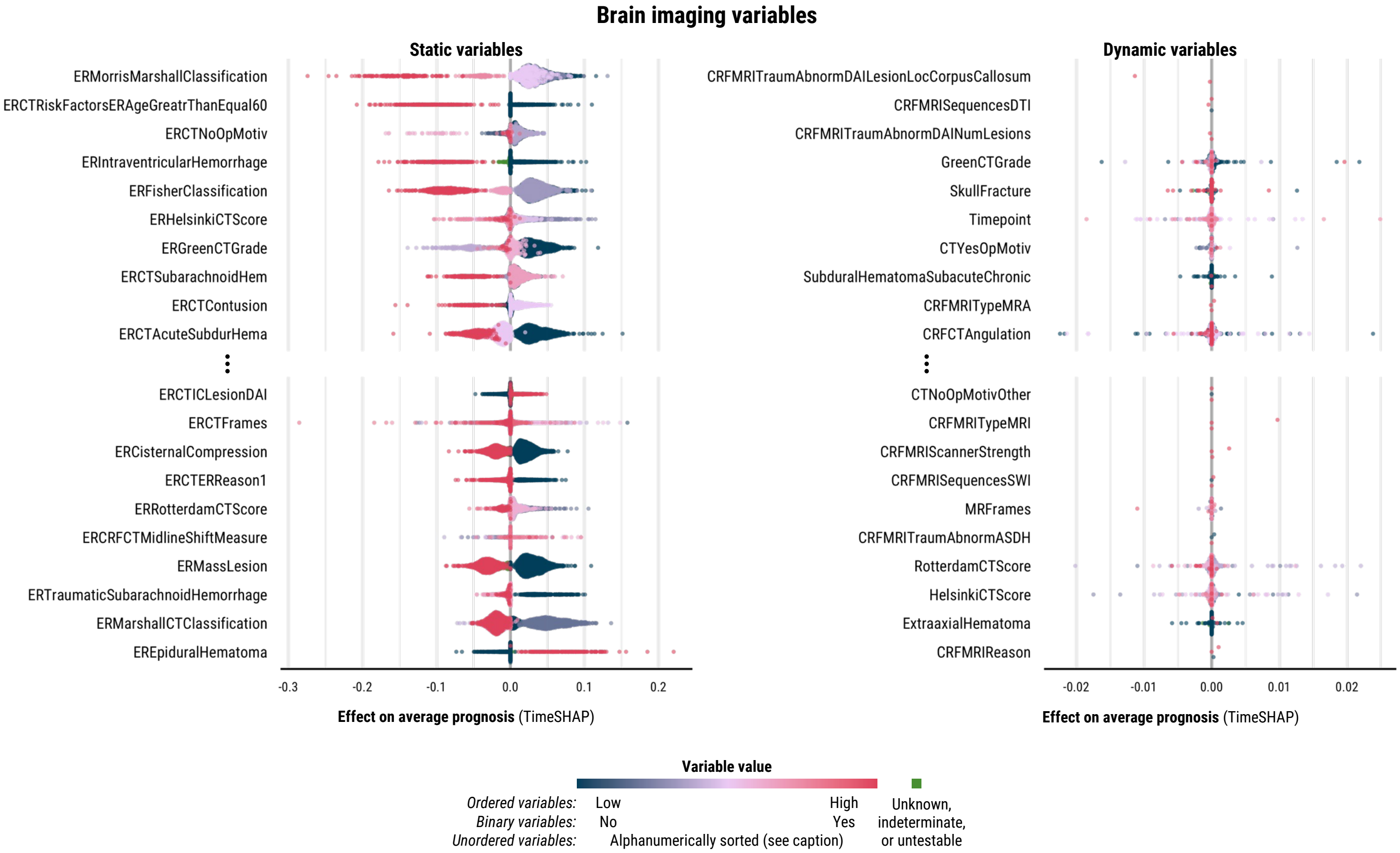

F

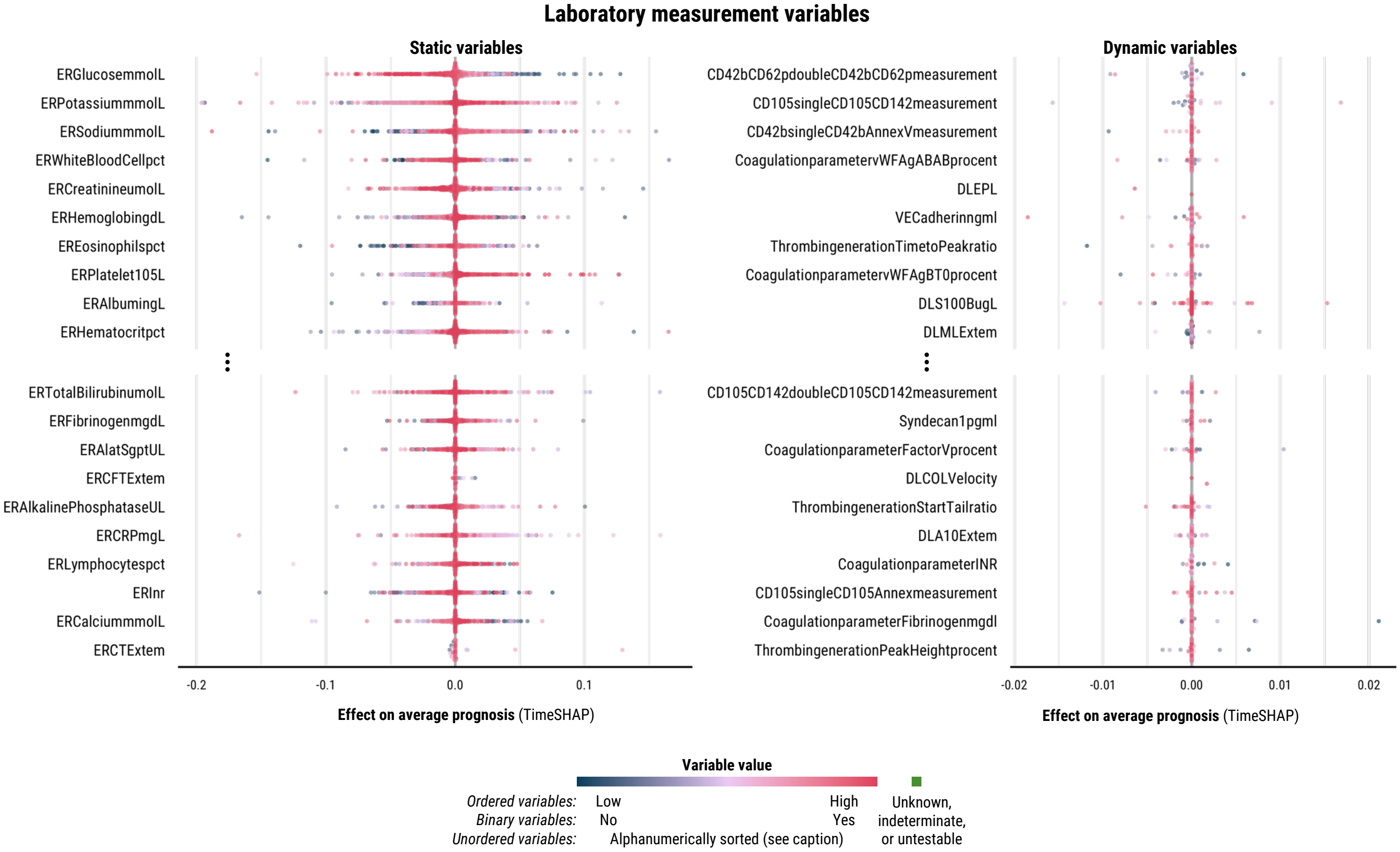

**Supplementary Figure 9** (*continued*).

G

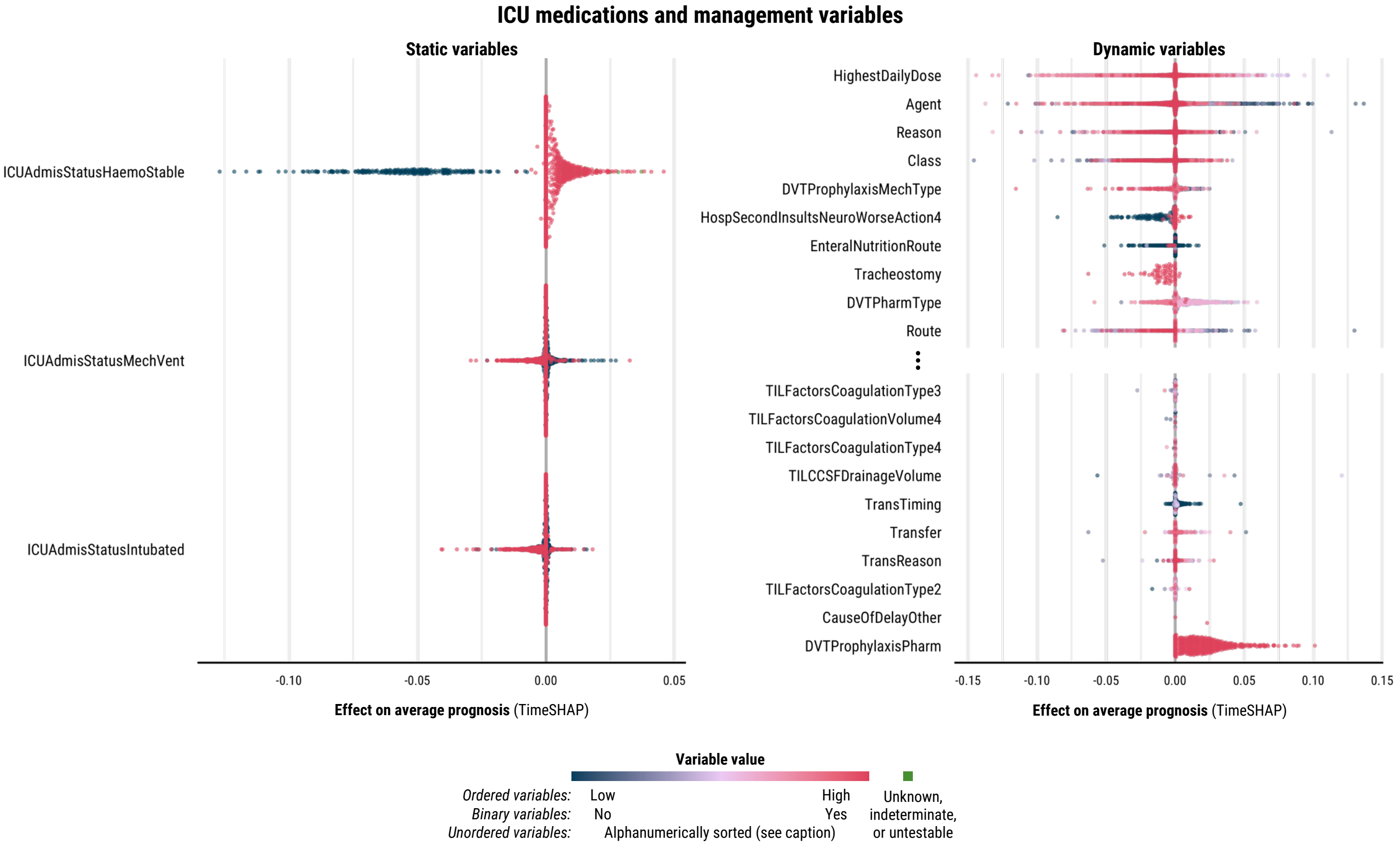

H

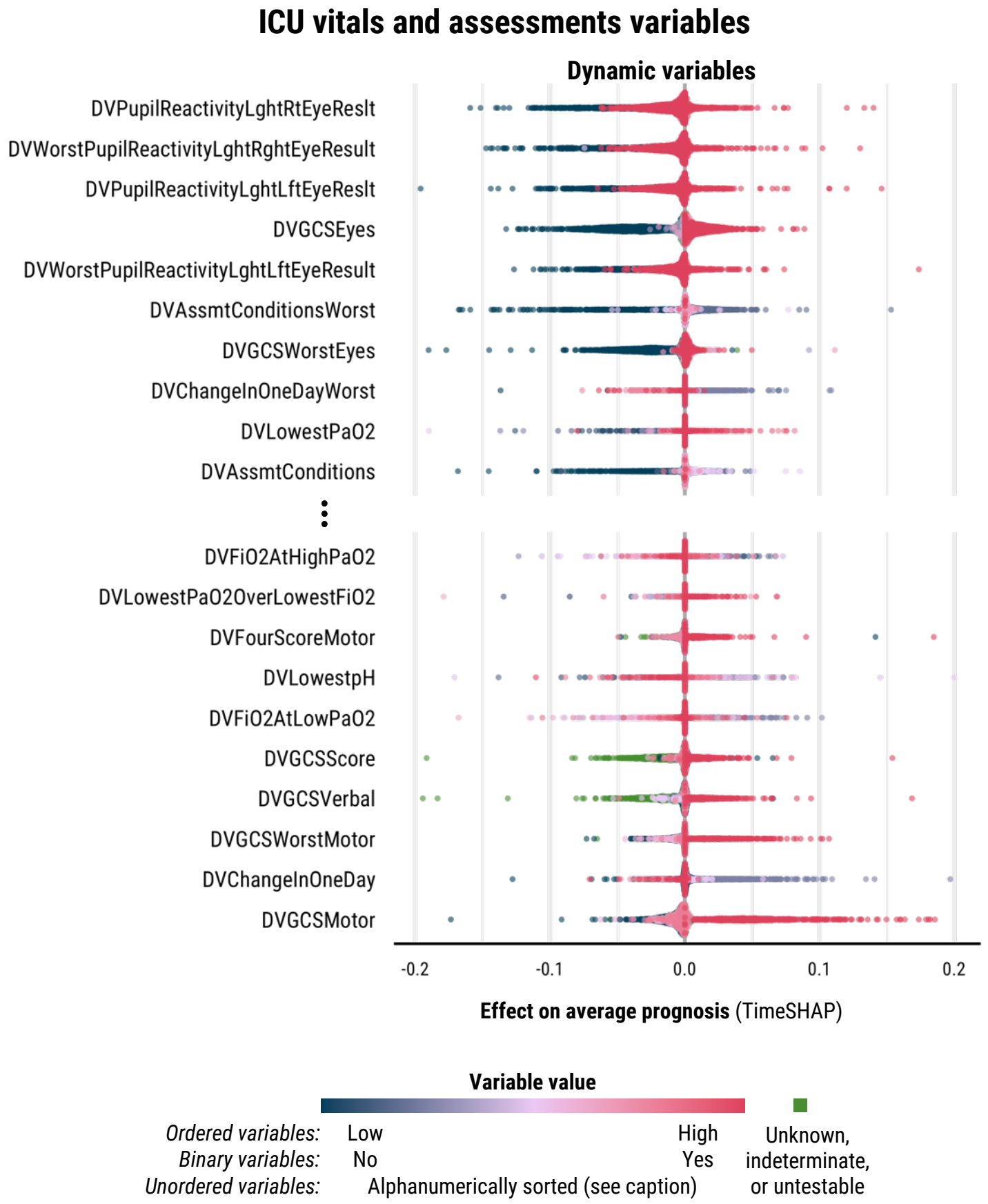

**Supplementary Figure 9** (*continued*).

I

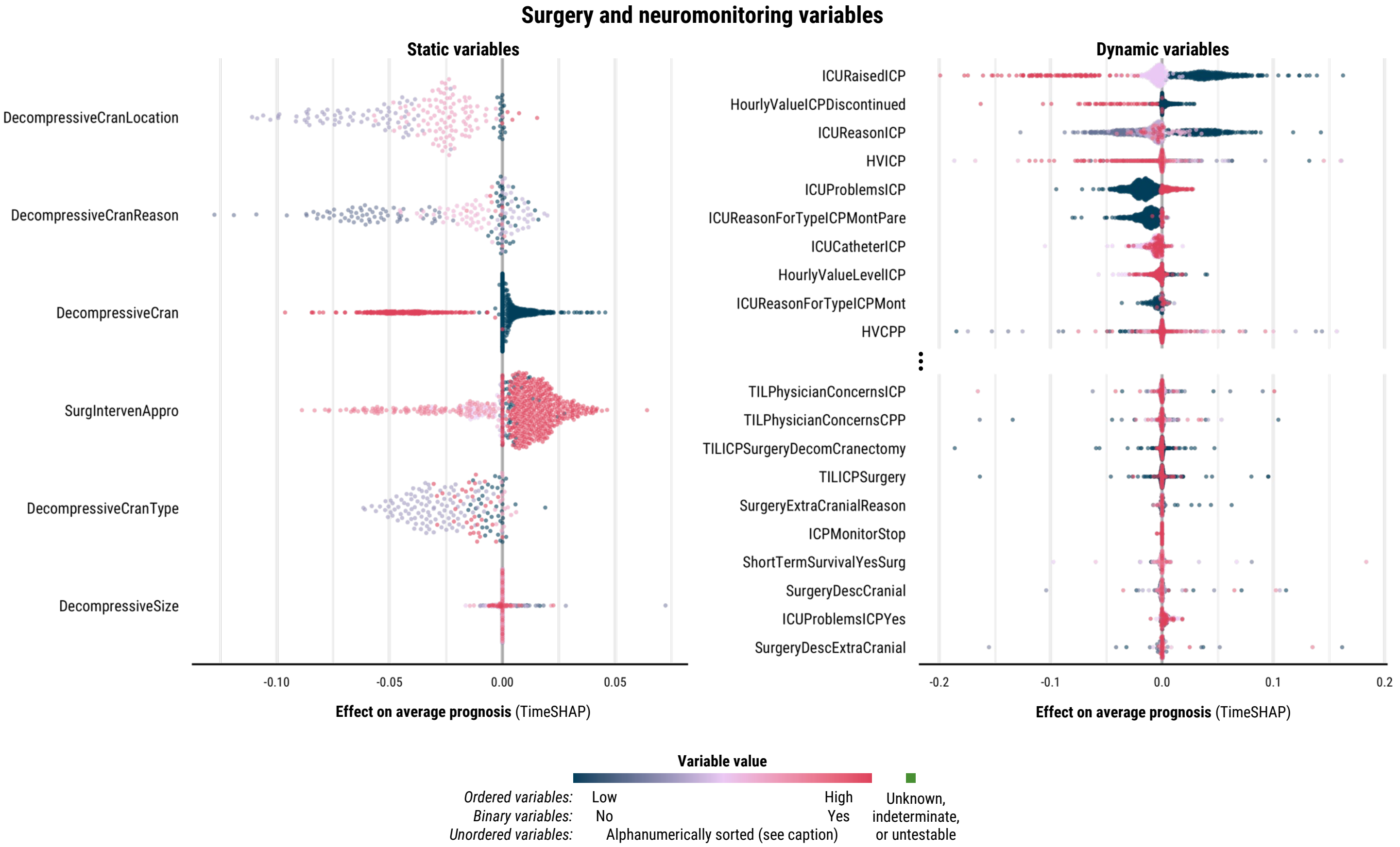

**Supplementary Figure 9. Population-level TimeSHAP values for each category of variables.** All abbreviated variable names are decoded in Supplementary Note 1. TimeSHAP values are interpreted as contributions of variables towards the difference in a patient's expected six-month functional outcome output from that of the average patient (Supplementary Figure 13). The variables were selected – for each category of variables – by first identifying the ten variables with non-missing value tokens with the most negative median TimeSHAP values across the population (above the ellipses) and then, among the remaining variables, selecting the ten with non-missing value tokens with the most positive median TimeSHAP values (below the ellipses). Each point represents the mean TimeSHAP value for a token across an individual patient's high-magnitude transitions. The colour of the point represents the relative ordered value of a token within a variable, and for unordered variables (e.g., employment status before injury), tokens were sorted alphanumerically (the sort index per possible unordered variable token is provided in Supplementary Note 1). Green points represent variable tokens that are not missing but explicitly encode an unknown value.

**Supplementary Figure 10. TimeSHAP values stratified by WLST and inclusion of physician impressions.** Legend provided at end of figure (pp 16).

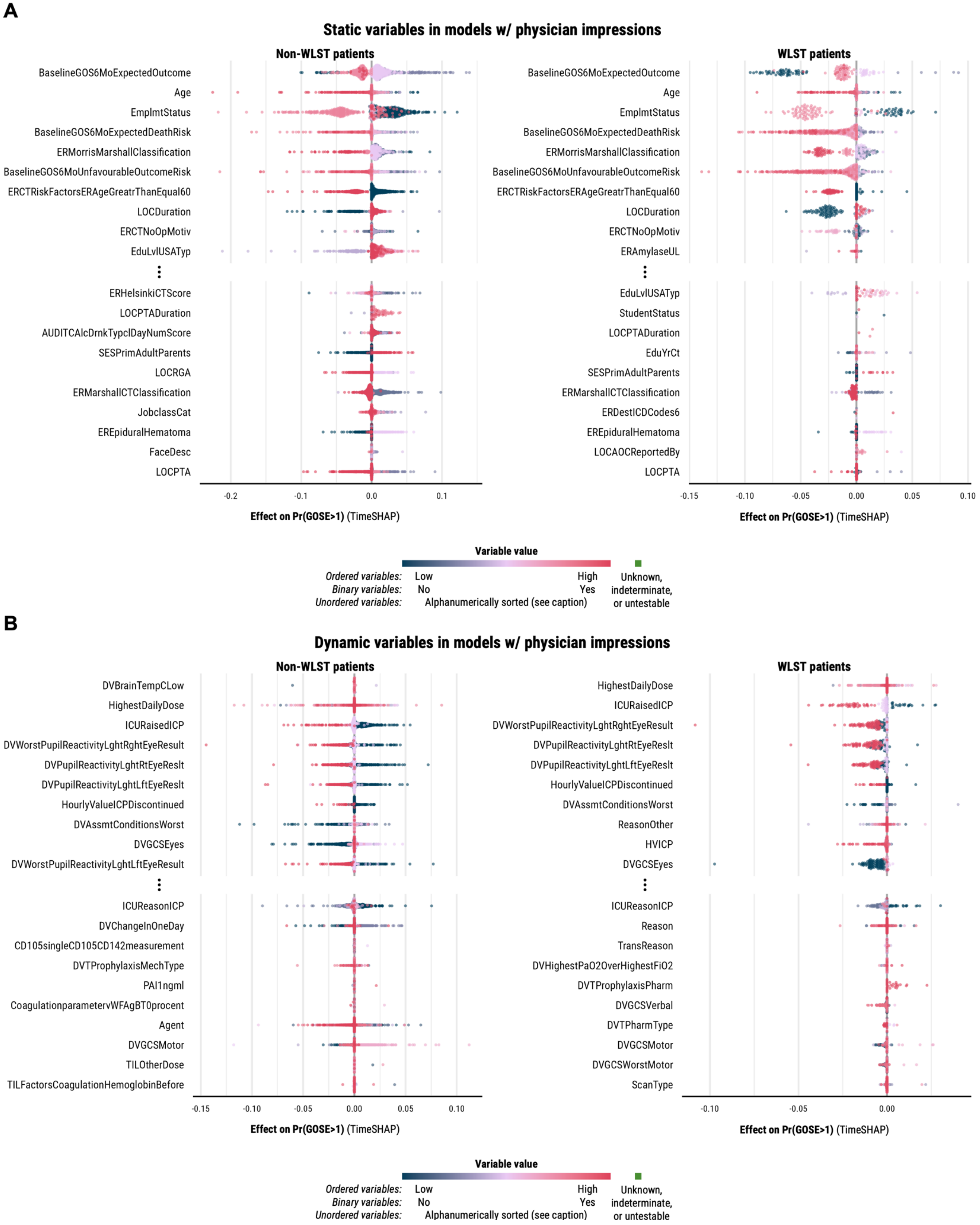

**Supplementary Figure 10** (*continued*).

**C**

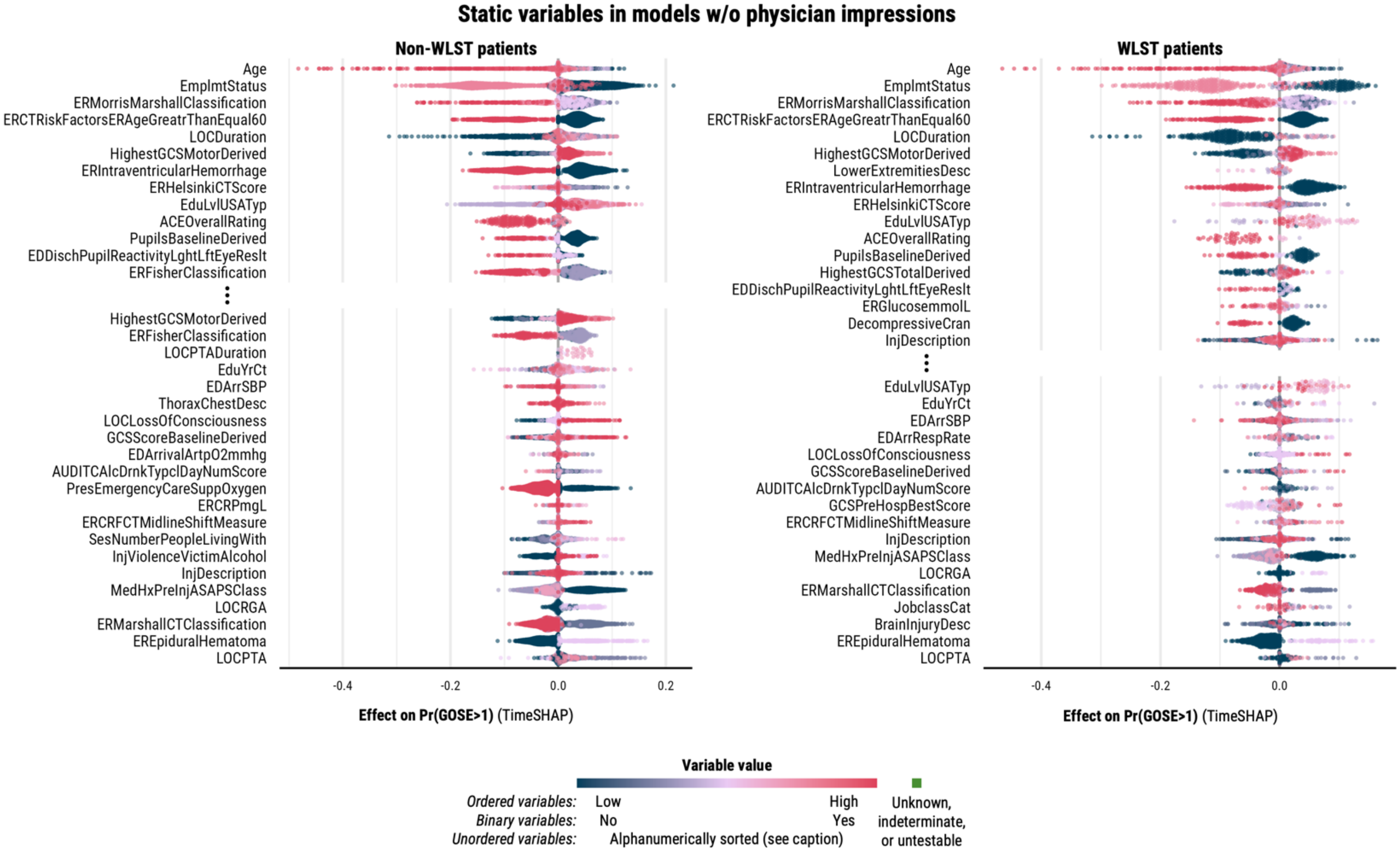

**D**

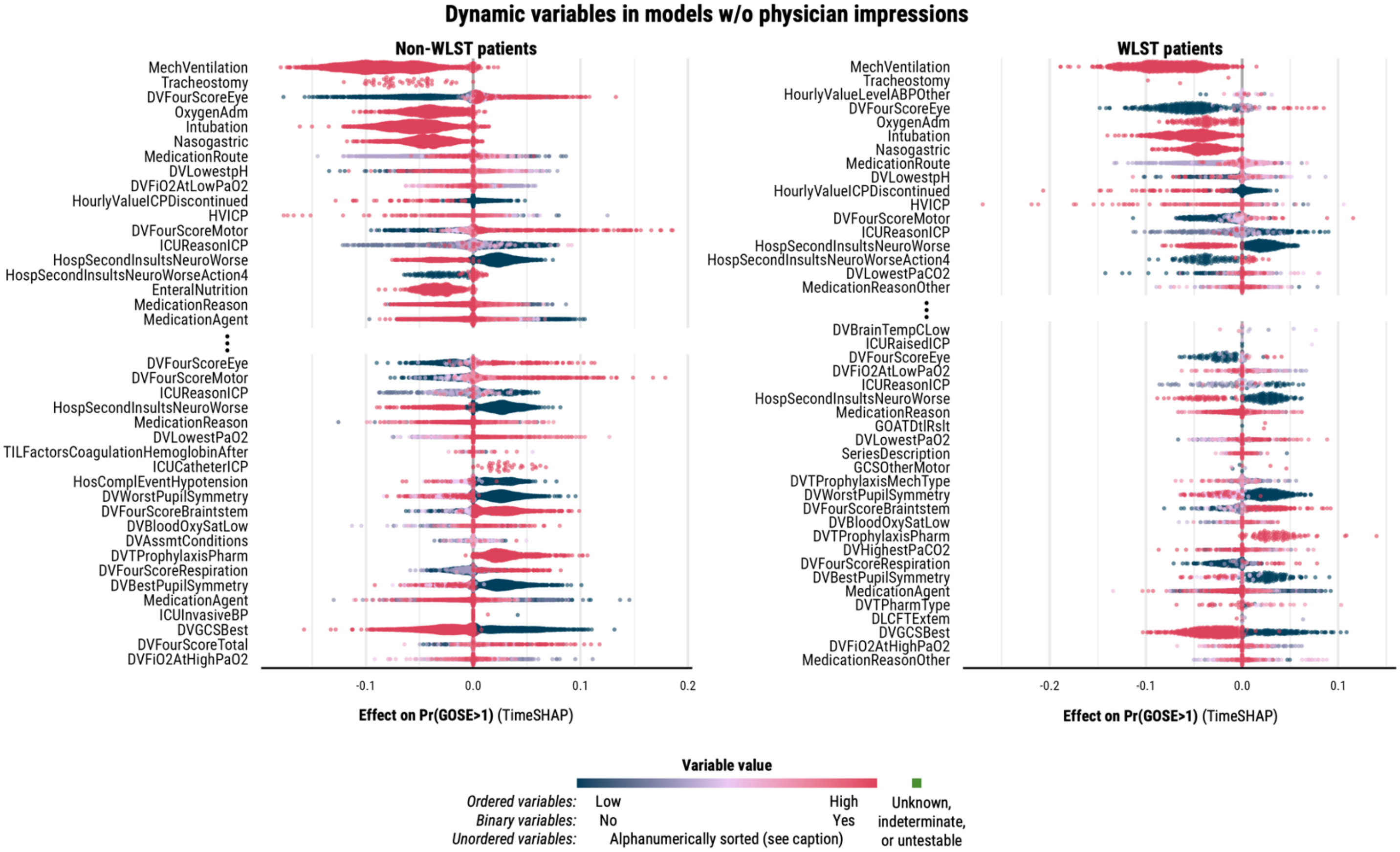

**Supplementary Figure 10. TimeSHAP values stratified by WLST and inclusion of physician impressions.** All abbreviated variable names are decoded in Supplementary Note 1. TimeSHAP values are interpreted as contributions of variables towards the difference in a patient's probability of survival (i.e., Pr(GOSE>1)) from that of the average patient (Supplementary Figure 13). The variables were selected – for each category of variables – by first identifying the ten variables with non-missing value tokens with the most negative median TimeSHAP values across the population (above the ellipses) and then, among the remaining variables, selecting the ten with non-missing value tokens with the most positive median TimeSHAP values (below the ellipses). Each point represents the mean TimeSHAP value for a token across an individual patient's high-magnitude transitions. The colour of the point represents the relative ordered value of a token within a variable, and for unordered variables (e.g., employment status before injury), tokens were sorted alphanumerically (the sort index per possible unordered variable token is provided in Supplementary Note 1). Green points represent variable tokens that are not missing but explicitly encode an unknown value.

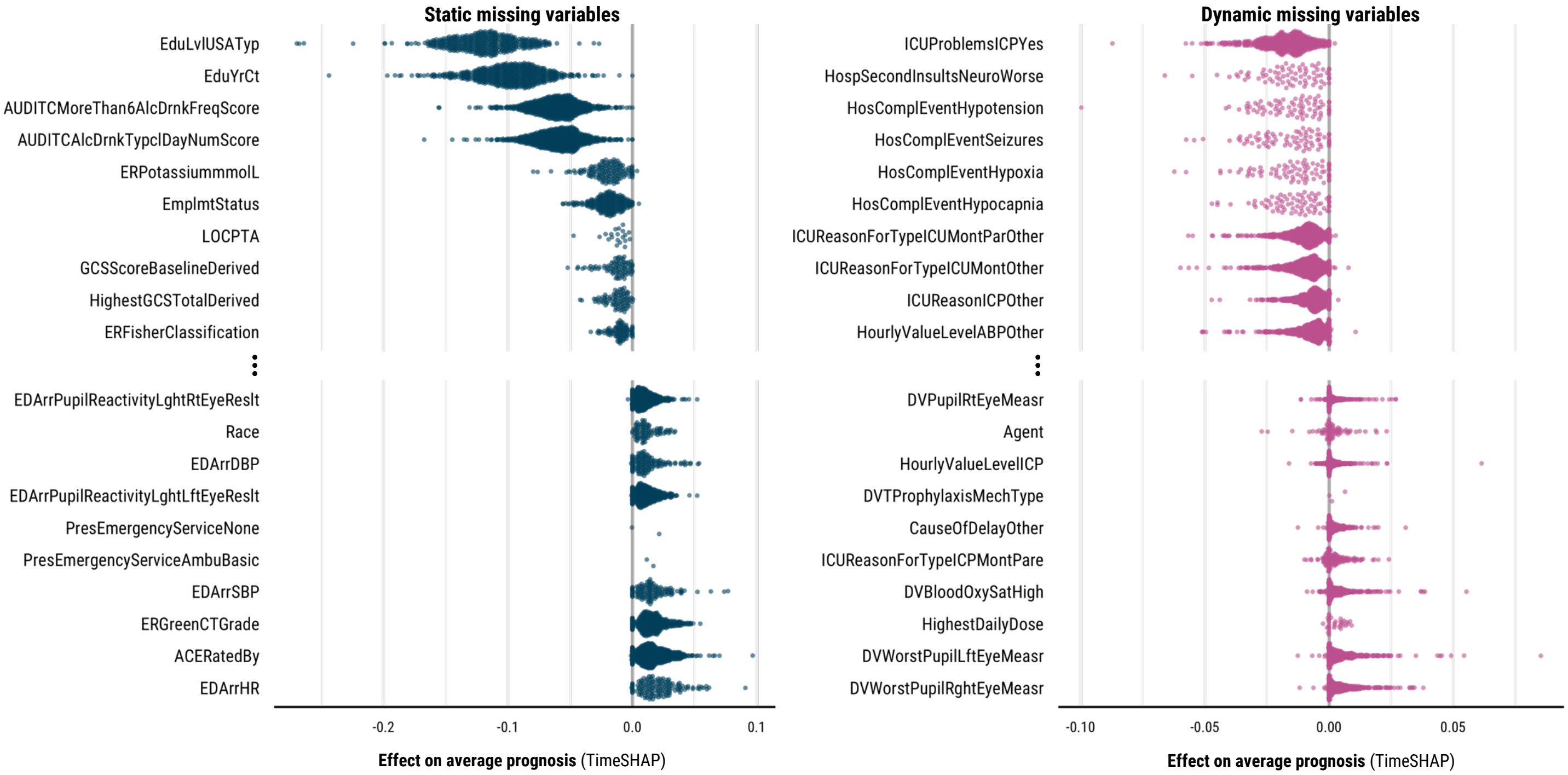

**Supplementary Figure 11. Population-level TimeSHAP values for missing value tokens.** All abbreviated variable names are decoded in Supplementary Note 1. TimeSHAP values are interpreted as contributions of a variable's missingness towards the difference in a patient's expected six-month functional outcome output from that of the average patient (Supplementary Figure 13). The variables were selected – for each category of variables – by first identifying the ten variables with missing value tokens with the most negative median TimeSHAP values across the population (above the ellipses) and then, among the remaining variables, selecting the ten with missing value tokens with the most positive median TimeSHAP values (below the ellipses). Each point represents the mean TimeSHAP value for a missing value token across an individual patient's high-magnitude transitions.

**Supplementary Figure 12. Example of individual ICU disease course for each level of six-month functional outcome.** Legend provided at end of figure (pp 19).

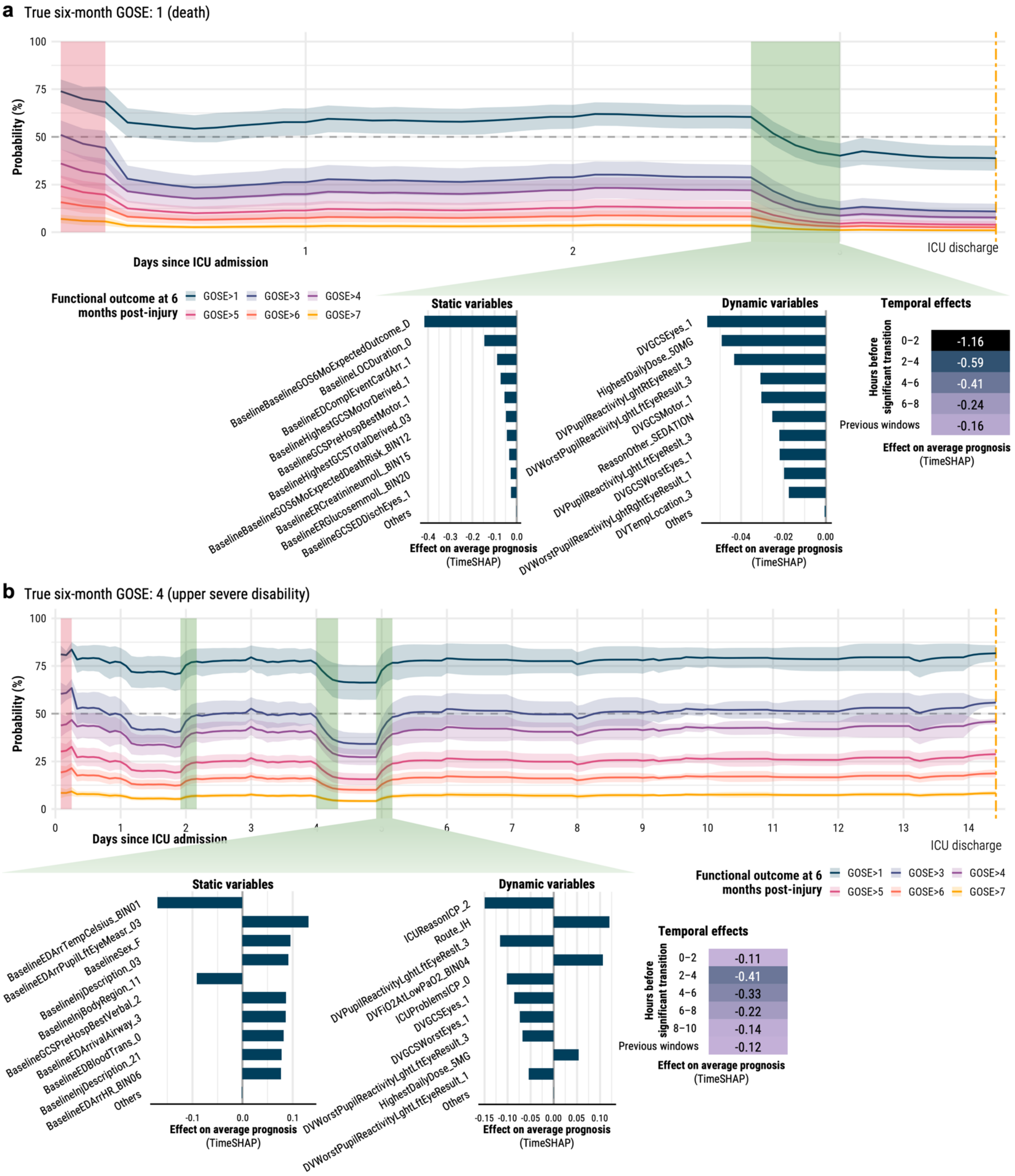

**Supplementary Figure 12** (*continued*).

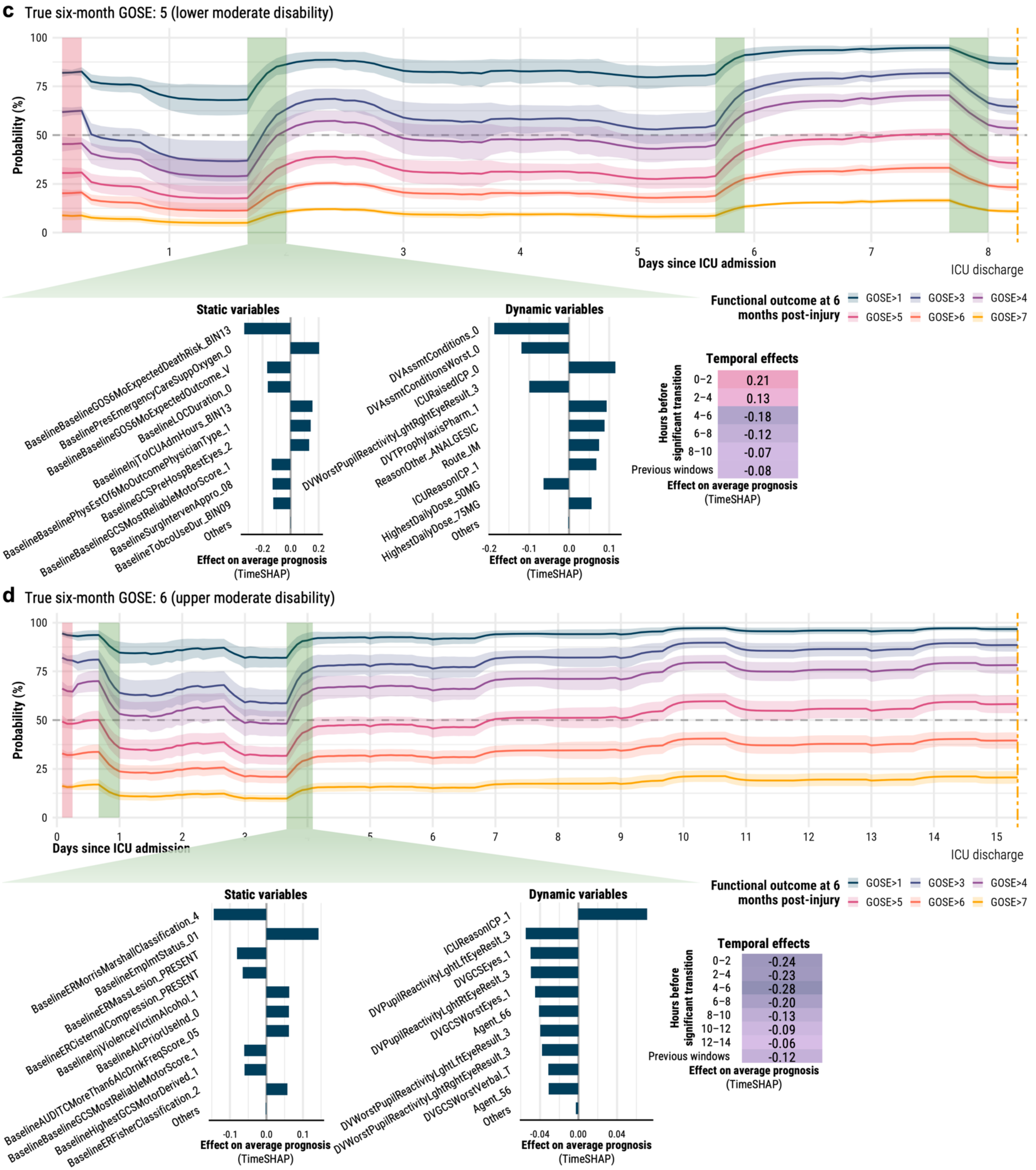

**Supplementary Figure 12** (*continued*).

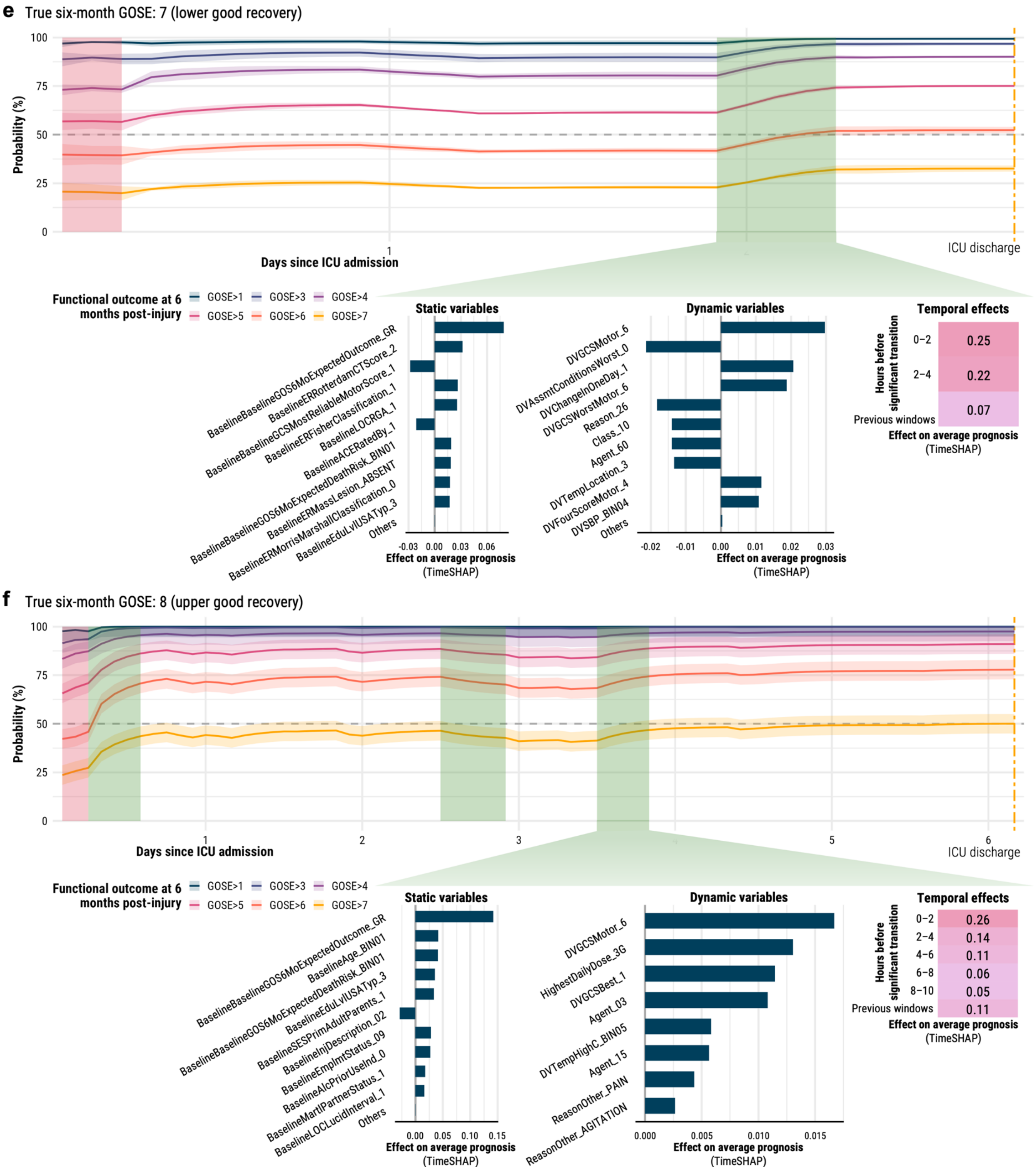

**Supplementary Figure 12. Example of individual ICU disease course for each level of six-month functional outcome.** All abbreviated variable names are decoded in Supplementary Note 1. Shaded red regions represent the pre-calibration range of model outputs (Figure 1) while shaded green regions represent high-magnitude transitions (as defined in the Methods). TimeSHAP values are interpreted as contributions of variables towards the difference in a patient's expected six-month functional outcome output from that of the average patient (Supplementary Figure 13). (**a**) A c. 40-year-old male, admitted to the ICU after a severe TBI (GCS 3), who died within six months of TBI (GOSE 1). (**b**) A c. 20-year-old female, admitted to the ICU after a severe TBI (GCS 7), who became severely disabled with partial functional dependency by six months post-injury (GOSE 4). (**c**) A c. 60-year-old male, admitted to the ICU after a

severe TBI (GCS 5), who became functionally independent but unable to participate in one or more life roles by six months post-injury (GOSE 5). (**d**) A c. 40-year-old male, admitted to the ICU after a severe TBI (GCS 3), who became functionally independent and able to participate in a limited capacity in life roles by six months post-injury (GOSE 6). (**e**) A c. 30-year-old female, admitted to the ICU after a moderate TBI (GCS 11), who returned to normal life with some neurological symptoms by six months post-injury (GOSE 7). (**f**) A c. 20-year-old male, admitted to the ICU after a mild TBI (GCS 15), who fully returned to normal life by six months post-injury (GOSE 8).

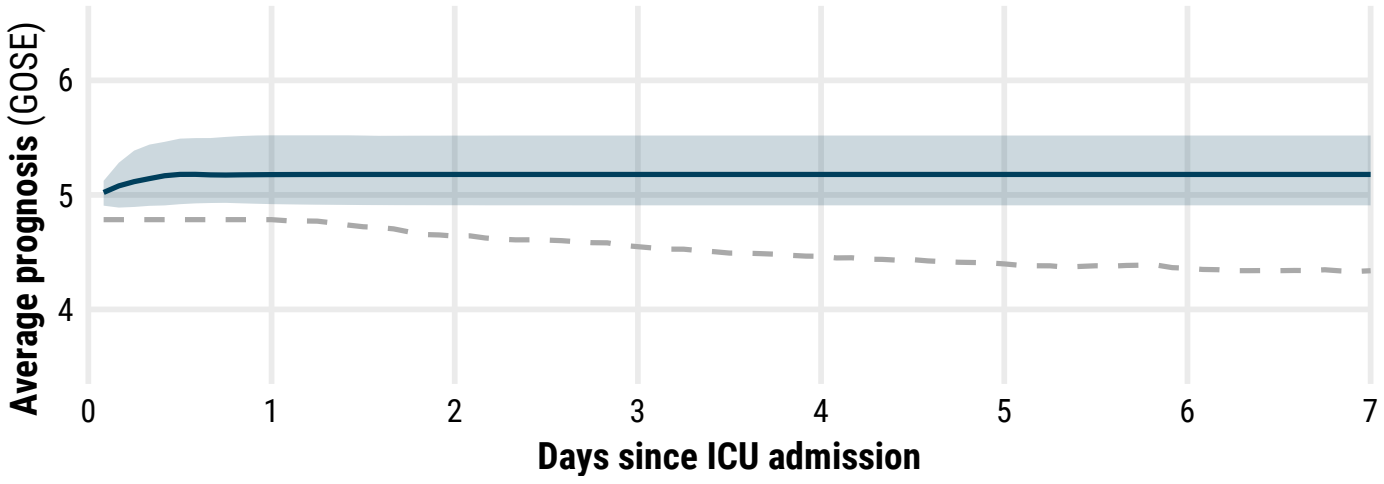

**Supplementary Figure 13. Expected six-month functional outcome output for the average patient token set.** The expected six-month functional outcome is defined as the GOSE corresponding to $\widehat{\mathbb{E}}[i]$, as defined in the Supplementary Methods. The average patient token set was defined to be one with tokens that are in 50+% of training set time windows, and the corresponding model output is the baseline for TimeSHAP calculation, against which variable contributions are calculated. The curve represents the median output across the 100 training sets of the repeated cross-validation partitions, and the shaded region represents the interquartile range. The dashed grey line represents the true average six-month functional outcome for the population over time, as patients left the ICU.

## SUPPLEMENTARY TABLES

**Supplementary Table 1. IMPACT extended model variables stratified by six-month functional outcome.**

| IMPACT extended model variables | Overall ($n$ = 1550) | Glasgow Outcome Scale – Extended (GOSE) at 6 months post-injury | | | | | | | $p$-value |
|---|---|---|---|---|---|---|---|---|---|
| | | **1** ($n$ = 318) | **2 or 3** ($n$ = 262) | **4** ($n$ = 120) | **5** ($n$ = 227) | **6** ($n$ = 200) | **7** ($n$ = 206) | **8** ($n$ = 217) | |
| Age [years] | 51 (31–66) | 66 (50–76) | 55 (36–68) | 48 (29–61) | 44 (31–56) | 41 (27–53) | 48 (31–65) | 41 (24–61) | <0.0001 |
| GCS motor score ($n$ = 1509) | 5 (1–6) | 2 (1–5) | 3 (1–5) | 5 (1–6) | 5 (1–6) | 5 (2–6) | 5 (3–6) | 6 (5–6) | <0.0001 |
| (1) No response | 484 (32.1%) | 152 (50.0%) | 104 (40.6%) | 35 (29.9%) | 63 (28.5%) | 46 (23.6%) | 47 (23.0%) | 37 (17.5%) | |
| (2) Abnormal extension | 54 (3.6%) | 17 (5.6%) | 20 (7.8%) | 4 (3.4%) | 6 (2.7%) | 3 (1.5%) | 2 (1.0%) | 2 (0.9%) | |
| (3) Abnormal flexion | 63 (4.2%) | 14 (4.6%) | 12 (4.7%) | 8 (6.8%) | 11 (5.0%) | 8 (4.1%) | 4 (2.0%) | 6 (2.8%) | |
| (4) Withdrawal from stimulus | 114 (7.6%) | 27 (8.9%) | 23 (9.0%) | 8 (6.8%) | 20 (9.0%) | 21 (10.8%) | 8 (3.9%) | 7 (3.3%) | |
| (5) Movement localised to stimulus | 305 (20.2%) | 52 (17.1%) | 47 (18.4%) | 24 (20.5%) | 50 (22.6%) | 46 (23.6%) | 44 (21.6%) | 42 (19.8%) | |
| (6) Obeys commands | 489 (32.4%) | 42 (13.8%) | 50 (19.5%) | 38 (32.5%) | 71 (32.1%) | 71 (36.4%) | 99 (48.5%) | 118 (55.7%) | |
| Unreactive pupils ($n$ = 1465) | | | | | | | | | <0.0001 |
| One | 111 (7.6%) | 31 (10.5%) | 31 (12.3%) | 7 (6.3%) | 20 (9.3%) | 5 (2.6%) | 8 (4.1%) | 9 (4.4%) | |
| Two | 168 (11.5%) | 84 (28.5%) | 33 (13.0%) | 8 (7.2%) | 14 (6.5%) | 8 (4.2%) | 16 (8.2%) | 5 (2.4%) | |
| Hypoxia | 207 (13.4%) | 60 (18.9%) | 33 (12.6%) | 14 (11.7%) | 35 (15.4%) | 33 (16.5%) | 16 (7.8%) | 16 (7.4%) | 0.37 |
| Hypotension | 210 (13.5%) | 56 (17.6%) | 51 (19.5%) | 21 (17.5%) | 32 (14.1%) | 22 (11.0%) | 15 (7.3%) | 13 (6.0%) | 0.0038 |
| Marshall CT score ($n$ = 1255) | VI (II–VI) | III (II–VI) | II (II–VI) | II (II–VI) | II (II–II) | II (II–III) | II (II–II) | VI (II–VI) | 0.043 |
| No visible pathology (I) | 118 (9.4%) | 8 (3.3%) | 11 (5.3%) | 5 (5.2%) | 17 (8.7%) | 25 (15.2%) | 24 (13.6%) | 28 (16.5%) | |
| Diffuse injury II | 592 (47.2%) | 56 (22.8%) | 84 (40.6%) | 54 (56.2%) | 92 (47.2%) | 100 (60.6%) | 103 (58.5%) | 103 (60.6%) | |
| Diffuse injury III | 108 (8.6%) | 42 (17.1%) | 17 (8.2%) | 10 (10.4%) | 14 (7.2%) | 9 (5.5%) | 6 (3.4%) | 10 (5.9%) | |
| Diffuse injury IV | 16 (1.3%) | 7 (2.8%) | 1 (0.5%) | 1 (1.0%) | 4 (2.1%) | 1 (0.6%) | 1 (0.6%) | 1 (0.6%) | |
| Mass lesion (V & VI) | 421 (33.5%) | 133 (54.0%) | 94 (45.4%) | 26 (27.1%) | 68 (34.9%) | 30 (18.2%) | 42 (23.9%) | 28 (16.5%) | |
| Traumatic subarachnoid haemorrhage ($n$ = 1254) | 957 (76.3%) | 221 (90.2%) | 176 (84.2%) | 73 (76.0%) | 150 (76.9%) | 106 (63.9%) | 125 (71.4%) | 106 (63.1%) | 0.16 |
| Extradural haematoma ($n$ = 1257) | 244 (19.4%) | 31 (12.7%) | 32 (15.3%) | 21 (21.9%) | 46 (23.6%) | 32 (19.3%) | 42 (23.9%) | 40 (23.5%) | 0.016 |
| Glucose [mmol/L] ($n$ = 1062) | 7.7 (6.6–9.4) | 8.8 (7.3–11) | 8.0 (6.5–9.8) | 7.6 (6.5–9.3) | 7.8 (6.6–9.6) | 7.7 (6.5–8.7) | 7.3 (6.3–8.5) | 7.1 (6.3–8.1) | 0.013 |
| Haemoglobin [g/dL] ($n$ = 1140) | 13 (12–14) | 13 (11–14) | 13 (11–14) | 14 (12–14) | 13 (12–14) | 14 (12–15) | 13 (12–15) | 14 (13–15) | 0.038 |

Data are median (IQR) for continuous variables and $n$ (% of column group) for categorical variables. Units of variables are provided in square brackets. If a characteristic had missing values for some patients in the population, the non-missing sample size was provided in parentheses – e.g., Marshall CT score ($n$ = 1255). $p$-values are determined from proportional odds logistic regression (POLR) analysis trained on all IMPACT variables concurrently[A1] and are combined across 100 missing value imputations via $z$-transformation[A2]. For categorical variables with $k > 2$ categories (e.g., GCSm), $p$-values were calculated with a likelihood ratio test (with $k$-1 degrees of freedom) on POLR.

**Supplementary Table 2. Calculated cut-offs to define high-magnitude transitions.**

| Threshold | Cutoffs (Δ%) | |
|---|---|---|
| | **Negative transitions**, i.e., worsening (≤) | **Positive transitions**, i.e., improvement (≥) |
| GOSE>1 | -6.258903 | 4.017577 |
| GOSE>3 | -6.168803 | 4.727354 |
| GOSE>4 | -5.831151 | 4.435974 |
| GOSE>5 | -5.253576 | 3.712104 |
| GOSE>6 | -4.371744 | 2.740238 |
| GOSE>7 | -3.044348 | 1.758345 |

For negative transitions, data represent the $1^{st}$ percentile of negative differences in consecutive time-window probabilities (in percentage) for the given GOSE threshold across the population. High-magnitude negative transitions were defined as those with probability differences less than or equal to the cut-off value for the given GOSE threshold. For positive transitions, data represent the $99^{th}$ percentile of positive differences in consecutive time-window probabilities (in percentage) for the given GOSE threshold across the population. High-magnitude positive transitions were defined as those with probability differences greater than or equal to the cut-off value for the given GOSE threshold.

**Supplementary Table 3. CENTER-TBI study sites, ethical committees, approval numbers, and approval dates.**

| Country | City | Insitute | Ethical Commitee | Number Approval | Date Approval |
|---|---|---|---|---|---|
| **Austria** | Vienna | Medizinische Universität Wien Universitätsklinik für Unfallchirurgie | Ethikkommission der Medizinischen Universität Wien | 1646/2014 | 3/05/2016 |
| **Austria** | Innsbruck | Medizinische Universität Innsbruck Universitätsklinik für Neurologie | Ethikkommission der Medizinischen Universität Innsbruck | AN2014-0336 343/4.22 | 1/12/2014 |
| **Belgium** | Antwerp | Antwerp University Hospital | Centraal Ethisch Comité - Ethisch Comité Universitair Ziekenhuis Antwerpen en de Universiteit Antwerpen | B300201422714 | 17/11/2014 |
| **Belgium** | Liège | CHR Citadelle | - Centraal Ethisch Comité - Ethisch Comité Universitair Ziekenhuis Antwerpen en de Universiteit Antwerpen<br>- Comité d'Ethique 412 | - B300201422714<br>- 1427 | -17/11/2014<br>- 25/11/2014 |
| **Belgium** | Liège | CHU | - Centraal Ethisch Comité - Ethisch Comité Universitair Ziekenhuis Antwerpen en de Universiteit Antwerpen<br>- Comité d'Ethique hospitalo-facultaire niversitaire de Liège (707) | - B300201422714<br>- B707201422102 / 2014-244 | - 17/11/2014<br>- 07/10/2014 |
| **Belgium** | Leuven | UZ Leuven | - Centraal Ethisch Comité - Ethisch Comité Universitair Ziekenhuis Antwerpen en de Universiteit Antwerpen<br>- Comissie Medische Ethiek UZ KU Leuven / Onderzoek | - B300201422714<br>- B322201523981 / S57019 (ML11365) | - 06/13/2015<br>- 07/10/2014 |
| **Denmark** | Odense | Odense Universiteitshospital - Neurokirurgisk afdeling | De Videnskabsetiske Komitéer for Region Syddanmark | S-20140215 | 16/03/2015 |
| **Denmark** | Copenhagen | Region Hovedstaden Rigshopitalet | De Videnskabsetiske Komitéer for Region Syddanmark | S-20140215 | 16/03/2015 |
| **Finland** | Turku | Turku University Hospital | Varsinais suomen sairaanhoitopiirin kuntayhtyma - Eettinen Toimikunta | 95/1801/2014 | 24/10/2014 |
| **Finland** | Helsinki | Helsinki University Central Hospital | Varsinais suomen sairaanhoitopiirin kuntayhtyma - Eettinen Toimikunta | 95/1801/2014 | 24/10/2014 |
| **France** | Paris | APHP | Agence Nationale de Sécurité du Médicament et des Produits de Santé ANSM | 141421B-31 | 19/12/2014 |
| **France** | Besançon | CHRU de Besançon | Agence Nationale de Sécurité du Médicament et des Produits de Santé ANSM | 141421B-31 | 19/12/2014 |
| **France** | Lille | Lille University Hospital | Agence Nationale de Sécurité du Médicament et des Produits de Santé ANSM | 141421B-31 | 19/12/2014 |
| **France** | Grenoble | University Hospital of Grenoble | Agence Nationale de Sécurité du Médicament et des Produits de Santé ANSM | 141421B-31 | 19/12/2014 |
| **France** | Nancy | University Hospital Nancy | Agence Nationale de Sécurité du Médicament et des Produits de Santé ANSM | 141421B-31 | 19/12/2014 |
| **France** | Poitiers | CHU Poitiers | Agence Nationale de Sécurité du Médicament et des Produits de Santé ANSM | 141421B-31 | 19/12/2014 |
| **Germany** | Heidelberg | Universitätsklinikum Heidelberg Neurochirurgische Kliniki | Ethikkommission Medizinsche Fakultät Heidelberg | S-435/2014 | 27/05/2015 |
| **Germany** | Berlin | Charité \| Campus Virchow Klinikum | Ethikkommission an der Medizinsche Fakultät Der rheinsch-Westfälischen Technischen Hocgschule Aachen | 1098/15 | 26/02/2016 |
| **Germany** | Aachen | Uniklinik RWTH Aachen | Ethikkommission an der Medizinsche Fakultät Der rheinsch-Westfälischen Technischen Hocgschule Aachen | EK 174/15 | 2/07/2015 |
| **Germany** | Ludwigsburg | Klinikum Ludwigsburg | Ethikkommission Medizinsche Fakultät Heidelberg | S-435/2014 | 29/01/2016 |
| **Hungary** | Pecs | Pecsi Tudomanyegyetem Klinikai Kozpont | - ETT TUKEB Egészségügyi Tudományos Tanács<br>- Pécsi Tudományegyetem | - 42558-3/2014/EKU<br>- 5421 | - 2/10/2014<br>- 28/03/2015 |
| **Hungary** | Szeged | zegedi tudományegyetem orvostudományi kar Szent Györgyi Albert Klinikai Központ | -ETT TUKEB Egészségügyi Tudományos Tanács<br>- Szegedi Tudományegyetem | - 42558-3/2014/EKU<br>- 3803 | - 2/10/2014<br>- 23/05/2016 |
| **Israel** | Haifa | Rambam Medical Center | Helsinki Committee, Rambam Health Care Campus | RMB 373-14 | 22/07/2015 |
| **Israel** | Jeruzalem | Hadassah-hebrew University Medical Center | Hadassah Medical Organization IRB | 0590-16 HMO | 6/12/2016 |
| **Italy** | Milan | Fondazione IRCCS Ca' Granda Ospedale Maggiore Policlinico | Fondazione IRCCS Ca' Granda Ospedale Maggiore Policlinico - Direzione Scientifica Comitato Etico | 542/2014 | 20/10/2014 |
| **Italy** | Milan | Ospedale San Raffaele | Comitato Etico - Ospedale San Raffaele | 217/2014 | 10/04/2015 |
| **Italy** | Torino | AOU Città della Salute e della Scienza di Torino | Comitato Etico Interaziendale A.O.U. Città della Salute e della Scienza di Torino - A.O. Ordine Mauriziano - A.S.L. | 0015269 | 15/02/2016 |
| **Italy** | Cesena | Bufalini Hospital | Comitato Etico IRST IRCCS AVR | 1675/2015 I.5/207 | 18/03/2015 |
| **Italy** | Padova | Azienda Ospedaliera Università di Padova | Comitato Etico - Ospedale San Raffaele | 217/2014 | 10/04/2015 |
| **Italy** | Monza | San Gerardo Hospital/ASST | Comitato Etico Della Provincia Monza Brianza | 1978/2014 | 22/12/2014 |
| **Italy** | Novara | Maggiore Della Carità Hospital | Comitato Etico Interaziendale A.O.U. 'Maggiore della Carità' | CE 46/15 | 3/07/2015 |
| **Italy** | Milan | Niguarda Hospital | Comitato Etico - Ospedale Niguarda Ca' Granda | 636-122015 | 23/12/2015 |
| **Latvia** | Riga | Pauls Stradins Clinical University Hospital | Ethics Commiitee for Clinical Research at Pauls Stradins Clinical University Hospital Development Society | 171215-1E | 17/12/2015 |
| **Latvia** | Riga | Riga Eastern Clinical University Hospital | Ethics Commiitee for Clinical Research at Pauls Stradins Clinical University Hospital Development Society | 171215-1E | 17/12/2015 |
| **Latvia** | Rezekne | Rezekne Hospital | Ethics Commiitee for Clinical Research at Pauls Stradins Clinical University Hospital Development Society | 171215-1E | 17/12/2015 |
| **Lithuania** | Vilnius | Vilniaus Universiteto Ligonine | VILNIAUS REGIONINIS BIOMEDICININIŲ TYRIMŲ ETIKOS KOMITETAS | 158200-15-801-323 | 6/10/2015 |
| **Lithuania** | Kaunas | LSMUL Kauno klinikos Skubio pagalbos skyrius | KAUNO REGIONINIS BIOMEDICININIŲ TYRIMŲ ETIKOS KOMITETAS | BE-2-6 | 6/01/2015 |
| **Netherlands** | Leiden | Het Leids Universitair Medisch Centrum te Leiden | Leids Universitair Centrum - Commissie Medische Ethiek | P14.222/NV/nv | 3/12/2014 |
| **Netherlands** | Rotterdam | Erasmus MC | Leids Universitair Centrum - Commissie Medische Ethiek | P14.222/NV/nv | 3/12/2014 |
| **Netherlands** | The Hague | Medisch Centrum Haaglanden | Leids Universitair Centrum - Commissie Medische Ethiek | P14.222/NV/nv | 3/12/2014 |
| **Netherlands** | The Hague | Het Haga Hospital | Leids Universitair Centrum - Commissie Medische Ethiek | P14.222/NV/nv | 25/11/2015 |
| **Netherlands** | Nijmegen | Radboud UMC | Leids Universitair Centrum - Commissie Medische Ethiek | P14.222/NV/nv | 2/07/2015 |
| **Netherlands** | Tilburg | St. Elisabeth Ziekenhuis | Leids Universitair Centrum - Commissie Medische Ethiek | P14.222/NV/nv | 2/07/2015 |
| **Netherlands** | Groningen | UMC Groningen | Leids Universitair Centrum - Commissie Medische Ethiek | P14.222/NV/nv | 3/12/2014 |
| **Norway** | Tromso | Universitetssykehuset Nord-Norge | Regional komité for medisinsk og helsefaglig forskningsetikk REK midt-Norge (REK midt) | 2014/1454 | 23/03/2015 |
| **Norway** | Trondheim | St.Olavs Hospital | Regional komité for medisinsk og helsefaglig forskningsetikk REK midt-Norge (REK midt) | 2014/1454 | 23/03/2015 |
| **Norway** | Oslo | Oslo Universitetssykehus | Regional komité for medisinsk og helsefaglig forskningsetikk REK midt-Norge (REK midt) | 2014/1454 | 23/03/2015 |
| **Romania** | Timisoara | Clinica de Neurochirurgie Universitatea de Medicina si Farmacie "Victor Babes" Timisoara | Comitetului de Etica a Spitalului Clinic Judeteam de Urgenta Timisoara | 73 | 16/10/2014 |
| **Serbia** | Novi Sad | Klinidkog centra Vojvodine | Etidkog odbora Klinidkog centra Vojvodine | 00-08/332 | 11/07/2014 |
| **Spain** | Madrid | Hospital Universitario 12 de Octubre | Comité Etico de Investigacion Clinica del Hospital Universitario 12 de Octubre | 14/262 | 3/11/2014 |
| **Spain** | Barcelona | Vall d'Hebron University Hospital | Comité ético de investigación clínica y comisión de proyectos de investigación del hospital universitari Vall d'Hebron | ID-RTF080 | 13/11/2014 |
| **Spain** | Bilbao | Clínico Universitario de Cruces | Comité Etico de Investigacion Clinica de Euskadi | PI2014158 | 24/02/2015 |
| **Spain** | Valencia | Clínico Universitario de Valencia | Comité Etico de Investigacion Clinica del Clínico Universitario de Valencia | F-CE-GEva-15 | 23/04/2015 |
| **Sweden** | Stockholm | Karolinksa University Hospital | EPN (Regionala Etikprövningsnämnden i Stockholm) | 2014/1473-31/4 | 24/09/2014 |
| **Sweden** | Umea | Umea University Hosptital | EPN (Regionala Etikprövningsnämnden i Stockholm) | 2014/1473-31/4 | 24/09/2014 |
| **Switzerland** | Lausanne | Centre hospitalier universitaire Vaudois | La Commission cantonale (VD) d'éthique de la recherche sur l'être humain (CER-VD) | 473/11 | 9/12/2014 |
| **UK** | Birmingham | Queen Elizabeth Hospital | - NHS HRA<br>- UHB Research Governance Office - Queen Elizabeth Hospital | - 14/SC/1370<br>- RRK5224 | - 22/12/2014<br>- 24/02/2016 |
| **UK** | Cambridge | Cambridge University Hospital NHS Foundation Trust | - NHS HRA<br>- Research and Development Department - Cambridge University Hospital NHS Foundation Trust | - 14/SC/1370<br>- AO93184 | - 22/12/2014<br>- 30/01/2015 |
| **UK** | Southampton | University Hospitals Southhampton NHS Trust | - NHS HRA<br>- Research Governance Office - University Hospitals Southhampton NHS Trust | - 14/SC/1370<br>- RHM CRI0294 | - 22/12/2014<br>- 13/03/2015 |
| **UK** | Sheffield | Sheffield Teaching Hospitals NHS Foundation Trust | - NHS HRA<br>- Research and Development Department - Sheffield Teaching Hospitals NHS Foundation Trust | - 14/SC/1370<br>- STH18187 | - 22/12/2014<br>- 25/03/2015 |
| **UK** | London | Kings college London | - NHS HRA<br>- Research & Innovation Office - Kings college London NHS Foundation Trust | - 14/SC/1370<br>- KCH15-204 | -<br>22/12/2014<br>-<br>30/12/2015 |
| **UK** | Salford | Salford Royal Hospital | - NHS HRA<br>- Research and Development Department -Salford Royal Hospital NHS Foundation Trust | - 14/SC/1370<br>- 2015/025ET | -<br>22/12/2014<br>-<br>27/07/2015 |
| **UK** | Liverpool | The Walton centre NHS Foundation Trust | - NHS HRA<br>- Research & Innovation Office - The Walton centre NHS Foundation Trust | - 14/SC/1370<br>- RG154-15 | -<br>22/12/2014<br>-<br>11/05/2015 |
| **UK** | Bristol | Southmead Hospital Bristol | - NHS HRA<br>- Research & Innovation - North Bristol NHS Trust | - 14/SC/1370<br>- 3427 | -<br>22/12/2014<br>-<br>23/12/2014 |
| **UK/Scotland** | Edinburgh | Lothian Health Board | - NHS Scotland<br>- Research and Development Department - University Hospitals Division NHS Lothian | - 14/SS/1086<br>- 2015/0171 | -<br>23/12/2014<br>-<br>19/06/2015 |

**SUPPLEMENTARY NOTES**

NOTE: Static variables are those with values fixed at ICU admission (e.g., helmet on during accident?). Intervention variable directly represent a treatment or management decision performed during a patient's ICU stay (i.e., administration of hypertonic saline). Since an intervention variable must take place during a patient's ICU stay, a variable cannot be both a static and an intervention variable. However, a variable can be both not static (i.e., dynamic) and not an intervention (e.g., a result from an ICU lab test or imaging report).

**Supplementary Note 1. Variables by category.**

*Demographics and socioeconomic status*

| Name | Static | Intervention | Format | Description | Possible Values |
|---|---|---|---|---|---|
| Age | TRUE | FALSE | integer | Age, recorded in years. | Not Applicable |
| EduLvlEUROFather | TRUE | FALSE | integer | Father Highest level of education completed. Only if patient age <18 | 0=None, not currently in school;1=Currently in diploma or degree-oriented program;2=Primary school;3=Secondary school / High school;4=Post-high school training (e.g. trade/technical certificate);5=College / University (diploma or degree);88=Unknown |
| EduLvlEUROMother | TRUE | FALSE | integer | Mother Highest level of education completed. Only if patient age <18 | 0=None, not currently in school;1=Currently in diploma or degree-oriented program;2=Primary school;3=Secondary school / High school;4=Post-high school training (e.g. trade/technical certificate);5=College / University (diploma or degree);88=Unknown |
| EduLvlUSATyp | TRUE | FALSE | integer | Highest level of education of patient. | 0=None, not currently in school;1=Currently in diploma or degree-oriented program;2=Primary school;3=Secondary school / High school;4=Post-high school training (e.g. trade/technical certificate);5=College / University (diploma or degree);88=Unknown |
| EduYrCt | TRUE | FALSE | integer | Number of years of education completed. | Not Applicable |
| EmplmtStatus | TRUE | FALSE | integer | Employment status before injury. | 1=Working (35 hours or more per week);2=Working (20-34 hours per week);3=Working (less than 20 hours per week);4=In working force, but currently on sick leave;5=Special employment / sheltered employment;6=Looking for work, unemployed;7=Unable to work;8=Retired;9=Student / schoolgoing;10=Homemaker, keeping house;88=Unknown |
| JobclassCat | TRUE | FALSE | integer | Reflects job category. Only displayed if employment status (Subject.EmplmtStatus) is 1-4 | 0=None;1=Manager / Professional;2=Technician / Supervisor / Associate Professional;3=Clerk / Sales;4=Skilled manual worker;5=Manual worker;99=Other |
| JobclassCatOther | TRUE | FALSE | text | Specifies the "other" job category than predefined list. Only displayed if employment status (Subject.EmplmtStatus) is 1-4 | Not Applicable |
| MartlPartnerStatus | TRUE | FALSE | integer | Marital status. | 1=Never been married;2=Married;3=Living together/common law;4=Divorced;5=Separated;6=Widowed;88=Unknown;99=Other |
| Race | TRUE | FALSE | text | Race of Subject | Asian=Asian;Black=Black;NotAllowed=Not allowed; Unknown=Unknown;White=White |
| SesEduNoFather | TRUE | FALSE | integer | Father's number of years education completed. Only if patient age <18 | Not Applicable |
| SesEduNoMother | TRUE | FALSE | integer | Mother's number of years of education completed. Only if patient age <18 | Not Applicable |
| SesNumberPeopleLivingWith | TRUE | FALSE | integer | Living together and number of people living with provides an indication of "social support" - important to recovery and social re-integration. | Not Applicable |
| SESPrimAdultAlone | TRUE | FALSE | integer | Reflects living situation prior to the injury: --> Alone | 0=No;1=Yes |
| SESPrimAdultChildren | TRUE | FALSE | integer | Reflects living situation prior to the injury: --> with Child/children | 0=No;1=Yes |
| SESPrimAdultParents | TRUE | FALSE | integer | Reflects living situation prior to the injury: --> with parents | 0=No;1=Yes |
| SESPrimAdultSiblings | TRUE | FALSE | integer | Reflects living situation prior to the injury: --> with siblings | 0=No;1=Yes |
| SESPrimAdultSignOther | TRUE | FALSE | integer | Reflects living situation prior to the injury: --> with a significant other partner | 0=No;1=Yes |
| SESPrimAdultSpousePartner | TRUE | FALSE | integer | Reflects living situation prior to the injury: --> with spouse | 0=No;1=Yes |
| Sex | TRUE | FALSE | text | Sex of subject | F=Female;M=Male |
| SiteCode | TRUE | FALSE | text | Anonymized site code. | Not Applicable |
| StudentStatus | TRUE | FALSE | integer | Only displayed if employment status is student/school going | 0=None;1=Full time, diploma / degree oriented;2=Part time, diploma / degree oriented;3=Other school;88=Unknown |

*Medical and behavioural history*

| Name | Static | Intervention | Format | Description | Possible Values |
|---|---|---|---|---|---|
| AlcPriorUseInd | TRUE | FALSE | integer | On presentation the behavioral history of the patient was recorded. This reflects his past use of alcoholic beverages (beer, wine, spirits). | 0=No;1=Yes;88=Unknown |
| AlcUseDur | TRUE | FALSE | integer | On presentation the behavioral history of the patient was recorded. This reflects the number of years of alcohol use, if past use of alcoholic beverages (beer, wine, spirits) was 'yes'. | Not Applicable |
| AlcUseLstMoDaysDrankNum | TRUE | FALSE | integer | On presentation the behavioral history of the patient was recorded. This reflects his use in the past three months of alcoholic beverages (beer, wine, spirits) (>2/day) | 0=No;1=Yes;88=Unknown |
| AnticoagAntiThrombinProtein | TRUE | FALSE | integer | Medical history. Use of anticoagulant or platelet aggregation inhibitor by the patient. This variable describes the use of antithrombin protein therapeutics (Atryn). | 0=No;1=Yes |
| AnticoagCoumarin | TRUE | FALSE | integer | Medical history. Use of anticoagulant or platelet aggregation inhibitor by the patient. This variable describes the use of Coumarin derivative (Coumadin, Warfarin). | 0=No;1=Yes |
| AnticoagDirectThrombinInhib | TRUE | FALSE | integer | Medical history. Use of anticoagulant or platelet aggregation inhibitor by the patient. This variable describes the use of direct thrombin inhibitors (eg. dabigatran, argatroban, melagatran). | 0=No;1=Yes |
| AnticoagFactorXaInhib | TRUE | FALSE | integer | Medical history. Use of anticoagulant or platelet aggregation inhibitor by the patient. This variable describes the use of inhibitor of factor Xa (eg. rivaroxaban). | 0=No;1=Yes |
| AnticoagHeparin | TRUE | FALSE | integer | Medical history. Use of anticoagulant or platelet aggregation inhibitor by the patient. This variable describes the use of heparin. | 0=No;1=Yes |
| AnticoagLowMolHeparin | TRUE | FALSE | integer | Medical history. Use of anticoagulant or platelet aggregation inhibitor by the patient. This variable describes the use of low molecular weight heparin | 0=No;1=Yes |
| AnticoagulantOther | TRUE | FALSE | integer | Medical history. Use of anticoagulant or platelet aggregation inhibitor by the patient. This variable describes the use of another type of anticoagulant or platelet aggregation inhibitor, not specified elsewhere. | 0=No;1=Yes |
| AnticoagulantOtherTxt | TRUE | FALSE | text | Medical history. Use of anticoagulant or platelet aggregation inhibitor by the patient. This variable describes the use of another type of anticoagulant or platelet aggregation inhibitor, not specified elsewhere. Text field. | Not Applicable |
| AnticoagulantReasonCardiac | TRUE | FALSE | integer | Medical history. This variable is populated when the reason for using anticoagulants or platelet aggregation inhibitors by the patient is cardiac. | 0=No;1=Yes |
| AnticoagulantReasonCardiacCABG | TRUE | FALSE | integer | Medical history. This variable is populated when the reason for using anticoagulants or platelet aggregation inhibitors by the patient is cardiac, specifically CABG. | 0=No;1=Yes |
| AnticoagulantReasonCardiacFibrill | TRUE | FALSE | integer | Medical history. This variable is populated when the reason for using anticoagulants or platelet aggregation inhibitors by the patient is cardiac, specifically atrial fibrillation/flutter. | 0=No;1=Yes |
| AnticoagulantReasonCardiacStent | TRUE | FALSE | integer | Medical history. This variable is populated when the reason for using anticoagulants or platelet aggregation inhibitors by the patient is cardiac, specifically a cardiac stent. | 0=No;1=Yes |
| AnticoagulantReasonCardiacValve | TRUE | FALSE | integer | Medical history. This variable is populated when the reason for using anticoagulants or platelet aggregation inhibitors by the patient is cardiac, specifically a valve prosthesis. | 0=No;1=Yes |
| AnticoagulantReasonCardiovas | TRUE | FALSE | integer | Medical history. This variable is populated when the reason for using anticoagulants or platelet aggregation inhibitors by the patient is cardiovascular. | 0=N0;1=Yes |

| Name | Static | Intervention | Format | Description | Possible Values |
|---|---|---|---|---|---|
| AnticoagulantReasonCardiovasCarotidStent | TRUE | FALSE | integer | Medical history. This variable is populated when the reason for using anticoagulants or platelet aggregation inhibitors by the patient is cardiovascular, specifically a carotid or cerebral stent. | 0=No;1=Yes |
| AnticoagulantReasonCardiovasLimbIsch | TRUE | FALSE | integer | Medical history. This variable is populated when the reason for using anticoagulants or platelet aggregation inhibitors by the patient is cardiovascular, specifically limb ischaemia. | 0=No;1=Yes |
| AnticoagulantReasonCardiovasOtherStent | TRUE | FALSE | integer | Medical history. This variable is populated when the reason for using anticoagulants or platelet aggregation inhibitors by the patient is cardiovascular, specifically a stent not specified elsewhere. | 0=No;1=Yes |
| AnticoagulantReasonCardiovasStenosis | TRUE | FALSE | integer | Medical history. This variable is populated when the reason for using anticoagulants or platelet aggregation inhibitors by the patient is cardiovascular, specifically a cardiovascular stenosis | 0=No;1=Yes |
| AnticoagulantReasonCardiovasTIS | TRUE | FALSE | integer | Medical history. This variable is populated when the reason for using anticoagulants or platelet aggregation inhibitors by the patient is cardiovascular, specifically a transient ischaemic attack/stroke | 0=No;1=Yes |
| AnticoagulantReasonOther | TRUE | FALSE | integer | Medical history. This variable is populated when the reason for using anticoagulants or platelet aggregation inhibitors by the patient is not specified elsewhere. | 0=No;1=Yes |
| AnticoagulantReasonOtherTxt | TRUE | FALSE | text | Medical history. This variable is populated when the reason for using anticoagulants or platelet aggregation inhibitors by the patient is not specified elsewhere (text field). | Not Applicable |
| AnticoagulantReasonThrombo | TRUE | FALSE | integer | Medical history. This variable is populated when the reason for using anticoagulants or platelet aggregation inhibitors by the patient is thromboembolic. | 0=No;1=Yes |
| AnticoagulantReasonThromboDVTLess6 | TRUE | FALSE | integer | Medical history. This variable is populated when the reason for using anticoagulants or platelet aggregation inhibitors by the patient is thromboembolic, specifically a single episode of DVT (deep venous thrombosis) or PE (pulmonary embolism) <6 months. | 0=No;1=Yes |
| AnticoagulantReasonThromboDVTMore6 | TRUE | FALSE | integer | Medical history. This variable is populated when the reason for using anticoagulants or platelet aggregation inhibitors by the patient is thromboembolic, specifically a single episode of DVT (deep venous thrombosis) or PE (pulmonary embolism) >6 months. | 0=No;1=Yes |
| AnticoagulantReasonThromboMultipleEpisode | TRUE | FALSE | integer | Medical history. This variable is populated when the reason for using anticoagulants or platelet aggregation inhibitors by the patient is thromboembolic, specifically two or more episodes of DVT (deep venous thrombosis) or PE (pulmonary embolism). | 0=No;1=Yes |
| AnticoagXarelto | TRUE | FALSE | integer | Medical history. Use of anticoagulant or platelet aggregation inhibitor by the patient. This variable describes the use of Xarelto | 0=No;1=Yes |
| AUDITCAlcDrnkTypclDayNumScore | TRUE | FALSE | integer | On presentation the behavioral history of the patient was recorded. In case of past use of alcoholic beverages (beer, wine, spirits), this reflects the alcohol frequency: average number of drinks on a "drinking" day | 1=1-2;2=3-4;3=5-6;4=7-9;5=10 or more;88=Unknown |
| AUDITCDrnkContainAlcFreqScore | TRUE | FALSE | integer | Detailed questions on the use of alcohol are derived from the first 3 questions of the "AUDIT" questionnaire, a screening tool developed by the World Health Organization (WHO) to assess alcohol consumption, drinking behaviors, and alcohol-related problems. The same questions are asked post-injury during full follow-up assessments (including cognitive testing.) This reflects the frequency of having a drink containing alcohol. | 0=Never;1=Monthly or less;2=2-4 times a month;3=2-3 times a week;4=4 or more times a week;88=Unknown |

| Name | Static | Intervention | Format | Description | Possible Values |
|---|---|---|---|---|---|
| AUDITCMoreThan6AlcDrnkFreqScore | TRUE | FALSE | integer | Detailed questions on the use of alcohol are derived from the first 3 questions of the "AUDIT" questionnaire, a screening tool developed by the World Health Organization (WHO) to assess alcohol consumption, drinking behaviors, and alcohol-related problems. The same questions are asked post-injury during full follow-up assessments (including cognitive testing.) This reflects the frequency of having six or more drinks on one occasion. | 0=Never;1=Monthly or less (incorrect please correct);2=2-4 times a month (incorrect please correct);3=2-3 times a week (incorrect please correct);4=4 or more times a week (incorrect please correct);5=Less than monthly;6=Monthly;7=Weekly;8=Daily or almost daily;88=Unknown |
| BetaBlocker | TRUE | FALSE | integer | A specific question on the use of beta-blockers is included as some reports indicate better outcome with the use of beta blockers (Research interest Rotterdam). If yes, specification is requested and differentiated into: Non-selective blockers/Selective beta-1 blockers/alpha-1 and beta blockers. | 0=No;1=Yes;88=Unknown |
| BetaBlockerAlphaBucundolol | TRUE | FALSE | integer | Medical history. Use of anticoagulant or platelet aggregation inhibitor by the patient. This variable describes the use of a beta blocker, specifically Bucindolol. | 0=No;1=Yes |
| BetaBlockerAlphaCarvedilol | TRUE | FALSE | integer | Medical history. Use of anticoagulant or platelet aggregation inhibitor by the patient. This variable describes the use of a beta blocker, specifically Carvedilol (Eucardic). | 0=No;1=Yes |
| BetaBlockerAlphaLabetolol | TRUE | FALSE | integer | Medical history. Use of anticoagulant or platelet aggregation inhibitor by the patient. This variable describes the use of a beta blocker, specifically Labetalol (Trandate). | 0=No;1=Yes |
| BetaBlockerAlphaOther | TRUE | FALSE | integer | Medical history. Use of anticoagulant or platelet aggregation inhibitor by the patient. This variable describes the use of an alpha 1 and beta-blocker, not specified elsewhere. | 0=No;1=Yes |
| BetaBlockerAlphaOtherTxt | TRUE | FALSE | text | Medical history. Use of anticoagulant or platelet aggregation inhibitor by the patient. This variable describes the use of an alpha 1 and beta-blocker, not specified elsewhere (text field). | Not Applicable |
| BetaBlockerNonSelectCarteolol | TRUE | FALSE | integer | Medical history. Use of anticoagulant or platelet aggregation inhibitor by the patient. This variable describes the use of beta blockers, specifically Carteolol. | 0=No;1=Yes |
| BetaBlockerNonSelectNadolol | TRUE | FALSE | integer | Medical history. Use of anticoagulant or platelet aggregation inhibitor by the patient. This variable describes the use of beta blockers, specifically Nadolol | 0=No;1=Yes |
| BetaBlockerNonSelectOther | TRUE | FALSE | integer | Medical history. Use of anticoagulant or platelet aggregation inhibitor by the patient. This variable describes the use of beta blockers, nonselective, not specified elsewhere. | 0=No;1=Yes |
| BetaBlockerNonSelectOtherTxt | TRUE | FALSE | text | Medical history. Use of anticoagulant or platelet aggregation inhibitor by the patient. This variable describes the use of beta blockers, nonselective, not specified elsewhere (textfield). | Not Applicable |
| BetaBlockerNonSelectPenbutolol | TRUE | FALSE | integer | Medical history. Use of anticoagulant or platelet aggregation inhibitor by the patient. This variable describes the use of beta blockers, specifically Penbutolo. | 0=No;1=Yes |
| BetaBlockerNonSelectPindolol | TRUE | FALSE | integer | Medical history. Use of anticoagulant or platelet aggregation inhibitor by the patient. This variable describes the use of beta blockers, specifically Pindolol (Viskeen) | 0=No;1=Yes |
| BetaBlockerNonSelectPropranolol | TRUE | FALSE | integer | Medical history. This variable describes the use of beta blockers, specifically Propranolol | 0=No;1=Yes |
| BetaBlockerNonSelectSotalol | TRUE | FALSE | integer | Medical history. This variable describes the use of beta blockers, specifically Sotalol (Sotacor) | 0=No;1=Yes |
| BetaBlockerSelectAcebutolol | TRUE | FALSE | integer | Medical history. This variable describes the use of beta blockers, specifically Acebutolol (Sectral) | 0=No;1=Yes |

| Name | Static | Intervention | Format | Description | Possible Values |
|---|---|---|---|---|---|
| BetaBlockerSelectAtenolol | TRUE | FALSE | integer | Medical history. This variable describes the use of beta blockers, specifically Atenolol (Tenormin) | 0=No;1=Yes |
| BetaBlockerSelectBetaxolol | TRUE | FALSE | integer | Medical history. This variable describes the use of beta blockers, specifically Betaxolol (Kerlon) | 0=No;1=Yes |
| BetaBlockerSelectBisoprolol | TRUE | FALSE | integer | Medical history. This variable describes the use of beta blockers, specifically Bisoprolol (Emcor) | 0=No;1=Yes |
| BetaBlockerSelectCeliprolol | TRUE | FALSE | integer | Medical history. This variable describes the use of beta blockers, specifically Celiprolol (Dilanorm) | 0=No;1=Yes |
| BetaBlockerSelectEsmolol | TRUE | FALSE | integer | Medical history. This variable describes the use of beta blockers, specifically Esmolol (Brevibloc) | 0=No;1=Yes |
| BetaBlockerSelectMetoprolol | TRUE | FALSE | integer | Medical history. This variable describes the use of beta blockers, specifically Metoprolol (Selokeen). | 0=No;1=Yes |
| BetaBlockerSelectNebivolol | TRUE | FALSE | integer | Medical history. This variable describes the use of beta blockers, specifically Nebivolol (Nebilet) | 0=No;1=Yes |
| BetaBlockerSelectOther | TRUE | FALSE | integer | Medical history. This variable describes the use of beta blockers, specifically selective beta1blockers not specified elsewhere. | 0=No;1=Yes |
| CannabisCurrentUse | TRUE | FALSE | integer | On presentation the behavioral history of the patient was recorded. This reflects the use in the past three months of Cannabis (marijuana, pot, grass, hash, etc.) | 0=No;1=Yes;88=Unknown |
| CannabisPriorUse | TRUE | FALSE | integer | On presentation the behavioral history of the patient was recorded. This reflects his past use of Cannabis (marijuana, pot, grass, hash, etc.) | 0=No;1=Yes;88=Unknown |
| CannabisPriorUseDuration | TRUE | FALSE | text | On presentation the behavioral history of the patient was recorded. This reflects the number of years of past use of Cannabis if applicable. | Not Applicable |
| DrgSubIllctCurntUseInd | TRUE | FALSE | integer | On presentation the behavioral history of the patient was recorded. This reflects the use in the past three months of Other recreational drugs (than Cannabis) | 0=No;1=Yes;88=Unknown |
| DrgSubIllctUseCatOther | TRUE | FALSE | text | On presentation the behavioral history of the patient was recorded. This reflects his past use of which type of drugs. | Not Applicable |
| DrgSubIllctUseDur | TRUE | FALSE | integer | On presentation the behavioral history of the patient was recorded. This reflects the number of years of his past use of recreational drugs, if applicable. | Not Applicable |
| DrgSubPriorIllctUseInd | TRUE | FALSE | integer | On presentation the behavioral history of the patient was recorded. This reflects his past use of Other recreational drugs (other than Cannabis) | 0=No;1=Yes;88=Unknown |
| DrugIllicitCurrentUseOther | TRUE | FALSE | text | On presentation the behavioral history of the patient was recorded. This reflects the use in the past three months of which type of drugs | Not Applicable |
| MedHxAnticoagulantsOrPlatelet | TRUE | FALSE | integer | Summary question to document if the subject was taking anticoagulants or platelet aggregation inhibitors prior to injury. If yes, details are requested concerning which (groups of) agents were used. This information is of high relevance in relation to the shift of epidemiologic patterns in TBI towards higher age (with more co-morbidities and medication). | 0=No;1=Yes anticoagulants;2=Yes platelet aggregation inhibitors;3=Yes, both;88=Unknown |
| MedHxCardio | TRUE | FALSE | integer | Details on Medical History are captured for 15 body regions/disease area's. If overall question for the body region/disease is "yes", more detailed info is requested and captured in the sub-domains. This variable is used for documenting cardiovascular medical history. | 0=No;1=Yes;88=Unknown |
| MedHxCardioArrhythmia | TRUE | FALSE | integer | Details on Medical History are captured for 15 body regions/disease area's. If overall question for the body region/disease is "yes", more detailed info is requested and | 0=No;1=Yes |

| Name | Static | Intervention | Format | Description | Possible Values |
|---|---|---|---|---|---|
| | | | | captured in the sub-domains. This variable is used for documenting cardiovascular medical history, specifically arrhythmia. | |
| MedHxCardioCongenitalHD | TRUE | FALSE | integer | Details on Medical History are captured for 15 body regions/disease area's. If overall question for the body region/disease is "yes", more detailed info is requested and captured in the sub-domains. This variable is used for documenting cardiovascular medical history, specifically congenital heart disease. | 0=No;1=Yes |
| MedHxCardioHTN | TRUE | FALSE | integer | Details on Medical History are captured for 15 body regions/disease area's. If overall question for the body region/disease is "yes", more detailed info is requested and captured in the sub-domains. This variable is used for documenting cardiovascular medical history, specifically hypertension | 0=No;1=Yes |
| MedHxCardioIschemicHD | TRUE | FALSE | integer | Details on Medical History are captured for 15 body regions/disease area's. If overall question for the body region/disease is "yes", more detailed info is requested and captured in the sub-domains. This variable is used for documenting cardiovascular medical history, specifically Ischemic heart disease | 0=No;1=Yes |
| MedHxCardioNYHA | TRUE | FALSE | text | Details on Medical History are captured for 15 body regions/disease area's. If overall question for the body region/disease is "yes", more detailed info is requested and captured in the sub-domains. This variable is used for documenting cardiovascular medical history, specifically NYHA, a classification system for severity of cardiac disease - generally used for ischaemia, but used here in broader sense. | I=I;II=II;III=III;IV=IV |
| MedHxCardioOther | TRUE | FALSE | integer | Details on Medical History are captured for 15 body regions/disease area's. If overall question for the body region/disease is "yes", more detailed info is requested and captured in the sub-domains. This variable is used for documenting cardiovascular medical history, not specified elsewhere. | 0=No;1=Yes |
| MedHxCardioOtherTxt | TRUE | FALSE | text | Details on Medical History are captured for 15 body regions/disease area's. If overall question for the body region/disease is "yes", more detailed info is requested and captured in the sub-domains. This variable is used for documenting cardiovascular medical history, not specified elsewhere (textfield) | Not Applicable |
| MedHxCardioPeripheralVascular | TRUE | FALSE | integer | Details on Medical History are captured for 15 body regions/disease area's. If overall question for the body region/disease is "yes", more detailed info is requested and captured in the sub-domains. This variable is used for documenting cardiovascular medical history, specifically peripheral vascular disease. | 0=No;1=Yes |
| MedHxCardioThromboembolic | TRUE | FALSE | integer | Details on Medical History are captured for 15 body regions/disease area's. If overall question for the body region/disease is "yes", more detailed info is requested and captured in the sub-domains. This variable is used for documenting cardiovascular medical history, specifically thromboembolic | 0=No;1=Yes |
| MedHxCardioValvularHD | TRUE | FALSE | integer | Details on Medical History are captured for 15 body regions/disease area's. If overall question for the body region/disease is "yes", more detailed info is requested and captured in the sub-domains. This variable is used for documenting cardiovascular | 0=No;1=Yes |

| Name | Static | Intervention | Format | Description | Possible Values |
|---|---|---|---|---|---|
| | | | | medical history, specifically valvular heart disease | |
| MedHxDevelopmental | TRUE | FALSE | integer | Details on Medical History are captured for 15 body regions/disease area's. If overall question for the body region/disease is "yes", more detailed info is requested and captured in the sub-domains. This variable is used for documenting developmental diseases. | 0=No;1=Yes;88=Unknown |
| MedHxDevelopmentalADDandADHD | TRUE | FALSE | integer | Details on Medical History are captured for 15 body regions/disease area's. If overall question for the body region/disease is "yes", more detailed info is requested and captured in the sub-domains. This variable is used for documenting developmental diseases, specifically attention deficit/hyperactivity disorder. | 0=No;1=Yes |
| MedHxDevelopmentalLearningDisability | TRUE | FALSE | integer | Details on Medical History are captured for 15 body regions/disease area's. If overall question for the body region/disease is "yes", more detailed info is requested and captured in the sub-domains. This variable is used for documenting developmental diseases, specifically learning disability | 0=No;1=Yes |
| MedHxDevelopmentalOther | TRUE | FALSE | integer | Details on Medical History are captured for 15 body regions/disease area's. If overall question for the body region/disease is "yes", more detailed info is requested and captured in the sub-domains. This variable is used for documenting developmental diseases that are not specified elsewhere. | 0=No;1=Yes |
| MedHxDevelopmentalOtherTxt | TRUE | FALSE | text | Details on Medical History are captured for 15 body regions/disease area's. If overall question for the body region/disease is "yes", more detailed info is requested and captured in the sub-domains. This variable is used for documenting developmental diseases that are not specified elsewhere (textfield). | Not Applicable |
| MedHxEndocrine | TRUE | FALSE | integer | Details on Medical History are captured for 15 body regions/disease area's. If overall question for the body region/disease is "yes", more detailed info is requested and captured in the sub-domains. This variable is used for documenting endocrine diseases | 0=No;1=Yes;88=Unknown |
| MedHxEndocrineIDDM | TRUE | FALSE | integer | Details on Medical History are captured for 15 body regions/disease area's. If overall question for the body region/disease is "yes", more detailed info is requested and captured in the sub-domains. This variable is used for documenting endocrine diseases, specifically IDDM (Insulin dependent diabetes mellitus) | 0=No;1=Yes |
| MedHxEndocrineIDDMControl | TRUE | FALSE | integer | Details on Medical History are captured for 15 body regions/disease area's. If overall question for the body region/disease is "yes", more detailed info is requested and captured in the sub-domains. This variable is used for documenting for endocrine diseases, specifically IDDM (Insulin dependent diabetes mellitus) - how well it is controlled. | 1=Well controlled;2=Difficult controlled;88=Unknown |
| MedHxEndocrineNIDDM | TRUE | FALSE | integer | Details on Medical History are captured for 15 body regions/disease area's. If overall question for the body region/disease is "yes", more detailed info is requested and captured in the sub-domains. This variable is used for documenting endocrine diseases, specifically NIDDM (Non-insulin dependent diabetes mellitus) | 0=No;1=Yes |
| MedHxEndocrineNIDDMControl | TRUE | FALSE | integer | Details on Medical History are captured for 15 body regions/disease area's. If overall question for the body region/disease is | 1=Well controlled;2=Difficult controlled;88=Unknown |

| Name | Static | Intervention | Format | Description | Possible Values |
|---|---|---|---|---|---|
| | | | | "yes", more detailed info is requested and captured in the sub-domains. This variable is used for documenting for endocrine diseases, specifically NIDDM (Non-insulin dependent diabetes mellitus), how well it is controlled. | |
| MedHxEndocrineOther | TRUE | FALSE | integer | Details on Medical History are captured for 15 body regions/disease area's. If overall question for the body region/disease is "yes", more detailed info is requested and captured in the sub-domains. This variable is used for documenting endocrine diseases not mentioned elsewhere. | 0=No;1=Yes |
| MedHxEndocrineOtherTxt | TRUE | FALSE | text | Details on Medical History are captured for 15 body regions/disease area's. If overall question for the body region/disease is "yes", more detailed info is requested and captured in the sub-domains. This variable is used for documenting endocrine diseases not specified elsewhere (textfield). | Not Applicable |
| MedHxEndocrineThyroid | TRUE | FALSE | integer | Details on Medical History are captured for 15 body regions/disease area's. If overall question for the body region/disease is "yes", more detailed info is requested and captured in the sub-domains. This variable is used for documenting endocrine diseases, specifically thyroid disorder. | 0=No;1=Yes |
| MedHxENT | TRUE | FALSE | integer | Details on Medical History are captured for 15 body regions/disease area's. If overall question for the body region/disease is "yes", more detailed info is requested and captured in the sub-domains. This variable is used for documenting ENT (Eye, Ear, Nose & Throat) disease | 0=No;1=Yes;88=Unknown |
| MedHxENTHearing | TRUE | FALSE | integer | Details on Medical History are captured for 15 body regions/disease area's. If overall question for the body region/disease is "yes", more detailed info is requested and captured in the sub-domains. This variable is used for documenting ENT (Eye, Ear, Nose & Throat) disease, specifically hearing deficits. | 0=No;1=Yes |
| MedHxENTOther | TRUE | FALSE | integer | Details on Medical History are captured for 15 body regions/disease area's. If overall question for the body region/disease is "yes", more detailed info is requested and captured in the sub-domains. This variable is used for documenting ENT (Eye, Ear, Nose & Throat) diseases not specified elsewhere. | 0=No;1=Yes |
| MedHxENTOtherTxt | TRUE | FALSE | text | Details on Medical History are captured for 15 body regions/disease area's. If overall question for the body region/disease is "yes", more detailed info is requested and captured in the sub-domains. This variable is used for documenting ENT (Eye, Ear, Nose & Throat) diseases not specified elsewhere (textfield). | Not Applicable |
| MedHxENTSinusitis | TRUE | FALSE | integer | Details on Medical History are captured for 15 body regions/disease area's. If overall question for the body region/disease is "yes", more detailed info is requested and captured in the sub-domains. This variable is used for documenting ENT (Eye, Ear, Nose & Throat) diseases, specifically sinusitis. | 0=No;1=Yes |
| MedHxENTVisionAbn | TRUE | FALSE | integer | Details on Medical History are captured for 15 body regions/disease area's. If overall question for the body region/disease is "yes", more detailed info is requested and captured in the sub-domains. This variable is used for documenting ENT (Eye, Ear, Nose & Throat) diseases, specifically vision. | 0=No;1=Yes |
| MedHxGastro | TRUE | FALSE | integer | Details on Medical History are captured for 15 body regions/disease area's. If overall question for the body region/disease is | 0=No;1=Yes;88=Unknown |

| Name | Static | Intervention | Format | Description | Possible Values |
|---|---|---|---|---|---|
| | | | | "yes", more detailed info is requested and captured in the sub-domains. This variable is used for documenting gastrointestinal disease. | |
| MedHxGastroGERD | TRUE | FALSE | integer | Details on Medical History are captured for 15 body regions/disease area's. If overall question for the body region/disease is "yes", more detailed info is requested and captured in the sub-domains. This variable is used for documenting gastrointestinal disease, specifically GERD (Gastroesophageal Reflux Disease). | 0=No;1=Yes |
| MedHxGastroGIBleed | TRUE | FALSE | integer | Details on Medical History are captured for 15 body regions/disease area's. If overall question for the body region/disease is "yes", more detailed info is requested and captured in the sub-domains. This variable is used for documenting gastrointestinal disease, specifically gastrointestinal bleeding. | 0=No;1=Yes |
| MedHxGastroIBS | TRUE | FALSE | integer | Details on Medical History are captured for 15 body regions/disease area's. If overall question for the body region/disease is "yes", more detailed info is requested and captured in the sub-domains. This variable is used for documenting gastrointestinal disease, specifically inflammatory bowel disease. | 0=No;1=Yes |
| MedHxGastroOther | TRUE | FALSE | integer | Details on Medical History are captured for 15 body regions/disease area's. If overall question for the body region/disease is "yes", more detailed info is requested and captured in the sub-domains. This variable is used for documenting gastrointestinal disease not specified elsewhere. | 0=No;1=Yes |
| MedHxGastroOtherTxt | TRUE | FALSE | text | Details on Medical History are captured for 15 body regions/disease area's. If overall question for the body region/disease is "yes", more detailed info is requested and captured in the sub-domains. This variable is used for documenting gastrointestinal disease not specified elsewhere (textfield). | Not Applicable |
| MedHxHematologic | TRUE | FALSE | integer | Details on Medical History are captured for 15 body regions/disease area's. If overall question for the body region/disease is "yes", more detailed info is requested and captured in the sub-domains. This variable is used for documenting hematologic diseases. | 0=No;1=Yes;88=Unknown |
| MedHxHematologicAIDS | TRUE | FALSE | integer | Details on Medical History are captured for 15 body regions/disease area's. If overall question for the body region/disease is "yes", more detailed info is requested and captured in the sub-domains. This variable is used for documenting hematologic diseases, specifically AIDS | 0=No;1=Yes |
| MedHxHematologicAnemia | TRUE | FALSE | integer | Details on Medical History are captured for 15 body regions/disease area's. If overall question for the body region/disease is "yes", more detailed info is requested and captured in the sub-domains. This variable is used for documenting hematologic diseases, like anemia. | 0=No;1=Yes |
| MedHxHematologicHIV | TRUE | FALSE | integer | Details on Medical History are captured for 15 body regions/disease area's. If overall question for the body region/disease is "yes", more detailed info is requested and captured in the sub-domains. This variable is used for documenting hematologic diseases, like HIV positive. | 0=No;1=Yes |
| MedHxHematologicOther | TRUE | FALSE | integer | Details on Medical History are captured for 15 body regions/disease area's. If overall question for the body region/disease is "yes", more detailed info is requested and captured in the sub-domains. This variable is | 0=No;1=Yes |

| Name | Static | Intervention | Format | Description | Possible Values |
| --- | --- | --- | --- | --- | --- |
| | | | | used for documenting hematologic diseases, not specified elsewhere. | |
| MedHxHematologicOtherTxt | TRUE | FALSE | text | Details on Medical History are captured for 15 body regions/disease area's. If overall question for the body region/disease is "yes", more detailed info is requested and captured in the sub-domains. This variable is used for documenting hematologic diseases, not specified elsewhere (textfield). | Not Applicable |
| MedHxHematologicSickleCell | TRUE | FALSE | integer | Details on Medical History are captured for 15 body regions/disease area's. If overall question for the body region/disease is "yes", more detailed info is requested and captured in the sub-domains. This variable is used for documenting hematologic diseases, specifically sickle cell disease. | 0=No;1=Yes |
| MedHxHepatic | TRUE | FALSE | integer | Details on Medical History are captured for 15 body regions/disease area's. If overall question for the body region/disease is "yes", more detailed info is requested and captured in the sub-domains. This variable is used for documenting hepatic diseases. | 0=No;1=Yes;88=Unknown |
| MedHxHepaticCirrhosis | TRUE | FALSE | integer | Details on Medical History are captured for 15 body regions/disease area's. If overall question for the body region/disease is "yes", more detailed info is requested and captured in the sub-domains. This variable is used for documenting hepatic diseases, specifically cirrhosis. | 0=No;1=Yes |
| MedHxHepaticFailure | TRUE | FALSE | integer | Details on Medical History are captured for 15 body regions/disease area's. If overall question for the body region/disease is "yes", more detailed info is requested and captured in the sub-domains. This variable is used for documenting hepatic diseases, specifically hepatic failure | 0=No;1=Yes |
| MedHxHepaticHepatitis | TRUE | FALSE | integer | Details on Medical History are captured for 15 body regions/disease area's. If overall question for the body region/disease is "yes", more detailed info is requested and captured in the sub-domains. This variable is used for documenting hepatic diseases, specifically hepatitis. | 0=No;1=Yes |
| MedHxHepaticInsufficiency | TRUE | FALSE | integer | Details on Medical History are captured for 15 body regions/disease area's. If overall question for the body region/disease is "yes", more detailed info is requested and captured in the sub-domains. This variable is used for documenting hepatic diseases, specifically hepatic insufficiency. | 0=No;1=Yes |
| MedHxHepaticOther | TRUE | FALSE | integer | Details on Medical History are captured for 15 body regions/disease area's. If overall question for the body region/disease is "yes", more detailed info is requested and captured in the sub-domains. This variable is used for documenting hepatic diseases not specified elsewhere | 0=No;1=Yes |
| MedHxHepaticOtherTxt | TRUE | FALSE | text | Details on Medical History are captured for 15 body regions/disease area's. If overall question for the body region/disease is "yes", more detailed info is requested and captured in the sub-domains. This variable is used for documenting hepatic diseases not specified elsewhere (textfield) | Not Applicable |
| MedHxMusculoskeletal | TRUE | FALSE | integer | Details on Medical History are captured for 15 body regions/disease area's. If overall question for the body region/disease is "yes", more detailed info is requested and captured in the sub-domains. This variable is used for documenting musculoskeletal diseases | 0=No;1=Yes;88=Unknown |
| MedHxMusculoskeletalArthritis | TRUE | FALSE | integer | Details on Medical History are captured for 15 body regions/disease area's. If overall question for the body region/disease is | 0=No;1=Yes |

| Name | Static | Intervention | Format | Description | Possible Values |
|---|---|---|---|---|---|
| | | | | "yes", more detailed info is requested and captured in the sub-domains. This variable is used for documenting musculoskeletal diseases, specifically arthritis | |
| MedHxMusculoskeletalOther | TRUE | FALSE | integer | Details on Medical History are captured for 15 body regions/disease area's. If overall question for the body region/disease is "yes", more detailed info is requested and captured in the sub-domains. This variable is used for documenting musculoskeletal diseases not specified elsewhere. | 0=No;1=Yes |
| MedHxMusculoskeletalOtherTxt | TRUE | FALSE | text | Details on Medical History are captured for 15 body regions/disease area's. If overall question for the body region/disease is "yes", more detailed info is requested and captured in the sub-domains. This variable is used for documenting musculoskeletal diseases not specified elsewhere (textfield) | Not Applicable |
| MedHxNeuro | TRUE | FALSE | integer | Details on Medical History are captured for 15 body regions/disease area's. If overall question for the body region/disease is "yes", more detailed info is requested and captured in the sub-domains. This variable is used for documenting neurological diseases. | 0=No;1=Yes;88=Unknown |
| MedHxNeuroCerebrovascularAccident | TRUE | FALSE | integer | Details on Medical History are captured for 15 body regions/disease area's. If overall question for the body region/disease is "yes", more detailed info is requested and captured in the sub-domains. This variable is used for documenting neurological diseases, specifically cerebrovascular accidents. | 0=No;1=Yes |
| MedHxNeuroEpilepsyGeneralized | TRUE | FALSE | integer | Details on Medical History are captured for 15 body regions/disease area's. If overall question for the body region/disease is "yes", more detailed info is requested and captured in the sub-domains. This variable is used for documenting neurological diseases, specifically epilepsy (generalized). | 0=No;1=Yes |
| MedHxNeuroEpilepsyOther | TRUE | FALSE | integer | Details on Medical History are captured for 15 body regions/disease area's. If overall question for the body region/disease is "yes", more detailed info is requested and captured in the sub-domains. This variable is used for documenting neurological diseases, specifically epilepsy (other). | 0=No;1=Yes |
| MedHxNeuroEpilepsyPartial | TRUE | FALSE | integer | Details on Medical History are captured for 15 body regions/disease area's. If overall question for the body region/disease is "yes", more detailed info is requested and captured in the sub-domains. This variable is used for documenting neurological diseases, specifically epilepsy (partial). | 0=No;1=Yes |
| MedHxNeuroFebrileSeizures | TRUE | FALSE | integer | Details on Medical History are captured for 15 body regions/disease area's. If overall question for the body region/disease is "yes", more detailed info is requested and captured in the sub-domains. This variable is used for documenting neurological diseases, specifically febrile seizures (children). | 0=No;1=Yes |
| MedHxNeuroHeadache | TRUE | FALSE | integer | Details on Medical History are captured for 15 body regions/disease area's. If overall question for the body region/disease is "yes", more detailed info is requested and captured in the sub-domains. This variable is used for documenting neurological diseases, specifically headache (non migraine). | 0=No;1=Yes |
| MedHxNeuroMigraine | TRUE | FALSE | integer | Details on Medical History are captured for 15 body regions/disease area's. If overall question for the body region/disease is "yes", more detailed info is requested and captured in the sub-domains. This variable is used for documenting neurological diseases, specifically migraines. | 0=No;1=Yes |

| Name | Static | Intervention | Format | Description | Possible Values |
|---|---|---|---|---|---|
| MedHxNeuroMigraineFamHist | TRUE | FALSE | integer | Details on Medical History are captured for 15 body regions/disease area's. If overall question for the body region/disease is "yes", more detailed info is requested and captured in the sub-domains. This variable is used for documenting neurological diseases, specifically family history of migraine. | 0=No;1=Yes |
| MedHxNeuroOther | TRUE | FALSE | integer | Details on Medical History are captured for 15 body regions/disease area's. If overall question for the body region/disease is "yes", more detailed info is requested and captured in the sub-domains. This variable is used for documenting neurological diseases not specified elsewhere. | 0=No;1=Yes |
| MedHxNeuroOtherTxt | TRUE | FALSE | text | Details on Medical History are captured for 15 body regions/disease area's. If overall question for the body region/disease is "yes", more detailed info is requested and captured in the sub-domains. This variable is used for documenting neurological diseases not specified elsewhere (textfield) | Not Applicable |
| MedHxNeuroPain | TRUE | FALSE | integer | Details on Medical History are captured for 15 body regions/disease area's. If overall question for the body region/disease is "yes", more detailed info is requested and captured in the sub-domains. This variable is used for documenting neurological diseases not specified elsewhere (textfield) | 0=No;1=Yes;88=Unknown |
| MedHxNeuroTIA | TRUE | FALSE | integer | Details on Medical History are captured for 15 body regions/disease area's. If overall question for the body region/disease is "yes", more detailed info is requested and captured in the sub-domains. This variable is used for documenting neurological diseases, specifically transient ischemic attacks | 0=No;1=Yes |
| MedHxOncologic | TRUE | FALSE | integer | Details on Medical History are captured for 15 body regions/disease area's. If overall question for the body region/disease is "yes", more detailed info is requested and captured in the sub-domains. This variable is used for documenting oncologic diseases. | 0=No;1=Yes;88=Unknown |
| MedHxOncologicBreast | TRUE | FALSE | integer | Details on Medical History are captured for 15 body regions/disease area's. If overall question for the body region/disease is "yes", more detailed info is requested and captured in the sub-domains. This variable is used for documenting oncologic diseases, like breast cancer. | 0=No;1=Yes |
| MedHxOncologicGI | TRUE | FALSE | integer | Details on Medical History are captured for 15 body regions/disease area's. If overall question for the body region/disease is "yes", more detailed info is requested and captured in the sub-domains. This variable is used for documenting oncologic diseases, specifically gastrointestinal cancer. | 0=No;1=Yes |
| MedHxOncologicKidney | TRUE | FALSE | integer | Details on Medical History are captured for 15 body regions/disease area's. If overall question for the body region/disease is "yes", more detailed info is requested and captured in the sub-domains. This variable is used for documenting oncologic diseases, specifically kidney cancer. | 0=No;1=Yes |
| MedHxOncologicLeukemia | TRUE | FALSE | integer | Details on Medical History are captured for 15 body regions/disease area's. If overall question for the body region/disease is "yes", more detailed info is requested and captured in the sub-domains. This variable is used for documenting oncologic diseases, specifically leukemia. | 0=No;1=Yes |
| MedHxOncologicLung | TRUE | FALSE | integer | Details on Medical History are captured for 15 body regions/disease area's. If overall question for the body region/disease is "yes", more detailed info is requested and captured in the sub-domains. This variable is | 0=No;1=Yes |

| Name | Static | Intervention | Format | Description | Possible Values |
|---|---|---|---|---|---|
| | | | | used for documenting oncologic diseases, specifically lung cancer. | |
| MedHxOncologicLymphoma | TRUE | FALSE | integer | Details on Medical History are captured for 15 body regions/disease area's. If overall question for the body region/disease is "yes", more detailed info is requested and captured in the sub-domains. This variable is used for documenting oncologic diseases, specifically lymphoma. | 0=No;1=Yes |
| MedHxOncologicOther | TRUE | FALSE | integer | Details on Medical History are captured for 15 body regions/disease area's. If overall question for the body region/disease is "yes", more detailed info is requested and captured in the sub-domains. This variable is used for documenting oncologic diseases not specified elsewhere | 0=No;1=Yes |
| MedHxOncologicOtherTxt | TRUE | FALSE | text | Details on Medical History are captured for 15 body regions/disease area's. If overall question for the body region/disease is "yes", more detailed info is requested and captured in the sub-domains. This variable is used for documenting oncologic diseases not specified elsewhere (textfield). | Not Applicable |
| MedHxOncologicProstate | TRUE | FALSE | integer | Details on Medical History are captured for 15 body regions/disease area's. If overall question for the body region/disease is "yes", more detailed info is requested and captured in the sub-domains. This variable is used for documenting oncologic, specifically prostate cancer. | 0=No;1=Yes |
| MedHxOther | TRUE | FALSE | integer | Details on Medical History are captured for 15 body regions/disease area's. If overall question for the body region/disease is "yes", more detailed info is requested and captured in the sub-domains. This variable is used for documenting other medical history, not specified elsewhere. | 0=No;1=Yes |
| MedHxOtherTxt | TRUE | FALSE | text | Details on Medical History are captured for 15 body regions/disease area's. If overall question for the body region/disease is "yes", more detailed info is requested and captured in the sub-domains. This variable is used for documenting other medical history, not specified elsewhere (textfield) | Not Applicable |
| MedHxPreInjASAPSClass | TRUE | FALSE | integer | Preinjury ASAPS classification. Common classification system used in anaesthesia; denotes overall health | 1=A normal healthy patient;2=A patient with mild systemic disease;3=A patient with severe systemic disease;4=A patient with a severe systemic disease that is a constant threat to life;88=Unknown |
| MedHxPreTBIConcussions | TRUE | FALSE | integer | Details on Medical History are captured for 15 body regions/disease area's. If overall question for the body region/disease is "yes", more detailed info is requested and captured in the sub-domains. This variable is used for documenting previous TBI/ concussions | 0=No;1=Yes;88=Unknown |
| MedHxPreTBIConcussionsTotal | TRUE | FALSE | integer | Details on Medical History are captured for 15 body regions/disease area's. If overall question for the body region/disease is "yes", more detailed info is requested and captured in the sub-domains. This variable is used for documenting total number of previous TBI/ concussions | Not Applicable |
| MedHxPreTBIConcussionsTotalHosAdmit | TRUE | FALSE | integer | Details on Medical History are captured for 15 body regions/disease area's. If overall question for the body region/disease is "yes", more detailed info is requested and captured in the sub-domains. This variable is used for documenting total number of hospital admissions for previous TBI/ concussions | Not Applicable |
| MedHxPsychiatric | TRUE | FALSE | integer | Details on Medical History are captured for 15 body regions/disease area's. If overall question for the body region/disease is "yes", more detailed info is requested and | 0=No;1=Yes;88=Unknown |

| Name | Static | Intervention | Format | Description | Possible Values |
|---|---|---|---|---|---|
| | | | | captured in the sub-domains. This variable is used for documenting psychiatric diseases. | |
| MedHxPsychiatricAnx | TRUE | FALSE | integer | Details on Medical History are captured for 15 body regions/disease area's. If overall question for the body region/disease is "yes", more detailed info is requested and captured in the sub-domains. This variable is used for documenting psychiatric diseases, specifically anxiety. | 0=No;1=Yes |
| MedHxPsychiatricDep | TRUE | FALSE | integer | Details on Medical History are captured for 15 body regions/disease area's. If overall question for the body region/disease is "yes", more detailed info is requested and captured in the sub-domains. This variable is used for documenting psychiatric diseases, specifically depression. | 0=No;1=Yes |
| MedHxPsychiatricOther | TRUE | FALSE | integer | Details on Medical History are captured for 15 body regions/disease area's. If overall question for the body region/disease is "yes", more detailed info is requested and captured in the sub-domains. This variable is used for documenting psychiatric diseases not documented elsewhere. | 0=No;1=Yes |
| MedHxPsychiatricOtherTxt | TRUE | FALSE | text | Details on Medical History are captured for 15 body regions/disease area's. If overall question for the body region/disease is "yes", more detailed info is requested and captured in the sub-domains. This variable is used for documenting psychiatric diseases not documented elsewhere (textfield). | Not Applicable |
| MedHxPsychiatricSchiz | TRUE | FALSE | integer | Details on Medical History are captured for 15 body regions/disease area's. If overall question for the body region/disease is "yes", more detailed info is requested and captured in the sub-domains. This variable is used for documenting psychiatric diseases, specifically schizophrenia. | 0=No;1=Yes |
| MedHxPsychiatricSleep | TRUE | FALSE | integer | Details on Medical History are captured for 15 body regions/disease area's. If overall question for the body region/disease is "yes", more detailed info is requested and captured in the sub-domains. This variable is used for documenting psychiatric diseases, specifically sleep disorders. | 0=No;1=Yes |
| MedHxPsychiatricSubstanceAbuse | TRUE | FALSE | integer | Details on Medical History are captured for 15 body regions/disease area's. If overall question for the body region/disease is "yes", more detailed info is requested and captured in the sub-domains. This variable is used for documenting psychiatric diseases, specifically substance abuse disorders. | 0=No;1=Yes |
| MedHxPulmonary | TRUE | FALSE | integer | Details on Medical History are captured for 15 body regions/disease area's. If overall question for the body region/disease is "yes", more detailed info is requested and captured in the sub-domains. This variable is used for documenting pulmonary diseases | 0=No;1=Yes;88=Unknown |
| MedHxPulmonaryAsthma | TRUE | FALSE | integer | Details on Medical History are captured for 15 body regions/disease area's. If overall question for the body region/disease is "yes", more detailed info is requested and captured in the sub-domains. This variable is used for documenting pulmonary diseases, specifically asthma. | 0=No;1=Yes |
| MedHxPulmonaryCOPD | TRUE | FALSE | integer | Details on Medical History are captured for 15 body regions/disease area's. If overall question for the body region/disease is "yes", more detailed info is requested and captured in the sub-domains. This variable is used for documenting pulmonary diseases, specifically COPD (Chronic Obstructive Pulmonary Disease) | 0=No;1=Yes |
| MedHxPulmonaryOther | TRUE | FALSE | integer | Details on Medical History are captured for 15 body regions/disease area's. If overall | 0=No;1=Yes |

| Name | Static | Intervention | Format | Description | Possible Values |
|---|---|---|---|---|---|
| | | | | question for the body region/disease is "yes", more detailed info is requested and captured in the sub-domains. This variable is used for documenting pulmonary diseases not specified elsewhere. | |
| MedHxPulmonaryOtherTxt | TRUE | FALSE | text | Details on Medical History are captured for 15 body regions/disease area's. If overall question for the body region/disease is "yes", more detailed info is requested and captured in the sub-domains. This variable is used for documenting pulmonary diseases not specified elsewhere (textfield). | Not Applicable |
| MedHxPulmonaryPneumonia | TRUE | FALSE | integer | Details on Medical History are captured for 15 body regions/disease area's. If overall question for the body region/disease is "yes", more detailed info is requested and captured in the sub-domains. This variable is used for documenting pulmonary diseases, specifically pneumonia. | 0=No;1=Yes |
| MedHxPulmonaryTB | TRUE | FALSE | integer | Details on Medical History are captured for 15 body regions/disease area's. If overall question for the body region/disease is "yes", more detailed info is requested and captured in the sub-domains. This variable is used for documenting pulmonary diseases, specifically tuberculosis. | 0=No;1=Yes |
| MedHxRenal | TRUE | FALSE | integer | Details on Medical History are captured for 15 body regions/disease area's. If overall question for the body region/disease is "yes", more detailed info is requested and captured in the sub-domains. This variable is used for documenting renal diseases. | 0=No;1=Yes;88=Unknown |
| MedHxRenalFailure | TRUE | FALSE | integer | Details on Medical History are captured for 15 body regions/disease area's. If overall question for the body region/disease is "yes", more detailed info is requested and captured in the sub-domains. This variable is used for documenting renal diseases, specifically renal failure. | 0=No;1=Yes |
| MedHxRenalInsufficiency | TRUE | FALSE | integer | Details on Medical History are captured for 15 body regions/disease area's. If overall question for the body region/disease is "yes", more detailed info is requested and captured in the sub-domains. This variable is used for documenting renal diseases, specifically renal insufficiency. | 0=No;1=Yes |
| MedHxRenalOther | TRUE | FALSE | integer | Details on Medical History are captured for 15 body regions/disease area's. If overall question for the body region/disease is "yes", more detailed info is requested and captured in the sub-domains. This variable is used for documenting renal diseases not specified elsewhere. | 0=No;1=Yes |
| MedHxRenalOtherTxt | TRUE | FALSE | text | Details on Medical History are captured for 15 body regions/disease area's. If overall question for the body region/disease is "yes", more detailed info is requested and captured in the sub-domains. This variable is used for documenting renal diseases not specified elsewhere (textfield). | Not Applicable |
| MedHxRenalUTI | TRUE | FALSE | integer | Details on Medical History are captured for 15 body regions/disease area's. If overall question for the body region/disease is "yes", more detailed info is requested and captured in the sub-domains. This variable is used for documenting renal diseases, specifically chronic UTI (urinary tract infection). | 0=No;1=Yes |
| PlateletAggreOther | TRUE | FALSE | integer | Medical history. Variable documents the use of anticoagulants or platelet aggregation inhibitors. This variable contains the medication of this type not specified elsewhere. | 0=No;1=Yes |

| Name | Static | Intervention | Format | Description | Possible Values |
|---|---|---|---|---|---|
| PlateletAggreOtherTxt | TRUE | FALSE | text | Medical history. Variable documents the use of anticoagulants or platelet aggregation inhibitors. This variable contains the medication of this type not specified elsewhere (textfield). | Not Applicable |
| PltAggregAdenosineInhib | TRUE | FALSE | integer | Medical history. Use of platelet aggregation inhibitor by the patient. This variable describes the use of an adenosine reuptake inhibitor (eg. Persantin, dipyridamole). | 0=No;1=Yes |
| PltAggregADPReceptInhib | TRUE | FALSE | integer | Medical history. Use of platelet aggregation inhibitor by the patient. This variable describes the use of ADP receptor inhibitors. | 0=No;1=Yes |
| PltAggregADPReceptInhibEffient | TRUE | FALSE | integer | Medical history. Use of platelet aggregation inhibitor by the patient. This variable describes the use of Parasugrel (Effient) | 0=No;1=Yes |
| PltAggregADPReceptInhibOther | TRUE | FALSE | integer | Medical history. Use of platelet aggregation inhibitor by the patient. This variable describes the use of ADP receptor inhibitors, not specified elsewhere. | 0=No;1=Yes |
| PltAggregADPReceptInhibOtherTxt | TRUE | FALSE | text | Medical history. Use of platelet aggregation inhibitor by the patient. This variable describes the use of ADP receptor inhibitors, not specified elsewhere (textfield). | Not Applicable |
| PltAggregADPReceptInhibPlavix | TRUE | FALSE | integer | Medical history. Use of platelet aggregation inhibitor by the patient. This variable describes the use of Clopidogrel (Plavix). | 0=No;1=Yes |
| PltAggregADPReceptInhibTiclid | TRUE | FALSE | integer | Medical history. Use of platelet aggregation inhibitor by the patient. This variable describes the use of Ticlopidine (Ticlid). | 0=No;1=Yes |
| PltAggregAspirin | TRUE | FALSE | integer | Medical history. Use of platelet aggregation inhibitor by the patient. This variable describes the use of Aspirin | 0=No;1=Yes |
| PltAggregGlycoproteinInhib | TRUE | FALSE | integer | Medical history. Use of platelet aggregation inhibitor by the patient. This variable describes the use of glycoprotein IIB/IIIA inhibitors (eg. Aggrastat). | 0=No;1=Yes |
| PmMedicalCode | TRUE | FALSE | integer | Medical History of patients has been recorded under MedHx.MedHx* A number of predefined medical conditions were coded in the e-CRF. If the patient was taking medication for this predefined medical condition, the same code needed to be entered here. An option other was also available for conditions not listed. Please check MedHx.MedHx* for the information on the medical condition(s) corresponding to this variable. | Not Applicable |
| PmMedicationName | TRUE | FALSE | text | These fields capture information on medication taken pre-injury. For each medication taken, the corresponding Medical History code is listed to document the reason for which this med was taken. If the medication taken was not linked to one of the recorded medication history codes, Investigators were asked to enter "777" as related history code. | Not Applicable |
| PriorMeds | TRUE | FALSE | | | Not Applicable |
| SedativeCurrentUse | TRUE | FALSE | integer | On presentation the behavioral history of the patient was recorded. This reflects if in the past three months the subjects used sedatives or sleeping pills. | 0=No;1=Yes;88=Unknown |
| SedativePriorUse | TRUE | FALSE | integer | On presentation the behavioral history of the patient was recorded. This reflects his past use of Sedatives or sleeping pill. | 0=No;1=Yes;88=Unknown |
| SedativePriorUseDuration | TRUE | FALSE | text | On presentation the behavioral history of the patient was recorded. This reflects the number sof years of his past use sedatives (if applicable). | Not Applicable |
| TobcoCurntUseInd | TRUE | FALSE | integer | On presentation the behavioral history of the patient was recorded. This reflects his use in the past three months of Tobacco products (cigarettes, cigars, pipe, chewing tobacco, etc.) | 0=No;1=Yes;88=Unknown |

| Name | Static | Intervention | Format | Description | Possible Values |
|---|---|---|---|---|---|
| TobcoPriorUseInd | TRUE | FALSE | integer | The form "Behavioral History" captures information on past and current use of alcohol, tobacco, sedatives/sleeping pills, cannabis and other recreational drugs. Use is differentiated as "Past user" (eg stopped) versus "use in the past 3 months. Note: These variables do not reflect use of these substances at the time of injury. | 0=No;1=Yes;88=Unknown |
| TobcoUseDur | TRUE | FALSE | integer | On presentation the behavioral history of the patient was recorded. This reflects the number of years of his past use of Tobacco (if applicable). | Not Applicable |

*Injury characteristics and severity*

| Name | Static | Intervention | Format | Description | Possible Values |
|---|---|---|---|---|---|
| AbdomenPelvicContentsAIS | TRUE | FALSE | integer | AIS score for the Abdomen/Pelvic Contents In the original AIS classification of injury severity, the grading is from 1 (minor) to 6 (unsurvivable). We added a score of 0 to designate absence of injuries. | 0=None;1=Minor: no treatment needed;2=Moderate: requires only outpatient treatment;3=Serious: requires non-ICU hospital admission;4=Severe: requires ICU observation and/or basic treatment;5=Critical: requires intubation, mechanical ventilation or vasopressors for blood pressure support;6=Unsurvivable: not survivable |
| AbdomenPelvicContentsDesc | TRUE | FALSE | text | Injury description related to the AIS/ISS score for the Abdomen/Pelvic Contents | Not Applicable |
| BestOfAbdomenPelvicLumbarISS | TRUE | FALSE | integer | AbdomenPelvicLumbar region (Highest AIS of the region)^2 compare AbdomenPelvicContentsAIS, LumbarSpineAIS. This score is taken forward for ISS calculation | Not Applicable |
| BestOfChestSpineISS | TRUE | FALSE | integer | (highest AIS of the region)^2 Compare ThoraxChestAIS, ThoracicSpineAIS and select the highest for ISS calculation | Not Applicable |
| BestOfExternaISS | TRUE | FALSE | integer | External region (ExternaAIS)^2 select the highest external AIS severity code for ISS calculation. | Not Applicable |
| BestOfExtremitiesISS | TRUE | FALSE | integer | Extremities region (Highest AIS of the region)^2 compare UpperExtremitiesAIS, LowerExtremitiesAIS, PelvicGirdleAIS select the highest for ISS calculation | Not Applicable |
| BestOfFaceISS | TRUE | FALSE | integer | Face region (FaceAIS)^2 select the highest facial injury for ISS calculation | Not Applicable |
| BestOfHeadBrainCervicalISS | TRUE | FALSE | integer | HeadBrainCervical region (Highest AIS of the region)^2 Compare HeadNeckAIS, InjuryHx.BrainInjuryAIS, CervicalSpineAIS select the highest scoring injury in any of these 3 areas for ISS calculation | Not Applicable |
| BrainInjuryAIS | TRUE | FALSE | integer | AIS score for the Brain Injury In the original AIS classification of injury severity, the grading is from 1 (minor) to 6 (unsurvivable). We added a score of 0 to designate absence of injuries. | 0=None;1=Minor: no treatment needed;2=Moderate: requires only outpatient treatment;3=Serious: requires non-ICU hospital admission;4=Severe: requires ICU observation and/or basic treatment;5=Critical: requires intubation, mechanical ventilation or vasopressors for blood pressure support;6=Unsurvivable: not survivable |
| BrainInjuryDesc | TRUE | FALSE | text | Injury description related to the AIS/ISS score for the Brain Injury. Injury description is coded by drop-down menus for each body region | Not Applicable |
| CervicalSpineAIS | TRUE | FALSE | integer | AIS score for the Cervical Spine region. In the original AIS classification of injury severity, the grading is from 1 (minor) to 6 (unsurvivable). We added a score of 0 to designate absence of injuries. | 0=None;1=Minor: no treatment needed;2=Moderate: requires only outpatient treatment;3=Serious: requires non-ICU hospital admission;4=Severe: requires ICU observation and/or basic treatment;5=Critical: requires intubation, mechanical ventilation or vasopressors for blood pressure support;6=Unsurvivable: not survivable |
| CervicalSpineDesc | TRUE | FALSE | text | Injury description related to the AIS/ISS score for the Cervical Spine region. | Not Applicable |
| DayInjury | TRUE | FALSE | text | Day of injury. (Sunday - Saturday) | 0=Sunday;1=Monday;2=Tuesday;3=Wednesday;4=Thursday;5=Friday;6=Saturday |
| ExternaAIS | TRUE | FALSE | integer | AIS score for the External skin In the original AIS classification of injury severity, the grading is from 1 (minor) to 6 (unsurvivable). We added a score of 0 to designate absence of injuries. | 0=None;1=Minor: no treatment needed;2=Moderate: requires only outpatient treatment;3=Serious: requires non-ICU hospital admission;4=Severe: requires ICU observation and/or basic treatment;5=Critical: requires intubation, mechanical ventilation or vasopressors for blood pressure support;6=Unsurvivable: not survivable |

| Name | Static | Intervention | Format | Description | Possible Values |
| --- | --- | --- | --- | --- | --- |
| ExternaDesc | TRUE | FALSE | text | Injury description related to the AIS/ISS score for the Externa (skin) region. | Not Applicable |
| FaceAIS | TRUE | FALSE | integer | AIS score for Face (incl.maxillofacial) In the original AIS classification of injury severity, the grading is from 1 (minor) to 6 (unsurvivable). We added a score of 0 to designate absence of injuries. | 0=None;1=Minor: no treatment needed;2=Moderate: requires only outpatient treatment;3=Serious: requires non-ICU hospital admission;4=Severe: requires ICU observation and/or basic treatment;5=Critical: requires intubation, mechanical ventilation or vasopressors for blood pressure support;6=Unsurvivable: not survivable |
| FaceDesc | TRUE | FALSE | text | Injury description related to the AIS/ISS score for the Face (incl.maxillofacial) region | Not Applicable |
| HeadNeckAIS | TRUE | FALSE | integer | AIS score for the Head Neck region In the original AIS classification of injury severity, the grading is from 1 (minor) to 6 (unsurvivable). We added a score of 0 to designate absence of injuries. | 0=None;1=Minor: no treatment needed;2=Moderate: requires only outpatient treatment;3=Serious: requires non-ICU hospital admission;4=Severe: requires ICU observation and/or basic treatment;5=Critical: requires intubation, mechanical ventilation or vasopressors for blood pressure support;6=Unsurvivable: not survivable |
| HeadNeckDesc | TRUE | FALSE | text | Injury description for the Head and Neck AIS. | Not Applicable |
| InjAIS | TRUE | FALSE | integer | In the original AIS classification of injury severity, the grading is from 1 (minor) to 6 (unsurvivable). We added a score of 0 to designate absence of injuries. This is the AIS score for body regions as specified by AIS.InjBodyRegion. | 0=None;1=Minor: no treatment needed;2=Moderate: requires only outpatient treatment;3=Serious: requires non-ICU hospital admission;4=Severe: requires ICU observation and/or basic treatment;5=Critical: requires intubation, mechanical ventilation or vasopressors for blood pressure support;6=Unsurvivable: not survivable |
| InjArea | TRUE | FALSE | integer | Reflects the area where the injury took place (urban or rural). | 1=Urban (city);2=Rural;88=Unknown |
| InjBodyRegion | TRUE | FALSE | integer | Injuries and their severity are recorded according to a (modification of) the AIS. The AIS recognizes 6 main body regions. We included a further subdivision for some body regions, resulting in a total of 12 regions. For example, spine is not considered separately in the original AIS classification, but included under neck/chest and abdomen regions. For TBI, however, we considered it important to record spine separately. | 1=Externa;2=Head and Neck-Other;3=Brain Injury;4=Cervical Spine;5=Face;6=Thorax/Chest;7=Thoracic Spine;8=Abdomen/Pelvic Contents;9=Lumbar Spine;10=Upper Extremities;11=Lower Extremities;12=Pelvic Girdle |
| InjCause | TRUE | FALSE | integer | Reflects the cause of injury. | 1=Road traffic incident;2=Incidental fall;3=Other non-intentional injury;4=Violence/assault;5=Act of mass violence;6=Suicide attempt;88=Unknown;99=Other |
| InjCauseOther | TRUE | FALSE | text | Reflects if the cause of injury was "other" than the pre-listed causes. See also InjuryHx.InjCause | Not Applicable |
| InjDescription | TRUE | FALSE | integer | List of body regions with 55 subcategories describing the injury. | 1=Brain Injury: Concussion;2=Brain Injury: Contusions;3=Brain Injury: EDH;4=Brain Injury: Diffuse Injury;5=Brain Injury: ASDH;6=Brain Injury: Other;7=Head and Neck-Other: Specify in comments box;8=Cervical Spine: Fracture;9=Cervical Spine: Dislocation;10=Cervical Spine: Other;11=Face: Maxillo-facial fracture le Fort I;12=Face: Maxillo-facial fracture le Fort II;13=Face: Maxillo-facial fracture le Fort III;14=Face: Orbital fracture;15=Face: Zygomatic arch fracture;16=Face: Other;17=Thorax/Chest: Rib fracture;18=Thorax/Chest: Lung contusion;19=Thorax/Chest: Cardiac contusion;20=Thorax/Chest: Aorta dissection;21=Thorax/Chest: Pneumo-thorax;22=Thorax/Chest: Hemato-thorax;23=Thorax/Chest: Other;24=Thoracic Spine: Fracture;25=Thoracic Spine: Dislocation;26=Abdomen/Pelvic Contents: Spleen rupture;27=Abdomen/Pelvic Contents: Liver rupture;28=Abdomen/Pelvic Contents: Perforating abdominal injury;29=Abdomen/Pelvic Contents: Kidney contusion;30=Abdomen/Pelvic Contents: Retroperitoneal hematoma;31=Abdomen/Pelvic Contents: Other;32=Lumbar Spine: Fracture;33=Lumbar Spine: Dislocation;34=Lumbar Spine: Sacral fracture;35=Lumbar Spine: Other;36=Upper Extremities: Humerus fracture;37=Upper Extremities: Radial and/or ulnar fracture;38=Upper Extremities: Dislocation;39=Upper Extremities: Hand;40=Upper Extremities: Finger;41=Lower Extremities: Femoral fracture;42=Lower Extremities: Tibia plateau fracture;43=Lower Extremities: Tibia fracture;44=Lower Extremities: Ankle fracture;45=Lower Extremities: Calcaneus fracture;46=Lower Extremities: Metatarsal/tarsal fracture (toe fracture);47=Lower Extremities: Fibula fracture;48=Pelvic Girdle: Pelvic fracture;49=Pelvic Girdle: Hip fracture;50=Pelvic Girdle: Hip dislocation;51=Externa: Other;52=Thoracic Spine: Other;53=Upper |

| Name | Static | Intervention | Format | Description | Possible Values |
|---|---|---|---|---|---|
| | | | | | Extremities: Other;54=Lower Extremities: Other;55=Pelvic Girdle: Other |
| InjDesOther | TRUE | FALSE | text | Free text specifying the injury when AIS.InjDescription is "other" | Not Applicable |
| InjIndContactSportType | TRUE | FALSE | integer | Reflects the kind of contact sport involved as cause of injury - Only applicable for sports/recreational injuries | 1=Boxing;2=Martial Arts;99=Other |
| InjIndSportTypeOther | TRUE | FALSE | text | Reflects if the kind of contact sport involved as cause of injury was "other" than the pre-defined list - Only applicable for sports/recreational injuries. See also InjuryHx.InjIndContactSportType | Not Applicable |
| InjIntention | TRUE | FALSE | integer | Reflects if the cause on injury was intentional or unintentional. | 1=Intentional;2=Unintentional;3=Undetermined |
| InjMech1 | TRUE | FALSE | integer | Reflects the mechanism of injury - only applicable for Closed TBI | 1=High velocity trauma (acceleration/deceleration); 2=Direct impact: blow to head; 3=Direct impact: head against object; 6=Ground level fall; 7=Fall from height > 1 meter/5 stairs; 99=Other closed head injury; |
| InjMech2 | TRUE | FALSE | integer | Reflects the mechanism of injury - only applicable for Closed TBI | 1=High velocity trauma (acceleration/deceleration); 2=Direct impact: blow to head; 3=Direct impact: head against object; 6=Ground level fall; 7=Fall from height > 1 meter/5 stairs; 99=Other closed head injury; |
| InjMech3 | TRUE | FALSE | integer | Reflects the mechanism of injury - only applicable for Closed TBI | 1=High velocity trauma (acceleration/deceleration); 2=Direct impact: blow to head; 3=Direct impact: head against object; 6=Ground level fall; 7=Fall from height > 1 meter/5 stairs; 99=Other closed head injury; |
| InjMech6 | TRUE | FALSE | integer | Reflects the mechanism of injury - only applicable for Closed TBI | 1=High velocity trauma (acceleration/deceleration); 2=Direct impact: blow to head; 3=Direct impact: head against object; 6=Ground level fall; 7=Fall from height > 1 meter/5 stairs; 99=Other closed head injury; |
| InjMech7 | TRUE | FALSE | integer | Reflects the mechanism of injury - only applicable for Closed TBI | 1=High velocity trauma (acceleration/deceleration); 2=Direct impact: blow to head; 3=Direct impact: head against object; 6=Ground level fall; 7=Fall from height > 1 meter/5 stairs; 99=Other closed head injury; |
| InjMech99 | TRUE | FALSE | integer | Reflects the mechanism of injury - only applicable for Closed TBI | 1=High velocity trauma (acceleration/deceleration); 2=Direct impact: blow to head; 3=Direct impact: head against object; 6=Ground level fall; 7=Fall from height > 1 meter/5 stairs; 99=Other closed head injury; |
| InjMechOther | TRUE | FALSE | text | Reflects if the mechanism of injury was "other" than the pre-defined list - only applicable for Closed TBI. See also InjuryHx.InjMech | Not Applicable |
| InjOtherPartyInvolved | TRUE | FALSE | integer | Reflects that "another party involved in the cause of injury = N/A". | 0=No;1=Yes |
| InjOtherPartySleepingPills | TRUE | FALSE | integer | Reflects if sedatives or sleeping pills were involved in the cause of injury. | 0=No;1=Suspect;2=Definite;88=Unknown |
| InjPenetratingType | TRUE | FALSE | integer | Reflects the mechanism of injury - only applicable if Penetrating brain injury | 1=Gunshot wound;2=Fragment (incl. shell/shrapnel);99=Other penetrating brain injury |
| InjPenetratingTypeOther | TRUE | FALSE | text | Reflects if the mechanism of injury was other than the pre-defined list - only applicable if Penetrating brain injury. See also InjuryHx.InjPenetratingType | Not Applicable |
| InjPlace | TRUE | FALSE | integer | Reflects the place where the TBI injury occurred. | 1=Street/highway;2=Home/domestic;3=Work/school;4=Sport/Recreational;5=Military deployment;6=Public location (eg. bar, station;88=Unknown;99=Other |
| InjPlaceOther | TRUE | FALSE | text | Reflects if the place where the TBI injury occurred was "other" than the pre-defined list. See also InjuryHx.InjPlace | Not Applicable |
| InjRecSportType | TRUE | FALSE | integer | Reflects of the cause of injury was "Other Sport & Recreational Activities" - Only applicable for sports/recreational injuries | 1=Rollerblading/Skateboarding/Scootering;2=Skiing;3=Snowboarding;4=Hiking/Climbing;5=Horseriding;6=Golf;7=Cycling;8=Off-road vehicular sports;9=Water sports;10=Playground activity;88=Unknown;99=Other |
| InjRecSportTypeOther | TRUE | FALSE | text | Describes which was the cause of injury if "Other Sport & Recreational Activities" - Only applicable for sports/recreational injuries | Not Applicable |
| InjRoadAccEjectedFromVehicle | TRUE | FALSE | integer | Reflects if the subject was ejected from the vehicle - Only applicable if subject was motor vehicle occupant | 0=No;1=Yes;88=Unknown |
| InjRoadAccOtherParty | TRUE | FALSE | integer | Reflects if another party was involved in the Cause of Injury in case of a Road Traffic accident | 1=Motor vehicle;2=Pedestrian;3=Cyclist;4=Moped/Scooter;5=Tram/Bus;6=Train/Metro;7=Obstacle;10=Motor Bike;11=Lorry (camion);88=Unknown;99=Other |

| Name | Static | Intervention | Format | Description | Possible Values |
|---|---|---|---|---|---|
| InjRoadAccOtherPartyInvolved | TRUE | FALSE | integer | Describes if another party than the pre-defined list was involved in the Cause of Injury in case of a Road Traffic accident | 0=No;1=Yes;88=Unknown |
| InjRoadAccOtherPartyOther | TRUE | FALSE | text | Describes which other party than the pre-defined list was involved in the Cause of Injury in case of a Road Traffic accident | Not Applicable |
| InjRoadAccVictim | TRUE | FALSE | integer | Describes the type of victim in case of a Road traffic accident. | 1=Motor vehicle occupant;2=Pedestrian;3=Cyclist;4=Moped/Scooter;5=Motor Bike;99=Other |
| InjRoadAccVictimOther | TRUE | FALSE | text | Reflects if the type of victim was "other" than the predefined list in case of a Road traffic accident. | Not Applicable |
| InjRoadAccVictimVehiclePlace | TRUE | FALSE | integer | Reflects the occupant placement of the victim in the vehicle in case of a Road Traffic Accident | 1=Driver;2=Front seat passenger;3=Back seat passenger |
| InjSafetyAirbag | TRUE | FALSE | integer | Perfects if the airbag was deployed - Only applicable if subject was motor vehicle occupant | 0=No;1=Yes;77=Not Applicable;88=Unknown |
| InjSafetyHelmet | TRUE | FALSE | integer | Reflects if the victim was wearing a safety helmet. Only applicable in case of cyclist, scooter, motorbike incident. However, may also have been scored for various sports injuries. | 0=No;1=Yes;77=Not Applicable;88=Unknown |
| InjSafetySeatbelt | TRUE | FALSE | integer | Reflects if the victim was wearing a seat-belt. Only applicable if subject was motor vehicle occupant | 0=No;1=Yes;77=Not Applicable;88=Unknown |
| InjTeamSportType | TRUE | FALSE | integer | Reflects the type of team sport that was the cause of the injury - Only applicable for sports/recreational injuries | 1=Football (soccer);2=Rugby;3=Field Hockey;4=Ice Hockey;5=Lacrosse;99=Other |
| InjTeamSportTypeOther | TRUE | FALSE | text | Reflects if the type of team sport that was the cause of the injury was "other" than the predefined list- Only applicable for sports/recreational injuries | Not Applicable |
| InjToICUAdmHours | TRUE | FALSE | decimal | Time between injury and admission to ICU | 1=High velocity trauma (acceleration/deceleration); 2=Direct impact: blow to head; 3=Direct impact: head against object; 6=Ground level fall; 7=Fall from height > 1 meter/5 stairs; 99=Other closed head injury; |
| InjType | TRUE | FALSE | integer | Details of Injury are captured in 3 different variables: Type of Injury, Place of Injury and Mechanism of injury. This reflects the type of injury. | 1=Closed;2=Blast;3=Crush;5=Penetrating;6=Penetrating-perforating;7=Penetrating-tangential;8=Closed with open depressed skull fracture;88=Unknown |
| InjVictimAlcoholTestType | TRUE | FALSE | text | Reflects type of alcohol test used (breath test or blood test) for the victim | Blood=Blood Test;Breath=Breath Test |
| InjVictimBloodAlcoholmgdL | TRUE | FALSE | decimal | Reflects the level of mg/dL alcohol recorded in the victim during the alcohol test in case alcohol was related to the cause of injury. | Not Applicable |
| InjVictimBloodAlcoholpermil | TRUE | FALSE | decimal | Reflects the level of alcohol per mil (0/00) recorded in the victim during the alcohol test in case alcohol was related to the cause of injury. | Not Applicable |
| InjVictimDrugsTypeOther | TRUE | FALSE | text | Describes which other drugs where involved for the victim in the cause of injury. | Not Applicable |
| InjVictimSleepingPills | TRUE | FALSE | integer | Reflects if for the victim use of sedatives of sleeping pills were involved in the cause of injury. | 0=No;1=Suspect;2=Definite;88=Unknown |
| InjVictimTypeDrugs | TRUE | FALSE | integer | Rf elects which kind of drugs were involved in the cause of injury at the victims site. | 1=Cannabis;2=Cocaine;3=Methamphetamine's;4=Opioids;5=XTC;88=Unknown;99=Other |
| InjViolence | TRUE | FALSE | integer | Reflects the type of violence used as cause of injury - Only applicable if violence was the cause of injury | 1=Robbery;2=Interpersonal violence (fight);3=Domestic assault;4=Child abuse;5=Gang violence;6=Military deployment;88=Unknown;99=Other |
| InjViolenceOther | TRUE | FALSE | text | Reflects the "other" type of violence than the predefined list used as cause of injury - Only applicable if violence was the cause of injury | Not Applicable |
| InjViolenceOtherPartyAlcohol | TRUE | FALSE | integer | Information on drug and alcohol abuse of possible influence on the incident is different for victim versus "other party" (if involved) This reflects if alcohol was involved in the cause of injury for the other party involved | 0=No;1=Definite;2=Suspect;88=Unknown |
| InjViolenceOtherPartyDrugs | TRUE | FALSE | integer | Information on drug and alcohol abuse of possible influence on the incident is different for victim versus "other party" (if | 0=No;1=Definite;2=Suspect;88=Unknown |

| Name | Static | Intervention | Format | Description | Possible Values |
|---|---|---|---|---|---|
| | | | | involved). This reflects if drugs was involved as cause of injury for the other party involved | |
| InjViolenceVictimAlcohol | TRUE | FALSE | integer | Information on drug and alcohol abuse of possible influence on the incident is different for victim versus "other party" (if involved) This reflects alcohol involvement for the victim. | 0=No;1=Definite;2=Suspect;88=Unknown |
| InjViolenceVictimDrugs | TRUE | FALSE | integer | Information on drug and alcohol abuse of possible influence on the incident is different for victim versus "other party" (if involved) This reflects drugs involvement for the victim. | 0=No;1=Definite;2=Suspect;88=Unknown |
| LowerExtremitiesAIS | TRUE | FALSE | integer | AIS score for the Lower extremities. In the original AIS classification of injury severity, the grading is from 1 (minor) to 6 (unsurvivable). We added a score of 0 to designate absence of injuries. | 0=None;1=Minor: no treatment needed;2=Moderate: requires only outpatient treatment;3=Serious: requires non-ICU hospital admission;4=Severe: requires ICU observation and/or basic treatment;5=Critical: requires intubation, mechanical ventilation or vasopressors for blood pressure support;6=Unsurvivable: not survivable |
| LowerExtremitiesDesc | TRUE | FALSE | text | Injury Description for the AIS score for Lower extremities as subdomain of Extremities and pelvic girdle. | Not Applicable |
| LumbarSpineAIS | TRUE | FALSE | integer | AIS score for the Lumbar Spine region. In the original AIS classification of injury severity, the grading is from 1 (minor) to 6 (unsurvivable). We added a score of 0 to designate absence of injuries. | 0=None;1=Minor: no treatment needed;2=Moderate: requires only outpatient treatment;3=Serious: requires non-ICU hospital admission;4=Severe: requires ICU observation and/or basic treatment;5=Critical: requires intubation, mechanical ventilation or vasopressors for blood pressure support;6=Unsurvivable: not survivable |
| LumbarSpineDesc | TRUE | FALSE | text | Injury description for AIS score for Lumbar spine as subdomain of Abdomen/Pelvic contents. | Not Applicable |
| MonthInjury | TRUE | FALSE | integer | Month of Injury (January - December) | 1=January;2=February;3=March;4=April;5=May;6=June;7=July;8=August;9=September;10=October;11=November;12=December |
| PelvicGirdleAIS | TRUE | FALSE | integer | AIS score for the Pelvic Girdle region. In the original AIS classification of injury severity, the grading is from 1 (minor) to 6 (unsurvivable). We added a score of 0 to designate absence of injuries. | 0=None;1=Minor: no treatment needed;2=Moderate: requires only outpatient treatment;3=Serious: requires non-ICU hospital admission;4=Severe: requires ICU observation and/or basic treatment;5=Critical: requires intubation, mechanical ventilation or vasopressors for blood pressure support;6=Unsurvivable: not survivable |
| PelvicGirdleDesc | TRUE | FALSE | text | Injury description for AIS score for Pelvic Girdle as subdomain of Extremities and pelvic girdle. | Not Applicable |
| ThoracicSpineAIS | TRUE | FALSE | integer | AIS score for the Thoracic spine Region In the original AIS classification of injury severity, the grading is from 1 (minor) to 6 (unsurvivable). We added a score of 0 to designate absence of injuries. | 0=None;1=Minor: no treatment needed;2=Moderate: requires only outpatient treatment;3=Serious: requires non-ICU hospital admission;4=Severe: requires ICU observation and/or basic treatment;5=Critical: requires intubation, mechanical ventilation or vasopressors for blood pressure support;6=Unsurvivable: not survivable |
| ThoracicSpineDesc | TRUE | FALSE | text | Injury description for the AIS of Thoracic spine as subdomain of Thorax/Chest | Not Applicable |
| ThoraxChestAIS | TRUE | FALSE | integer | AIS score for the Thorax Chest region. In the original AIS classification of injury severity, the grading is from 1 (minor) to 6 (unsurvivable). We added a score of 0 to designate absence of injuries. | 0=None;1=Minor: no treatment needed;2=Moderate: requires only outpatient treatment;3=Serious: requires non-ICU hospital admission;4=Severe: requires ICU observation and/or basic treatment;5=Critical: requires intubation, mechanical ventilation or vasopressors for blood pressure support;6=Unsurvivable: not survivable |
| ThoraxChestDesc | TRUE | FALSE | text | Injury description for the AIS of the Thorax/Chest region. | Not Applicable |
| TotalISS | TRUE | FALSE | integer | The Injury Severity Score is calculated as the sum of the squares of the the 3 body regions with the highest AIS score. The max score for the ISS = 75. If any body region AIS is assigned a score of "6", the ISS is automatically set to 75 (highest score). In the calculation of the ISS, only the 6 main body regions are taken into consideration. | Not Applicable |
| UpperExtremitiesAIS | TRUE | FALSE | integer | AIS score for the Upper extremities. In the original AIS classification of injury severity, the grading is from 1 (minor) to 6 (unsurvivable). We added a score of 0 to designate absence of injuries. | 0=None;1=Minor: no treatment needed;2=Moderate: requires only outpatient treatment;3=Serious: requires non-ICU hospital admission;4=Severe: requires ICU observation and/or basic treatment;5=Critical: requires intubation, mechanical ventilation or vasopressors for blood pressure support;6=Unsurvivable: not survivable |

| Name | Static | Intervention | Format | Description | Possible Values |
|---|---|---|---|---|---|
| UpperExtremitiesDesc | TRUE | FALSE | text | Injury description for the AIS score of the Upper extremities as subdomain of Extremities and pelvic girdle. | Not Applicable |

*Emergency care and ICU admission*

| Name | Static | Intervention | Format | Description | Possible Values |
|---|---|---|---|---|---|
| ACEFocalNeuroDeficit | TRUE | FALSE | integer | On the neurological assessment at presentation overall rating was recorded on Focal neurological deficit (eg paresis or dysphasia) | 0=No;1=Yes;88=Unknown |
| ACEFocalNeuroDeficitDysphasia | TRUE | FALSE | integer | On the neurological assessment at presentation overall rating was recorded on Focal neurological deficit (eg paresis or dysphasia). If this was rate as "yes", details were recorded on whether it was paresis or dysphasia. | 0=No;1=Yes;88=Unknown |
| ACEFocalNeuroDeficitOther | TRUE | FALSE | integer | On the neurological assessment at presentation overall rating was recorded on Focal neurological deficit (eg paresis or dysphasia) If this was rate as "yes", details were recorded on whether it was paresis or dysphasia or other. | 0=No;1=Yes |
| ACEFocalNeuroDeficitOtherTxt | TRUE | FALSE | text | On the neurological assessment at presentation overall rating was recorded on Focal neurological deficit (eg paresis or dysphasia) If this was rate as "yes", details were recorded on whether it was paresis or dysphasia or other, and for other it was specified which one. | Not Applicable |
| ACEFocalNeuroDeficitParesis | TRUE | FALSE | integer | On the neurological assessment at presentation overall rating was recorded on Focal neurological deficit (eg paresis or dysphasia) If this was rate as "yes", details were recorded on whether it was paresis or dysphasia. | 0=No;1=Yes;88=Unknown |
| ACEOverallRating | TRUE | FALSE | integer | On the neurological assessment at presentation overall rating was recorded on "How different the person is acting compared to his/her usual self" The rating was on a scale from 1 (normal) to 6 (very different). | 1=1;2=2;3=3;4=4;5=5;6=6 |
| ACERatedBy | TRUE | FALSE | integer | On the neurological assessment at presentation overall rating was recorded on "How different the person is acting compared to his/her usual self" This variable reflects by whom the rating was performed. | 1=Proxy;2=Subject;3=Both proxy and subject;4=Not done |
| BaselineGCSMostReliableAssessmentCondition | TRUE | FALSE | integer | A baseline risk assessment was performed at the hospital (ER). This reflects the Conditions of assessment for the Most reliable Motor Score for risk assessment. | 0=No sedation/paralysis;1=Under sedation;3=After stopping sedation;4=After pharmacological reversal |
| BaselineGCSMostReliableAssessmentTime | TRUE | FALSE | integer | A baseline risk assessment was performed at the hospital (ER). This reflects the Time of assessment for the Most reliable Motor Score for risk assessment. | 1=Admission;2=Post-stabilization;3=First hospital;4=Scene of accident;5=Other |
| BaselineGCSMostReliableMotorScore | TRUE | FALSE | text | A baseline risk assessment was performed at the hospital (ER). This reflects the Most reliable baseline Motor score of the GCS as given by sites - for use in prognostic models. | 1=1-None;2=2-Abnormal extension;3=3-Abnormal flexion;4=4-Normal flexion/withdrawal;5=5-Localizes to pain;6=6-Obeys command;O=Untestable (Other);P=Untestable (Deep sedation/paralyzed);UN=Unknown |
| BaselineGOS6MoExpectedDeathRisk | TRUE | FALSE | text | At ER discharge, physician estimate of six month outcome was recorded as a baseline risk assessment: "Given all current available information, what is, in your subjective opinion, the most likely 6-month outcome of this patient? To be based upon information on discharge ER or admission to hospital/ICU". This reflects the Risk of death in % | Not Applicable |
| BaselineGOS6MoExpectedOutcome | TRUE | FALSE | text | At ER discharge, physician estimate of six month outcome was recorded as a baseline risk assessment: "Given all current available information, what is, in your subjective opinion, the most likely 6-month outcome of | D=D - Death;GR=GR - Good Recovery;MD=MD - Moderate Disability;SD=SD - Severe Disability;V=V - Vegetative State |

| Name | Static | Intervention | Format | Description | Possible Values |
|---|---|---|---|---|---|
| | | | | this patient? To be based upon information on discharge ER or admission to hospital/ICU". This reflects the Expected outcome (GOS) | |
| BaselineGOS6MoUnfavourableOutcomeRisk | TRUE | FALSE | text | At ER discharge, physician estimate of six month outcome was recorded as a baseline risk assessment: "Given all current available information, what is, in your subjective opinion, the most likely 6-month outcome of this patient? To be based upon information on discharge ER or admission to hospital/ICU". This reflects the Risk of unfavorable outcome (D, VS, SD) in % | Not Applicable |
| BaselinePhysEstOf6MoOutcomePhysicianQual | TRUE | FALSE | integer | At ER discharge, physician estimate of six month outcome was recorded as a baseline risk assessment: "Given all current available information, what is, in your subjective opinion, the most likely 6-month outcome of this patient? To be based upon information on discharge ER or admission to hospital/ICU". This reflects the qualification of the physician who provided prognostic estimate on ER discharge/admission to hospital/ICU | 1=Resident;2=Junior staff (< 5 years);3=Senior staff (>= 5 years);4=Head of department |
| BaselinePhysEstOf6MoOutcomePhysicianType | TRUE | FALSE | integer | At ER discharge, physician estimate of six month outcome was recorded as a baseline risk assessment: "Given all current available information, what is, in your subjective opinion, the most likely 6-month outcome of this patient? To be based upon information on discharge ER or admission to hospital/ICU". This reflects the type of the physician who provided prognostic estimate on ER discharge/admission to hospital/ICU | 1=ER Physician;2=Intensive Care;3=Neurology;4=Neurosurgery;5=Traumatology;88=Unknown |
| DispER | TRUE | FALSE | integer | Destination of the patient at ER discharge. | 1=Discharge home;2=Discharge other facility;3=Hospital admission--Ward;4=Hospital admission--Intermediate/high care unit;5=Hospital admission--ICU;6=Hospital admission--OR for immediate surgical procedure;7=Death;8=Hospital admission--Other (e.g. observation unit);88=Unknown |
| EDAirway | TRUE | FALSE | integer | Records treatment performed in the ER/on admission with regard to Airways. Pre-hospital inteventions are documented at: InjuryHx.PresEmergencyCare | 1=No specific treatment;2=Supplemental oxygen (via nasal tube or mask);3=Adjunctive airway (eg. Mayo tube);4=Temporary support with bag, valve;5=Intubation;6=Mechanical ventilation;88=Unknown |
| EDArrDBP | TRUE | FALSE | integer | Reflects Clinical Exam (on arrival at Study Center) --> BP (mmHg) --> Diastolic | Not Applicable |
| EDArrHR | TRUE | FALSE | integer | Reflects the vital signs at ER arrival: Heart rate --> Beats per min. | Not Applicable |
| EDArrivalAirway | TRUE | FALSE | integer | The ABC status on arrival documents the status of Airway, Breathing and Circulation upon arrival to Study Hospital (ER); In addition, administration of supplemental oxygen and spinal immobilization is documented. | 1=Clear;2=Obstructed;3=Adjunctive Airway;4=Intubated;88=Unknown |
| EDArrivalArtpCO2mmhg | TRUE | FALSE | decimal | Reflects the vital signs at ER arrival: Arterial pCO2 --> mmHg | Not Applicable |
| EDArrivalArtpO2mmhg | TRUE | FALSE | decimal | Reflects the vital signs at ER arrival: Arterial pO2 --> mmHg | Not Applicable |
| EDArrivalBaseExcess | TRUE | FALSE | decimal | Reflects the vital signs at ER arrival: Base excess --> mEq/l | Not Applicable |
| EDArrivalBMIKgCm | TRUE | FALSE | decimal | BMI using height in cm (EDArrivalHeightCm) and weight (EDArrivalBodyWeightKg) BMI= (Weight/(Height*Height))) * 10000 | Not Applicable |
| EDArrivalBodyWeightKg | TRUE | FALSE | decimal | Body weight in KG at ER arrival | Not Applicable |
| EDArrivalBodyWeightMeasured | TRUE | FALSE | integer | Provides an indication of accuracy of reported body weight at ER arrival | 1=Estimated;2=Self reported;3=Measured;4=Proxy reported;88=Unknown |
| EDArrivalBreathing | TRUE | FALSE | integer | The ABC status on arrival documents the status of Airway, Breathing and Circulation upon arrival to Study Hospital (ER); In addition, administration of supplemental oxygen and spinal immobilization is documented. | 1=Spontaneous, adequate;2=Spontaneous, insufficient;3=Manual support with bag, valve;4=Mechanical ventilation;88=Unknown |

| Name | Static | Intervention | Format | Description | Possible Values |
|---|---|---|---|---|---|
| EDArrivalCirculation | TRUE | FALSE | integer | The ABC status on arrival documents the status of Airway, Breathing and Circulation upon arrival to Study Hospital (ER); In addition, administration of supplemental oxygen and spinal immobilization is documented. | 0=No specific therapy;1=IV Fluids;2=Vasopressors;3=CPR;88=Unknown |
| EDArrivalFiO2 | TRUE | FALSE | integer | Reflects the vital signs at ER arrival: FiO2 (in %) Information on FiO2 (Fraction of Inspired oxygen) at time of arterial blood gas sampling is requested in order to be able to calculate PaO2/FiO2 as measure of severity of hypoxaemia; dependent on altitude; at sea-level, normal values are > 500 mmHg | Not Applicable |
| EDArrivalHeightCm | TRUE | FALSE | decimal | Reflects Height in cm at ER arrival | Not Applicable |
| EDArrivalHeightMeasured | TRUE | FALSE | integer | Provides an indication of accuracy of reported height at ER arrival | 1=Estimated;2=Self reported;3=Measured;4=Proxy reported;88=Unknown |
| EDArrivalLactate | TRUE | FALSE | decimal | Reflects the vital signs at ER arrival: Lactate --> mEq/l | Not Applicable |
| EDArrivalpH | TRUE | FALSE | decimal | Reflects the vital signs at ER arrival: pH | Not Applicable |
| EDArrivalSpinalImmob | TRUE | FALSE | integer | The ABC status on arrival documents the status of Airway, Breathing and Circulation upon arrival to Study Hospital (ER); In addition, administration of supplemental oxygen and spinal immobilization is documented. | 0=No;1=Yes;88=Unknown |
| EDArrivalSupplementalOxygen | TRUE | FALSE | integer | The ABC status on arrival documents the status of Airway, Breathing and Circulation upon arrival to Study Hospital (ER); In addition, administration of supplemental oxygen and spinal immobilization is documented. | 0=No;1=Yes;88=Unknown |
| EDArrPupilLftEyeMeasr | TRUE | FALSE | integer | Reflects Clinical Exam (on arrival at Study Center) --> Left eye Size | Not Applicable |
| EDArrPupilReactivityLghtLftEyeReslt | TRUE | FALSE | integer | Neurological assessment (GCS and pupils) was recorded for the scene of accident, the first hospital (if applicable), the Arrival to ER of the study hospital and post-stabilization. This reflects the reactivity of the left eye pupil for the assessment at Arrival to ER of the study hospital. | 1=+ (Brisk);2=+ (Sluggish);3=- (Negative) |
| EDArrPupilReactivityLghtRtEyeReslt | TRUE | FALSE | integer | Neurological assessment (GCS and pupils) was recorded for the scene of accident, the first hospital (if applicable), the Arrival to ER of the study hospital and post-stabilization. This reflects the reactivity of the right eye pupil for the assessment at Arrival to ER of the study hospital. | 1=+ (Brisk);2=+ (Sluggish);3=- (Negative) |
| EDArrPupilRtEyeMeasr | TRUE | FALSE | integer | Reflects Clinical Exam (on arrival at Study Center) --> Right eye size | Not Applicable |
| EDArrPupilSymmetry | TRUE | FALSE | integer | Neurological assessment (GCS and pupils) was recorded for the scene of accident, the first hospital (if applicable), the Arrival to ER of the study hospital and post-stabilization. This reflects the Pupil symmetry for the assessment at Arrival to ER of the study hospital. | 1=Equal;2=Unequal R>L;3=Unequal L>R;66=Untestable;88=Unknown |
| EDArrRespRate | TRUE | FALSE | integer | Reflects the vital signs at ER arrival: Respiratory rate --> cycles per min | Not Applicable |
| EDArrSBP | TRUE | FALSE | integer | Reflects Clinical Exam (on arrival at Study Center) --> BP (mmHg) --> Systolic | Not Applicable |
| EDArrSpO2 | TRUE | FALSE | decimal | Reflects Clinical Exam (on arrival at Study Center) --> Oxygen saturation (in %) | Not Applicable |
| EDArrTempCelsius | TRUE | FALSE | decimal | Reflects the vital signs at ER arrival: Temperature --> Celcius | Not Applicable |
| EDBloodGasConditions | TRUE | FALSE | integer | Reflects the vital signs at ER arrival: Conditions for First arterial blood gas done (if applicable) | 1=Pre-intubation, room air;2=Pre-intubation, +O2;3=Post-intubation, not ventilated;4=Post-intubation, ventilated |
| EDBloodTrans | TRUE | FALSE | integer | Reflects of blood transfusion was done in the ER of the study hospital | 0=No;1=Yes;88=Unknown |
| EDCircCPR | TRUE | FALSE | integer | Records treatment performed in the ER/on admission with regard to Circulation: CPR. Pre-hospital inteventions are documented at: InjuryHx.PresEmergencyCare | 0=No;1=Yes |

| Name | Static | Intervention | Format | Description | Possible Values |
|---|---|---|---|---|---|
| EDCircIV | TRUE | FALSE | integer | Records treatment performed in the ER/on admission with regard to Circulation: IV fluids. Pre-hospital inteventions are documented at: InjuryHx.PresEmergencyCare | 0=No;1=Yes |
| EDCircNone | TRUE | FALSE | integer | Records treatment performed in the ER/on admission with regard to Circulation: no specific treatment Pre-hospital inteventions are documented at: InjuryHx.PresCirculationTreatmentNone | 0=No;1=Yes |
| EDCircVaso | TRUE | FALSE | integer | Records treatment performed in the ER/on admission with regard to Circulation: vasopressors. Pre-hospital inteventions are documented at: InjuryHx.PresEmergencyCare | 0=No;1=Yes |
| EDCoagulopathyType1 | TRUE | FALSE | integer | Reflects the type of transfusion or coagulopathy treatment given in the ER of the study hospital (if applicable) | 1=Packed red blood cell concentrates (pRBCs);2=Fresh whole blood;3=Fresh frozen plasma (FFP);4=Freeze dried plasma / lypholized plasma;5=Platelet concentrates;6=PCC (prothrombin complex concentrates);7=Fibrinogen concentrate;8=Albumin;9=Recombinant factor FVIIa;10=Tranexamic acid (TXA);11=Cryoprecipitate;12=Desmopression (DDAVP);13=Factor XIII;14=Calcium;15=Vitamin K (Konakion) |
| EDCoagulopathyType2 | TRUE | FALSE | integer | Reflects the type of transfusion or coagulopathy treatment given in the ER of the study hospital (if applicable) | 1=Packed red blood cell concentrates (pRBCs);2=Fresh whole blood;3=Fresh frozen plasma (FFP);4=Freeze dried plasma / lypholized plasma;5=Platelet concentrates;6=PCC (prothrombin complex concentrates);7=Fibrinogen concentrate;8=Albumin;9=Recombinant factor FVIIa;10=Tranexamic acid (TXA);11=Cryoprecipitate;12=Desmopression (DDAVP);13=Factor XIII;14=Calcium;15=Vitamin K (Konakion) |
| EDCoagulopathyType3 | TRUE | FALSE | integer | Reflects the type of transfusion or coagulopathy treatment given in the ER of the study hospital (if applicable) | 1=Packed red blood cell concentrates (pRBCs);2=Fresh whole blood;3=Fresh frozen plasma (FFP);4=Freeze dried plasma / lypholized plasma;5=Platelet concentrates;6=PCC (prothrombin complex concentrates);7=Fibrinogen concentrate;8=Albumin;9=Recombinant factor FVIIa;10=Tranexamic acid (TXA);11=Cryoprecipitate;12=Desmopression (DDAVP);13=Factor XIII;14=Calcium;15=Vitamin K (Konakion) |
| EDCoagulopathyType4 | TRUE | FALSE | integer | Reflects the type of transfusion or coagulopathy treatment given in the ER of the study hospital (if applicable) | 1=Packed red blood cell concentrates (pRBCs);2=Fresh whole blood;3=Fresh frozen plasma (FFP);4=Freeze dried plasma / lypholized plasma;5=Platelet concentrates;6=PCC (prothrombin complex concentrates);7=Fibrinogen concentrate;8=Albumin;9=Recombinant factor FVIIa;10=Tranexamic acid (TXA);11=Cryoprecipitate;12=Desmopression (DDAVP);13=Factor XIII;14=Calcium;15=Vitamin K (Konakion) |
| EDCoagulopathyVolume1 | TRUE | FALSE | text | Reflects the volume of transfusion or coagulopathy treatment given in the ER of the study hospital (if applicable) | Not Applicable |
| EDCoagulopathyVolume2 | TRUE | FALSE | text | Reflects the volume of transfusion or coagulopathy treatment given in the ER of the study hospital (if applicable) | Not Applicable |
| EDCoagulopathyVolume3 | TRUE | FALSE | text | Reflects the volume of transfusion or coagulopathy treatment given in the ER of the study hospital (if applicable) | Not Applicable |
| EDCoagulopathyVolume4 | TRUE | FALSE | text | Reflects the volume of transfusion or coagulopathy treatment given in the ER of the study hospital (if applicable) | Not Applicable |
| EDCompEventHypothermia | TRUE | FALSE | integer | Second Insults reported here relate to the pre-hospital and ER phase. Hypothermia is defined as a documented core temperature of < 35 C. | 0=No;1=Definite;2=Suspect;88=Unknown |
| EDComplEventCardArr | TRUE | FALSE | integer | Second Insults reported here relate to the pre-hospital and ER phase: Cardiac Arrest | 0=No;1=Yes |
| EDComplEventHypotension | TRUE | FALSE | integer | Second Insults reported here relate to the pre-hospital and ER phase. Definite hypotension is defined as a documented systolic BP < 90 mm Hg (adults); "Suspected" was scored if the patient did not have a documented blood pressure, but was reported to be in shock or have an absent brachial pulse (not related to injury of the extremity) | 0=No;1=Definite;2=Suspect;88=Unknown |
| EDComplEventHypoxia | TRUE | FALSE | integer | Second Insults reported here relate to the pre-hospital and ER phase. Definite hypoxia is defined as a documented PaO2 <8 kPa (60 mm Hg) and/or SaO2<90%; "Suspected" | 0=No;1=Definite;2=Suspect;88=Unknown |

| Name | Static | Intervention | Format | Description | Possible Values |
|---|---|---|---|---|---|
| | | | | was scored if the patient did not have documented hypoxia by PaO2 or SaO2, but there was a clinical suspicion , as evidenced by for example cyanosis, apnoea or respiratory distress | |
| EDComplEventSeizures | TRUE | FALSE | integer | Second Insults reported here relate to the pre-hospital and ER phase: seizures | 0=No;1=Partial/Focal;2=Generalized;3=Status epilepticus;88=Unknown |
| EDCorrCoagulopathy | TRUE | FALSE | integer | Documents blood transfusions and treatment of coagulopathy in the acute phase at presentation. | 0=No;1=Yes;88=Unknown |
| EDDischPupilLftEyeMeasr | TRUE | FALSE | integer | Neurological assessment (GCS and pupils) was recorded for the scene of accident, the first hospital (if applicable), the Arrival to ER of the study hospital and post-stabilization. This reflects the size of the left eye pupil for the assessment POST¬≠-STABILIZATION. | Not Applicable |
| EDDischPupilReactivityLghtLftEyeReslt | TRUE | FALSE | integer | Neurological assessment (GCS and pupils) was recorded for the scene of accident, the first hospital (if applicable), the Arrival to ER of the study hospital and post-stabilization. This reflects the reactivity of the left eye pupil for the assessment POST¬≠-STABILIZATION. | 1=+ (Brisk);2=+ (Sluggish);3=- (Negative) |
| EDDischPupilReactivityLghtRtEyeReslt | TRUE | FALSE | integer | Neurological assessment (GCS and pupils) was recorded for the scene of accident, the first hospital (if applicable), the Arrival to ER of the study hospital and post-stabilization. This reflects the reactivity of the right eye pupil for the assessment POST¬≠-STABILIZATION. | 1=+ (Brisk);2=+ (Sluggish);3=- (Negative) |
| EDDischPupilRtEyeMeasr | TRUE | FALSE | integer | Neurological assessment (GCS and pupils) was recorded for the scene of accident, the first hospital (if applicable), the Arrival to ER of the study hospital and post-stabilization. This reflects the size of the right eye pupil for the assessment POST¬≠-STABILIZATION. | Not Applicable |
| EDDischPupilSymmetry | TRUE | FALSE | integer | Neurological assessment (GCS and pupils) was recorded for the scene of accident, the first hospital (if applicable), the Arrival to ER of the study hospital and post-stabilization. This reflects pupil symmetry for the assessment POST¬≠-STABILIZATION. | 1=Equal;2=Unequal R>L;3=Unequal L>R;66=Untestable;88=Unknown |
| EDICPMonitoring | TRUE | FALSE | integer | Scheduled for ICP monitoring; e.g. may not accurately reflect if ICP monitoring was indeed performed | 0=No;1=Yes;88=Unknown |
| EDIVAlbumin | TRUE | FALSE | integer | Records treatment performed in the ER/on admission with regard to Circulation: IV fluids --> Albumin | 0=No;1=Yes |
| EDIVBlood | TRUE | FALSE | integer | Records treatment performed in the ER/on admission with regard to Circulation: IV fluids --> Blood | 0=No;1=Yes |
| EDIVColloids | TRUE | FALSE | integer | Records treatment performed in the ER/on admission with regard to Circulation: IV fluids --> Colloids | 0=No;1=Yes |
| EDIVCrystalloids | TRUE | FALSE | integer | Records treatment performed in the ER/on admission with regard to Circulation: IV fluids --> Crystalloids | 0=No;1=Yes |
| EDIVMannitol | TRUE | FALSE | integer | Records treatment performed in the ER/on admission with regard to Circulation: IV fluids --> Mannitol | 0=No;1=Yes |
| EDIVSaline | TRUE | FALSE | integer | Records treatment performed in the ER/on admission with regard to Circulation: IV fluids --> Hypertonic saline | 0=No;1=Yes |
| EDSecondInsultsNeuroWorse | TRUE | FALSE | integer | The importance of Neuroworsening was first described by Morris and Marshall. The occurrence of neuroworsening is related to poorer outcome in subjects with moderate to severe TBI. Neuroworsening is defined as 1) a decrease in GCS motor score of 2 or more points; 2) a new loss of pupillary reactivity or development of pupillary assymmetry >= | 0=No;1=Yes;88=Unknown |

| Name | Static | Intervention | Format | Description | Possible Values |
|---|---|---|---|---|---|
| | | | | 2mm; 3) deterioration in neurological or CT status sufficient to warrant immediate medical or surgical intervention | |
| EDSecondInsultsNeuroWorseYes1 | TRUE | FALSE | integer | This variable provides a specification of the type of neuroworsening if it occurs. | 1=Decrease in motor score >= 2 points; 2=Development of pupillary abnormalities; 3=Other neurological and/or CT deterioration |
| EDSecondInsultsNeuroWorseYes2 | TRUE | FALSE | integer | This variable provides a specification of the type of neuroworsening if it occurs. | 1=Decrease in motor score >= 2 points; 2=Development of pupillary abnormalities; 3=Other neurological and/or CT deterioration |
| EDSecondInsultsNeuroWorseYes3 | TRUE | FALSE | integer | This variable provides a specification of the type of neuroworsening if it occurs. | 1=Decrease in motor score >= 2 points; 2=Development of pupillary abnormalities; 3=Other neurological and/or CT deterioration |
| EDSecondInsultsPreAdmisCourse | TRUE | FALSE | integer | The pre-admission course should only be considered an intracranial second insult in case of Deterioration. The nature of deterioration will in most cases be further detailed under the variable "Neuroworsening". | 0=Deterioration;1=Stable;2=Improving;88=Unknown |
| EDSpinalImmob | TRUE | FALSE | integer | Records treatment performed in the ER/on admission with regard to Spinal immobilization. Pre-hospital inteventions are documented at: InjuryHx.PresEmergencyCare | 0=No;1=Yes;88=Unknown |
| EmergSurgInterventionsExtraCran | TRUE | FALSE | integer | Documents emergency extracranial surgery performed in study hospital. Procedures performed in "first" hospital (in case of secondary referral) are documented at: InjuryHx.PresERextracranialSurg | 0=No;1=Yes;88=Unknown |
| EmergSurgInterventionsExtraCranYes | TRUE | FALSE | integer | Documents emergency extracranial surgery performed in study hospital. Procedures performed in "first" hospital (in case of secondary referral) are documented at: InjuryHx.PresERextracranialSurg | 1=Damage control thoracotomy;2=Damage control laparotomy;3=Extraperitoneal pelvic packing;4=External fixation limb;5=Cranio-maxillo-facial reconstruction;99=Other |
| EmergSurgInterventionsExtraCranYesOther | TRUE | FALSE | text | Documents emergency extracranial surgery performed in study hospital. Procedures performed in "first" hospital (in case of secondary referral) are documented at: InjuryHx.PresERextracranialSurg | Not Applicable |
| EmergSurgInterventionsIntraCran | TRUE | FALSE | integer | Documents emergency intracranial surgery performed in study hospital. Procedures performed in "first" hospital (in case of secondary referral) are documented at: InjuryHx.PresERIntracranialSurg | 0=No;1=Yes;88=Unknown |
| EmergSurgInterventionsIntraCranYes | TRUE | FALSE | integer | Documents emergency intracranial surgery performed in study hospital. Procedures performed in "first" hospital (in case of secondary referral) are documented at: InjuryHx.PresERIntracranialSurg | 1=Craniotomy for haematoma/contusion;2=Decompressive Craniectomy;3=Depressed skull fracture;99=Other intracranial procedure |
| EmerSurgIntraCranSurviveNoSurg | TRUE | FALSE | integer | "InjuryHx.EmerSurgIntraCranSurviveNoSurg" and "InjuryHx.EmerSurgIntraCranSurviveYesSurg" These 2 variables aim to capture information on the surgeon's expectations, eg if the surgeon considers a realistic expectation of benefit, or performs the surgery as a "last resort" in a likely hopeless case. 'The short term survival chances of the patients if I DO NOT operate will be (in %)' | Not Applicable |
| EmerSurgIntraCranSurviveYesSurg | TRUE | FALSE | integer | "InjuryHx.EmerSurgIntraCranSurviveNoSurg" and "InjuryHx.EmerSurgIntraCranSurviveYesSurg" These 2 variables aim to capture information on the surgeon's expectations, eg if the surgeon considers a realistic expectation of benefit, or performs the surgery as a "last resort" in a likely hopeless case. 'The short term survival chances of the patients if I DO operate will be (in %)' | Not Applicable |
| ERDestICDCodes1 | TRUE | FALSE | text | Up to 16 fields available to enter diagnosis as recorded by hospital administration according to ICD codes; applicable to patients discharged directly from the ER. For patients admitted to hospital or ICU, ICD codes are documented in: Hospital.ICDCode1 and Hospital.ICUDischargeICDCode1 | Not Applicable |

| Name | Static | Intervention | Format | Description | Possible Values |
|---|---|---|---|---|---|
| ERDestICDCodes2 | TRUE | FALSE | text | Up to 16 fields available to enter diagnosis as recorded by hospital administration according to ICD codes; applicable to patients discharged directly from the ER. For patients admitted to hospital or ICU, ICD codes are documented in: Hospital.ICDCode1 and Hospital.ICUDischargeICDCode1 | Not Applicable |
| ERDestICDCodes3 | TRUE | FALSE | text | Up to 16 fields available to enter diagnosis as recorded by hospital administration according to ICD codes; applicable to patients discharged directly from the ER. For patients admitted to hospital or ICU, ICD codes are documented in: Hospital.ICDCode1 and Hospital.ICUDischargeICDCode1 | Not Applicable |
| ERDestICDCodes4 | TRUE | FALSE | text | Up to 16 fields available to enter diagnosis as recorded by hospital administration according to ICD codes; applicable to patients discharged directly from the ER. For patients admitted to hospital or ICU, ICD codes are documented in: Hospital.ICDCode1 and Hospital.ICUDischargeICDCode1 | Not Applicable |
| ERDestICDCodes5 | TRUE | FALSE | text | Up to 16 fields available to enter diagnosis as recorded by hospital administration according to ICD codes; applicable to patients discharged directly from the ER. For patients admitted to hospital or ICU, ICD codes are documented in: Hospital.ICDCode1 and Hospital.ICUDischargeICDCode1 | Not Applicable |
| ERDestICDCodes6 | TRUE | FALSE | text | Up to 16 fields available to enter diagnosis as recorded by hospital administration according to ICD codes; applicable to patients discharged directly from the ER. For patients admitted to hospital or ICU, ICD codes are documented in: Hospital.ICDCode1 and Hospital.ICUDischargeICDCode1 | Not Applicable |
| ERDestICDCodesVersion | TRUE | FALSE | integer | Reflects the version used: ICD-9 or ICD-10. Up to 16 fields available to enter diagnosis as recorded by hospital administration according to ICD codes; applicable to patients discharged directly from the ER. For patients admitted to hospital or ICU, ICD codes are documented in: Hospital.ICDCode1 and Hospital.ICUDischargeICDCode1 | 9=ICD-9;10=ICD-10 |
| ERDischMotivForDestChoice | TRUE | FALSE | integer | WHY Question: documents main reason for choice of destination at ER discharge. | 1=Normal CT;2=Medical necessity;3=Social circumstances;4=No (ICU-) beds available;5=Requiring specialized facilities;88=Unknown;99=Other |
| ERDischMotivForDestChoiceOther | TRUE | FALSE | text | WHY Question: documents main reason for choice of destination after ER discharge --> Other | Not Applicable |
| FirstGCS | TRUE | FALSE | | | Not Applicable |
| FirstHospAssmtCondition | TRUE | FALSE | integer | Neurological assessment (GCS and pupils) was recorded for the scene of accident, the first hospital (if applicable), the Arrival to ER of the study hospital and post-stabilization. This Describes the condition under which the GCS was assessed at First Hospital. | 0=No sedation or paralysis;1=Sedated;2=Paralyzed;3=Temporary stop of sedation/paralysis;4=Reversal of sedation/paralysis;5=Active reversal (pharmacologic) of sedation/paralysis;99=Other |
| GcsEDArrAssmtCond | TRUE | FALSE | integer | Neurological assessment (GCS and pupils) was recorded for the scene of accident, the first hospital (if applicable), the Arrival to ER of the study hospital and post-stabilization. This describes the condition under which the GCS was assessed at Arrival to ER of the study hospital. | 0=No sedation or paralysis;1=Sedated;2=Paralyzed;3=Temporary stop of sedation/paralysis;4=Reversal of sedation/paralysis;5=Active reversal (pharmacologic) of sedation/paralysis;99=Other |
| GCSEDArrEyes | TRUE | FALSE | text | Neurological assessment (GCS and pupils) was recorded for the scene of accident, the first hospital (if applicable), the Arrival to ER of the study hospital and post- | 1=1-None;2=2-To pain;3=3-To speech;4=4-Spontaneously;O=Untestable (other);S=Untestable (swollen);UN=Unknown |

| Name | Static | Intervention | Format | Description | Possible Values |
|---|---|---|---|---|---|
| | | | | stabilization. This reflects the GCS¬†Eye¬†opening at Arrival to ER of study hospital. | |
| GCSEDArrMotor | TRUE | FALSE | text | Neurological assessment (GCS and pupils) was recorded for the scene of accident, the first hospital (if applicable), the Arrival to ER of the study hospital and post-stabilization. This reflects the GCS Motor score at Arrival to ER of study hospital. | 1=1-None;2=2-Abnormal extension;3=3-Abnormal flexion;4=4-Normal flexion/withdrawal;5=5-Localizes to pain;6=6-Obeys command;O=Untestable (Other);P=Untestable (Deep sedation/paralyzed);UN=Unknown |
| GCSEDArrScore | TRUE | FALSE | text | Neurological assessment (GCS and pupils) was recorded for the scene of accident, the first hospital (if applicable), the Arrival to ER of the study hospital and post-stabilization. This is a Calculated score for Arrival at ER of study hospital: GCSEDArrEyes + GCSEDArrMotor + GCSEDArrVerbal. If one or more of these is Untestable or unknown then = "No Sum" | 3=3;4=4;5=5;6=6;7=7;8=8;9=9;10=10;11=11;12=12;13=13;14=14;15=15;No Sum=Untestable/Unknown |
| GCSEDArrVerbal | TRUE | FALSE | text | Neurological assessment (GCS and pupils) was recorded for the scene of accident, the first hospital (if applicable), the Arrival to ER of the study hospital and post-stabilization. This reflects GCS Verbal score at Arrival to ER of study hospital | 1=1-None;2=2-Incomprehensible sound;3=3-Inappropriate words;4=4-Confused;5=5-Oriented;O=Untestable (Other);T=Untestable (Tracheotomy/endotracheal tube);UN=Unknown |
| GcsEDDischAssmtCond | TRUE | FALSE | integer | Neurological assessment (GCS and pupils) was recorded for the scene of accident, the first hospital (if applicable), the Arrival to ER of the study hospital and post-stabilization. This describes the condition under which the GCS was assessed POST-STABILIZATION. | 0=No Sedation or Paralysis;1=Sedated;2=Paralyzed;3=Temporary stop of sedation/paralysis;4=Reversal of sedation/paralysis;5=Active reversal (pharmacologic) of sedation/paralysis;99=Other |
| GCSEDDischEyes | TRUE | FALSE | text | Neurological assessment (GCS and pupils) was recorded for the scene of accident, the first hospital (if applicable), the Arrival to ER of the study hospital and post-stabilization. This reflects GCS eye opening for the assessment POST¬≠-STABILIZATION. | 1=1-None;2=2-To pain;3=3-To speech;4=4-Spontaneously;O=Untestable (other);S=Untestable (swollen);UN=Unknown |
| GCSEDDischMotor | TRUE | FALSE | text | Neurological assessment (GCS and pupils) was recorded for the scene of accident, the first hospital (if applicable), the Arrival to ER of the study hospital and post-stabilization. This reflects GCS Motor score for the assessment POST¬≠-STABILIZATION. | 1=1-None;2=2-Abnormal extension;3=3-Abnormal flexion;4=4-Normal flexion/withdrawal;5=5-Localizes to pain;6=6-Obeys command;O=Untestable (Other);P=Untestable (Deep sedation/paralyzed);UN=Unknown |
| GCSEDDischScore | TRUE | FALSE | text | Neurological assessment (GCS and pupils) was recorded for the scene of accident, the first hospital (if applicable), the Arrival to ER of the study hospital and post-stabilization. This is the Calculated score for the POST-STABILIZATION assessment: GCSEDDischEyes + GCSEDDischMotor + GCSEDDischVerbal. If one or more of these is Untestable or unknown then = "No Sum" | 3=3;4=4;5=5;6=6;7=7;8=8;9=9;10=10;11=11;12=12;13=13;14=14;15=15;No Sum=Untestable/Unknown |
| GCSEDDischVerbal | TRUE | FALSE | text | Neurological assessment (GCS and pupils) was recorded for the scene of accident, the first hospital (if applicable), the Arrival to ER of the study hospital and post-stabilization. This reflects GCS Verbal score for the assessment POST¬≠-STABILIZATION. | 1=1-None;2=2-Incomprehensible sound;3=3-Inappropriate words;4=4-Confused;5=5-Oriented;O=Untestable (Other);T=Untestable (Tracheotomy/endotracheal tube);UN=Unknown |
| GCSFirstHospEyes | TRUE | FALSE | text | Neurological assessment (GCS and pupils) was recorded for the scene of accident, the first hospital (if applicable), the Arrival to ER of the study hospital and post-stabilization. This reflects GCS eye opening for the assessment at First Hospital. | 1=1-None;2=2-To pain;3=3-To speech;4=4-Spontaneously;O=Untestable (other);S=Untestable (swollen);UN=Unknown |
| GCSFirstHospMotor | TRUE | FALSE | text | Neurological assessment (GCS and pupils) was recorded for the scene of accident, the first hospital (if applicable), the Arrival to ER of the study hospital and post-stabilization. This reflects GCS Motor score for the assessment at First Hospital. | 1=1-None;2=2-Abnormal extension;3=3-Abnormal flexion;4=4-Normal flexion/withdrawal;5=5-Localizes to pain;6=6-Obeys command;O=Untestable (Other);P=Untestable (Deep sedation/paralyzed);UN=Unknown |

| Name | Static | Intervention | Format | Description | Possible Values |
|---|---|---|---|---|---|
| GCSFirstHospPupilLftEyeMeasure | TRUE | FALSE | integer | Neurological assessment (GCS and pupils) was recorded for the scene of accident, the first hospital (if applicable), the Arrival to ER of the study hospital and post-stabilization. This reflects Left Pupil Size for the assessment at First Hospital. | Not Applicable |
| GCSFirstHospPupilReactivityLightLftEyeReslt | TRUE | FALSE | integer | Neurological assessment (GCS and pupils) was recorded for the scene of accident, the first hospital (if applicable), the Arrival to ER of the study hospital and post-stabilization. This reflects the reactivity of the Left Pupil for the assessment at First Hospital. | 1=+ (Sluggish);2=+ (Brisk);3=- (Negative) |
| GCSFirstHospPupilReactivityLightRghtEyeReslt | TRUE | FALSE | integer | Neurological assessment (GCS and pupils) was recorded for the scene of accident, the first hospital (if applicable), the Arrival to ER of the study hospital and post-stabilization. This reflects the reactivity of the Right Pupil for the assessment at First Hospital. | 1=+ (Sluggish);2=+ (Brisk);3=- (Negative) |
| GCSFirstHospPupilRightEyeMeasure | TRUE | FALSE | integer | Neurological assessment (GCS and pupils) was recorded for the scene of accident, the first hospital (if applicable), the Arrival to ER of the study hospital and post-stabilization. This reflects Right Pupil Size for the assessment at First Hospital. | Not Applicable |
| GCSFirstHospPupilSymmetry | TRUE | FALSE | integer | Neurological assessment (GCS and pupils) was recorded for the scene of accident, the first hospital (if applicable), the Arrival to ER of the study hospital and post-stabilization. This reflects the pupil symmetry for the assessment at First Hospital. | 1=Equal;2=Unequal R>L;3=Unequal L>R;66=Untestable;88=Unknown |
| GCSFirstHospScore | TRUE | FALSE | text | Neurological assessment (GCS and pupils) was recorded for the scene of accident, the first hospital (if applicable), the Arrival to ER of the study hospital and post-stabilization. This is the Calculated Score for the assessment at First Hospital: GCSFirstHospEyes + GCSFirstHospMotor + GCSFirstHospVerbal. If one or more of these is Untestable or unknown then = "No Sum" Sum Score may be recorded with no components | 3=3;4=4;5=5;6=6;7=7;8=8;9=9;10=10;11=11;12=12;13=13;14=14;15=15;No Sum=Untestable/Unknown |
| GCSFirstHospVerbal | TRUE | FALSE | text | Neurological assessment (GCS and pupils) was recorded for the scene of accident, the first hospital (if applicable), the Arrival to ER of the study hospital and post-stabilization. This reflects the GCS Verbal score for the assessment at First Hospital. | 1=1-None;2=2-Incomprehensible sound;3=3-Inappropriate words;4=4-Confused;5=5-Oriented;O=Untestable (Other);T=Untestable (tracheotomy/endotracheal tube);UN=Unknown |
| GCSMotorBaselineDerived | TRUE | FALSE | integer | This is a derived variable calculated centrally. It represents the GCS motor score for baseline risk adjustment with missing values imputed using IMPACT methodology - take Poststabilisation value and if absent work back in time towards prehospital values until non-missing value found. RECOMMENDED FOR BASELINE RISK ADJUSTMENT. | 1=1-None;2=2-Abnormal extension;3=3-Abnormal flexion;4=4-Normal flexion/withdrawal;5=5-Localizes to pain;6=6-Obeys command |
| GCSOtherAssmtConditions | TRUE | FALSE | integer | Neurological assessment (GCS and pupils) was recorded for the scene of accident, the first hospital (if applicable), the Arrival to ER of the study hospital and post-stabilization. There was also an additional option "Other" assessment. This describes the condition under which the GCS was assessed for the assessment "Other". | 0=No Sedation or Paralysis;1=Sedated;2=Paralyzed;3=Temporary stop of sedation/paralysis;4=Reversal of sedation/paralysis;5=Active reversal (pharmacologic) of sedation/paralysis;99=Other |
| GCSOtherEyes | TRUE | FALSE | text | Neurological assessment (GCS and pupils) was recorded for the scene of accident, the first hospital (if applicable), the Arrival to ER of the study hospital and post-stabilization. There was also an additional option "Other" assessment. This describes | 1=1-None;2=2-To pain;3=3-To speech;4=4-Spontaneously;O=Untestable (other);S=Untestable (swollen);UN=Unknown |

| Name | Static | Intervention | Format | Description | Possible Values |
|---|---|---|---|---|---|
| | | | | the GCS Eye opening for the assessment "Other". | |
| GCSOtherMotor | TRUE | FALSE | text | Neurological assessment (GCS and pupils) was recorded for the scene of accident, the first hospital (if applicable), the Arrival to ER of the study hospital and post-stabilization. There was also an additional option "Other" assessment. This describes GCS Motor score for the assessment "Other". | 1=1-None;2=2-Abnormal extension;3=3-Abnormal flexion;4=4-Normal flexion/withdrawal;5=5-Localizes to pain;6=6-Obeys command;O=Untestable (Other);P=Untestable (Deep sedation/paralyzed);UN=Unknown |
| GCSOtherPupilLftEyeMeasure | TRUE | FALSE | integer | Neurological assessment (GCS and pupils) was recorded for the scene of accident, the first hospital (if applicable), the Arrival to ER of the study hospital and post-stabilization. There was also an additional option "Other" assessment. This describes Left Pupil Size for the assessment "Other". | Not Applicable |
| GCSOtherPupilReactivityLightLftEyeReslt | TRUE | FALSE | integer | Neurological assessment (GCS and pupils) was recorded for the scene of accident, the first hospital (if applicable), the Arrival to ER of the study hospital and post-stabilization. There was also an additional option "Other" assessment. This describes the reactivity of the LEFT pupil for the assessment "Other". | 1=+ (Sluggish);2=+ (Brisk);3=- (Negative) |
| GCSOtherPupilRightEyeMeasure | TRUE | FALSE | integer | Neurological assessment (GCS and pupils) was recorded for the scene of accident, the first hospital (if applicable), the Arrival to ER of the study hospital and post-stabilization. There was also an additional option "Other" assessment. This describes the Right pupil size for the assessment "Other". | Not Applicable |
| GCSOtherPupilSymmetry | TRUE | FALSE | integer | Neurological assessment (GCS and pupils) was recorded for the scene of accident, the first hospital (if applicable), the Arrival to ER of the study hospital and post-stabilization. There was also an additional option "Other" assessment. This describes the pupil symmetry for the assessment "Other". | 1=Equal;2=Unequal R>L;3=Unequal L>R;66=Untestable;88=Unknown |
| GCSOtherReactivityLightRghtEyeReslt | TRUE | FALSE | integer | Neurological assessment (GCS and pupils) was recorded for the scene of accident, the first hospital (if applicable), the Arrival to ER of the study hospital and post-stabilization. There was also an additional option "Other" assessment. This describes the reactivity of the Right Pupil for the assessment "Other". | 1=+ (Sluggish);2=+ (Brisk);3=- (Negative) |
| GCSOtherScore | TRUE | FALSE | text | Neurological assessment (GCS and pupils) was recorded for the scene of accident, the first hospital (if applicable), the Arrival to ER of the study hospital and post-stabilization. There was also an additional option "Other" assessment. This is the Calculated Score for the assessment "Other": GCSOtherEyes + GCSOtherMotor + GCSOtherVerbal. If one or more of these is Untestable or unknown then = "No Sum" Sum score may be reported when components not available | 3=3;4=4;5=5;6=6;7=7;8=8;9=9;10=10;11=11;12=12;13=13;14=14;15=15;No Sum=Untestable/Unknown |
| GCSOtherVerbal | TRUE | FALSE | text | Neurological assessment (GCS and pupils) was recorded for the scene of accident, the first hospital (if applicable), the Arrival to ER of the study hospital and post-stabilization. There was also an additional option "Other" assessment. This describes the GCS Verbal score for the assessment "Other". | 1=1-None;2=2-Incomprehensible sound;3=3-Inappropriate words;4=4-Confused;5=5-Oriented;O=Untestable (Other);T=Untestable (tracheotomy/endotracheal tube);UN=Unknown |
| GCSPreHospBestEyes | TRUE | FALSE | text | Neurological assessment (GCS and pupils) was recorded for the scene of accident, the first hospital (if applicable), the Arrival to ER of the study hospital and post-stabilization. This describes the GCS Eye | 1=1-None;2=2-To pain;3=3-To speech;4=4-Spontaneously;O=Untestable (Other);S=Untestable (swollen);UN=Unknown |

| Name | Static | Intervention | Format | Description | Possible Values |
| --- | --- | --- | --- | --- | --- |
| | | | | opening for the assessment at Scene Of Accident. | |
| GCSPreHospBestMotor | TRUE | FALSE | text | Neurological assessment (GCS and pupils) was recorded for the scene of accident, the first hospital (if applicable), the Arrival to ER of the study hospital and post-stabilization. This describes the GCS Motor score for the assessment at Scene Of Accident. | 1=1-None;2=2-Abnormal extension;3=3-Abnormal flexion;4=4-Normal flexion/withdrawal;5=5-Localizes to pain;6=6-Obeys command;O=Untestable (Other);P=Untestable (Deep sedation/paralyzed);UN=Unknown |
| GCSPreHospBestScore | TRUE | FALSE | text | Neurological assessment (GCS and pupils) was recorded for the scene of accident, the first hospital (if applicable), the Arrival to ER of the study hospital and post-stabilization. This is the Calculated Score for the assessment at Scene Of Accident: GCSPreHospBestEyes + GCSPreHospBestMotor + GCSPreHospBestVerbal. If one or more of these is Untestable or unknown then = "No Sum" GCS sum score may be recorded when components not available - check "InjuryHx.GCSPreHospBestReportedTotalScore" | 3=3;4=4;5=5;6=6;7=7;8=8;9=9;10=10;11=11;12=12;13=13;14=14;15=15;No Sum=Untestable/Unknown |
| GCSPreHospBestVerbal | TRUE | FALSE | text | Neurological assessment (GCS and pupils) was recorded for the scene of accident, the first hospital (if applicable), the Arrival to ER of the study hospital and post-stabilization. This describes the GCS Verbal score for the assessment at Scene Of Accident. | 1=1-None;2=2-Incomprehensible sound;3=3-Inappropriate words;4=4-Confused;5=5-Oriented;O=Untestable (Other);T=Untestable (Tracheotomy/endotracheal tube);UN=Unknown |
| GcsPreHospLftEyeMeasr | TRUE | FALSE | text | Neurological assessment (GCS and pupils) was recorded for the scene of accident, the first hospital (if applicable), the Arrival to ER of the study hospital and post-stabilization. This describes the Left Pupil Size for the assessment at Scene Of Accident. | Not Applicable |
| GCSPreHospPupilReactivityLghtLftEyeResult | TRUE | FALSE | integer | Neurological assessment (GCS and pupils) was recorded for the scene of accident, the first hospital (if applicable), the Arrival to ER of the study hospital and post-stabilization. This describes the reactivity of the Left pupil for the assessment at Scene Of Accident. | 1=+ (Sluggish);2=+ (Brisk);3=- (Negative) |
| GCSPreHospPupilReactivityLghtRghtEyeResult | TRUE | FALSE | integer | Neurological assessment (GCS and pupils) was recorded for the scene of accident, the first hospital (if applicable), the Arrival to ER of the study hospital and post-stabilization. This describes the reactivity of the Right pupil for the assessment at Scene Of Accident. | 1=+ (Sluggish);2=+ (Brisk);3=- (Negative) |
| GCSPreHospPupilSymmetry | TRUE | FALSE | integer | Neurological assessment (GCS and pupils) was recorded for the scene of accident, the first hospital (if applicable), the Arrival to ER of the study hospital and post-stabilization. This describes the Pupil symmetry for the assessment at Scene Of Accident. | 1=Equal;2=Unequal R>L;3=Unequal L>R;66=Untestable;88=Unknown |
| GcsPreHospRghtEyeMeasr | TRUE | FALSE | text | Neurological assessment (GCS and pupils) was recorded for the scene of accident, the first hospital (if applicable), the Arrival to ER of the study hospital and post-stabilization. This describes the Right Pupil Size for the assessment at Scene Of Accident. | Not Applicable |
| GCSScoreBaselineDerived | TRUE | FALSE | integer | This is a derived variable calculated centrally. It represents the total GCS (single timepoint) for baseline risk adjustment with missing values imputed using IMPACT methodology - take Poststabilisation value and if absent work back in time towards prehospital values until non-missing value found. Intubated / untestable V score treated | 3=3;4=4;5=5;6=6;7=7;8=8;9=9;10=10;11=11;12=12;13=13;14=14;15=15 |

| Name | Static | Intervention | Format | Description | Possible Values |
|---|---|---|---|---|---|
| | | | | as unknown. RECOMMENDED FOR BASELINE RISK ADJUSTMENT. | |
| GOAT | TRUE | FALSE | | | Not Applicable |
| GOATBrnDate | TRUE | FALSE | integer | Reflects for the GOAT outcome test on the question "When were you born?" if the subject made an error or not in his reply. | -4=Error (-4);0=No Error |
| GOATBuldngLoc | TRUE | FALSE | integer | Reflects for the GOAT outcome test on the question "Where are you now?" if the subject made an error or not in his reply. | -5=Error (-5);0=No Error |
| GOATCityLoc | TRUE | FALSE | integer | Reflects for the GOAT outcome test on the question "Where are you now (which city)?" if the subject made an error or not in his reply. | -5=Error (-5);0=No Error |
| GOATCrntTm | TRUE | FALSE | integer | Reflects for the GOAT outcome test on the question "What time is it now?" if the subject made an error or not in his reply. | -5=Two and one-half hour + error (-5);-4=Two hour error (-4);-3=One and one-half hour error (-3);-2=One hour error (-2);-1=Half-hour error (-1);0=No Error |
| GOATDayDate | TRUE | FALSE | integer | Reflects for the GOAT outcome test on the question "What day of the week is it?" if the subject made an error or not in his reply. | -3=Three day error (-3);-2=Two day error (-2);-1=One day error (-1);0=No Error |
| GOATDayMnthDate | TRUE | FALSE | integer | Reflects for the GOAT outcome test on the question "What day of the month is it? (i.e. the date)" if the subject made an error or not in his reply. | -5=Five day + error (-5);-4=Four day error (-4);-3=Three day error (-3);-2=Two day error (-2);-1=One day error (-1);0=No Error |
| GOATDtlRslt | TRUE | FALSE | integer | Reflects for the GOAT outcome test on the question " What is the first event you can remember after the injury?" if the subject made an error or not in his reply. | -5=Error (-5);0=No Error |
| GOATFirEvntRslt | TRUE | FALSE | integer | Reflects for the GOAT outcome test on the question "What is the first event you can remember after the injury?" if the subject made an error or not in his reply. | -5=Error (-5);0=No Error |
| GOATFirEvntRsltTxt | TRUE | FALSE | text | Reflects for the GOAT outcome test on the question "What is the first event you can remember after the injury?" the details described by the subject. | Not Applicable |
| GOATHospAdmDate | TRUE | FALSE | integer | Reflects for the GOAT outcome test on the question "On what date were you admitted to the hospital?" if the subject made an error or not in his reply. | -5=Error (-5);0=No Error |
| GOATLiveLoc | TRUE | FALSE | integer | Reflects for the GOAT outcome test on the question "Where do you live?" if the subject made an error or not in his reply. | -4=Error (-4);0=No Error |
| GOATLstEvntDtlRslt | TRUE | FALSE | integer | Reflects for the GOAT outcome test on the question "What is the last event you can recall before the injury - Can you describe in detail?" if the subject made an error or not in his reply. | -5=Error (-5);0=No Error |
| GOATLstEvntRslt | TRUE | FALSE | integer | Reflects for the GOAT outcome test on the question "What is the last event you can recall before the injury?" if the subject made an error or not in his reply. | -5=Error (-5);0=No Error |
| GOATLstEvntRsltTxt | TRUE | FALSE | text | Reflects for the GOAT outcome test the reply on the question "What is the last event you can recall before the injury - Can you describe in detail?" | Not Applicable |
| GOATMnthDate | TRUE | FALSE | integer | Reflects for the GOAT outcome test on the question "What is the month?" if the subject made an error or not in his reply. | -15=Three or more month error (-15);-10=Two month error (-10);-5=One month error (-5);0=No Error |
| GOATNeuroPsychCompCode | TRUE | FALSE | integer | Reflects if GOAT test was done or not. Neuropsych testing was conducted during scheduled follow-up visits to hospital according to study protocol. Cross-sectional assessments across all strata was performed at 6 mnths after injury. For patients included in the MR substudy full testing was conducted for ER patients at: 2-3 weeks, 3 mnths and 6 mnths; Adm and ICU strata: 6 mnths, 12 mnths and 24 mnths Subjects were only considered testable if the GOAT was >=65. Neuropsych testing included: GOAT, RAVLT, TMT, CANTAB subtests, 10 m walk, Timed up-and-go, as well as Participant Questionnaire B.. "Untestable" | 1=1.0 Test not done;2=2.0 Test attempted but not completed;3=3.0 Test completed |

| Name | Static | Intervention | Format | Description | Possible Values |
|---|---|---|---|---|---|
| | | | | patients received the GOAT and JFK-Coma Recovery Scale-revised.. | |
| GOATNm | TRUE | FALSE | integer | Reflects for the GOAT outcome test on the question "What is your name?" if the subject made an error or not in his reply. | -2=Error (-2);0=No Error |
| GOATOutcome | TRUE | FALSE | text | The total GOAT score is calculated as 100 - Total error points; Interpretation: 76-100 = Normal / 66-75 = Borderline / <66 = Impaired | Borderline=Borderline;Impaired=Impaired;Normal=Normal |
| GOATTestAttemptd NotCompOptions | TRUE | FALSE | decimal | Documents reasons why GOAT was not completed when it was initially attempted (7 categories) | 2.1=Not completed - Cognitive/neurological deficits;2.2=Not completed - Non-neurological/physical reason;2.3=Not completed - Lack of effort/uncooperative;2.4=Not completed - Language;2.5=Not completed - Illness/fatigue;2.6=Not completed - Logistical reasons, other reasons;2.7=Not completed - Examiner error |
| GOATTestCompleted Options | TRUE | FALSE | decimal | Reflects the completion of the GOAT outcome test (Test completed in full results valid/Test completed Nonstandard, results valid/Nonstandard administration/Other) | 3=Test completed in full - results valid;3.1=Test completed - Non-standard, results valid;3.2=Non-standard administration - Other |
| GOATTestComplNon StandAdminOTHER | TRUE | FALSE | text | | Not Applicable |
| GOATTestNotDoneO ptions | TRUE | FALSE | decimal | Documents reasons why GOAT was not attempted (8 categories) | 1.1=Not attempted - Cognitive/neurological deficits;1.2=Not attempted - Non-neurological/physical reasons;1.3=Not attempted - Lack of effort/uncooperative;1.4=Not attempted - Language;1.5=Not attempted - Illness/fatigue;1.6=Not attempted - Logistical reasons, other reasons;1.7=Not attempted- Return to all normal activities;1.8=Not attempted - Patient not available |
| GOATTestNotDoneO ptionsOTHER | TRUE | FALSE | text | Documents the "other" reason why GOAT was not attempted | Not Applicable |
| GOATTotError | TRUE | FALSE | integer | Total error score for the GOAT outcome test. The total error score is calculated automatically. | Not Applicable |
| GOATTotScr | TRUE | FALSE | integer | Total actual score for the GOAT test. The total GOAT score is calculated as 100 - Total error points; Interpretation: 76-100 = Normal / 66-75 = Borderline / <66 = Impaired | Not Applicable |
| GOATTranspTyp | TRUE | FALSE | integer | Reflects for the GOAT outcome test on the question "How did you get to the hospital?" if the subject made an error or not in his reply. | -5=Error (-5);0=No Error |
| GOATYrDate | TRUE | FALSE | integer | Reflects for the GOAT outcome test on the question "What is the year?" if the subject made an error or not in his reply. | -30=Three or more year error (-30);-20=Two year error (-20);-10=One year error (-10);0=No Error |
| HighestGCSMotorDe rived | TRUE | FALSE | integer | This is a derived variable calculated centrally. It represents the GCS motor score for baseline risk adjustment with missing values imputed using ‚Äòhighest‚Äô value (best neurology) methodology- take best neurology of any of prehospital to Poststabilisation time points. DEPRECATED: we recommend using GCSMotorBaselineDerived for baseline risk adjustment instead (higher pseudo-R-squared in proportional odds model). | 1=1-None;2=2-Abnormal extension;3=3-Abnormal flexion;4=4-Normal flexion/withdrawal;5=5-Localizes to pain;6=6-Obeys command |
| HighestGCSTotalDeri ved | TRUE | FALSE | integer | This is a derived variable calculated centrally. Total GCS (single timepoint) for baseline risk adjustment with missing values imputed using ‚Äòhighest‚Äô value (best neurology) methodology - take best neurology of any of prehospital to Poststabilisation time points. Intubated / untestable V score treated as unknown. DEPRECATED: we recommend using GCSScoreBaselineDerived for baseline risk adjustment instead (higher pseudo-R-squared in proportional odds model). | 3=3;4=4;5=5;6=6;7=7;8=8;9=9;10=10;11=11;12=12;13=13;14=14;15=15 |
| HighestPupilsDerived | TRUE | FALSE | integer | This is a derived variable calculated centrally. Number of unreactive pupils for baseline risk adjustment with missing values imputed using ‚Äòhighest‚Äô value (best neurology) methodology- take best neurology of any of prehospital to Poststabilisation time points. Untestable pupil ignored: I.e. 1 reactive + 1 untestable | 0=Both reacting;1=One reacting;2=Both unreacting |

| Name | Static | Intervention | Format | Description | Possible Values |
| --- | --- | --- | --- | --- | --- |
| | | | | = 1 reactive (this assumption applies only to a small proportion of the data). DEPRECATED: we recommend using PupilsBaselineDerived for baseline risk adjustment instead (higher pseudo-R-squared in proportional odds model). | |
| ICUAdmReason | TRUE | FALSE | integer | Main reason for admission to ICU | 1=Mechanical ventilation;2=Frequent neurological observations;3=Haemodynamic invasive monitoring;4=Extracranial injuries;5=Neurological operation;6=Clinical deterioration;99=Other |
| ICUAdmReasonOther | TRUE | FALSE | text | Specifies the "other" if the main reason for admission to the ICU was other than the pre-defined list. | Not Applicable |
| InterventRadiology | TRUE | FALSE | integer | Reflects if at time of discharge from the ER some Interventional Radiology was scheduled | 0=No;1=Yes;88=Unknown |
| LOCAOC | TRUE | FALSE | integer | TBI may be present in the absence of LOC or PTA. Alteration of Consciousness (AOC) is then the main presenting symptom and considered diagnostic of TBI. Details of symptoms are captured in the Rivermead Questionnaire. | 0=No;1=Yes, immediate;2=Not tested due to LOC;3=Suspected;4=Yes, delayed onset;88=Unknown |
| LOCAOCDelayedHrs | TRUE | FALSE | integer | TBI may be present in the absence of LOC or PTA. Alteration of Consciousness (AOC) is then the main presenting symptom and considered diagnostic of TBI. This reflects the Number of hours after injury that alteration of consciousness occurred - Only in case of delayed onset. Details of symptoms are captured in the Rivermead Questionnaire. | Not Applicable |
| LOCAOCDuration | TRUE | FALSE | integer | TBI may be present in the absence of LOC or PTA. Alteration of Consciousness (AOC) is then the main presenting symptom and considered diagnostic of TBI. This reflects the Duration of alteration of consciousness. Details of symptoms are captured in the Rivermead Questionnaire. | 0=None;2=<1 minute;3=1-29 minutes;4=30-59 minutes;5=1-24 hours;6=1-7 days;7=>7 days;88=Unknown |
| LOCAOCReportedBy | TRUE | FALSE | integer | TBI may be present in the absence of LOC or PTA. Alteration of Consciousness (AOC) is then the main presenting symptom and considered diagnostic of TBI. This reflects by whom the alteration of consciousness was reported. Details of symptoms are captured in the Rivermead Questionnaire. | 1=Patient;2=Witness;3=Clinical interview;4=Medical chart;5=Not available |
| LOCDuration | TRUE | FALSE | integer | LOC and PTA are reported as part of the neurological assessment. This reflects the duration of Loss of Consciousness (LOC). Note: for patients admitted to hospital, the time to obeying commands is documented on hospital discharge: Hospital.HospDischargeTimeToObeyCommands | 0=No return of consciousness;2=<1 minute;3=1-29 minutes;4=30-59 minutes;5=1-24 hours;6=1-7 days;7=>7 days;88=Unknown |
| LOCGCSSumDet | TRUE | FALSE | integer | LOC and PTA are reported as part of the neurological assessment. This reflects for the Loss of Consciousness (LOC) the GCS sum score deterioration within one hour after presentation. | 0=None;1=1 point;2=2 or more points;88=Unknown |
| LOCLossOfConsciousness | TRUE | FALSE | integer | LOC and PTA are reported as part of the neurological assessment. Loss of Consciousness (LOC) is a definite sign of TBI. However, TBI may be present without any LOC. Presence and duration is captured. | 0=No;1=Yes;3=Suspected;88=Unknown |
| LOCLucidInterval | TRUE | FALSE | integer | LOC and PTA are reported as part of the neurological assessment. Lucid Interval is defined as a temporary improvement in a patient's condition after a traumatic brain injury, after which the condition deteriorates. A lucid interval is especially indicative of an epidural hematoma. An estimated 20 to 50% of patients with epidural hematoma experience such a lucid interval. | 0=No;1=Yes;88=Unknown |
| LOCLucidIntervalHrs | TRUE | FALSE | decimal | Lucid Interval is defined as a temporary improvement in a patient's condition after a | Not Applicable |

| Name | Static | Intervention | Format | Description | Possible Values |
|---|---|---|---|---|---|
| | | | | traumatic brain injury, after which the condition deteriorates. This reflects the Number of hours after injury that secondary deterioration occurred (in case Lucid Interval = Yes) | |
| LOCPTA | TRUE | FALSE | integer | LOC and PTA are reported as part of the neurological assessment. Post-traumatic amnesia (PTA) is the period after the injury that the patient cannot remember. In contrast to retrograde amnesia, the duration of PTA remains constant over time. To document presence/absence of PTA on discharge from the ER, the GOAT questionnaire is requested: Outcomes.GOATDate | 0=No;1=Yes, ongoing;2=Yes, resolved;3=Suspected;88=Unknown |
| LOCPTADuration | TRUE | FALSE | integer | LOC and PTA are reported as part of the neurological assessment. This variable is recorded only Only if PTA is yes. The duration of PTA reflects the severity of TBI. In patients with more sever TBI, the duration of PTA cannot be determined on presentation. For patients admitted to hospital, the duration of TBI in days is also captured on hospital discharge:.Hospital.HospDischPTADays | 0=None;2=<1 hour;5=1-24 hours;6=1-7 days;7=7-28 days;8=1-2 hours;9=2-4 hours;10=4-24 hours;11=>1 day;28=>28;77=N/A (e.g. death);88=Unknown |
| LOCPTAReportedBy | TRUE | FALSE | integer | LOC and PTA are reported as part of the neurological assessment. This reflects by whom PTA is reported. | 1=Patient;2=Witness;3=Retrospective assessment/ clinical interview;4=Medical chart;5=Not available;6=Prospective assessment with PTA scale |
| LOCPTAScale | TRUE | FALSE | integer | LOC and PTA are reported as part of the neurological assessment. In some centres, prospective assessment of amnesia (PTA) after TBI is performed using a dedicated scale. This variable documents the scale used. | 1=GOAT;2=Westmead;3=O-Log;4=Nijmegen PTA scale;99=Other |
| LOCReportedBy | TRUE | FALSE | integer | LOC and PTA are reported as part of the neurological assessment. This variable reflects by whom LOC was reported. | 1=Self report;2=Witness;3=Clinical interview;4=Medical chart;5=Not available |
| LOCRGA | TRUE | FALSE | integer | This reflects presence or absence of retrograde amnesia during neurological assessment. Amnesia after injury is a sign of TBI. Retrograde amnesia is the period before the injury that the patient cannot remember. The duration of retrograde amnesia becomes shorter as the injury is longer ago. The duration of retrograde amnesia is therefore dependent on time after injury at which it was assessed. | 0=No;1=Yes;88=Unknown |
| LOCRGADur | TRUE | FALSE | integer | This reflects the duration of retrograde amnesia is present during neurological assessment. | 0=None;1=<30;2=>= 30 minutes;88=Unknown |
| LOCRGAReportBy | TRUE | FALSE | integer | This reflects by whom Retrograde amnesia was reported if present during neurological assessment. | 1=Self report;2=Witness;3=Clinical interview;4=Medical chart;5=Not available |
| NeuroAssmtsAVPU | TRUE | FALSE | text | AVPU is scored as part of the neurological assessment on arrival to the ER. The AVPU scale (an acronym from "alert, voice, pain, unresponsive") is a system by which a health care professional can measure and record a patient's responsiveness, indicating their level of consciousness. | 88=Unknown;A=Patient is awake;P=The patient responds to painful stimulation;U=The patient is completely unresponsive;V=Patient responds to verbal stimulation |
| OtherGCS | TRUE | FALSE | | | Not Applicable |
| PainScale | TRUE | FALSE | integer | During neurological assessment at arrival to ER an overall rating was recorded for pain intensity going from 0 (zero pain) to 100 (unbearable pain). | Not Applicable |
| PreHospAssmtConditions | TRUE | FALSE | integer | Neurological assessment (GCS and pupils) was recorded for the scene of accident, the first hospital (if applicable), the Arrival to ER of the study hospital and post-stabilization. This describes the condition under which the GCS was assessed for the assessment at Scene Of Accident. | 0=No sedation or paralysis;1=Sedated;2=Paralyzed;3=Temporary stop of sedation/paralysis;4=Reversal of sedation/paralysis;5=Active reversal (pharmacologic) of sedation/paralysis;99=Other |
| PresArrivalMethod | TRUE | FALSE | integer | Reflects the mode of transportation used to transport the subject from the scene of accident to the hospital. | 1=Ambulance;2=Helicopter;3=Medical mobile team;4=Walk in or drop off;99=Other |

| Name | Static | Intervention | Format | Description | Possible Values |
|---|---|---|---|---|---|
| PresCirculationTreatmentCPR | TRUE | FALSE | integer | The status of Airway, Breathing and Circulation on scene are documented. This records Emergency care treatment on scene performed with regard to Circulation: CPR (Cardio-pulmonary resuscitation) ER arrival status is documented at: InjuryHX.EDArrivalCirculation | 0=No;1=Yes |
| PresCirculationTreatmentIVFluids | TRUE | FALSE | integer | The status of Airway, Breathing and Circulation on scene are documented. This records Emergency care treatment on scene performed with regard to Circulation: IV Fluids ER arrival status is documented at: InjuryHX.EDArrivalCirculation | 0=No;1=Yes |
| PresCirculationTreatmentNone | TRUE | FALSE | integer | The status of Airway, Breathing and Circulation on scene are documented. This records Emergency care treatment on scene performed with regard to Circulation: None ER arrival status is documented at: InjuryHX.EDArrivalCirculation | 0=No;1=Yes |
| PresCTBrain | TRUE | FALSE | integer | Only applicable in case of secondary referral. Reflects CT brain procedure performed at first hospital (not study hospital). | 0=No;1=Yes;88=Unknown |
| PresEmergencyCare | TRUE | FALSE | integer | Reflects if and by whom emergency medical care was given at the scene of accident (highest level of assistance) | 0=None;1=Untrained person (by stander);2=Trainer/coach;3=Military, non-medic;4=Paramedic;5=Nurse;6=Physician;7=Medical rescue team;99=Other |
| PresEmergencyCareIntubation | TRUE | FALSE | integer | Reflects if intubation was performed on scene. | 0=No;1=Yes;88=Unknown |
| PresEmergencyCareSuppOxygen | TRUE | FALSE | integer | Reflects if supplemental oxygen was given on scene. | 0=No;1=Yes;88=Unknown |
| PresEmergencyCareVentilation | TRUE | FALSE | integer | Reflects if Mechanical Ventilation was done on scene. | 0=No;1=Yes;88=Unknown |
| PresEmergencyServiceAmbuBasic | TRUE | FALSE | integer | Reflects type of Emergency service involved at accident scene --> Ambulance (basic EMT¬≠B) | 0=No;1=Yes |
| PresEmergencyServiceAmbuSpec | TRUE | FALSE | integer | Reflects type of Emergency service involved at accident scene --> Ambulance specialized (EMT¬≠P) | 0=No;1=Yes |
| PresEmergencyServiceFirefighter | TRUE | FALSE | integer | Reflects type of Emergency service involved at accident scene --> Firefighter | 0=No;1=Yes |
| PresEmergencyServiceHelicopter | TRUE | FALSE | integer | Reflects type of Emergency service involved at accident scene --> Helicopter | 0=No;1=Yes |
| PresEmergencyServiceNone | TRUE | FALSE | integer | Reflects type of mergency service involved at accident scene --> None | 0=No;1=Yes |
| PresEmergencyServicePolice | TRUE | FALSE | integer | Reflects type of Emergency service involved at accident scene --> Police | 0=No;1=Yes |
| PresERExtracranialSurg | TRUE | FALSE | integer | Only applicable in case of secondary referral. Reflects Emergency Extracranial surgery procedure performed at first hospital (not study hospital). | 0=No;1=Yes;88=Unknown |
| PresERIntracranialSurg | TRUE | FALSE | integer | Only applicable in case of secondary referral. Reflects the Emergency intracranial surgery Procedure performed at first hospital (not study hospital). | 0=No;1=Yes;88=Unknown |
| PresIntubation | TRUE | FALSE | integer | Only applicable in case of secondary referral. Reflects procedure of intubation performed at first hospital (not study hospital). | 0=No;1=Yes;88=Unknown |
| PresTBIRef | TRUE | FALSE | integer | Reflects if the subject was transported from the scene of accident immediately to the study hospital (primary referral) or if secondary referral occurred from a first hospital to the Study hospital. | 1=Primary;2=Secondary |
| PupilsBaselineDerived | TRUE | FALSE | integer | This is a derived variable calculated centrally. Number of unreactive pupils for baseline risk adjustment with missing values imputed using IMPACT methodology- take Poststabilisation value and if absent work back in time towards prehospital values until non-missing value found. Untestable pupil ignored: I.e. 1 reactive + 1 untestable = 1 reactive (this assumption applies only to a small proportion of the data). | 0=Both reacting;1=One reacting;2=Both unreacting |

| Name | Static | Intervention | Format | Description | Possible Values |
|---|---|---|---|---|---|
| | | | | RECOMMENDED FOR BASELINE RISK ADJUSTMENT. | |
| PupilsNonSymmetric | TRUE | FALSE | integer | Pupil symmetry derived variable calculated from (GCSFirstHospPupilSymmetry,EDArrPupilSymmetry,GCSPreHospPupilSymmetry,EDDischPupilSymmetry,PupilsNonSymmetric) | 0=No;1=Yes |
| RPQ | TRUE | FALSE | | | Not Applicable |
| RPQ13Score | TRUE | FALSE | integer | Score for the Rivermead Assessment RPQ-13. Score for the Not scored by investigator, calculated score. Baseline RPQ documents complaints and symptoms commonly reported after TBI. Baseline RPQ can only be assessed in conscious subjects who are out of PTA. Provides further details on AOC. | Not Applicable |
| RPQ3Score | TRUE | FALSE | integer | Score for the Rivermead Assessment RPQ-3. Not scored by investigator, calculated score. Baseline RPQ documents complaints and symptoms commonly reported after TBI. Baseline RPQ can only be assessed in conscious subjects who are out of PTA. Provides further details on AOC. | Not Applicable |
| RPQBlurredVision | TRUE | FALSE | integer | This variable reflects if the subject, compared with before the accident, now (over the last 24 hours) suffers from blurred vision, as part of the Rivermead RPQ assessment. | 0=0-Not experienced at all;1=1- No more of a problem;2=2- A mild problem;3=3- A moderate problem;4=4- A severe problem |
| RPQDepressed | TRUE | FALSE | integer | This variable reflects if the subject, compared with before the accident, now (over the last 24 hours) suffers from feeling depressed or tearful, as part of the Rivermead RPQ assessment. | 0=0-Not experienced at all;1=1- No more of a problem;2=2- A mild problem;3=3- A moderate problem;4=4- A severe problem |
| RPQDizziness | TRUE | FALSE | integer | This variable reflects if the subject, compared with before the accident, now (over the last 24 hours) suffers from feelings of dizziness, as part of the Rivermead RPQ assessment. | 0=0-Not experienced at all;1=1- No more of a problem;2=2- A mild problem;3=3- A moderate problem;4=4- A severe problem |
| RPQDoubleVision | TRUE | FALSE | integer | This variable reflects if the subject, compared with before the accident, now (over the last 24 hours) suffers from double vision, as part of the Rivermead RPQ assessment. | 0=0-Not experienced at all;1=1- No more of a problem;2=2- A mild problem;3=3- A moderate problem;4=4- A severe problem |
| RPQFatigue | TRUE | FALSE | integer | This variable reflects if the subject, compared with before the accident, now (over the last 24 hours) suffers from Fatigue, tiring more easily, as part of the Rivermead RPQ assessment. | 0=0-Not experienced at all;1=1- No more of a problem;2=2- A mild problem;3=3- A moderate problem;4=4- A severe problem |
| RPQForgetful | TRUE | FALSE | integer | This variable reflects if the subject, compared with before the accident, now (over the last 24 hours) suffers from Forgetfulness, poor memory, as part of the Rivermead RPQ assessment. | 0=0-Not experienced at all;1=1- No more of a problem;2=2- A mild problem;3=3- A moderate problem;4=4- A severe problem |
| RPQFrustrated | TRUE | FALSE | integer | This variable reflects if the subject, compared with before the accident, now (over the last 24 hours) suffers from Feeling frustrated or impatient, as part of the Rivermead RPQ assessment. | 0=0-Not experienced at all;1=1- No more of a problem;2=2- A mild problem;3=3- A moderate problem;4=4- A severe problem |
| RPQHeadaches | TRUE | FALSE | integer | This variable reflects if the subject, compared with before the accident, now (over the last 24 hours) suffers from Headaches, as part of the Rivermead RPQ assessment. | 0=0-Not experienced at all;1=1- No more of a problem;2=2- A mild problem;3=3- A moderate problem;4=4- A severe problem |
| RPQIrritable | TRUE | FALSE | integer | This variable reflects if the subject, compared with before the accident, now (over the last 24 hours) suffers from Being irritable, easily angered, as part of the Rivermead RPQ assessment. | 0=0-Not experienced at all;1=1- No more of a problem;2=2- A mild problem;3=3- A moderate problem;4=4- A severe problem |
| RPQLightSensitivity | TRUE | FALSE | integer | This variable reflects if the subject, compared with before the accident, now (over the last 24 hours) suffers from Light sensitivity (easily upset by bright light), as part of the Rivermead RPQ assessment. | 0=0-Not experienced at all;1=1- No more of a problem;2=2- A mild problem;3=3- A moderate problem;4=4- A severe problem |

| Name | Static | Intervention | Format | Description | Possible Values |
|---|---|---|---|---|---|
| RPQLongerToThink | TRUE | FALSE | integer | This variable reflects if the subject, compared with before the accident, now (over the last 24 hours) suffers from Taking longer to think, as part of the Rivermead RPQ assessment. | 0=0-Not experienced at all;1=1- No more of a problem;2=2- A mild problem;3=3- A moderate problem;4=4- A severe problem |
| RPQNausea | TRUE | FALSE | integer | This variable reflects if the subject, compared with before the accident, now (over the last 24 hours) suffers from Nausea and/or vomiting, as part of the Rivermead RPQ assessment. | 0=0-Not experienced at all;1=1- No more of a problem;2=2- A mild problem;3=3- A moderate problem;4=4- A severe problem |
| RPQNoiseSensitivity | TRUE | FALSE | integer | This variable reflects if the subject, compared with before the accident, now (over the last 24 hours) suffers from Noise sensitivity (easily upset by loud noise), as part of the Rivermead RPQ assessment. | 0=0-Not experienced at all;1=1- No more of a problem;2=2- A mild problem;3=3- A moderate problem;4=4- A severe problem |
| RPQOther1 | TRUE | FALSE | integer | This variable reflects if the subject, compared with before the accident, now (over the last 24 hours) suffers from Other difficulties than the pre-defined list, as part of the Rivermead RPQ assessment. | 0=0-Not experienced at all;1=1-No more of a problem;2=2-A mild problem;3=3-A moderate problem;4=4-A severe problem |
| RPQOther1Text | TRUE | FALSE | text | This variable reflects which other difficulties, if the subject, compared with before the accident, now (over the last 24 hours) suffers from Other difficulties than the pre-defined list, as part of the Rivermead RPQ assessment. | Not Applicable |
| RPQOther2 | TRUE | FALSE | integer | This variable reflects if the subject, compared with before the accident, now (over the last 24 hours) suffers from Other difficulties than the pre-defined list, as part of the Rivermead RPQ assessment. | 0=0-Not experienced at all;1=1-No more of a problem;2=2-A mild problem;3=3-A moderate problem;4=4-A severe problem |
| RPQOther2Text | TRUE | FALSE | text | This variable reflects which other difficulties, if the subject, compared with before the accident, now (over the last 24 hours) suffers from Other difficulties than the pre-defined list, as part of the Rivermead RPQ assessment. | Not Applicable |
| RPQPoorConcentration | TRUE | FALSE | integer | This variable reflects if the subject, compared with before the accident, now (over the last 24 hours) suffers from Poor concentration, as part of the Rivermead RPQ assessment. | 0=0-Not experienced at all;1=1- No more of a problem;2=2- A mild problem;3=3- A moderate problem;4=4- A severe problem |
| RPQQuestionnaireMode | TRUE | FALSE | integer | The mode in which the Rivermead RPQ assessment was performed; could be "Personal interview", "Postal questionnaire", "Telephone interview", "Web-based completion" | 1=Telephone interview;2=Postal questionnaire;3=Web-based completion;4=Personal interview |
| RPQRestless | TRUE | FALSE | integer | This variable reflects if the subject, compared with before the accident, now (over the last 24 hours) suffers from Restlessness, as part of the Rivermead RPQ assessment. | 0=0-Not experienced at all;1=1- No more of a problem;2=2- A mild problem;3=3- A moderate problem;4=4- A severe problem |
| RPQSleepDisturbance | TRUE | FALSE | integer | This variable reflects if the subject, compared with before the accident, now (over the last 24 hours) suffers from Sleep disturbance, as part of the Rivermead RPQ assessment. | 0=0-Not experienced at all;1=1- No more of a problem;2=2- A mild problem;3=3- A moderate problem;4=4- A severe problem |
| RPQTotalScore | TRUE | FALSE | integer | RPQ (Rivermead post-concussion symptoms questionnaire) Total Score. Calculated centrally. | Not Applicable |
| SympSkullFract | TRUE | FALSE | integer | During neurological assessment at arrival in the ER, Clinical signs of skull base fracture (e.g. raccoon eyes, battle sign, hemotympanun, CSF otorrhea, CRF rhinorrhea, bleeding from ear) were recorded. | 0=No;1=Yes;88=Unknown |
| SympVomiting | TRUE | FALSE | integer | During neurological assessment at arrival in the ER, Vomiting was recorded. | 0=No;1=Once;2=More than once;88=Unknown |

*Brain imaging reports*

| Name | Static | Intervention | Format | Description | Possible Values |
|---|---|---|---|---|---|
| AnyIntracranTraumaticAbnormality | FALSE | FALSE | text | This variable was assessed by a central review panel, according the TBI-Common Data Elements. This variable indicates whether any of the 12 following CDEs is present (Mass lesion, ExtraaxialHematoma, EpiduralHematoma, SubduralHematomaAcute, SubduralHematomaSubacuteChronic, SubduralCollectionMixedDensity, Contusion, TAI, traumaticSubarachnoidHemorrhage, IntraventricularHemorrhage, MidlineShift or CisternalCompression. | absent=Absent;present=Present;indeteminate=Indeterminate;uninterpretable=Uninterpretable |
| CisternalCompression | FALSE | FALSE | text | This variable was assessed by a central review panel, according the TBI-Common Data Elements. This variable indicates whether there is compression of one or more the basal cisterns (i.e. suprasellar, quadrigeminal, prepontine, ambient, cisterna magna). More descriptive information (location, compression vs. absence) can be found in the JSON files. | absent=Absent;present=Present;indeteminate=Indeterminate;uninterpretable=Uninterpretable |
| Contusion | FALSE | FALSE | text | This variable was assessed by a central review panel, according the TBI-Common Data Elements. This variable indicates whether one or more contusions are present. More descriptive information (location, volume, number) and advanced information (e.g. hemorrhagic, non-hemorrhagic, intracerebral hemorrhage, etc.) can be found in the JSON files. Volume was estimated using the AxBxC/2 method. Edema was included in the measurements. | absent=Absent;present=Present;indeteminate=Indeterminate;uninterpretable=Uninterpretable |
| CRFCTAngulation | FALSE | FALSE | integer | This variable describes if a CT scan is performed with or without angulation. There are three options: no angulation (volume scan), orbital-meatal line and other. There is a risk of different interpretation, since there was no definition. | 1=No angulation (volume scan);2=Orbital-meatal line;99=Other |
| CRFCTManuf | FALSE | FALSE | text | This variable describes the CT scans manufacturer. | 99=Other;AGFA=Agfa;CARE=Carestream;GE=GE;HITA=Hitachi;KONI=Konica Minolta;PHIL=Philips;SIEM=Siemens;TOSH=Toshiba |
| CRFCTMidlineShift | FALSE | FALSE | integer | CT parameters scored by the investigator: reflects if there was midline shift on the CT. | 0=No;1=Yes |
| CRFCTMidlineShiftMeasure | FALSE | FALSE | decimal | CT parameters scored by the investigator: reflects the volume in mm if there was a midline shift on the CT Check also "Imaging.CRFCTMidlineShift" | Not Applicable |
| CRFCTReason | FALSE | FALSE | text | This variable contains the main reason why a CT-scan, during hospital stay, was performed. One of following options: standard follow-up, post-operative control, clinical deterioration, (suspicion of) increasing ICP, lack of improvement, unknown, other (specified in CTMRI.CTReasonOther) The reason for making an early CT-scan/ER scan can be found in: CTMRI.CTERReason | CD=Clinical deterioration;ICUADM88=Unknown;ICUADM99=Other;IICP=(Suspicion of) Increasing ICP;LOP=Lack of improvement;POC=Post-operative control;SFU=Standard follow-up |
| CRFCTScannerType | FALSE | FALSE | integer | This variable specifies the type of CT-scanner by the number of slices. | 16=16-slice;32=32-slice;64=64-slice;99=Other;128=128-slice;256=256-slice;320=320-slice |
| CRFCTTypeCCT | FALSE | FALSE | integer | This variable describes the type of CT scan that has been made. Multiple options can be selected from: Non-contrast CT, Contrast CT, CT Angiography, Perfusion CT. | 0=No;1=Yes |
| CRFCTTypeCTA | FALSE | FALSE | integer | This variable describes the type of CT scan that has been made. Multiple options can be selected from: Non-contrast CT, Contrast CT, CT Angiography, Perfusion CT. | 0=No;1=Yes |
| CRFCTTypeNCCT | FALSE | FALSE | integer | This variable describes the type of CT scan that has been made. Multiple options can be selected from: Non-contrast CT, Contrast CT, CT Angiography, Perfusion CT. | 0=No;1=Yes |

| Name | Static | Intervention | Format | Description | Possible Values |
|---|---|---|---|---|---|
| CRFCTTypePCT | FALSE | FALSE | integer | This variable describes the type of CT scan that has been made. Multiple options can be selected from: Non-contrast CT, Contrast CT, CT Angiography, Perfusion CT. | 0=No;1=Yes |
| CRFForm | FALSE | FALSE | text | Acute or follow up form in the e-CRF. | CTMRI=CT/MRI;FollowUp=Follow-up |
| CRFMRIManuf | FALSE | FALSE | text | This variable describes the MRI-scan manufacturer. This variable combines data from CTMRI.MRIManuf and FollowUp.MRIManuf | 99=Other;GE=GE;PHIL=Philips;SIEM=Siemens;TOSH=Toshiba |
| CRFMRIReason | FALSE | FALSE | text | This variable contains the main reason why an MRI was performed. This variable combines data from CTMRI.MRIReason and FollowUp.MRIReason | 88=Unknown;99=Other;ICUADM1=Discrepancy between CT and clinical condition;ICUADM2=Standard Care;ICUADM3=Detection of brainstem lesions;STUDYPROT=Study protocol |
| CRFMRIReasonOther | FALSE | FALSE | text | This variable contains the main reason why an MRI was performed, when in the variable "MRIReason", the option "other" was chosen. This is a text field. This variable combines data from CTMRI.MRIReasonOther and FollowUp.MRIReasonOther | Not Applicable |
| CRFMRIResultPreExistAbnorm | FALSE | FALSE | integer | This variable describes whether or not there are pre-existing abnormalities present on MRI scan (yes, no, unknown). Assessment by investigator and or physician. This variable combines data from CTMRI.MRIResultPreExistAbnorm and FollowUp.MRIResultPreExistAbnorm | 0=No;1=Yes;88=Unknown |
| CRFMRIResultTraumaticAbnorm | FALSE | FALSE | integer | This variable describes whether or not there are traumatic abnormalities present on MRI scan (yes, no, unknown). Assessment by investigator and/or physician. This variable combines data from CTMRI.MRIResultTraumaticAbnorm and FollowUp.MRIResultTraumaticAbnorm | 0=No;1=Yes;88=Unknown |
| CRFMRIScannerStrength | FALSE | FALSE | text | This variable describes the MRI scanner strength. This is a text field. This variable combines data from CTMRI.MRIScannerStrength and FollowUp.MRIScannerStrength | Not Applicable |
| CRFMRISequences99 | FALSE | FALSE | integer | This variable describes the MRI sequence. Options are: T1, T2, FLAIR, DWI, GRE, SWI, DTI, MRSI, PWI, Other. Multiple options can be selected. This variable combines data from CTMRI.MRISequences and FollowUp.MRISequences | 0=No;1=Yes |
| CRFMRISequencesDTI | FALSE | FALSE | integer | This variable describes the MRI sequence. Options are: T1, T2, FLAIR, DWI, GRE, SWI, DTI, MRSI, PWI, Other. Multiple options can be selected. This variable combines data from CTMRI.MRISequences and FollowUp.MRISequences | 0=No;1=Yes |
| CRFMRISequencesDWI | FALSE | FALSE | integer | This variable describes the MRI sequence. Options are: T1, T2, FLAIR, DWI, GRE, SWI, DTI, MRSI, PWI, Other. Multiple options can be selected. This variable combines data from CTMRI.MRISequences and FollowUp.MRISequences | 0=No;1=Yes |
| CRFMRISequencesFLAIR | FALSE | FALSE | integer | This variable describes the MRI sequence. Options are: T1, T2, FLAIR, DWI, GRE, SWI, DTI, MRSI, PWI, Other. Multiple options can be selected. This variable combines data from CTMRI.MRISequences and FollowUp.MRISequences | 0=No;1=Yes |
| CRFMRISequencesGRE | FALSE | FALSE | integer | This variable describes the MRI sequence. Options are: T1, T2, FLAIR, DWI, GRE, SWI, DTI, MRSI, PWI, Other. Multiple options can be selected. This variable combines data from CTMRI.MRISequences and FollowUp.MRISequences | 0=No;1=Yes |
| CRFMRISequencesMRSI | FALSE | FALSE | integer | This variable describes the MRI sequence. Options are: T1, T2, FLAIR, DWI, GRE, SWI, DTI, MRSI, PWI, Other. Multiple options can be selected. This variable | 0=No;1=Yes |

| Name | Static | Intervention | Format | Description | Possible Values |
| --- | --- | --- | --- | --- | --- |
| | | | | combines data from CTMRI.MRISequences and FollowUp.MRISequences | |
| CRFMRISequencesPWI | FALSE | FALSE | integer | This variable describes the MRI sequence. Options are: T1, T2, FLAIR, DWI, GRE, SWI, DTI, MRSI, PWI, Other. Multiple options can be selected. This variable combines data from CTMRI.MRISequences and FollowUp.MRISequences | 0=No;1=Yes |
| CRFMRISequencesSWI | FALSE | FALSE | integer | This variable describes the MRI sequence. Options are: T1, T2, FLAIR, DWI, GRE, SWI, DTI, MRSI, PWI, Other. Multiple options can be selected. This variable combines data from CTMRI.MRISequences and FollowUp.MRISequences | 0=No;1=Yes |
| CRFMRISequencesT1 | FALSE | FALSE | integer | This variable describes the MRI sequence. Options are: T1, T2, FLAIR, DWI, GRE, SWI, DTI, MRSI, PWI, Other. Multiple options can be selected. This variable combines data from CTMRI.MRISequences and FollowUp.MRISequences | 0=No;1=Yes |
| CRFMRISequencesT2 | FALSE | FALSE | integer | This variable describes the MRI sequence. Options are: T1, T2, FLAIR, DWI, GRE, SWI, DTI, MRSI, PWI, Other. Multiple options can be selected. This variable combines data from CTMRI.MRISequences and FollowUp.MRISequences | 0=No;1=Yes |
| CRFMRITraumAbnormASDH | FALSE | FALSE | integer | This variable describes whether or not there is an acute subdural hematoma present on MRI scan (yes, no, unknown). Assessment by investigator and/or physician. This variable combines data from CTMRI.MRITraumAbnormASDH and FollowUp.MRITraumAbnormASDH | 0=No;1=Yes;88=Unknown |
| CRFMRITraumAbnormContusion | FALSE | FALSE | integer | This variable describes whether or not there is a contusion present on MRI scan (yes, no, unknown). Assessment by investigator and/or physician. This variable combines data from CTMRI.MRITraumAbnormContusion and FollowUp.MRITraumAbnormContusion | 0=No;1=Yes;88=Unknown |
| CRFMRITraumAbnormDAI | FALSE | FALSE | integer | This variable describes whether or not there is DAI (diffuse axonal injury) present on MRI scan (yes, no, unknown). Assessment by investigator and/or physician. This variable combines data from CTMRI.MRITraumAbnormDAI and FollowUp.MRITraumAbnormDAI | 0=No;1=Yes;88=Unknown |
| CRFMRITraumAbnormDAILesionLocBrainstem | FALSE | FALSE | integer | This variable describes whether or not there is a brain stem lesion present on MRI scan (yes, no, unknown). Assessment by investigator and/or physician. This variable combines data from CTMRI.MRITraumAbnormDAILesionLocBrainstem and FollowUp.MRITraumAbnormDAILesionLocBrainstem | 0=No;1=Yes;88=Unknown |
| CRFMRITraumAbnormDAILesionLocCorpusCallosum | FALSE | FALSE | integer | This variable describes whether or not there is a corpus callosum lesion present on MRI scan (yes, no, unknown). Assessment by investigator and/or physician. This variable combines data from CTMRI.MRITraumAbnormDAILesionLocCorpusCallosum and FollowUp.MRITraumAbnormDAILesionLocCorpusCallosum | 0=No;1=Yes;88=Unknown |
| CRFMRITraumAbnormDAILesionLocDiffuseWhiteMatter | FALSE | FALSE | integer | This variable describes whether or not there is a lesion in diffuse white matter present on MRI scan (yes, no, unknown). Assessment by investigator and/or physician. This variable combines data from CTMRI.MRITraumAbnormDAILesionLocDiffuseWhiteMatter and FollowUp.MRITraumAbnormDAILesionLocDiffuseWhiteMatter | 0=No;1=Yes;88=Unknown |

| Name | Static | Intervention | Format | Description | Possible Values |
|---|---|---|---|---|---|
| CRFMRITraumAbnormDAINumLesions | FALSE | FALSE | integer | This variable describes how many DAI lesions are present on MRI scan (1,2,3,4,>=5). Assessment by investigator and/or physician. This variable combines data from CTMRI.MRITraumAbnormDAINumLesions and FollowUp.MRITraumAbnormDAINumLesions | 1=1;2=2;3=3;4=4;5=>= 5 |
| CRFMRITraumAbnormEDH | FALSE | FALSE | integer | This variable describes whether or not there is an epidural hematoma present on MRI scan (yes, no, unknown). Assessment by investigator and/or physician. This variable combines data from CTMRI.MRITraumAbnormEDH and FollowUp.MRITraumAbnormEDH | 0=No;1=Yes;88=Unknown |
| CRFMRITypeMRA | FALSE | FALSE | integer | This variable describes the type of MRI scan that has been made. Options are: MRI, MRA This variable combines data from CTMRI.MRIType and FollowUp.MRIType | 0=No;1=Yes |
| CRFMRITypeMRI | FALSE | FALSE | integer | This variable describes the type of MRI scan that has been made. Options are: MRI, MRA This variable combines data from CTMRI.MRIType and FollowUp.MRIType | 0=No;1=Yes |
| CTAcuteSubdurHema | FALSE | FALSE | integer | Assessment by clinician/investigator whether or not in his/her interpretation an acute subdural hematoma is present. Also, the size (not quantified) is requested to be estimated. | 0=No;1=Small;2=Large (mass);88=Unknown |
| CTBasalCisternsAbsentCompressed | FALSE | FALSE | integer | Assessment by clinician/investigator whether or not in his/her interpretation the basal cisterns are compressed. | 0=No;1=Yes |
| CTContusion | FALSE | FALSE | integer | Assessment by clinician/investigator whether or not in his/her interpretation an intracerebral hematoma/contusion is present. Also, the size (not quantified) is requested to be estimated. | 0=No;1=Small;2=Large (mass);88=Unknown |
| CTDeprSkullFract | FALSE | FALSE | integer | Assessment by clinician/investigator whether or not in his/her interpretation a depressed skull fracture is present. In addition, when present, clinician/investigator has to document whether the fracture is associated with an open wound or not (compound vs closed) | 0=No;1=Closed;2=Open (compound) |
| CTExtraduralHema | FALSE | FALSE | integer | Assessment by clinician/investigator whether or not in his/her interpretation an acute extradural/epidural hematoma is present. Also, the size (not quantified) is requested to be estimated. | 0=No;1=Small;2=Large (mass);88=Unknown |
| CTFrames | FALSE | FALSE | | | Not Applicable |
| CTICLesionDAI | FALSE | FALSE | integer | Assessment by clinician/investigator whether or not in his/her interpretation diffuse axonal injury is present. | 0=No;1=Yes;88=Unknown |
| CTIschemia | FALSE | FALSE | integer | Assessment by clinician/investigator whether or not in his/her interpretation ischemia is present. Only applicable for clinical follow-up CT, not applicable to initial CT. In addition, the severity, in terms of how many arterial territories have ischemia, is requested to be answered. | 0=No;1=Single arterial territory;2=Multiple territories;3=Hemisphere |
| CTNoOpMotiv | FALSE | FALSE | integer | WHY question: documents reason for not having an indication for (intra)cranial surgery. | 0=No surgical lesion;1=Lesion present, but Acceptable/good neurologic condition;2=Lesion present, but Guideline adherence;3=Lesion present, but Little/no mass effect;4=Lesion present, but Not hospital policy;5=Lesion present, but Extremely poor prognosis;6=Lesion present, but Brain Death;7=Lesion present, but Old age;8=Lesion present, but Wish family;88=Unknown;99=Lesion present, but Other |
| CTNoOpMotivOther | FALSE | FALSE | text | Specification, only applicable if "CTMRI.CTNoOpMotiv" was "other" | Not Applicable |
| CTSchedForOp | FALSE | FALSE | integer | Whether or not the patient is scheduled for (intra)cranial surgery. The main aim here is to capture whether the clinical team taking | 0=No;1=Yes |

| Name | Static | Intervention | Format | Description | Possible Values |
|---|---|---|---|---|---|
| | | | | care of the patient sees a neurosurgical indication. | |
| CTSubarachnoidHem | FALSE | FALSE | integer | Assessment by clinician/investigator whether or not in his/her interpretation subarachnoid hemorrhage is present. In addition, the location of the hemorrhage is requested to be answered. | 0=No;1=Basal;2=Cortical;3=Basal and Cortical |
| CTYesOpMotiv | FALSE | FALSE | integer | WHY question: documents reason for having an indication for (intra)cranial surgery. | 1=Emergency/life saving;2=Clinical deterioration;3=Mass effect on CT;4=Radiological progression;5=(Suspicion of) raised ICP;6=Guideline adherence;7=To prevent deterioration;8=Depressed skull fracture;99=Other |
| CTYesOpMotivOther | FALSE | FALSE | text | Free text if "CTMRI.CTYesOpMotiv" was marked as 'Other'. Relates to the WHY question: documents reason for having an indication for (intra)cranial surgery. | Not Applicable |
| EpiduralHematoma | FALSE | FALSE | text | This variable was assessed by a central review panel, according the TBI-Common Data Elements. This variable indicates whether an epidural hematoma or multiple epidural hematomas are present. More descriptive information (location, volume, number) and advanced information (e.g. arterial versus venous) can be found in the JSON files. Note: Volume was estimated using the AxBxC/2 method. In the JSON files, "descriptive_volume" is the estimated volume of the lesion and can be used for analysis, "descriptive_length", "descriptive_width", "descriptive_max_thickness" are measurements that should not be used in any analysis. | absent=Absent;present=Present;indeteminate=Indeterminate;uninterpretable=Uninterpretable |
| ERAnyIntracranTraumaticAbnormality | TRUE | FALSE | text | In emergency room: This variable was assessed by a central review panel, according the TBI-Common Data Elements. This variable indicates whether any of the 12 following CDEs is present (Mass lesion, ExtraaxialHematoma, EpiduralHematoma, SubduralHematomaAcute, SubduralHematomaSubacuteChronic, SubduralCollectionMixedDensity, Contusion, TAI, traumaticSubarachnoidHemorrhage, IntraventricularHemorrhage, MidlineShift or CisternalCompression. | absent=Absent;present=Present;indeteminate=Indeterminate;uninterpretable=Uninterpretable |
| ERCisternalCompression | TRUE | FALSE | text | In emergency room: This variable was assessed by a central review panel, according the TBI-Common Data Elements. This variable indicates whether there is compression of one or more the basal cisterns (i.e. suprasellar, quadrigeminal, prepontine, ambient, cisterna magna). More descriptive information (location, compression vs. absence) can be found in the JSON files. | absent=Absent;present=Present;indeteminate=Indeterminate;uninterpretable=Uninterpretable |
| ERContusion | TRUE | FALSE | text | In emergency room: This variable was assessed by a central review panel, according the TBI-Common Data Elements. This variable indicates whether one or more contusions are present. More descriptive information (location, volume, number) and advanced information (e.g. hemorrhagic, non-hemorrhagic, intracerebral hemorrhage, etc.) can be found in the JSON files. Volume was estimated using the AxBxC/2 method. Edema was included in the measurements. | absent=Absent;present=Present;indeteminate=Indeterminate;uninterpretable=Uninterpretable |
| ERCRFCTAngulation | TRUE | FALSE | integer | In emergency room: This variable describes if a CT scan is performed with or without angulation. There are three options: no angulation (volume scan), orbital-meatal line and other. There is a risk of different interpretation, since there was no definition. | 1=No angulation (volume scan);2=Orbital-meatal line;99=Other |

| Name | Static | Intervention | Format | Description | Possible Values |
|---|---|---|---|---|---|
| ERCRFCTManuf | TRUE | FALSE | text | In emergency room: This variable describes the CT scans manufacturer. | 99=Other;AGFA=Agfa;CARE=Carestream;GE=GE;HITA=Hitachi;KONI=Konica Minolta;PHIL=Philips;SIEM=Siemens;TOSH=Toshiba |
| ERCRFCTMidlineShift | TRUE | FALSE | integer | In emergency room: CT parameters scored by the investigator: reflects if there was midline shift on the CT. | 0=No;1=Yes |
| ERCRFCTMidlineShiftMeasure | TRUE | FALSE | decimal | In emergency room: CT parameters scored by the investigator: reflects the volume in mm if there was a midline shift on the CT Check also "Imaging.CRFCTMidlineShift" | Not Applicable |
| ERCRFCTScannerType | TRUE | FALSE | integer | In emergency room: This variable specifies the type of CT-scanner by the number of slices. | 16=16-slice;32=32-slice;64=64-slice;99=Other;128=128-slice;256=256-slice;320=320-slice |
| ERCRFCTTypeCCT | TRUE | FALSE | integer | In emergency room: This variable describes the type of CT scan that has been made. Multiple options can be selected from: Non-contrast CT, Contrast CT, CT Angiography, Perfusion CT. | 0=No;1=Yes |
| ERCRFCTTypeCTA | TRUE | FALSE | integer | In emergency room: This variable describes the type of CT scan that has been made. Multiple options can be selected from: Non-contrast CT, Contrast CT, CT Angiography, Perfusion CT. | 0=No;1=Yes |
| ERCRFCTTypeNCCT | TRUE | FALSE | integer | In emergency room: This variable describes the type of CT scan that has been made. Multiple options can be selected from: Non-contrast CT, Contrast CT, CT Angiography, Perfusion CT. | 0=No;1=Yes |
| ERCRFCTTypePCT | TRUE | FALSE | integer | In emergency room: This variable describes the type of CT scan that has been made. Multiple options can be selected from: Non-contrast CT, Contrast CT, CT Angiography, Perfusion CT. | 0=No;1=Yes |
| ERCRFForm | TRUE | FALSE | text | In emergency room: Acute or follow up form in the e-CRF. | CTMRI=CT/MRI;FollowUp=Follow-up |
| ERCRFMRIManuf | TRUE | FALSE | text | In emergency room: This variable describes the MRI-scan manufacturer. This variable combines data from CTMRI.MRIManuf and FollowUp.MRIManuf | 99=Other;GE=GE;PHIL=Philips;SIEM=Siemens;TOSH=Toshiba |
| ERCRFMRIResultPreExistAbnorm | TRUE | FALSE | integer | In emergency room: This variable describes whether or not there are pre-existing abnormalities present on MRI scan (yes, no, unknown). Assessment by investigator and or physician. This variable combines data from CTMRI.MRIResultPreExistAbnorm and FollowUp.MRIResultPreExistAbnorm | 0=No;1=Yes;88=Unknown |
| ERCRFMRIResultTraumaticAbnorm | TRUE | FALSE | integer | In emergency room: This variable describes whether or not there are traumatic abnormalities present on MRI scan (yes, no, unknown). Assessment by investigator and/or physician. This variable combines data from CTMRI.MRIResultTraumaticAbnorm and FollowUp.MRIResultTraumaticAbnorm | 0=No;1=Yes;88=Unknown |
| ERCRFMRIScannerStrength | TRUE | FALSE | text | In emergency room: This variable describes the MRI scanner strength. This is a text field. This variable combines data from CTMRI.MRIScannerStrength and FollowUp.MRIScannerStrength | Not Applicable |
| ERCRFMRISequences99 | TRUE | FALSE | integer | In emergency room: This variable describes the MRI sequence. Options are: T1, T2, FLAIR, DWI, GRE, SWI, DTI, MRSI, PWI, Other. Multiple options can be selected. This variable combines data from CTMRI.MRISequences and FollowUp.MRISequences | 0=No;1=Yes |
| ERCRFMRISequencesDTI | TRUE | FALSE | integer | In emergency room: This variable describes the MRI sequence. Options are: T1, T2, FLAIR, DWI, GRE, SWI, DTI, MRSI, PWI, Other. Multiple options can be selected. This variable combines data from CTMRI.MRISequences and FollowUp.MRISequences | 0=No;1=Yes |

| Name | Static | Intervention | Format | Description | Possible Values |
|---|---|---|---|---|---|
| ERCRFMRISequencesDWI | TRUE | FALSE | integer | In emergency room: This variable describes the MRI sequence. Options are: T1, T2, FLAIR, DWI, GRE, SWI, DTI, MRSI, PWI, Other. Multiple options can be selected. This variable combines data from CTMRI.MRISequences and FollowUp.MRISequences | 0=No;1=Yes |
| ERCRFMRISequencesFLAIR | TRUE | FALSE | integer | In emergency room: This variable describes the MRI sequence. Options are: T1, T2, FLAIR, DWI, GRE, SWI, DTI, MRSI, PWI, Other. Multiple options can be selected. This variable combines data from CTMRI.MRISequences and FollowUp.MRISequences | 0=No;1=Yes |
| ERCRFMRISequencesGRE | TRUE | FALSE | integer | In emergency room: This variable describes the MRI sequence. Options are: T1, T2, FLAIR, DWI, GRE, SWI, DTI, MRSI, PWI, Other. Multiple options can be selected. This variable combines data from CTMRI.MRISequences and FollowUp.MRISequences | 0=No;1=Yes |
| ERCRFMRISequencesMRSI | TRUE | FALSE | integer | In emergency room: This variable describes the MRI sequence. Options are: T1, T2, FLAIR, DWI, GRE, SWI, DTI, MRSI, PWI, Other. Multiple options can be selected. This variable combines data from CTMRI.MRISequences and FollowUp.MRISequences | 0=No;1=Yes |
| ERCRFMRISequencesPWI | TRUE | FALSE | integer | In emergency room: This variable describes the MRI sequence. Options are: T1, T2, FLAIR, DWI, GRE, SWI, DTI, MRSI, PWI, Other. Multiple options can be selected. This variable combines data from CTMRI.MRISequences and FollowUp.MRISequences | 0=No;1=Yes |
| ERCRFMRISequencesSWI | TRUE | FALSE | integer | In emergency room: This variable describes the MRI sequence. Options are: T1, T2, FLAIR, DWI, GRE, SWI, DTI, MRSI, PWI, Other. Multiple options can be selected. This variable combines data from CTMRI.MRISequences and FollowUp.MRISequences | 0=No;1=Yes |
| ERCRFMRISequencesT1 | TRUE | FALSE | integer | In emergency room: This variable describes the MRI sequence. Options are: T1, T2, FLAIR, DWI, GRE, SWI, DTI, MRSI, PWI, Other. Multiple options can be selected. This variable combines data from CTMRI.MRISequences and FollowUp.MRISequences | 0=No;1=Yes |
| ERCRFMRISequencesT2 | TRUE | FALSE | integer | In emergency room: This variable describes the MRI sequence. Options are: T1, T2, FLAIR, DWI, GRE, SWI, DTI, MRSI, PWI, Other. Multiple options can be selected. This variable combines data from CTMRI.MRISequences and FollowUp.MRISequences | 0=No;1=Yes |
| ERCRFMRITraumAbnormASDH | TRUE | FALSE | integer | In emergency room: This variable describes whether or not there is an acute subdural hematoma present on MRI scan (yes, no, unknown). Assessment by investigator and/or physician. This variable combines data from CTMRI.MRITraumAbnormASDH and FollowUp.MRITraumAbnormASDH | 0=No;1=Yes;88=Unknown |
| ERCRFMRITraumAbnormContusion | TRUE | FALSE | integer | In emergency room: This variable describes whether or not there is a contusion present on MRI scan (yes, no, unknown). Assessment by investigator and/or physician. This variable combines data from CTMRI.MRITraumAbnormContusion and FollowUp.MRITraumAbnormContusion | 0=No;1=Yes;88=Unknown |
| ERCRFMRITraumAbnormDAI | TRUE | FALSE | integer | In emergency room: This variable describes whether or not there is DAI (diffuse axonal injury) present on MRI scan (yes, no, | 0=No;1=Yes;88=Unknown |

| Name | Static | Intervention | Format | Description | Possible Values |
|---|---|---|---|---|---|
| | | | | unknown). Assessment by investigator and/or physician. This variable combines data from CTMRI.MRITraumAbnormDAI and FollowUp.MRITraumAbnormDAI | |
| ERCRFMRITraumAbnormDAILesionLocBrainstem | TRUE | FALSE | integer | In emergency room: This variable describes whether or not there is a brain stem lesion present on MRI scan (yes, no, unknown). Assessment by investigator and/or physician. This variable combines data from CTMRI.MRITraumAbnormDAILesionLocBrainstem and FollowUp.MRITraumAbnormDAILesionLocBrainstem | 0=No;1=Yes;88=Unknown |
| ERCRFMRITraumAbnormDAILesionLocCorpusCallosum | TRUE | FALSE | integer | In emergency room: This variable describes whether or not there is a corpus callosum lesion present on MRI scan (yes, no, unknown). Assessment by investigator and/or physician. This variable combines data from CTMRI.MRITraumAbnormDAILesionLocCorpusCallosum and FollowUp.MRITraumAbnormDAILesionLocCorpusCallosum | 0=No;1=Yes;88=Unknown |
| ERCRFMRITraumAbnormDAILesionLocDiffuseWhiteMatter | TRUE | FALSE | integer | In emergency room: This variable describes whether or not there is a lesion in diffuse white matter present on MRI scan (yes, no, unknown). Assessment by investigator and/or physician. This variable combines data from CTMRI.MRITraumAbnormDAILesionLocDiffuseWhiteMatter and FollowUp.MRITraumAbnormDAILesionLocDiffuseWhiteMatter | 0=No;1=Yes;88=Unknown |
| ERCRFMRITraumAbnormDAINumLesions | TRUE | FALSE | integer | In emergency room: This variable describes how many DAI lesions are present on MRI scan (1,2,3,4,>=5). Assessment by investigator and/or physician. This variable combines data from CTMRI.MRITraumAbnormDAINumLesions and FollowUp.MRITraumAbnormDAINumLesions | 1=1;2=2;3=3;4=4;5=>= 5 |
| ERCRFMRITraumAbnormEDH | TRUE | FALSE | integer | In emergency room: This variable describes whether or not there is an epidural hematoma present on MRI scan (yes, no, unknown). Assessment by investigator and/or physician. This variable combines data from CTMRI.MRITraumAbnormEDH and FollowUp.MRITraumAbnormEDH | 0=No;1=Yes;88=Unknown |
| ERCRFMRITypeMRA | TRUE | FALSE | integer | In emergency room: This variable describes the type of MRI scan that has been made. Options are: MRI, MRA This variable combines data from CTMRI.MRIType and FollowUp.MRIType | 0=No;1=Yes |
| ERCRFMRITypeMRI | TRUE | FALSE | integer | In emergency room: This variable describes the type of MRI scan that has been made. Options are: MRI, MRA This variable combines data from CTMRI.MRIType and FollowUp.MRIType | 0=No;1=Yes |
| ERCTAcuteSubdurHema | TRUE | FALSE | integer | In emergency room: Assessment by clinician/investigator whether or not in his/her interpretation an acute subdural hematoma is present. Also, the size (not quantified) is requested to be estimated. | 0=No;1=Small;2=Large (mass);88=Unknown |
| ERCTBasalCisternsAbsentCompressed | TRUE | FALSE | integer | In emergency room: Assessment by clinician/investigator whether or not in his/her interpretation the basal cisterns are compressed. | 0=No;1=Yes |
| ERCTContusion | TRUE | FALSE | integer | In emergency room: Assessment by clinician/investigator whether or not in his/her interpretation an intracerebral hematoma/contusion is present. Also, the size (not quantified) is requested to be estimated. | 0=No;1=Small;2=Large (mass);88=Unknown |

| Name | Static | Intervention | Format | Description | Possible Values |
|---|---|---|---|---|---|
| ERCTDeprSkullFract | TRUE | FALSE | integer | In emergency room: Assessment by clinician/investigator whether or not in his/her interpretation a depressed skull fracture is present. In addition, when present, clinician/investigator has to document whether the fracture is associated with an open wound or not (compound vs closed) | 0=No;1=Closed;2=Open (compound) |
| ERCTERReason1 | TRUE | FALSE | integer | In emergency room: WHY question: reason for performing CT; only applicable to initial scan (presentation). | 1=GCS <= 14; 2=GCS = 15 + risk factors; 3=Head wound; 4=Exclusion of abnormalities prior to discharge; 5=Suspicion of maxillofacial injury; 88=Unknown; 99=Other |
| ERCTERReason2 | TRUE | FALSE | integer | In emergency room: WHY question: reason for performing CT; only applicable to initial scan (presentation). | 1=GCS <= 14; 2=GCS = 15 + risk factors; 3=Head wound; 4=Exclusion of abnormalities prior to discharge; 5=Suspicion of maxillofacial injury; 88=Unknown; 99=Other |
| ERCTERReason3 | TRUE | FALSE | integer | In emergency room: WHY question: reason for performing CT; only applicable to initial scan (presentation). | 1=GCS <= 14; 2=GCS = 15 + risk factors; 3=Head wound; 4=Exclusion of abnormalities prior to discharge; 5=Suspicion of maxillofacial injury; 88=Unknown; 99=Other |
| ERCTERReason4 | TRUE | FALSE | integer | In emergency room: WHY question: reason for performing CT; only applicable to initial scan (presentation). | 1=GCS <= 14; 2=GCS = 15 + risk factors; 3=Head wound; 4=Exclusion of abnormalities prior to discharge; 5=Suspicion of maxillofacial injury; 88=Unknown; 99=Other |
| ERCTERReason5 | TRUE | FALSE | integer | In emergency room: WHY question: reason for performing CT; only applicable to initial scan (presentation). | 1=GCS <= 14; 2=GCS = 15 + risk factors; 3=Head wound; 4=Exclusion of abnormalities prior to discharge; 5=Suspicion of maxillofacial injury; 88=Unknown; 99=Other |
| ERCTERReason88 | TRUE | FALSE | integer | In emergency room: WHY question: reason for performing CT; only applicable to initial scan (presentation). | 1=GCS <= 14; 2=GCS = 15 + risk factors; 3=Head wound; 4=Exclusion of abnormalities prior to discharge; 5=Suspicion of maxillofacial injury; 88=Unknown; 99=Other |
| ERCTERReason99 | TRUE | FALSE | integer | In emergency room: WHY question: reason for performing CT; only applicable to initial scan (presentation). | 1=GCS <= 14; 2=GCS = 15 + risk factors; 3=Head wound; 4=Exclusion of abnormalities prior to discharge; 5=Suspicion of maxillofacial injury; 88=Unknown; 99=Other |
| ERCTERReasonOther | TRUE | FALSE | text | In emergency room: Specification, only applicable if "CTMRI.CTERReason" was "other" | Not Applicable |
| ERCTExtraduralHema | TRUE | FALSE | integer | In emergency room: Assessment by clinician/investigator whether or not in his/her interpretation an acute extradural/epidural hematoma is present. Also, the size (not quantified) is requested to be estimated. | 0=No;1=Small;2=Large (mass);88=Unknown |
| ERCTFrames | TRUE | FALSE | | | Not Applicable |
| ERCTICLesionDAI | TRUE | FALSE | integer | In emergency room: Assessment by clinician/investigator whether or not in his/her interpretation diffuse axonal injury is present. | 0=No;1=Yes;88=Unknown |
| ERCTImaging | TRUE | FALSE | | | Not Applicable |
| ERCTNoOpMotiv | TRUE | FALSE | integer | In emergency room: WHY question: documents reason for not having an indication for (intra)cranial surgery. | 0=No surgical lesion;1=Lesion present, but Acceptable/good neurologic condition;2=Lesion present, but Guideline adherence;3=Lesion present, but Little/no mass effect;4=Lesion present, but Not hospital policy;5=Lesion present, but Extremely poor prognosis;6=Lesion present, but Brain Death;7=Lesion present, but Old age;8=Lesion present, but Wish family;88=Unknown;99=Lesion present, but Other |
| ERCTNoOpMotivOther | TRUE | FALSE | text | In emergency room: Specification, only applicable if "CTMRI.CTNoOpMotiv" was "other" | Not Applicable |
| ERCTRiskFactorsERAgeGreatrThanEqual60 | TRUE | FALSE | integer | In emergency room: This variable describes the presence of risk factors (here: age greater than or equal to 60 years) for structural abnormalities on an early CT/MRI. Information for this variable was meant to be only entered for early CT, when CTMRI.CTERReason = "GCS = 15 + risk factors". However, due to the e-CRF lay-out, it will also be populated for patients with CTMRI.CTERReason is not "GCS = 15 + risk factors" or patients with an MRI. | 0=No;1=Yes |
| ERCTRiskFactorsERAlterationOfConsc | TRUE | FALSE | integer | In emergency room: This variable describes the presence of risk factors (here: alteration of consciousness) for structural abnormalities on an initial/ER CT/MRI. Information for this variable was meant to be only entered for initial/ER CT, when CTMRI.CTERReason = "GCS = 15 + risk factors". However, due to the e-CRF lay-out, it will also be populated for patients with | 0=No;1=Yes |

| Name | Static | Intervention | Format | Description | Possible Values |
|---|---|---|---|---|---|
| | | | | CTMRI.CTERReason is not "GCS = 15 + risk factors" or patients with an MRI. | |
| ERCTRiskFactorsERAnticoagTx | TRUE | FALSE | integer | In emergency room: This variable describes the presence of risk factors (here: use of anticoagulant Tx ) for structural abnormalities on an early CT/MRI. Information for this variable was meant to be only entered for early CT, when CTMRI.CTERReason = "GCS = 15 + risk factors". However due to the e-CRF lay-out, it will also be populated for patients with CTMRI.CTERReason is not "GCS = 15 + risk factors" or patients with an MRI. | 0=No;1=Yes |
| ERCTRiskFactorsERAnyNeuroDef | TRUE | FALSE | integer | In emergency room: This variable describes the presence of risk factors (here: 'any neurological deficit') for structural abnormalities on an initial/ER CT/MRI. Information for this variable was meant to be only entered for initial/ER CT, when CTMRI.CTERReason = "GCS = 15 + risk factors". However due to the e-CRF lay-out, it will also be populated for patients with CTMRI.CTERReason is not "GCS = 15 + risk factors" or patients with an MRI. | 0=No;1=Yes |
| ERCTRiskFactorsERClinSignsOfFractSkullBaseVault | TRUE | FALSE | integer | In emergency room: This variable describes the presence of risk factors (here: 'clinical signs of fracture skull base or vault') for structural abnormalities on an initial/ER CT/MRI. Information for this variable was meant to be only entered for initial/ER CT, when CTMRI.CTERReason = "GCS = 15+risk factors". However, due to the e-CRF lay-out, it will also be populated for patients with CTMRI.CTERReason is not "GCS = 15 + risk factors" or patients with an MRI. | 0=No;1=Yes |
| ERCTRiskFactorsERContusionFace | TRUE | FALSE | integer | In emergency room: This variable describes the presence of risk factors (here: 'contusion of the face') for structural abnormalities on an early CT/MRI. Information for this variable was meant to be only entered for early CT, when CTMRI.CTERReason = "GCS = 15 + risk factors". However due to the e-CRF lay-out, it will also be populated for patients with CTMRI.CTERReason is not "GCS = 15 + risk factors" or patients with an MRI. | 0=No;1=Yes |
| ERCTRiskFactorsERFallFromAnyElev | TRUE | FALSE | integer | In emergency room: This variable describes the presence of risk factors (here: fall from any elevation) for structural abnormalities on an early CT/MRI. Information for this variable was meant to be only entered for early CT, when CTMRI.CTERReason = "GCS = 15 + risk factors". However, due to the e-CRF lay-out, it will also be populated for patients with CTMRI.CTERReason is not "GCS = 15 + risk factors" or patients with an MRI. | 0=No;1=Yes |
| ERCTRiskFactorsERHeadache | TRUE | FALSE | integer | In emergency room: This variable describes the presence of risk factors (here: headache) for structural abnormalities on an early CT/MRI. Information for this variable was meant to be only entered for early CT, when CTMRI.CTERReason = "GCS = 15 + risk factors". However, due to the e-CRF lay-out, it will also be populated for patients with CTMRI.CTERReason is not "GCS = 15 + risk factors" or patients with an MRI. | 0=No;1=Yes |
| ERCTRiskFactorsERIntoxication | TRUE | FALSE | integer | In emergency room: This variable describes the presence of risk factors (here: intoxication) for structural abnormalities on an initial/ER CT/MRI. Information for this variable was meant to be only entered for initial/ER CT, when CTMRI.CTERReason = "GCS = 15+risk factors". However due to | 0=No;1=Yes |

| Name | Static | Intervention | Format | Description | Possible Values |
|---|---|---|---|---|---|
| | | | | the e-CRF lay-out, it will also be populated for patients with CTMRI.CTERReason is not "GCS = 15 + risk factors" or patients with an MRI. | |
| ERCTRiskFactorsERLOC | TRUE | FALSE | integer | In emergency room: This variable describes the presence of risk factors (here: loss of consciousness) for structural abnormalities on an initial/ER CT/MRI. Information for this variable was meant to be only entered for initial/ER CT, when CTMRI.CTERReason = "GCS = 15 + risk factors". However, due to the e-CRF lay-out, it will also be populated for patients with CTMRI.CTERReason is not "GCS = 15 + risk factors" or patients with an MRI. | 0=No;1=Yes |
| ERCTRiskFactorsEROther | TRUE | FALSE | integer | In emergency room: This variable describes the presence of risk factors (other reason, not specified elsewhere) for structural abnormalities on an early CT/MRI. Information for this variable was meant to be only entered for early CT, when CTMRI.CTERReason = "GCS = 15 + risk factors". However, due to the e-CRF lay-out, it will also be populated for patients with CTMRI.CTERReason is not "GCS = 15 + risk factors" or patients with an MRI. | 0=No;1=Yes |
| ERCTRiskFactorsEROtherTxt | TRUE | FALSE | text | In emergency room: This variable describes the presence of risk factors (other reason, not specified elsewhere: textfield) for structural abnormalities on an early CT/MRI. Information for this variable was meant to be only entered for early CT, when CTMRI.CTERReason = "GCS = 15 + risk factors". However, due to the e-CRF lay-out, it will also be populated for patients with CTMRI.CTERReason is not "GCS = 15 + risk factors" or patients with an MRI. | Not Applicable |
| ERCTRiskFactorsERPhysEvidTraumaHeadSkull | TRUE | FALSE | integer | In emergency room: This variable describes the presence of risk factors (here: physical evidence of trauma to head/skull) for structural abnormalities on an initial/ER CT/MRI. Information for this variable was meant to be only entered for initial/ER CT, when CTMRI.CTERReason = "GCS = 15 + risk factors". However, due to the e-CRF lay-out, it will also be populated for patients with CTMRI.CTERReason is not "GCS = 15 + risk factors" or patients with an MRI. | 0=No;1=Yes |
| ERCTRiskFactorsERPTAGreatrThanEqual4hrs | TRUE | FALSE | integer | In emergency room: This variable describes the presence of risk factors (here: PTA >= 4 hours) for structural abnormalities on an initial/ER CT/MRI. Information for this variable was meant to be only entered for initial/ER CT, when CTMRI.CTERReason = "GCS = 15 + risk factors". However, due to the e-CRF lay-out, it will also be populated for patients with CTMRI.CTERReason is not "GCS = 15 + risk factors" or patients with an MRI. | 0=No;1=Yes |
| ERCTRiskFactorsERSeizure | TRUE | FALSE | integer | In emergency room: This variable describes the presence of risk factors (here: seizure) for structural abnormalities on an initial/ER CT/MRI. Information for this variable was meant to be only entered for initial/ER CT, when CTMRI.CTERReason = "GCS = 15 + risk factors". However, due to the e-CRF lay-out, it will also be populated for patients with CTMRI.CTERReason is not "GCS = 15 + risk factors" or patients with an MRI. | 0=No;1=Yes |
| ERCTRiskFactorsERSignsFacialFract | TRUE | FALSE | integer | In emergency room: This variable describes the presence of risk factors (here: signs of facial fracture) for structural abnormalities on an initial/ER CT/MRI. Information for this variable was meant to be only entered | 0=No;1=Yes |

| Name | Static | Intervention | Format | Description | Possible Values |
|---|---|---|---|---|---|
| | | | | for initial/ER CT, when CTMRI.CTERReason = "GCS = 15 + risk factors". However, due to the e-CRF lay-out, it will also be populated for patients with CTMRI.CTERReason is not "GCS = 15 + risk factors" or patients with an MRI. | |
| ERCTRiskFactorsER Vomit | TRUE | FALSE | integer | In emergency room: This variable describes the presence of risk factors (here: vomiting) for structural abnormalities on an initial/ER CT/MRI. Information for this variable was meant to be only entered for initial/ER CT, when CTMRI.CTERReason = "GCS = 15 + risk factors". However, due to the e-CRF lay-out, it will also be populated for patients with CTMRI.CTERReason is not "GCS = 15 + risk factors" or patients with an MRI. | 0=No;1=Yes |
| ERCTRiskFactorsER VulnRoadUser | TRUE | FALSE | integer | In emergency room: This variable describes the presence of risk factors (here: vulnerable road users, like pedestrians or cyclists) for structural abnormalities on an early CT/MRI. Information for this variable was meant to be only entered for early CT, when CTMRI.CTERReason = "GCS = 15 + risk factors". However, due to the e-CRF lay-out, it will also be populated for patients with CTMRI.CTERReason is not "GCS = 15 + risk factors" or patients with an MRI. | 0=No;1=Yes |
| ERCTSchedForOp | TRUE | FALSE | integer | In emergency room: Whether or not the patient is scheduled for (intra)cranial surgery. The main aim here is to capture whether the clinical team taking care of the patient sees a neurosurgical indication. | 0=No;1=Yes |
| ERCTSubarachnoidH em | TRUE | FALSE | integer | In emergency room: Assessment by clinician/investigator whether or not in his/her interpretation subarachnoid hemorrhage is present. In addition, the location of the hemorrhage is requested to be answered. | 0=No;1=Basal;2=Cortical;3=Basal and Cortical |
| ERCTYesOpMotiv | TRUE | FALSE | integer | In emergency room: WHY question: documents reason for having an indication for (intra)cranial surgery. | 1=Emergency/life saving;2=Clinical deterioration;3=Mass effect on CT;4=Radiological progression;5=(Suspicion of) raised ICP;6=Guideline adherence;7=To prevent deterioration;8=Depressed skull fracture;99=Other |
| ERCTYesOpMotivOt her | TRUE | FALSE | text | In emergency room: Free text if "CTMRI.CTYesOpMotiv" was marked as 'Other'. Relates to the WHY question: documents reason for having an indication for (intra)cranial surgery. | Not Applicable |
| EREpiduralHematom a | TRUE | FALSE | text | In emergency room: This variable was assessed by a central review panel, according the TBI-Common Data Elements. This variable indicates whether an epidural hematoma or multiple epidural hematomas are present. More descriptive information (location, volume, number) and advanced information (e.g. arterial versus venous) can be found in the JSON files. Note: Volume was estimated using the AxBxC/2 method. In the JSON files, "descriptive_volume" is the estimated volume of the lesion and can be used for analysis, "descriptive_length", "descriptive_width", "descriptive_max_thickness" are measurements that should not be used in any analysis. | absent=Absent;present=Present;indeteminate=Indeterminate;uninterp retable=Uninterpretable |
| ERExtraaxialHemato ma | TRUE | FALSE | text | In emergency room: This variable was assessed by a central review panel, according the TBI-Common Data Elements. This variable indicates whether an extra-axial hematoma or multiple extra-axial hematomas are present. This term was mostly used for bleedings that were difficult to classify or for bleedings that were evacuated (i.e. after craniectomy). More descriptive information (location, volume, | absent=Absent;present=Present;indeteminate=Indeterminate;uninterp retable=Uninterpretable |

| Name | Static | Intervention | Format | Description | Possible Values |
| --- | --- | --- | --- | --- | --- |
| | | | | number) and advanced information (e.g. arterial versus venous) can be found in the JSON files. Note: Volume was estimated using the AxBxC/2 method. In the JSON files, "descriptive_volume" is the estimated volume of the lesion and can be used for analysis, "descriptive_length", "descriptive_width", "descriptive_max_thickness" are measurements that should not be used in any analysis. | |
| ERFisherClassification | TRUE | FALSE | integer | In emergency room: S: A CT grading scale for traumatic Subarachnoid Hemorrhage B: Useful in predicting the occurrence and severity of cerebral vasospasm. Ranges from 1-4 (No tSAH, no IVH (1), no IVH, trace or moderate tSAH (can be in multiple locations) (2), No IVH, full tSAH (3), IVH (4) R: Only available for CT scans. For each patient, the FisherClassification can be found in the Imaging.LesionData variable. For CENTER-TBI, only the older version of Fisher was scored, not the newer "modified" version. | 0=grade 0;1=grade 1;2=grade 2;3=grade 3;4=grade 4 |
| ERGreenCTGrade | TRUE | FALSE | integer | In emergency room: A CT grading scale for traumatic Subarachnoid Hemorrhage (tSAH). Useful for outcome prediction. Ranges from 1-4, with a subdivision of 3 into "31", "32" and 4 into "41" and "42". (thin tSAH (1), thick tSAH (2), thin tSAH with mass lesion and no MLS (31), thin tSAH with mass lesion and MLS (32), thick tSAH with mass lesion and no MLS (41), thick tSAH with mass lesion and MLS (42). Only available for CT scans. For each patient, the GreenCTgrade can be found in the Imaging.LesionData variable. Could only be filled out when tSAH was present. | Not Applicable |
| ERHelsinkiCTScore | TRUE | FALSE | integer | In emergency room: S: A general CT grading scale B: Useful for outcome prediction. Ranges from -3 to 14. Scoring is done as follows: subdural (+2), intracerebral hematoma (+2), epidural hematoma (-3), mass lesion size >25 cc (+2), IVH (+3), compressed cisterns (+1), obliterated cisterns (+5) R: Only available for CT scans. For each patient, the HelsinkiCTscore can be found in the Imaging.LesionData variable. Empty values for the HelsinkiCTscore stand for a normal CT scan! A score of 0 does not necessarily mean the patient had a normal CT. (e.g. epidural hematoma -3, + IVH = 3 = 0). | Not Applicable |
| ERIntraventricularHemorrhage | TRUE | FALSE | text | In emergency room: This variable was assessed by a central review panel, according the TBI-Common Data Elements. This variable indicates whether intraventricular blood is present. More descriptive information (location) can be found in the JSON files. | absent=Absent;present=Present;indeteminate=Indeterminate;uninterpretable=Uninterpretable |
| ERLesionData | TRUE | FALSE | text | In emergency room: This variable contains ALL lesion information as assessed by central review. There are 23 CDEs that have been evaluated for each patient and 6 classifications that have been filled out. For 13 CDEs and all 6 classifications separate variables have been made available. (see "imaging." variables: SkullFracture, Mass lesion, ExtraaxialHematoma, EpiduralHematoma, SubduralHematomaAcute, SubduralHematomaSubacuteChronic, SubduralCollectionMixedDensity, Contusion, TAI, | Not Applicable |

| Name | Static | Intervention | Format | Description | Possible Values |
|---|---|---|---|---|---|
| | | | | traumaticSubarachnoidHemorrhage, IntraventricularHemorrhage, MidlineShift, CisternalCompression. A derived variable was also created with AnyIntracranTraumaticAbnormality). These 13 CDEs only contain basic information (i.e. lesion is present, absent, indeterminate, uninterpretable, not interpreted). | |
| ERMarshallCTClassification | TRUE | FALSE | integer | In emergency room: A general CT grading scale. Useful for outcome prediction. Ranges from 1 to 6 in our dataset. (No visible pathology on CT (1), Cisterns present, MLS < 5 mm (2), Cisterns compressed or absent, MLS < 5 mm (3), MLS > 5 mm, no mass lesion > 25 cc (4), Evacuated mass lesion (5), Non-evacuated mass lesion (6)) Only available for CT scans. For each patient, the MarshallCTClassification can be found in a Imaging.LesionData variable. For the Marshall CT classification the admission scan was scored. (which is not necessarily the "worst"). | 1=Diffuse injury I (no visible pathology);2=Diffuse injury II;3=Diffuse injury III (swelling);4=Diffuse injury IV (shift);5=Evacuated mass lesion V;6=Non-evacuated mass lesion VI |
| ERMassLesion | TRUE | FALSE | text | In emergency room: This variable was assessed by a central review panel, according the TBI-Common Data Elements. This variable indicates whether a mass lesion is present. "Mass Lesion" in this case was defined as a total brain lesion volume > 25 cc. Note that this can mean that there is at least one large lesion or multiple co-existing lesions (contusions, subdural hematomas, etc.) that add up to > 25 cc. More information can be found in the JSON files. | absent=Absent;present=Present;indeteminate=Indeterminate;uninterpretable=Uninterpretable |
| ERMidlineShift | TRUE | FALSE | text | In emergency room: This variable was assessed by a central review panel, according the TBI-Common Data Elements. This variable indicates whether a midline shift of > 5 mm is present. More descriptive information (direction) can be found in the JSON files. | absent=Absent;present=Present;indeteminate=Indeterminate;uninterpretable=Uninterpretable |
| ERMorrisMarshallClassification | TRUE | FALSE | integer | In emergency room: S: A CT grading scale for traumatic Subarachnoid Hemorrhage (tSAH) B: Useful for outcome prediction. Ranges from 0-4. (no tSAH (0), trace or moderate tSAH in one location (basal, cortical or tentorial) (1), one location full, or 2 not full (2), two locations full (3), 3 locations or more (4)) R: Only available for CT scans. For each patient, the MorrisMarshallClassification can be found in a Imaging.LesionData variable. | Not Applicable |
| ERMRFrames | TRUE | FALSE | | | Not Applicable |
| ERMRIERReason88 | TRUE | FALSE | integer | In emergency room: This variable contains the main reason why an MRI, ultra-early MR (within 72 hrs), was performed. Possible options are ER only: Discrepancy between clinical symptomatology and (lack of) CT abnormalities; suspicion non¬≠metal foreign object; instead of CT (limiting radiation exposure); suspicion spinal cord lesion; unknown; other (further specified in text field CTMRI.MRIERReasonOther. Reasons for clinical MRI's can be found in variable CTMRI.MRIReason | 1=ER only: Discrepancy between clinical symptomatology and (lack of) CT abnormalities; 2=ER only: Suspicion non-metal foreign object; 3=ER only: Instead of CT (limiting radiation exposure); 4=ER only: Suspicion spinal cord lesion; 88=Unknown; 99=Other |
| ERMRIERReason99 | TRUE | FALSE | integer | In emergency room: This variable contains the main reason why an MRI, ultra-early MR (within 72 hrs), was performed. Possible options are ER only: Discrepancy between clinical symptomatology and (lack of) CT abnormalities; suspicion non¬≠metal foreign object; instead of CT (limiting radiation exposure); suspicion spinal cord lesion; unknown; other (further specified in | 1=ER only: Discrepancy between clinical symptomatology and (lack of) CT abnormalities; 2=ER only: Suspicion non-metal foreign object; 3=ER only: Instead of CT (limiting radiation exposure); 4=ER only: Suspicion spinal cord lesion; 88=Unknown; 99=Other |

| Name | Static | Intervention | Format | Description | Possible Values |
|---|---|---|---|---|---|
| | | | | text field CTMRI.MRIERReasonOther. Reasons for clinical MRI's can be found in variable CTMRI.MRIReason | |
| ERMRIERReasonER1 | TRUE | FALSE | integer | In emergency room: This variable contains the main reason why an MRI, ultra-early MR (within 72 hrs), was performed. Possible options are ER only: Discrepancy between clinical symptomatology and (lack of) CT abnormalities; suspicion non¬≠metal foreign object; instead of CT (limiting radiation exposure); suspicion spinal cord lesion; unknown; other (further specified in text field CTMRI.MRIERReasonOther. Reasons for clinical MRI's can be found in variable CTMRI.MRIReason | 1=ER only: Discrepancy between clinical symptomatology and (lack of) CT abnormalities; 2=ER only: Suspicion non-metal foreign object; 3=ER only: Instead of CT (limiting radiation exposure); 4=ER only: Suspicion spinal cord lesion; 88=Unknown; 99=Other |
| ERMRIERReasonER2 | TRUE | FALSE | integer | In emergency room: This variable contains the main reason why an MRI, ultra-early MR (within 72 hrs), was performed. Possible options are ER only: Discrepancy between clinical symptomatology and (lack of) CT abnormalities; suspicion non¬≠metal foreign object; instead of CT (limiting radiation exposure); suspicion spinal cord lesion; unknown; other (further specified in text field CTMRI.MRIERReasonOther. Reasons for clinical MRI's can be found in variable CTMRI.MRIReason | 1=ER only: Discrepancy between clinical symptomatology and (lack of) CT abnormalities; 2=ER only: Suspicion non-metal foreign object; 3=ER only: Instead of CT (limiting radiation exposure); 4=ER only: Suspicion spinal cord lesion; 88=Unknown; 99=Other |
| ERMRIERReasonER3 | TRUE | FALSE | integer | In emergency room: This variable contains the main reason why an MRI, ultra-early MR (within 72 hrs), was performed. Possible options are ER only: Discrepancy between clinical symptomatology and (lack of) CT abnormalities; suspicion non¬≠metal foreign object; instead of CT (limiting radiation exposure); suspicion spinal cord lesion; unknown; other (further specified in text field CTMRI.MRIERReasonOther. Reasons for clinical MRI's can be found in variable CTMRI.MRIReason | 1=ER only: Discrepancy between clinical symptomatology and (lack of) CT abnormalities; 2=ER only: Suspicion non-metal foreign object; 3=ER only: Instead of CT (limiting radiation exposure); 4=ER only: Suspicion spinal cord lesion; 88=Unknown; 99=Other |
| ERMRIERReasonER4 | TRUE | FALSE | integer | In emergency room: This variable contains the main reason why an MRI, ultra-early MR (within 72 hrs), was performed. Possible options are ER only: Discrepancy between clinical symptomatology and (lack of) CT abnormalities; suspicion non¬≠metal foreign object; instead of CT (limiting radiation exposure); suspicion spinal cord lesion; unknown; other (further specified in text field CTMRI.MRIERReasonOther. Reasons for clinical MRI's can be found in variable CTMRI.MRIReason | 1=ER only: Discrepancy between clinical symptomatology and (lack of) CT abnormalities; 2=ER only: Suspicion non-metal foreign object; 3=ER only: Instead of CT (limiting radiation exposure); 4=ER only: Suspicion spinal cord lesion; 88=Unknown; 99=Other |
| ERMRImaging | TRUE | FALSE | | | Not Applicable |
| ERRotterdamCTScore | TRUE | FALSE | integer | In emergency room: S: A general CT grading scale B: Useful for outcome prediction. Ranges from 1 to 6 in our dataset. Scoring was as follows: IVH or tSAH (+1), Epidural Mass not present (+1), MLS > 5 mm (+1), cisterns compressed (+1), cisterns absent (+1), SUM SCORE = +1 R: Only available for CT scans. For each patient, the RotterdamCTscore can be found in the Imaging.LesionData variable. | 1=Diffuse injury I (no visible pathology);2=Diffuse injury II;3=Diffuse injury III (swelling);4=Diffuse injury IV (shift);5=Evacuated mass lesion V;6=Non-evacuated mass lesion VI |
| ERScanType | TRUE | FALSE | text | In emergency room: S: MR Mapping for series description B: Since all centers use different series description, a mapping for T2, T1, FLAIR, T2*, DTI and rs-fMRI is provided. R: For CT, ScanType and Series Description should be the same. | Not Applicable |
| ERSeriesDescription | TRUE | FALSE | text | In emergency room: S: Scan Series description retrieved from DICOM header A: Serie Name Given by the image acquisition center. R: DICOM Tag (0008,103E) Series Description | Not Applicable |

| Name | Static | Intervention | Format | Description | Possible Values |
|---|---|---|---|---|---|
| ERSkullFracture | TRUE | FALSE | text | In emergency room: This variable was assessed by a central review panel, according the TBI-Common Data Elements. This variable indicates whether a skull fracture or multiple skull fractures are present. More descriptive information (location, number) and advanced information (morphology: e.g. depressed, compound, etc.) can be found in the JSON files. | absent=Absent;present=Present;indeteminate=Indeterminate;uninterpretable=Uninterpretable |
| ERSubduralCollectionMixedDensity | TRUE | FALSE | text | In emergency room: This variable was assessed by a central review panel, according the TBI-Common Data Elements. This variable indicates whether one or multiple subdural collections/mixed density hematomas are present. More descriptive information (location, volume, number) and advanced information (e.g. isodensity, hypodensity, acute on chronic, chronic recurrent etc.) can be found in the JSON files. Volume was estimated using the AxBxC/2 method. Note: cfr. epidural and subdural hematoms regarding volumes. Note: In the JSON files length, width and depth of lesions can not be used as metrics in any analysis! Only "descriptive volume" is a valid metric (in cc) | absent=Absent;present=Present;indeteminate=Indeterminate;uninterpretable=Uninterpretable |
| ERSubduralHematomaAcute | TRUE | FALSE | text | In emergency room: This variable was assessed by a central review panel, according the TBI-Common Data Elements. This variable indicates whether an acute subdural hematoma or multiple acute subdural hematomas are present. More descriptive information (location, volume, number) and advanced information (homogeneous versus heterogeneous) can be found in the JSON files.Volume was estimated using the AxBxC/2 method, however, small traces of acute subdural blood on the tentorium or interhemispheric were not measured. Note: cfr. epidural: only "descriptive_volume" can be used for analysis. Note: In the JSON files length, width and depth of lesions can not be used as metrics in any analysis! Only "descriptive volume" is a valid metric (in cc) | absent=Absent;present=Present;indeteminate=Indeterminate;uninterpretable=Uninterpretable |
| ERSubduralHematomaSubacuteChronic | TRUE | FALSE | text | In emergency room: This variable was assessed by a central review panel, according the TBI-Common Data Elements. This variable indicates whether one or multiple subacute/chronic subdural hematoma(s) are present. More descriptive information (location, volume, number) can be found in the JSON files.Note: Volume was estimated using the AxBxC/2 method. Note: this variable is best considered together with the "SubduralCollectionMixedDensity" variable, as the central reviewer did not have information to time of injury. Many subacute/chronic subdurals were therefore categorised as "mixed density". Note: cfr. epidural and acute subdural hematoma regarding volumes. Note: In the JSON files length, width and depth of lesions can not be used as metrics in any analysis! Only "descriptive volume" is a valid metric (in cc) | absent=Absent;present=Present;indeteminate=Indeterminate;uninterpretable=Uninterpretable |
| ERTAI | TRUE | FALSE | text | In emergency room: This variable was assessed by a central review panel, according the TBI-Common Data Elements. This variable indicates whether axonal injury is present. More descriptive information (location, TAI versus DAI) can be found in the JSON files. Note: be aware | absent=Absent;present=Present;indeteminate=Indeterminate;uninterpretable=Uninterpretable |

| Name | Static | Intervention | Format | Description | Possible Values |
|---|---|---|---|---|---|
| | | | | of the modality you are interested in (i.e. CT or MRI). | |
| ERTimepoint | TRUE | FALSE | text | In emergency room: Outcome timepoints are 2-3 weeks, 3 months, 6 months, 12 months and 24 months. Depending on the stratum and sub-studies of a patient, follow up was performed at some of these time points or at all timepoints. For an overview, please check the SOP Manual. | Clinical=Clinical;CT Early=CT Early;CT Followup=CT Follow-up;CT Post-Op=CT Post-Op;MR 12 months=MR 12 months;MR 2 weeks=MR 2 weeks;MR 24 months=MR 24 months;MR 3 months=MR 3 months;MR 6 months=MR 6 months;MR Early=MR Early |
| ERTraumaticSubarachnoidHemorrhage | TRUE | FALSE | text | In emergency room: This variable was assessed by a central review panel, according the TBI-Common Data Elements. This variable indicates whether traumatic subarachnoid blood is present. More descriptive information (location and amount (e.g. trace, moderate, severe)) can be found in the JSON files. * Location can be: cortical, basal, interhemispheric or tentorial. | absent=Absent;present=Present;indeteminate=Indeterminate;uninterpretable=Uninterpretable |
| ERWindowDetectionQuality | TRUE | FALSE | text | In emergency room: S: Automated Detection for bone versus brain. B: Possible options: Bone Window, Brain Window, Window Uncertainty R: Only available for CT scans. No brain window scans should be classified as bone window. | Not Applicable |
| ERXsiType | TRUE | FALSE | text | In emergency room: S: Type of imaging (CT/MR) A: Assess if the scan is CT or MR R: Possible values 'xnat:mrSessionData' and 'xnat:ctSessionData' | xnat:ctSessionData=xnat:ctSessionData;xnat:mrSessionData=xnat:mrSessionData |
| ExtraaxialHematoma | FALSE | FALSE | text | This variable was assessed by a central review panel, according the TBI-Common Data Elements. This variable indicates whether an extra-axial hematoma or multiple extra-axial hematomas are present. This term was mostly used for bleedings that were difficult to classify or for bleedings that were evacuated (i.e. after craniectomy). More descriptive information (location, volume, number) and advanced information (e.g. arterial versus venous) can be found in the JSON files. Note: Volume was estimated using the AxBxC/2 method. In the JSON files, "descriptive_volume" is the estimated volume of the lesion and can be used for analysis, "descriptive_length", "descriptive_width", "descriptive_max_thickness" are measurements that should not be used in any analysis. | absent=Absent;present=Present;indeteminate=Indeterminate;uninterpretable=Uninterpretable |
| FisherClassification | FALSE | FALSE | integer | S: A CT grading scale for traumatic Subarachnoid Hemorrhage B: Useful in predicting the occurrence and severity of cerebral vasospasm. Ranges from 1-4 (No tSAH, no IVH (1), no IVH, trace or moderate tSAH (can be in multiple locations) (2), No IVH, full tSAH (3), IVH (4) R: Only available for CT scans. For each patient, the FisherClassification can be found in the Imaging.LesionData variable. For CENTER-TBI, only the older version of Fisher was scored, not the newer "modified" version. | 0=grade 0;1=grade 1;2=grade 2;3=grade 3;4=grade 4 |
| GreenCTGrade | FALSE | FALSE | integer | A CT grading scale for traumatic Subarachnoid Hemorrhage (tSAH). Useful for outcome prediction. Ranges from 1-4, with a subdivision of 3 into "31", "32" and 4 into "41" and "42". (thin tSAH (1), thick tSAH (2), thin tSAH with mass lesion and no MLS (31), thin tSAH with mass lesion and MLS (32), thick tSAH with mass lesion and no MLS (41), thick tSAH with mass lesion and MLS (42). Only available for CT scans. For each patient, the GreenCTgrade can be found in the Imaging.LesionData | Not Applicable |

| Name | Static | Intervention | Format | Description | Possible Values |
|---|---|---|---|---|---|
| | | | | variable. Could only be filled out when tSAH was present. | |
| HelsinkiCTScore | FALSE | FALSE | integer | S: A general CT grading scale B: Useful for outcome prediction. Ranges from -3 to 14. Scoring is done as follows: subdural (+2), intracerebral hematoma (+2), epidural hematoma (-3), mass lesion size >25 cc (+2), IVH (+3), compressed cisterns (+1), obliterated cisterns (+5) R: Only available for CT scans. For each patient, the HelsinkiCTscore can be found in the Imaging.LesionData variable. Empty values for the HelsinkiCTscore stand for a normal CT scan! A score of 0 does not necessarily mean the patient had a normal CT. (e.g. epidural hematoma -3, + IVH = 3 = 0). | Not Applicable |
| IntraventricularHemorrhage | FALSE | FALSE | text | This variable was assessed by a central review panel, according the TBI-Common Data Elements. This variable indicates whether intraventricular blood is present. More descriptive information (location) can be found in the JSON files. | absent=Absent;present=Present;indeteminate=Indeterminate;uninterpretable=Uninterpretable |
| LesionData | FALSE | FALSE | text | This variable contains ALL lesion information as assessed by central review. There are 23 CDEs that have been evaluated for each patient and 6 classifications that have been filled out. For 13 CDEs and all 6 classifications separate variables have been made available. (see "imaging." variables: SkullFracture, Mass lesion, ExtraaxialHematoma, EpiduralHematoma, SubduralHematomaAcute, SubduralHematomaSubacuteChronic, SubduralCollectionMixedDensity, Contusion, TAI, traumaticSubarachnoidHemorrhage, IntraventricularHemorrhage, MidlineShift, CisternalCompression. A derived variable was also created with AnyIntracranTraumaticAbnormality). These 13 CDEs only contain basic information (i.e. lesion is present, absent, indeterminate, uninterpretable, not interpreted). | Not Applicable |
| MarshallCTClassification | FALSE | FALSE | integer | A general CT grading scale. Useful for outcome prediction. Ranges from 1 to 6 in our dataset. (No visible pathology on CT (1), Cisterns present, MLS < 5 mm (2), Cisterns compressed or absent, MLS < 5 mm (3), MLS > 5 mm, no mass lesion > 25 cc (4), Evacuated mass lesion (5), Non-evacuated mass lesion (6)) Only available for CT scans. For each patient, the MarshallCTClassification can be found in a Imaging.LesionData variable. For the Marshall CT classification the admission scan was scored. (which is not necessarily the "worst"). | 1=Diffuse injury I (no visible pathology);2=Diffuse injury II;3=Diffuse injury III (swelling);4=Diffuse injury IV (shift);5=Evacuated mass lesion V;6=Non-evacuated mass lesion VI |
| MassLesion | FALSE | FALSE | text | This variable was assessed by a central review panel, according the TBI-Common Data Elements. This variable indicates whether a mass lesion is present. "Mass Lesion" in this case was defined as a total brain lesion volume > 25 cc. Note that this can mean that there is at least one large lesion or multiple co-existing lesions (contusions, subdural hematomas, etc.) that add up to > 25 cc. More information can be found in the JSON files. | absent=Absent;present=Present;indeteminate=Indeterminate;uninterpretable=Uninterpretable |
| MidlineShift | FALSE | FALSE | text | This variable was assessed by a central review panel, according the TBI-Common Data Elements. This variable indicates whether a midline shift of > 5 mm is present. More descriptive information (direction) can be found in the JSON files. | absent=Absent;present=Present;indeteminate=Indeterminate;uninterpretable=Uninterpretable |

| Name | Static | Intervention | Format | Description | Possible Values |
|---|---|---|---|---|---|
| MorrisMarshallClassification | FALSE | FALSE | integer | S: A CT grading scale for traumatic Subarachnoid Hemorrhage (tSAH) B: Useful for outcome prediction. Ranges from 0-4. (no tSAH (0), trace or moderate tSAH in one location (basal, cortical or tentorial) (1), one location full, or 2 not full (2), two locations full (3), 3 locations or more (4)) R: Only available for CT scans. For each patient, the MorrisMarshallClassification can be found in a Imaging.LesionData variable. | Not Applicable |
| MRFrames | FALSE | FALSE | | | Not Applicable |
| RotterdamCTScore | FALSE | FALSE | integer | S: A general CT grading scale B: Useful for outcome prediction. Ranges from 1 to 6 in our dataset. Scoring was as follows: IVH or tSAH (+1), Epidural Mass not present (+1), MLS > 5 mm (+1), cisterns compressed (+1), cisterns absent (+1), SUM SCORE = +1 R: Only available for CT scans. For each patient, the RotterdamCTscore can be found in the Imaging.LesionData variable. | 1=Diffuse injury I (no visible pathology);2=Diffuse injury II;3=Diffuse injury III (swelling);4=Diffuse injury IV (shift);5=Evacuated mass lesion V;6=Non-evacuated mass lesion VI |
| ScanType | FALSE | FALSE | text | S: MR Mapping for series description B: Since all centers use different series description, a mapping for T2, T1, FLAIR, T2*, DTI and rs-fMRI is provided. R: For CT, ScanType and Series Description should be the same. | Not Applicable |
| SeriesDescription | FALSE | FALSE | text | S: Scan Series description retrieved from DICOM header A: Serie Name Given by the image acquisition center. R: DICOM Tag (0008,103E) Series Description | Not Applicable |
| SkullFracture | FALSE | FALSE | text | This variable was assessed by a central review panel, according the TBI-Common Data Elements. This variable indicates whether a skull fracture or multiple skull fractures are present. More descriptive information (location, number) and advanced information (morphology: e.g. depressed, compound, etc.) can be found in the JSON files. | absent=Absent;present=Present;indeteminate=Indeterminate;uninterpretable=Uninterpretable |
| SubduralCollectionMixedDensity | FALSE | FALSE | text | This variable was assessed by a central review panel, according the TBI-Common Data Elements. This variable indicates whether one or multiple subdural collections/mixed density hematomas are present. More descriptive information (location, volume, number) and advanced information (e.g. isodensity, hypodensity, acute on chronic, chronic recurrent etc.) can be found in the JSON files. Volume was estimated using the AxBxC/2 method. Note: cfr. epidural and subdural hematoms regarding volumes. Note: In the JSON files length, width and depth of lesions can not be used as metrics in any analysis! Only "descriptive volume" is a valid metric (in cc) | absent=Absent;present=Present;indeteminate=Indeterminate;uninterpretable=Uninterpretable |
| SubduralHematomaAcute | FALSE | FALSE | text | This variable was assessed by a central review panel, according the TBI-Common Data Elements. This variable indicates whether an acute subdural hematoma or multiple acute subdural hematomas are present. More descriptive information (location, volume, number) and advanced information (homogeneous versus heterogeneous) can be found in the JSON files.Volume was estimated using the AxBxC/2 method, however, small traces of acute subdural blood on the tentorium or interhemispheric were not measured. Note: cfr. epidural: only "descriptive_volume" can be used for analysis. Note: In the JSON files length, width and depth of lesions can not be | absent=Absent;present=Present;indeteminate=Indeterminate;uninterpretable=Uninterpretable |

| Name | Static | Intervention | Format | Description | Possible Values |
| --- | --- | --- | --- | --- | --- |
| | | | | used as metrics in any analysis! Only "descriptive volume" is a valid metric (in cc) | |
| SubduralHematomaSubacuteChronic | FALSE | FALSE | text | This variable was assessed by a central review panel, according the TBI-Common Data Elements. This variable indicates whether one or multiple subacute/chronic subdural hematoma(s) are present. More descriptive information (location, volume, number) can be found in the JSON files.Note: Volume was estimated using the AxBxC/2 method. Note: this variable is best considered together with the "SubduralCollectionMixedDensity" variable, as the central reviewer did not have information to time of injury. Many subacute/chronic subdurals were therefore categorised as "mixed density". Note: cfr. epidural and acute subdural hematoma regarding volumes. Note: In the JSON files length, width and depth of lesions can not be used as metrics in any analysis! Only "descriptive volume" is a valid metric (in cc) | absent=Absent;present=Present;indeteminate=Indeterminate;uninterpretable=Uninterpretable |
| TAI | FALSE | FALSE | text | This variable was assessed by a central review panel, according the TBI-Common Data Elements. This variable indicates whether axonal injury is present. More descriptive information (location, TAI versus DAI) can be found in the JSON files. Note: be aware of the modality you are interested in (i.e. CT or MRI). | absent=Absent;present=Present;indeteminate=Indeterminate;uninterpretable=Uninterpretable |
| Timepoint | FALSE | FALSE | text | Outcome timepoints are 2-3 weeks, 3 months, 6 months, 12 months and 24 months. Depending on the stratum and sub-studies of a patient, follow up was performed at some of these time points or at all timepoints. For an overview, please check the SOP Manual. | Clinical=Clinical;CT Early=CT Early;CT Followup=CT Follow-up;CT Post-Op=CT Post-Op;MR 12 months=MR 12 months;MR 2 weeks=MR 2 weeks;MR 24 months=MR 24 months;MR 3 months=MR 3 months;MR 6 months=MR 6 months;MR Early=MR Early |
| Timepoint | FALSE | FALSE | text | Outcome timepoints are 2-3 weeks, 3 months, 6 months, 12 months and 24 months. Depending on the stratum and sub-studies of a patient, follow up was performed at some of these time points or at all timepoints. For an overview, please check the SOP Manual. | 12mo=12 months;24mo=24 months;2wk=2 weeks;3mo=3 months;6mo=6 months;Base=Baseline |
| TraumaticSubarachnoidHemorrhage | FALSE | FALSE | text | This variable was assessed by a central review panel, according the TBI-Common Data Elements. This variable indicates whether traumatic subarachnoid blood is present. More descriptive information (location and amount (e.g. trace, moderate, severe)) can be found in the JSON files. * Location can be: cortical, basal, interhemispheric or tentorial. | absent=Absent;present=Present;indeteminate=Indeterminate;uninterpretable=Uninterpretable |
| WindowDetectionQuality | FALSE | FALSE | text | S: Automated Detection for bone versus brain. B: Possible options: Bone Window, Brain Window, Window Uncertainty R: Only available for CT scans. No brain window scans should be classified as bone window. | Not Applicable |
| XsiType | FALSE | FALSE | text | S: Type of imaging (CT/MR) A: Assess if the scan is CT or MR R: Possible values 'xnat:mrSessionData' and 'xnat:ctSessionData' | xnat:ctSessionData=xnat:ctSessionData;xnat:mrSessionData=xnat:mrSessionData |

*Laboratory measurements*

| Name | Static | Intervention | Format | Description | Possible Values |
| --- | --- | --- | --- | --- | --- |
| AnnexinVsingleCD105Annexmeasurement | FALSE | FALSE | decimal | Measurement of CD105- and AnnexinV-positive microparticles for quantification of endothelial derived microparticles (EDMP) - results for endothelial derived microparticles (EDMP) (single positive Annexin V+) - using flow cytometry techniques. Citrated plasma sample was centrifuged for 20 min at | Not Applicable |

| Name | Static | Intervention | Format | Description | Possible Values |
|---|---|---|---|---|---|
| | | | | 2,500 x g at RT. Supernatants were used to determine EDMP. Run on the BD Accuri C6 Plus flow cytometer (BD Biosciences; Heidelberg, Germany) at Institute for Research in Operative Medicine (IFOM, Cologne, Germany), University Witten/Herdecke. | |
| AnnexinVsingleCD42bAnnexmeasurement | FALSE | FALSE | decimal | Measurement of CD42b- and AnnexinV-positive microparticles for quantification of platelet-derived microparticles (PDMP) - results for platelet-derived microparticles (PDMP) (single positive Annexin V+) - using flow cytometry techniques. Citrated plasma sample was centrifuged for 20 min at 2,500 x g at RT. Supernatants were used to determine PDMP. Run on the BD Accuri C6 Plus flow cytometer (BD Biosciences; Heidelberg, Germany) at Institute for Research in Operative Medicine (IFOM, Cologne, Germany), University Witten/Herdecke. | Not Applicable |
| CD105AnnexinVdoubleCD105Annexmeasurement | FALSE | FALSE | decimal | Measurement of CD105- and AnnexinV-positive microparticles for quantification of endothelial derived microparticles (EDMP) - results for endothelial derived microparticles (EDMP) (double positive CD105+/Annexin V+) - using flow cytometry techniques. Citrated plasma sample was centrifuged for 20 min at 2,500 x g at RT. Supernatants were used to determine EDMP. Run on the BD Accuri C6 Plus flow cytometer (BD Biosciences; Heidelberg, Germany) at Institute for Research in Operative Medicine (IFOM, Cologne, Germany), University Witten/Herdecke. | Not Applicable |
| CD105CD142doubleCD105CD142measurement | FALSE | FALSE | decimal | Measurement of CD105- and CD142-positive microparticles for quantification of endothelial derived microparticles (EDMP) - results for endothelial derived microparticles (EDMP) (double positive CD105+/CD142+) - using flow cytometry techniques. Citrated plasma sample was centrifuged for 20 min at 2,500 x g at RT. Supernatants were used to determine EDMP. Run on the BD Accuri C6 Plus flow cytometer (BD Biosciences; Heidelberg, Germany) at Institute for Research in Operative Medicine (IFOM, Cologne, Germany), University Witten/Herdecke. | Not Applicable |
| CD105CD62EdoubleCD105CD62Emeasurement | FALSE | FALSE | decimal | Measurement of CD105- and CD62e-positive microparticles for quantification of endothelial derived microparticles (EDMP) - results for endothelial derived microparticles (EDMP) (double positive CD105+/CD62e+) - using flow cytometry techniques. Citrated plasma sample was centrifuged for 20 min at 2,500 x g at RT. Supernatants were used to determine EDMP. Run on the BD Accuri C6 Plus flow cytometer (BD Biosciences; Heidelberg, Germany) at Institute for Research in Operative Medicine (IFOM, Cologne, Germany), University Witten/Herdecke.. | Not Applicable |
| CD105singleCD105Annexmeasurement | FALSE | FALSE | decimal | Measurement of CD105- and AnnexinV-positive microparticles for quantification of endothelial derived microparticles (EDMP) - results for endothelial derived microparticles (EDMP) (single positive CD105+) - using flow cytometry techniques. Citrated plasma sample was centrifuged for 20 min at 2,500 x g at RT. Supernatants were used to determine EDMP. Run on the BD Accuri C6 Plus flow cytometer (BD Biosciences; Heidelberg, Germany) at Institute for | Not Applicable |

| Name | Static | Intervention | Format | Description | Possible Values |
| --- | --- | --- | --- | --- | --- |
| | | | | Research in Operative Medicine (IFOM, Cologne, Germany), University Witten/Herdecke. | |
| CD105singleCD105CD142measurement | FALSE | FALSE | decimal | Measurement of CD105- and CD142-positive microparticles for quantification of endothelial derived microparticles (EDMP) - results for endothelial derived microparticles (EDMP) (single positive CD105+) - using flow cytometry techniques. Citrated plasma sample was centrifuged for 20 min at 2,500 x g at RT. Supernatants were used to determine EDMP. Run on the BD Accuri C6 Plus flow cytometer (BD Biosciences; Heidelberg, Germany) at Institute for Research in Operative Medicine (IFOM, Cologne, Germany), University Witten/Herdecke. | Not Applicable |
| CD105singleCD105CD62Emeasurement | FALSE | FALSE | decimal | Measurement of CD105- and CD62e-positive microparticles for quantification of endothelial derived microparticles (EDMP) - results for endothelial derived microparticles (EDMP) (single positive CD105+) - using flow cytometry techniques. Citrated plasma sample was centrifuged for 20 min at 2,500 x g at RT. Supernatants were used to determine EDMP. Run on the BD Accuri C6 Plus flow cytometer (BD Biosciences; Heidelberg, Germany) at Institute for Research in Operative Medicine (IFOM, Cologne, Germany), University Witten/Herdecke. | Not Applicable |
| CD142singleCD105CD142measurement | FALSE | FALSE | decimal | Measurement of CD105- and CD142-positive microparticles for quantification of endothelial derived microparticles (EDMP) - results for endothelial derived microparticles (EDMP) (single positive CD142+) - using flow cytometry techniques. Citrated plasma sample was centrifuged for 20 min at 2,500 x g at RT. Supernatants were used to determine EDMP. Run on the BD Accuri C6 Plus flow cytometer (BD Biosciences; Heidelberg, Germany) at Institute for Research in Operative Medicine (IFOM, Cologne, Germany), University Witten/Herdecke. | Not Applicable |
| CD42bAnnexinVdoubleCD42bAnnexmeasurement | FALSE | FALSE | decimal | Measurement of CD42b- and AnnexinV-positive microparticles for quantification of platelet-derived microparticles (PDMP) - results for platelet-derived micro-particles (double positive for CD42b+/Annexin V+) - using flow cytometry techniques. Citrated plasma sample was centrifuged for 20 min at 2,500 x g at RT. Supernatants were used to determine PDMP. Run on the BD Accuri C6 Plus flow cytometer (BD Biosciences; Heidelberg, Germany) at Institute for Research in Operative Medicine (IFOM, Cologne, Germany), University Witten/Herdecke. | Not Applicable |
| CD42bCD62pdoubleCD42bCD62pmeasurement | FALSE | FALSE | decimal | Measurement of CD42b- and CD62p-positive microparticles for quantification of platelet-derived microparticles (PDMP) - results for platelet-derived microparticles (PDMP) (double positive CD42b+/CD62p+) - using flow cytometry techniques. Citrated plasma sample was centrifuged for 20 min at 2,500 x g at RT. Supernatants were used to determine PDMP. Run on the BD Accuri C6 Plus flow cytometer (BD Biosciences; Heidelberg, Germany) at Institute for Research in Operative Medicine (IFOM, Cologne, Germany), University Witten/Herdecke. | Not Applicable |

| Name | Static | Intervention | Format | Description | Possible Values |
|---|---|---|---|---|---|
| CD42bsingleCD42bAnnexVmeasurement | FALSE | FALSE | decimal | Measurement of CD42b and AnnexinV positive microparticles for quantification of platelet-derived microparticles (PDMP) - results for platelet-derived microparticles (PDMP) (single positive CD42b+) - using flow cytometry techniques. Citrated plasma sample was centrifuged for 20 min at 2,500 x g at RT. Supernatants were used to determine PDMP. Run on the BD Accuri C6 Plus flow cytometer (BD Biosciences; Heidelberg, Germany) at Institute for Research in Operative Medicine (IFOM, Cologne, Germany), University Witten/Herdecke. | Not Applicable |
| CD42bsingleCD42bCD62pmeasurement | FALSE | FALSE | decimal | Measurement of CD42b- and CD62p-positive microparticles for quantification of platelet-derived microparticles (PDMP) - results for platelet-derived microparticles (PDMP) (single positive CD42b+) - using flow cytometry techniques. Citrated plasma sample was centrifuged for 20 min at 2,500 x g at RT. Supernatants were used to determine PDMP. Run on the BD Accuri C6 Plus flow cytometer (BD Biosciences; Heidelberg, Germany) at Institute for Research in Operative Medicine (IFOM, Cologne, Germany), University Witten/Herdecke. | Not Applicable |
| CD62EsingleCD105CD62Emeasurement | FALSE | FALSE | decimal | Measurement of CD105- and CD62e-positive microparticles for quantification of endothelial derived microparticles (EDMP) - results for endothelial derived microparticles (EDMP) (single positive CD62e+) - using flow cytometry techniques. Citrated plasma sample was centrifuged for 20 min at 2,500 x g at RT. Supernatants were used to determine EDMP. Run on the BD Accuri C6 Plus flow cytometer (BD Biosciences; Heidelberg, Germany) at Institute for Research in Operative Medicine (IFOM, Cologne, Germany), University Witten/Herdecke. | Not Applicable |
| CD62psingleCD42bCD62pmeasurement | FALSE | FALSE | decimal | Measurement of CD42b- and CD62p-positive microparticles for quantification of platelet-derived microparticles (PDMP) - results for platelet-derived microparticles (PDMP) (single positive CD62p+) - using flow cytometry techniques. Citrated plasma sample was centrifuged for 20 min at 2,500 x g at RT. Supernatants were used to determine PDMP. Run on the BD Accuri C6 Plus flow cytometer (BD Biosciences; Heidelberg, Germany) at Institute for Research in Operative Medicine (IFOM, Cologne, Germany), University Witten/Herdecke. | Not Applicable |
| CoagulationparameterAntithrombinprocent | FALSE | FALSE | decimal | Assay results for standard coagulation test (Antithrombin) - Antithrombin in human citrated plasma measured with an automated chromogenic assay technology using the HemosIL¬Æ aliquid Antithrombin kit (Werfen, Bedford, USA). Run on the ACL TOP CTS 700 (Werfen, Barcelona, Spain). Coagulationparameter were performed by the Institute of Transfusion Medicine (ITM), Cologne-Merheim Medical Centre. Normal range: 83-128 Reference: HemosIL¬Æ package insert | Not Applicable |
| CoagulationparameterDdimersugl | FALSE | FALSE | decimal | Assay results for standard coagulation test (D-Dimers) - using the HemosIL¬Æ D-Dimer Controls kit (Werfen, Bedford, USA). Run on the ACL TOP CTS 700 (Werfen, Barcelona, Spain). Coagulationparameter were performed by the Institute of | Not Applicable |

| Name | Static | Intervention | Format | Description | Possible Values |
|---|---|---|---|---|---|
| | | | | Transfusion Medicine (ITM), Cologne-Merheim Medical Centre. Normal range: 0-232 Reference: HemosIL¬Æ package insert | |
| Coagulationparameter FactorIXprocent | FALSE | FALSE | decimal | Assay results for standard coagulation test (Factor IX) - human plasma immunodepleted of factor IX for the quantitaive determination of factor IX activity based on activated partial thromboplastin time (APTT) assay - using factor IX deficient plasma (Werfen, Barcelona, Spain). Run on the ACL TOP CTS 700 (Werfen, Barcelona, Spain). Coagulationparameter were performed by the Institute of Transfusion Medicine (ITM), Cologne-Merheim Medical Centre. Normal range: 65-150 Reference: HemosIL¬Æ package insert | Not Applicable |
| Coagulationparameter FactorVIIIprocent | FALSE | FALSE | decimal | Assay results for standard coagulation test (Factor VIII) - using a Coamatic factor VIII kit (Chromogenix, Bedford, USA) for chromogenic determination of factor VIII activity in human citrated plasma. Run on the ACL TOP CTS 700 (Werfen; Barcelona, Spain). Coagulationparameter were performed by the Institute of Transfusion Medicine (ITM), Cologne-Merheim Medical Centre. Normal range: 50-150 Reference: HemosIL¬Æ package insert | Not Applicable |
| Coagulationparameter FactorVprocent | FALSE | FALSE | decimal | Assay results for standard coagulation test (Factor V) - human plasma immunodepleted of factor V for the quantitaive determination of factor V activity based on the prothrombin time (PT) assay - using factor V deficient plasma (Werfen, Bedford, USA). Run on the ACL TOP CTS 700 (Werfen, Barcelona, Spain). Coagulationparameter were performed by the Institute of Transfusion Medicine (ITM), Cologne-Merheim Medical Centre. Normal range: 62-139 Reference: HemosIL¬Æ package insert | Not Applicable |
| Coagulationparameter FactorXIIIAgprocent | FALSE | FALSE | decimal | Assay results for standard coagulation test (Factor XIII) - measured with Chromogenix factor XIII Antigen kit (Chromogenix, Bedford, USA) based on an automated latex enhanced immunoassay techniques. Run on the ACL TOP CTS 700 (Werfen; Barcelona, Spain). Coagulationparameter were performed by the Institute of Transfusion Medicine (ITM), Cologne-Merheim Medical Centre. Normal range: 75.2-154.8 Reference: Chromogenix¬Æ package insert | Not Applicable |
| Coagulationparameter Fibrinogenmgdl | FALSE | FALSE | decimal | Assay results for standard coagulation test (Fibrinogen) - using a high sensitivity thromboplastin reagent (RecombiPasTin 2G (HemosIL¬Æ), Werfen, Bedford, USA) based on recombinant human tissue factor (RTF) for quantitative determination in citrated plasma of Prothrombin time (PT) and Fibrinogen. Run on the ACL TOP CTS 700 (Werfen, Barcelona, Spain). Coagulationparameter were performed by the Institute of Transfusion Medicine (ITM), Cologne-Merheim Medical Centre. Normal range: 276-471 Reference: HemosIL¬Æ package insert | Not Applicable |
| Coagulationparameter INR | FALSE | FALSE | decimal | Assay results for standard coagulation test (INR) - using a high sensitivity thromboplastin reagent (RecombiPasTin 2G (HemosIL¬Æ), Werfen, Bedford, USA) based on recombinant human tissue factor (RTF) for quantitative determination in citrated plasma of Prothrombin time (PT) and Fibrinogen. Run on the ACL TOP CTS 700 (Werfen, Barcelona, Spain). | Not Applicable |

| Name | Static | Intervention | Format | Description | Possible Values |
|---|---|---|---|---|---|
| | | | | Coagulationparameter were performed by the Institute of Transfusion Medicine (ITM), Cologne-Merheim Medical Centre. | |
| Coagulationparameter Plasminogenprocent | FALSE | FALSE | decimal | Assay results for standard coagulation test (Plasminogen) - Plasminogen in human citrated plasma measured with an automated chromogenic assay technology using the HemosIL¬Æ Plasminogen kit (Werfen, Bedford, USA). Run on the ACL TOP CTS 700 (Werfen, Barcelona, Spain). Coagulationparameter were performed by the Institute of Transfusion Medicine (ITM), Cologne-Merheim Medical Centre. Normal range: 80-133 Reference: HemosIL¬Æ package insert | Not Applicable |
| Coagulationparameter ProteinCprocent | FALSE | FALSE | decimal | Assay results for standard coagulation test (Protein C) - Protein C in human citrated plasma measured with an automated chromogenic assay technology using the HemosIL¬Æ Protein C kit (Werfen, Bedford, USA). Run on the ACL TOP CTS 700 (Werfen; Barcelona, Spain). Coagulationparameter were performed by the Institute of Transfusion Medicine (ITM), Cologne-Merheim Medical Centre. Normal range: 70-140 Reference: HemosIL¬Æ package insert | Not Applicable |
| Coagulationparameter ProteinSprocent | FALSE | FALSE | decimal | Assay results for standard coagulation test (Protein S) - using the HemosIL¬Æ Protein S Activity kit (Werfen, Bedford, USA). Determination of the functional acitivty of free Protein S by measuring the degree of prolongation of a prothrombin time in the presence of recombinant human tissue factor, phospholipids, calcium ions and protein C. Run on the ACL TOP CTS 700 (Werfen, Barcelona, Spain). Coagulationparameter were performed by the Institute of Transfusion Medicine (ITM), Cologne-Merheim Medical Centre. Normal range: 63.5-149 Reference: HemosIL¬Æ package insert | Not Applicable |
| Coagulationparameter PTTsec | FALSE | FALSE | decimal | Assay results for standard coagulation test (PTT) - using the HemosIL¬Æ APTT-SP kit (Werfen, Bedford, USA). Run on the ACL TOP CTS 700 (Werfen; Barcelona, Spain). Coagulationparameter were performed by the Institute of Transfusion Medicine (ITM), Cologne-Merheim Medical Centre. Normal range: 23-36 Reference: HemosIL¬Æ package insert | Not Applicable |
| Coagulationparameter Quickprocent | FALSE | FALSE | decimal | Assay results for standard coagulation test (Quick) - using a high sensitivity thromboplastin reagent (RecombiPasTin 2G (HemosIL¬Æ), Werfen, Bedford, USA) based on recombinant human tissue factor (RTF) for quantitative determination in citrated plasma of Prothrombin time (PT) and Fibrinogen. Run on the ACL TOP CTS 700 (Werfen, Barcelona, Spain). Coagulationparameter were performed by the Institute of Transfusion Medicine (ITM), Cologne-Merheim Medical Centre. Normal range: 70-130 Reference: HemosIL¬Æ package insert | Not Applicable |
| Coagulationparameter Thrombintimesec | FALSE | FALSE | decimal | Assay results for standard coagulation test (Thrombin Time) - using the HemosIL¬Æ Thrombin Time kit (Werfen, Bedford, USA). Fibrinogen in the citrated plasma sample is converted to fibrin by the addition of purified bovine thrombin and the time required to form the clot is measured. Run on the ACL TOP CTS 700 (Werfen; Barcelona, Spain). Coagulationparameter | Not Applicable |

| Name | Static | Intervention | Format | Description | Possible Values |
|---|---|---|---|---|---|
| | | | | were performed by the Institute of Transfusion Medicine (ITM), Cologne-Merheim Medical Centre. Normal range: 10-17 Reference: HemosIL¬Æ package insert | |
| Coagulationparameter vWFABT0procent | FALSE | FALSE | decimal | Assay results for standard coagulation test (von Willebrand Factor Activity) in citrated plasma samples in patients with bloodtype 0 - measured with the HemosIL¬Æ von Willebrand Activity kit (Werfen, Bedford, USA) based on an automated latex enhanced immunoassay technique. Run on the ACL TOP CTS 700 (Werfen; Barcelona, Spain). Coagulationparameter were performed by the Institute of Transfusion Medicine (ITM), Cologne-Merheim Medical Centre. Normal range: 40-126 Reference: HemosIL¬Æ package insert | Not Applicable |
| Coagulationparameter vWFABTABABprocent | FALSE | FALSE | decimal | Assay results for standard coagulation test (von Willebrand Factor Activity) in citrated plasma samples in patients with bloodtype A, B and AB - measured with the HemosIL¬Æ von Willebrand Activity kit (Werfen, Bedford, USA) based on an automated latex enhanced immunoassay technique. Run on the ACL TOP CTS 700 (Werfen, Barcelona, Spain). Coagulationparameter were performed by the Institute of Transfusion Medicine (ITM), Cologne-Merheim Medical Centre. Normal range: 49-163 Reference: HemosIL¬Æ package insert | Not Applicable |
| Coagulationparameter vWFAgABABprocent | FALSE | FALSE | decimal | Assay results for standard coagulation test (von Willebrand Factor Antigen) in citrated plasma samples in patients with bloodtype A, B and AB - measured with the HemosIL¬Æ von Willebrand Antigen kit (Werfen, Bedford, USA) based on an automated latex enhanced immunoassay technique. Run on the ACL TOP CTS 700 (Werfen, Barcelona, Spain). Coagulationparameter were performed by the Institute of Transfusion Medicine (ITM), Cologne-Merheim Medical Centre. Normal range: 66-176 Reference: HemosIL¬Æ package insert | Not Applicable |
| Coagulationparameter vWFAgBT0procent | FALSE | FALSE | decimal | Assay results for standard coagulation test (von Willebrand Factor Antigen) in citrated plasma samples in patients with bloodtype 0 - measured with the HemosIL¬Æ von Willebrand Antigen kit (Werfen, Bedford, USA) based on an automated latex enhanced immunoassay technique. Run on the ACL TOP CTS 700 (Werfen, Barcelona, Spain). Coagulationparameter were performed by the Institute of Transfusion Medicine (ITM), Cologne-Merheim Medical Centre. Normal range: 42-141 Reference: HemosIL¬Æ package insert | Not Applicable |
| DLA10Extem | FALSE | FALSE | text | Only applicable to sites doing ROTEM studies. ROTEM --> A10 --> EXTEM | Not Applicable |
| DLA10Fibtem | FALSE | FALSE | text | Only applicable to sites doing ROTEM studies. ROTEM --> A10 --> FIBTEM | Not Applicable |
| DLA5Extem | FALSE | FALSE | integer | Only applicable to sites doing ROTEM studies. ROTEM --> A5 --> EXTEM | Not Applicable |
| DLA5Fibtem | FALSE | FALSE | integer | Only applicable to sites doing ROTEM studies ROTEM --> A5 --> FIBTEM | Not Applicable |
| DLaAngleExtem | FALSE | FALSE | integer | Only applicable to sites doing ROTEM studies. ROTEM --> Œ±-angle --> EXTEM | Not Applicable |
| DLaAngleFibtem | FALSE | FALSE | integer | Only applicable to sites doing ROTEM studies. ROTEM --> Œ±-angle --> FIBTEM | Not Applicable |
| DLADPAggreg | FALSE | FALSE | decimal | Only applicable to sites doing multiplate studies. MULTIPLATE TEST --> ADP Test --> Aggregation | Not Applicable |

| Name | Static | Intervention | Format | Description | Possible Values |
|---|---|---|---|---|---|
| DLADPAUC | FALSE | FALSE | integer | Only applicable to sites doing multiplate studies. MULTIPLATE TEST --> ADP Test --> AUC (AU*min) | Not Applicable |
| DLADPVelocity | FALSE | FALSE | decimal | Only applicable to sites doing multiplate studies. MULTIPLATE TEST --> ADP Test --> Velocity | Not Applicable |
| DLAlatSgptUL | FALSE | FALSE | decimal | BLOOD CHEMISTRY --> ALAT/SGPT Recorded in "preferred" units (U/L) | Not Applicable |
| DLAlbumingL | FALSE | FALSE | decimal | BLOOD CHEMISTRY --> Albumin Recorded in "preferred" units (g/dL) | Not Applicable |
| DLAlkalinePhosphataseUL | FALSE | FALSE | decimal | BLOOD CHEMISTRY --> Alkaline Phosphatase Recorded in "preferred" units (U/L) | Not Applicable |
| DLAmylaseUL | FALSE | FALSE | decimal | BLOOD CHEMISTRY --> Amylase Recorded in "preferred" units (U/L) | Not Applicable |
| DLaPttsec | FALSE | FALSE | decimal | HAEMATOLOGY --> Activated thromboplastine time (aPTT) Recorded in "preferred" units (sec.) | Not Applicable |
| DLAsatSgotUL | FALSE | FALSE | decimal | BLOOD CHEMISTRY --> ASAT/SGOT Recorded in "preferred" units (U/L) | Not Applicable |
| DLASPIAggreg | FALSE | FALSE | decimal | Only applicable to sites doing multiplate studies. MULTIPLATE TEST --> ASPI Test --> Aggregation | Not Applicable |
| DLASPIAUC | FALSE | FALSE | integer | Only applicable to sites doing multiplate studies | Not Applicable |
| DLASPIVelocity | FALSE | FALSE | decimal | Only applicable to sites doing multiplate studies MULTIPLATE TEST --> ASPI Test --> Velocity (AU*min) | Not Applicable |
| DLCalciummmolL | FALSE | FALSE | decimal | BLOOD CHEMISTRY --> Calcium Recorded in "preferred" units (mmol/L) | Not Applicable |
| DLCFTExtem | FALSE | FALSE | integer | Only applicable to sites doing ROTEM studies. ROTEM --> CFT --> EXTEM | Not Applicable |
| DLCFTFibtem | FALSE | FALSE | integer | Only applicable to sites doing ROTEM studies. ROTEM --> CFT --> FIBTEM | Not Applicable |
| DLCL30 | FALSE | FALSE | integer | Only applicable to sites doing ROTEM/TEG studies. TEG --> CL30 | Not Applicable |
| DLCL60 | FALSE | FALSE | integer | Only applicable to sites doing ROTEM/TEG studies. TEG --> CL60 | Not Applicable |
| DLCOLAggreg | FALSE | FALSE | decimal | Only applicable to sites doing multiplate studies MULTIPLATE TEST --> COL Test --> Aggregation | Not Applicable |
| DLCOLAUC | FALSE | FALSE | integer | Only applicable to sites doing multiplate studies MULTIPLATE TEST --> COL Test --> AUC (AU*min) | Not Applicable |
| DLCOLVelocity | FALSE | FALSE | decimal | Only applicable to sites doing multiplate studies MULTIPLATE TEST --> COL Test --> Velocity (AU*min) | Not Applicable |
| DLCreatinineumolL | FALSE | FALSE | decimal | BLOOD CHEMISTRY --> Creatinine Recorded in "preferred" units (¬µmol/L) | Not Applicable |
| DLCRPmgL | FALSE | FALSE | decimal | HAEMATOLOGY --> C-reactive protein (CRP) Recorded in "preferred" units (mg/L) | Not Applicable |
| DLCTExtem | FALSE | FALSE | integer | Only applicable to sites doing ROTEM studies ROTEM --> CT --> EXTEM | Not Applicable |
| DLCTFibtem | FALSE | FALSE | integer | Only applicable to sites doing ROTEM studies ROTEM --> CT --> FIBTEM | Not Applicable |
| DLDdimersugL | FALSE | FALSE | decimal | HAEMATOLOGY --> D-dimers Recorded in "preferred" units ( ¬µg/L) | Not Applicable |
| DLEosinophilspct | FALSE | FALSE | decimal | HAEMATOLOGY --> Eosinophils Recorded in "preferred" units (%) | Not Applicable |
| DLEPL | FALSE | FALSE | integer | Only applicable to sites doing TEG/ROTEM studies TEG --> EPL | Not Applicable |
| DLFibrinogenmgdL | FALSE | FALSE | decimal | HAEMATOLOGY --> Fibrinogen Recorded in "preferred" units (mg/dL) | Not Applicable |
| DLGlucosemmolL | FALSE | FALSE | decimal | BLOOD CHEMISTRY --> Glucose Recorded in "preferred" units (mmol/L) | Not Applicable |
| DLHematocritpct | FALSE | FALSE | decimal | HAEMATOLOGY --> Hematocrit Recorded in "preferred" units (%) | Not Applicable |
| DLHemoglobingdL | FALSE | FALSE | decimal | HAEMATOLOGY --> Hemoglobin Recorded in "preferred" units (g/dL) | Not Applicable |
| DLInr | FALSE | FALSE | decimal | HAEMATOLOGY --> INR | Not Applicable |
| DLK | FALSE | FALSE | integer | Only applicable to sites doing TEG/ROTEM studies TEG --> K | Not Applicable |

| Name | Static | Intervention | Format | Description | Possible Values |
|---|---|---|---|---|---|
| DLLdhUL | FALSE | FALSE | decimal | BLOOD CHEMISTRY --> LDH (Lactate Dehydrogenase) Recorded in "preferred" units (U/L) | Not Applicable |
| DLLY30Extem | FALSE | FALSE | text | Only applicable to sites doing ROTEM studies ROTEM --> LY30 --> EXTEM | Not Applicable |
| DLLY30Fibtem | FALSE | FALSE | text | Only applicable to sites doing ROTEM studies ROTEM --> LY30 --> FIBTEM | Not Applicable |
| DLLymphocytespct | FALSE | FALSE | decimal | HAEMATOLOGY --> Lymphocytes Recorded in "preferred" units (%) | Not Applicable |
| DLMA | FALSE | FALSE | integer | Only applicable to sites doing TEG/ROTEM studies TEG --> MA | Not Applicable |
| DLMagnesiummmolL | FALSE | FALSE | decimal | BLOOD CHEMISTRY --> Magnesium Recorded in "preferred" units (mmol/L) | Not Applicable |
| DLMCFExtem | FALSE | FALSE | integer | Only applicable to sites doing ROTEM studies ROTEM --> MCF --> EXTEM | Not Applicable |
| DLMCFFibtem | FALSE | FALSE | integer | Only applicable to sites doing ROTEM studies ROTEM --> MCF --> FIBTEM | Not Applicable |
| DLMCFtExtem | FALSE | FALSE | integer | Only applicable to sites doing ROTEM studies ROTEM --> MCF-t --> EXTEM | Not Applicable |
| DLMCFtFibtem | FALSE | FALSE | integer | Only applicable to sites doing ROTEM studies ROTEM --> MCF-t --> FIBTEM | Not Applicable |
| DLMLExtem | FALSE | FALSE | text | Only applicable to sites doing ROTEM studies ROTEM --> ML --> EXTEM | Not Applicable |
| DLMLFibtem | FALSE | FALSE | text | Only applicable to sites doing ROTEM studies ROTEM --> ML --> FIBTEM | Not Applicable |
| DLNeutrophilspct | FALSE | FALSE | decimal | HAEMATOLOGY --> Neutrophils Recorded in "preferred" units (%) | Not Applicable |
| DLPlatelet105L | FALSE | FALSE | decimal | HAEMATOLOGY --> Platelet Recorded in "preferred" units (X10^9/L or X10^3/¬µL) | Not Applicable |
| DLPotassiummmolL | FALSE | FALSE | decimal | BLOOD CHEMISTRY --> Potassium Recorded in "preferred" units (mmol/L) | Not Applicable |
| DLProthrombineTimeSec | FALSE | FALSE | decimal | HAEMATOLOGY --> Prothrombine Time Recorded in "preferred" units (sec.) | Not Applicable |
| DLR | FALSE | FALSE | integer | Only applicable to sites doing TEG/ROTEM studies TEG --> R | Not Applicable |
| DLReason | FALSE | FALSE | integer | | 1=Study protocol;2=Unit policy;3=Clinical indication in specific patient |
| DLRISTOAggreg | FALSE | FALSE | decimal | Only applicable to sites doing multiplate studies MULTIPLATE TEST --> RISTO Test --> Aggregation | Not Applicable |
| DLRISTOAUC | FALSE | FALSE | integer | Only applicable to sites doing multiplate studies MULTIPLATE TEST --> RISTO Test --> AUC (AU*min) | Not Applicable |
| DLRISTOVelocity | FALSE | FALSE | decimal | Only applicable to sites doing multiplate studies MULTIPLATE TEST --> RISTO Test --> Velocity (AU*min) | Not Applicable |
| DLS100BugL | FALSE | FALSE | decimal | BLOOD CHEMISTRY --> S100B Recorded in "preferred" units (¬µg/L) | Not Applicable |
| DLSodiummmolL | FALSE | FALSE | decimal | BLOOD CHEMISTRY --> Sodium Recorded in "preferred" units (mmol/L) | Not Applicable |
| DLTEGaAngle | FALSE | FALSE | integer | Only applicable for sites doing TEG TEG --> Œ±-angle | Not Applicable |
| DLTEGType | FALSE | FALSE | text | Reflects type of TEG done - Only applicable for selected sites doing TEG | RAPID=Rapid;STAND=Standard |
| DLTMA | FALSE | FALSE | integer | Only applicable for sites doing TEG TEG --> TMA | Not Applicable |
| DLTotalBilirubinumolL | FALSE | FALSE | decimal | BLOOD CHEMISTRY --> Total Bilirubin Recorded in "preferred" units (¬µmol/L) | Not Applicable |
| DLToxScreen | FALSE | FALSE | text | Toxic Drug Screen Result Only if performed as part of clinical routine | NEG=Negative;POS=Positive |
| DLToxScreenPosAmphet | FALSE | FALSE | integer | Reflects if Toxic Drug Screen was positive for Amphetamines. Only if performed as part of clinical routine. | 0=No;1=Yes |
| DLToxScreenPosBarb | FALSE | FALSE | integer | Reflects if Toxic Drug Screen was positive for Barbiturates. Only if performed as part of clinical routine. | 0=No;1=Yes |
| DLToxScreenPosBenzo | FALSE | FALSE | integer | Reflects if Toxic Drug Screen was positive for Benzodiazepines. Only if performed as part of clinical routine. | 0=No;1=Yes |
| DLToxScreenPosCannabis | FALSE | FALSE | integer | Reflects if Toxic Drug Screen was positive for Cannabinoids. Only if performed as part of clinical routine. | 0=No;1=Yes |

| Name | Static | Intervention | Format | Description | Possible Values |
|---|---|---|---|---|---|
| DLToxScreenPosCocaine | FALSE | FALSE | integer | Reflects if Toxic Drug Screen was positive for Cocaine. Only if performed as part of clinical routine. | 0=No;1=Yes |
| DLToxScreenPosMeth | FALSE | FALSE | integer | Reflects if Toxic Drug Screen was positive for Methadone. Only if performed as part of clinical routine. | 0=No;1=Yes |
| DLToxScreenPosMethaqual | FALSE | FALSE | integer | Reflects if Toxic Drug Screen was positive for Methaqualone. Only if performed as part of clinical routine. | 0=No;1=Yes |
| DLToxScreenPosOpiate | FALSE | FALSE | integer | Reflects if Toxic Drug Screen was positive for Opiates. Only if performed as part of clinical routine. | 0=No;1=Yes |
| DLToxScreenPosOther | FALSE | FALSE | integer | Reflects if Toxic Drug Screen was positive for Other drugs than the predefined list. Only if performed as part of clinical routine. | 0=No;1=Yes |
| DLToxScreenPosOtherTxt | FALSE | FALSE | text | Specifies for which drugs, if Toxic Drug Screen was positive for Other drugs than the predefined list. Only if performed as part of clinical routine. | Not Applicable |
| DLToxScreenPosPhency | FALSE | FALSE | integer | Reflects if Toxic Drug Screen was positive for Phencyclidine. Only if performed as part of clinical routine. | 0=No;1=Yes |
| DLToxScreenType | FALSE | FALSE | text | Specifies the type of sample, Urine or Serum, if Toxic Drug Screen was performed. Only if performed as part of clinical routine. | SERUM=Serum;URINE=Urine |
| DLTRAPAggreg | FALSE | FALSE | decimal | Only applicable to sites doing multiplate studies. MULTIPLATE TEST --> TRAP Test --> Aggregation | Not Applicable |
| DLTRAPAUC | FALSE | FALSE | integer | Only applicable to sites doing multiplate studies. MULTIPLATE TEST --> TRAP Test --> AUC (AU*min) | Not Applicable |
| DLTRAPVelocity | FALSE | FALSE | decimal | Only applicable to sites doing multiplate studies. MULTIPLATE TEST --> TRAP Test --> Velocity (AU*min) | Not Applicable |
| DLUreammolL | FALSE | FALSE | decimal | BLOOD CHEMISTRY --> Urea Recorded in "preferred" units (mmol/L) | Not Applicable |
| DLWhiteBloodCellpct | FALSE | FALSE | decimal | HAEMATOLOGY --> White blood cell Recorded in "preferred" units (X10^9/L or X10^3/ŒºL) | Not Applicable |
| ERA10Extem | TRUE | FALSE | text | In emergency room: Only applicable to sites doing ROTEM studies. ROTEM --> A10 --> EXTEM | Not Applicable |
| ERA10Fibtem | TRUE | FALSE | text | In emergency room: Only applicable to sites doing ROTEM studies. ROTEM --> A10 --> FIBTEM | Not Applicable |
| ERA5Extem | TRUE | FALSE | integer | In emergency room: Only applicable to sites doing ROTEM studies. ROTEM --> A5 --> EXTEM | Not Applicable |
| ERA5Fibtem | TRUE | FALSE | integer | In emergency room: Only applicable to sites doing ROTEM studies ROTEM --> A5 --> FIBTEM | Not Applicable |
| ERaAngleExtem | TRUE | FALSE | integer | In emergency room: Only applicable to sites doing ROTEM studies. ROTEM --> Œ±-angle --> EXTEM | Not Applicable |
| ERaAngleFibtem | TRUE | FALSE | integer | In emergency room: Only applicable to sites doing ROTEM studies. ROTEM --> Œ±-angle --> FIBTEM | Not Applicable |
| ERADPAggreg | TRUE | FALSE | decimal | In emergency room: Only applicable to sites doing multiplate studies. MULTIPLATE TEST --> ADP Test --> Aggregation | Not Applicable |
| ERADPVelocity | TRUE | FALSE | decimal | In emergency room: Only applicable to sites doing multiplate studies. MULTIPLATE TEST --> ADP Test --> Velocity | Not Applicable |
| ERAlatSgptUL | TRUE | FALSE | decimal | In emergency room: BLOOD CHEMISTRY --> ALAT/SGPT Recorded in "preferred" units (U/L) | Not Applicable |
| ERAlbumingL | TRUE | FALSE | decimal | In emergency room: BLOOD CHEMISTRY --> Albumin Recorded in "preferred" units (g/dL) | Not Applicable |
| ERAlkalinePhosphataseUL | TRUE | FALSE | decimal | In emergency room: BLOOD CHEMISTRY --> Alkaline Phosphatase Recorded in "preferred" units (U/L) | Not Applicable |

| Name | Static | Intervention | Format | Description | Possible Values |
|---|---|---|---|---|---|
| ERAmylaseUL | TRUE | FALSE | decimal | In emergency room: BLOOD CHEMISTRY --> Amylase Recorded in "preferred" units (U/L) | Not Applicable |
| ERaPttsec | TRUE | FALSE | decimal | In emergency room: HAEMATOLOGY --> Activated thromboplastine time (aPTT) Recorded in "preferred" units (sec.) | Not Applicable |
| ERAsatSgotUL | TRUE | FALSE | decimal | In emergency room: BLOOD CHEMISTRY --> ASAT/SGOT Recorded in "preferred" units (U/L) | Not Applicable |
| ERASPIAggreg | TRUE | FALSE | decimal | In emergency room: Only applicable to sites doing multiplate studies. MULTIPLATE TEST --> ASPI Test --> Aggregation | Not Applicable |
| ERASPIVelocity | TRUE | FALSE | decimal | In emergency room: Only applicable to sites doing multiplate studies MULTIPLATE TEST --> ASPI Test --> Velocity (AU*min) | Not Applicable |
| ERCalciummmolL | TRUE | FALSE | decimal | In emergency room: BLOOD CHEMISTRY --> Calcium Recorded in "preferred" units (mmol/L) | Not Applicable |
| ERCFTExtem | TRUE | FALSE | integer | In emergency room: Only applicable to sites doing ROTEM studies. ROTEM --> CFT --> EXTEM | Not Applicable |
| ERCFTFibtem | TRUE | FALSE | integer | In emergency room: Only applicable to sites doing ROTEM studies. ROTEM --> CFT --> FIBTEM | Not Applicable |
| ERCL30 | TRUE | FALSE | integer | In emergency room: Only applicable to sites doing ROTEM/TEG studies. TEG --> CL30 | Not Applicable |
| ERCL60 | TRUE | FALSE | integer | In emergency room: Only applicable to sites doing ROTEM/TEG studies. TEG --> CL60 | Not Applicable |
| ERCOLAggreg | TRUE | FALSE | decimal | In emergency room: Only applicable to sites doing multiplate studies MULTIPLATE TEST --> COL Test --> Aggregation | Not Applicable |
| ERCOLVelocity | TRUE | FALSE | decimal | In emergency room: Only applicable to sites doing multiplate studies MULTIPLATE TEST --> COL Test --> Velocity (AU*min) | Not Applicable |
| ERCreatinineumolL | TRUE | FALSE | decimal | In emergency room: BLOOD CHEMISTRY --> Creatinine Recorded in "preferred" units (¬µmol/L) | Not Applicable |
| ERCRPmgL | TRUE | FALSE | decimal | In emergency room: HAEMATOLOGY --> C-reactive protein (CRP) Recorded in "preferred" units (mg/L) | Not Applicable |
| ERCTExtem | TRUE | FALSE | integer | In emergency room: Only applicable to sites doing ROTEM studies ROTEM --> CT --> EXTEM | Not Applicable |
| ERCTFibtem | TRUE | FALSE | integer | In emergency room: Only applicable to sites doing ROTEM studies ROTEM --> CT --> FIBTEM | Not Applicable |
| ERDdimersugL | TRUE | FALSE | decimal | In emergency room: HAEMATOLOGY --> D-dimers Recorded in "preferred" units ( ¬µg/L) | Not Applicable |
| EREosinophilspct | TRUE | FALSE | decimal | In emergency room: HAEMATOLOGY --> Eosinophils Recorded in "preferred" units (%) | Not Applicable |
| EREPL | TRUE | FALSE | integer | In emergency room: Only applicable to sites doing TEG/ROTEM studies TEG --> EPL | Not Applicable |
| ERFibrinogenmgdL | TRUE | FALSE | decimal | In emergency room: HAEMATOLOGY --> Fibrinogen Recorded in "preferred" units (mg/dL) | Not Applicable |
| ERGlucosemmolL | TRUE | FALSE | decimal | In emergency room: BLOOD CHEMISTRY --> Glucose Recorded in "preferred" units (mmol/L) | Not Applicable |
| ERHematocritpct | TRUE | FALSE | decimal | In emergency room: HAEMATOLOGY --> Hematocrit Recorded in "preferred" units (%) | Not Applicable |
| ERHemoglobingdL | TRUE | FALSE | decimal | In emergency room: HAEMATOLOGY --> Hemoglobin Recorded in "preferred" units (g/dL) | Not Applicable |
| ERInr | TRUE | FALSE | decimal | In emergency room: HAEMATOLOGY --> INR | Not Applicable |
| ERK | TRUE | FALSE | integer | In emergency room: Only applicable to sites doing TEG/ROTEM studies TEG --> K | Not Applicable |
| ERLabs | TRUE | FALSE | | | Not Applicable |

| Name | Static | Intervention | Format | Description | Possible Values |
|---|---|---|---|---|---|
| ERLdhUL | TRUE | FALSE | decimal | In emergency room: BLOOD CHEMISTRY --> LDH (Lactate Dehydrogenase) Recorded in "preferred" units (U/L) | Not Applicable |
| ERLY30Extem | TRUE | FALSE | text | In emergency room: Only applicable to sites doing ROTEM studies ROTEM --> LY30 --> EXTEM | Not Applicable |
| ERLY30Fibtem | TRUE | FALSE | text | In emergency room: Only applicable to sites doing ROTEM studies ROTEM --> LY30 --> FIBTEM | Not Applicable |
| ERLymphocytespct | TRUE | FALSE | decimal | In emergency room: HAEMATOLOGY --> Lymphocytes Recorded in "preferred" units (%) | Not Applicable |
| ERMA | TRUE | FALSE | integer | In emergency room: Only applicable to sites doing TEG/ROTEM studies TEG --> MA | Not Applicable |
| ERMagnesiummmolL | TRUE | FALSE | decimal | In emergency room: BLOOD CHEMISTRY --> Magnesium Recorded in "preferred" units (mmol/L) | Not Applicable |
| ERMCFExtem | TRUE | FALSE | integer | In emergency room: Only applicable to sites doing ROTEM studies ROTEM --> MCF --> EXTEM | Not Applicable |
| ERMCFFibtem | TRUE | FALSE | integer | In emergency room: Only applicable to sites doing ROTEM studies ROTEM --> MCF --> FIBTEM | Not Applicable |
| ERMCFtExtem | TRUE | FALSE | integer | In emergency room: Only applicable to sites doing ROTEM studies ROTEM --> MCF-t --> EXTEM | Not Applicable |
| ERMCFtFibtem | TRUE | FALSE | integer | In emergency room: Only applicable to sites doing ROTEM studies ROTEM --> MCF-t --> FIBTEM | Not Applicable |
| ERMLExtem | TRUE | FALSE | text | In emergency room: Only applicable to sites doing ROTEM studies ROTEM --> ML --> EXTEM | Not Applicable |
| ERMLFibtem | TRUE | FALSE | text | In emergency room: Only applicable to sites doing ROTEM studies ROTEM --> ML --> FIBTEM | Not Applicable |
| ERNeutrophilspct | TRUE | FALSE | decimal | In emergency room: HAEMATOLOGY --> Neutrophils Recorded in "preferred" units (%) | Not Applicable |
| ERPlatelet105L | TRUE | FALSE | decimal | In emergency room: HAEMATOLOGY --> Platelet Recorded in "preferred" units (X10^9/L or X10^3/¬µL) | Not Applicable |
| ERPotassiummmolL | TRUE | FALSE | decimal | In emergency room: BLOOD CHEMISTRY --> Potassium Recorded in "preferred" units (mmol/L) | Not Applicable |
| ERProthrombineTimeSec | TRUE | FALSE | decimal | In emergency room: HAEMATOLOGY --> Prothrombine Time Recorded in "preferred" units (sec.) | Not Applicable |
| ERR | TRUE | FALSE | integer | In emergency room: Only applicable to sites doing TEG/ROTEM studies TEG --> R | Not Applicable |
| ERS100BugL | TRUE | FALSE | decimal | In emergency room: BLOOD CHEMISTRY --> S100B Recorded in "preferred" units (¬µg/L) | Not Applicable |
| ERSodiummmolL | TRUE | FALSE | decimal | In emergency room: BLOOD CHEMISTRY --> Sodium Recorded in "preferred" units (mmol/L) | Not Applicable |
| ERTEGaAngle | TRUE | FALSE | integer | In emergency room: Only applicable for sites doing TEG TEG --> Œ±-angle | Not Applicable |
| ERTEGType | TRUE | FALSE | text | In emergency room: Reflects type of TEG done - Only applicable for selected sites doing TEG | RAPID=Rapid;STAND=Standard |
| ERTMA | TRUE | FALSE | integer | In emergency room: Only applicable for sites doing TEG TEG --> TMA | Not Applicable |
| ERTotalBilirubinumolL | TRUE | FALSE | decimal | In emergency room: BLOOD CHEMISTRY --> Total Bilirubin Recorded in "preferred" units (¬µmol/L) | Not Applicable |
| ERToxScreen | TRUE | FALSE | text | In emergency room: Toxic Drug Screen Result Only if performed as part of clinical routine | NEG=Negative;POS=Positive |
| ERToxScreenPosAmphet | TRUE | FALSE | integer | In emergency room: Reflects if Toxic Drug Screen was positive for Amphetamines. Only if performed as part of clinical routine. | 0=No;1=Yes |

| Name | Static | Intervention | Format | Description | Possible Values |
|---|---|---|---|---|---|
| ERToxScreenPosBarb | TRUE | FALSE | integer | In emergency room: Reflects if Toxic Drug Screen was positive for Barbiturates. Only if performed as part of clinical routine. | 0=No;1=Yes |
| ERToxScreenPosBenzo | TRUE | FALSE | integer | In emergency room: Reflects if Toxic Drug Screen was positive for Benzodiazepines. Only if performed as part of clinical routine. | 0=No;1=Yes |
| ERToxScreenPosCannabis | TRUE | FALSE | integer | In emergency room: Reflects if Toxic Drug Screen was positive for Cannabinoids. Only if performed as part of clinical routine. | 0=No;1=Yes |
| ERToxScreenPosCocaine | TRUE | FALSE | integer | In emergency room: Reflects if Toxic Drug Screen was positive for Cocaine. Only if performed as part of clinical routine. | 0=No;1=Yes |
| ERToxScreenPosMeth | TRUE | FALSE | integer | In emergency room: Reflects if Toxic Drug Screen was positive for Methadone. Only if performed as part of clinical routine. | 0=No;1=Yes |
| ERToxScreenPosMethaqual | TRUE | FALSE | integer | In emergency room: Reflects if Toxic Drug Screen was positive for Methaqualone. Only if performed as part of clinical routine. | 0=No;1=Yes |
| ERToxScreenPosOpiate | TRUE | FALSE | integer | In emergency room: Reflects if Toxic Drug Screen was positive for Opiates. Only if performed as part of clinical routine. | 0=No;1=Yes |
| ERToxScreenPosOther | TRUE | FALSE | integer | In emergency room: Reflects if Toxic Drug Screen was positive for Other drugs than the predefined list. Only if performed as part of clinical routine. | 0=No;1=Yes |
| ERToxScreenPosOtherTxt | TRUE | FALSE | text | In emergency room: Specifies for which drugs, if Toxic Drug Screen was positive for Other drugs than the predefined list. Only if performed as part of clinical routine. | Not Applicable |
| ERToxScreenPosPhency | TRUE | FALSE | integer | In emergency room: Reflects if Toxic Drug Screen was positive for Phencyclidine. Only if performed as part of clinical routine. | 0=No;1=Yes |
| ERToxScreenType | TRUE | FALSE | text | In emergency room: Specifies the type of sample, Urine or Serum, if Toxic Drug Screen was performed. Only if performed as part of clinical routine. | SERUM=Serum;URINE=Urine |
| ERTRAPAggreg | TRUE | FALSE | decimal | In emergency room: Only applicable to sites doing multiplate studies. MULTIPLATE TEST --> TRAP Test --> Aggregation | Not Applicable |
| ERTRAPVelocity | TRUE | FALSE | decimal | In emergency room: Only applicable to sites doing multiplate studies. MULTIPLATE TEST --> TRAP Test --> Velocity (AU*min) | Not Applicable |
| ERUreammolL | TRUE | FALSE | decimal | In emergency room: BLOOD CHEMISTRY --> Urea Recorded in "preferred" units (mmol/L) | Not Applicable |
| ERWhiteBloodCellpct | TRUE | FALSE | decimal | In emergency room: HAEMATOLOGY --> White blood cell Recorded in "preferred" units (X10^9/L or X10^3/ŒºL) | Not Applicable |
| FibrinolysisFibrinogenMonomerugml | FALSE | FALSE | decimal | Assay results for fibrin monomers -using an immunoturbidimetric determination technology (STA - Liatest FM-Kit, Stago; France). Measurement were performed at the Ludwig Boltzman Institute (Salzburg, Vienna). Reference interval of Liatest FM-Kit of 6 | Not Applicable |
| FibrinolysisregulatorAntiplasminProzent | FALSE | FALSE | decimal | Assay results for Antiplasmin - measured with a colorimetric assay technology (STA-Stachrom¬Æ-TAFI-Kit, Stago; France). Measurement were performed at the Ludwig Boltzman Institute (Salzburg, Vienna). Normal range: 80-120 Reference: STA-Stachrom¬Æ-TAFI-Kit package insert | Not Applicable |
| FibrinolysisregulatorTAFIprocent | FALSE | FALSE | decimal | Assay results for Thrombin-Activatable Fibrinolysis Inhibitor (TAFI) -measured with a colorimetric assay technology (STA-Stachrom-TAFI-Kit, Stago; France). Measurement were performed at the Ludwig Boltzman Institute (Salzburg, Vienna). Detection limit of Stachrom-TAFI-Kit: 5-195 | Not Applicable |
| GFAP | FALSE | FALSE | decimal | Assay results for Glial fibrillary acidic protein [GFAP] - measured with an ultrasensitive immunoassay using digital | Not Applicable |

| Name | Static | Intervention | Format | Description | Possible Values |
| --- | --- | --- | --- | --- | --- |
| | | | | array technology (Single Molecule Arrays, SiMoA)-based Human Neurology 4-Plex B assay (N4PB) run on the SR-X benchtop assay platform (Quanterix Corp., Lexington, MA) at the University of Florida (Gainesville, Florida). | |
| NFL | FALSE | FALSE | decimal | Assay results for Neurofilament protein-light (NFL) - measured with an ultrasensitive immunoassay using digital array technology (Single Molecule Arrays, SiMoA)-based Human Neurology 4-Plex B assay (N4PB) run on the SR-X benchtop assay platform (Quanterix Corp., Lexington, MA) at the University of Florida (Gainesville, Florida). | Not Applicable |
| NSE | FALSE | FALSE | decimal | Assay results for Neuron-specific enolase (NSE) - measured with a clinical-use automated system, using an electrochemiluminescence immunoassay kit (ECLIA) (Elecsys S100 and Elecsys NSE assays) run on the e 602 module of Cobas 8000 modular analyzer (Roche Diagnostics, Mannheim, Germany) at the University of Pecs (Pecs, Hungary). | Not Applicable |
| PAI1ngml | FALSE | FALSE | decimal | Assay results for Plasminogen-activator-inhibitor-1 (PAI-1) - using an ELISA Duo Set Kit from R&D Systems (Minneapolis, USA) based on the sandwich principle according to manufacturer‚Äôs instructions. EDTA plasma samples were used. Run on the EPOCH 2 (BioTek; Winooski, USA) microplate reader at Institute for Research in Operative Medicine (IFOM, Cologne, Germany), University Witten/Herdecke. | Not Applicable |
| S100B | FALSE | FALSE | decimal | Assay results for S100 calciumbinding protein B (S100B) - measured with a clinical-use automated system, using an electrochemiluminescence immunoassay kit (ECLIA) (Elecsys S100 and Elecsys NSE assays) run on the e 602 module of Cobas 8000 modular analyzer (Roche Diagnostics, Mannheim, Germany) at the University of Pecs (Pecs, Hungary). | Not Applicable |
| Syndecan1pgml | FALSE | FALSE | decimal | Assay results for Syndecan-1 - using an ELISA Duo Set Kit from R&D Systems (Minneapolis, USA) based on the sandwich principle according to manufacturer‚Äôs instructions. EDTA plasma samples were used. Run on the EPOCH 2 (BioTek; Winooski, USA) microplate reader at Institute for Research in Operative Medicine (IFOM, Cologne, Germany), University Witten/Herdecke. | Not Applicable |
| Tau | FALSE | FALSE | decimal | Assay results for T-TAU - measured with an ultrasensitive immunoassay using digital array technology (Single Molecule Arrays, SiMoA)-based Human Neurology 4-Plex B assay (N4PB) run on the SR-X benchtop assay platform (Quanterix Corp., Lexington, MA) at the University of Florida (Gainesville, Florida). | Not Applicable |
| ThrombingenerationETPnmmin | FALSE | FALSE | decimal | Assay results for endogenous thrombin potential (ETP) (STG¬Æ-BleedScreen, Stago, France) in citrated plasma - using a fluorogenic method (ST Genesia analyzer; Stago, France) - triggered by a low concentration of tissue factor. Measurement were performed at the Ludwig Boltzman Institute (Salzburg, Vienna). | Not Applicable |
| ThrombingenerationETPprocent | FALSE | FALSE | decimal | Assay results for endogenous thrombin potential (ETP) (STG¬Æ-BleedScreen, Stago, France) in citrated plasma - using a fluorogenic method (ST Genesia analyzer, Stago; France) - triggered by a low | Not Applicable |

| Name | Static | Intervention | Format | Description | Possible Values |
|---|---|---|---|---|---|
| | | | | concentration of tissue factor. Measurement were performed at the Ludwig Boltzman Institute (Salzburg, Vienna). | |
| ThrombingenerationLagTimemin | FALSE | FALSE | decimal | Assay results for quantitative determination of thrombin generation (STG¬Æ-BleedScreen, Stago, France) in citrated plasma (Lag Time) - using a fluorogenic method (ST Genesia analyzer, Stago; France) - triggered by a low concentration of tissue factor. Measurement were performed at the Ludwig Boltzman Institute (Salzburg, Vienna). | Not Applicable |
| ThrombingenerationLagTimeratio | FALSE | FALSE | decimal | Assay results for quantitative determination of thrombin generation (STG¬Æ-BleedScreen, Stago, France) in citrated plasma (Lag Time) - using a fluorogenic method (ST Genesia analyzer, Stago; France) - triggered by a low concentration of tissue factor. Measurement were performed at the Ludwig Boltzman Institute (Salzburg, Vienna). | Not Applicable |
| ThrombingenerationPeakHeightnm | FALSE | FALSE | decimal | Assay results for quantitative determination of thrombin generation (STG¬Æ-BleedScreen, Stago, France) in citrated plasma (Peak Height) - using a fluorogenic method (ST Genesia analyzer Stago; France) - triggered by a low concentration of tissue factor. Measurement were performed at the Ludwig Boltzman Institute (Salzburg, Vienna). | Not Applicable |
| ThrombingenerationPeakHeightprocent | FALSE | FALSE | decimal | Assay results for quantitative determination of thrombin generation (STG¬Æ-BleedScreen, Stago, France) in citrated plasma (Peak Height) - using a fluorogenic method (ST Genesia analyzer, Stago; France) - triggered by a low concentration of tissue factor. Measurement were performed at the Ludwig Boltzman Institute (Salzburg, Vienna). | Not Applicable |
| ThrombingenerationStartTailmin | FALSE | FALSE | decimal | Assay results for quantitative determination of thrombin generation in citrated plasma (Start Tail) - using a fluorogenic method (ST Genesia analyzer, Stago; France) - triggered by a low concentration of tissue factor. Measurement were performed at the Ludwig Boltzman Institute (Salzburg, Vienna). | Not Applicable |
| ThrombingenerationStartTailratio | FALSE | FALSE | decimal | Assay results for quantitative determination of thrombin generation in citrated plasma (Start Tail) - using a fluorogenic method (ST Genesia analyzer, Stago; France) - triggered by a low concentration of tissue factor. Measurement were performed at the Ludwig Boltzman Institute (Salzburg, Vienna). | Not Applicable |
| ThrombingenerationTimetoPeakmin | FALSE | FALSE | decimal | Assay results for quantitative determination of thrombin generation in citrated plasma (Time to Peak) - using a fluorogenic method (ST Genesia analyzer; Stago, France) - triggered by a low concentration of tissue factor. Measurement were performed at the Ludwig Boltzman Institute (Salzburg, Vienna). | Not Applicable |
| ThrombingenerationTimetoPeakratio | FALSE | FALSE | decimal | Assay results for quantitative determination of thrombin generation in citrated plasma (Time to Peak) - using a fluorogenic method (ST Genesia analyzer, Stago; France) - triggered by a low concentration of tissue factor. Measurement were performed at the Ludwig Boltzman Institute (Salzburg, Vienna). | Not Applicable |
| ThrombingenerationVelIndexnmmin | FALSE | FALSE | decimal | Assay results for quantitative determination of thrombin generation in citrated plasma (Velocity Index) - using a fluorogenic method (ST Genesia analyzer, Stago; France) - triggered by a low concentration of | Not Applicable |

| Name | Static | Intervention | Format | Description | Possible Values |
|---|---|---|---|---|---|
| | | | | tissue factor. Measurement were performed at the Ludwig Boltzman Institute (Salzburg, Vienna). | |
| ThrombingenerationVelIndexprocent | FALSE | FALSE | decimal | Assay results for quantitative determination of thrombin generation in in citrated plasma (Velocity index) - using a fluorogenic method (ST Genesia analyzer, Stago; France) - triggered by a low concentration of tissue factor. Measurement were performed at the Ludwig Boltzman Institute (Salzburg, Vienna). | Not Applicable |
| VECadherinngml | FALSE | FALSE | decimal | Assay results for VE-Cadherin - using an ELISA Duo Set Kit from R&D Systems (Minneapolis, USA) based on the sandwich principle according to manufacturer‚Äôs instructions. EDTA plasma samples were used. Run on the EPOCH 2 (BioTek; Winooski, USA) microplate reader at Institute for Research in Operative Medicine (IFOM, Cologne, Germany), University Witten/Herdecke. | Not Applicable |

*ICU medications and management*

| Name | Static | Intervention | Format | Description | Possible Values |
|---|---|---|---|---|---|
| Agent | FALSE | TRUE | integer | Details on medication captured information on Class, Agent, Reason, Highest daily dose, Route, start and stop date and whether or not medication has been ongoing after discharge. This variable describes the Agent. Agents commonly used for treatment of raised ICP are listed in the TIL section, but not included in the medication lists here. | 1=Analgesic: paracetamol;2=Analgesic: NSAIDs;3=Analgesic: tramadol;4=Analgesic: opioids (morphine, ect);5=Sedatives/treatment of agitation: barbiturates (penthothal, ect);6=Sedatives/treatment of agitation: clondine;7=Sedatives/treatment of agitation: dexmedetomidine;8=Sedatives/treatment of agitation: diazepam;9=Sedatives/treatment of agitation: fentanyl;10=Sedatives/treatment of agitation: haloperidol (haldol);11=Sedatives/treatment of agitation: lorazepam (tenesta, ect);12=Sedatives/treatment of agitation: midazolam;13=Sedatives/treatment of agitation: morphine;14=Sedatives/treatment of agitation: propofol;15=Sedatives/treatment of agitation: other;16=Neuromuscular blockade: pancuronium (pavulon);17=Neuromuscular blockade: atracurium (tracium);18=Neuromuscular blockade: cisatracurium (nimbex);19=Neuromuscular blockad: gallamine (flaxedil);20=Neuromuscular blockade: rocuronium (zemuron);21=Neuromuscular blockade: vecuronium (norcuron);22=Neuromuscular blockade: other;23=Anti- epileptic: carbamazepine (tegretol);24=Anti- epileptic: lamotrigine (lamectal);25=Anti- epileptic: levetirazetam (keppra);26=Anti- epileptic: phenytoine (diphantoine);27=Anti- epileptic: valproate (depakine);28=Anti- epileptic: other;29=Antibiotics: aminoglycoside (amikacine, gentamicine etc);30=Antibiotics: carbapemens (meronem etc);31=Antibiotics: cephalosporin 1st gen (cefalexin etc);32=Antibiotics: cephalosporin 2nd gen (cefuroxim etc);33=Antibiotics: cephalosporin 3rd gen (cefotaxine etc);34=Antibiotics: cephalosporin 4th gen (cefepime, maxipime etc);35=Antibiotics: cephalosporin 5th gen (ceftasoline etc);36=Antibiotics: glycopeptides (vancomycine);37=Antibiotics: lincosamides (clindamycine etc);38=Antibiotics: macrolidis (erythromycine etc);39=Antibiotics: nitrofurones (furoxone, furadantine etc);40=Antibiotics: penicillines (ampicilline, cloxacilline);41=Antibiotics: amoxycilline/clavulanic acid (augmentin, tyclav etc);42=Antibiotics: quinolones (ciprofloxacine etc);43=Antibiotics: sulfonamides (co-trimoxazole, doxycycline);44=Antibiotics: other;45=Anti-hypertensive: ACE blockers (captopril, perindopril etc);46=Anti-hypertensive: angiotensininhibitors (cardesartan etc);47=Anti- hypertensive: bA"tablockers (propanolol);48=Anti-hypertensive: clonidine;49=Anti-hypertensive: diuretics;50=Calcium channel blockers: nimodipine;51=Calcium channel blockers: nicardipine;52=Calcium channel blockers: verapamil;53=Steroids: methylprednisolone;54=Steroids: bA(c)tametasone;55=Steroids: dexametasone;56=Steroids: hydrocortisone/cortisone;57=Antacids: Aluminium hydroxide;58=Antacids: other;59=H2 receptor antagonist: Cimetidine;60=H2 receptor antagonist: Ranitidine (Zantac);61=Proton pump inhibitors: Omeprazol (Losec);62=Proton pump inhibitors: Esomeprazol (Nexium);63=Proton pump inhibitors: Pantoprazole (Pantozol);64=Prokinetics: Domperidon |

| Name | Static | Intervention | Format | Description | Possible Values |
|---|---|---|---|---|---|
| | | | | | (Motilium);65=Prokinetics: Erythromycin;66=Prokinetics: Metoclopramide (Primperan);67=Analgesic: other;68=Anti-hypertensive: other;69=Calcium channel blockers: other;70=Steroids: other;71=H2 receptor antagonist: other;72=Proton pump inhibitors: other;73=Prokinetics:other;99=Other, specify in Agent Other: Other |
| AgentOther | FALSE | TRUE | text | Details on medication captured information on Class, Agent, Reason, Highest daily dose, Route, start and stop date and whether or not medication has been ongoing after discharge. This variable describes if the Agent was "other" than the predefined list. | Not Applicable |
| CauseOfDelay | FALSE | FALSE | integer | WHY question: documents reason for delayed transition of care | 1=Unavailability of beds in the receiving unit;2=Unavailability of transport;3=Wish of patient/proxies;4=Need for isolation due to multi resistant bacteria;5=Funding issues;6=Bureaucratic causes;99=Other |
| CauseOfDelayOther | FALSE | FALSE | text | WHY question: documents "other" reason for delayed transition of care than predefined list | Not Applicable |
| Class | FALSE | TRUE | integer | Details on medication captured information on Class, Agent, Reason, Highest daily dose, Route, start and stop date and whether or not medication has been ongoing after discharge. This variable describes the Classes. Classes: analgesic, sedatives, neuromuscular blocking agents, anti-epileptic drugs, antibiotiucs, anti-hypertensive, calcium channel blockers, steroids, antacids, H2 receptor antagonists, proton pump inhibitors and prokinetics. Agents commonly used for treatment of raised ICP are listed in the TIL section, but not included in the medication lists here. | 1=Analgesic;2=Sedatives/treatment of agitation;3=Neuromuscular blockade;4=Anti-epileptic;5=Antibiotics;6=Anti-hypertensive;7=Calcium channel blockers;8=Steroids;9=Antacids;10=H2 receptor antagonist;11=Proton pump inhibitors;12=Prokinetics;99=Other, specify in Agent Other |
| DVTPharmType | FALSE | TRUE | integer | These variables aim to document specific information on the use of DVT prophylaxis. Little evidence exists on the use and timing of DVT prophylaxis after TBI, and considerable practice variation exists. This variable describes the Type of prophylaxis in case of Pharmacologic DVT. | 1=Heparin;2=Low molecular weight Heparin;3=Dalteparin (Fragmin);4=Enoxaparin;5=Nadroparin (Fraxiparine, Fraxodil);6=Parnaparin;7=Reviparin;8=Tinzaparin |
| DVTProphylaxisMech | FALSE | TRUE | integer | These variables aim to document specific information on the use of DVT prophylaxis. Little evidence exists on the use and timing of DVT prophylaxis after TBI, and considerable practice variation exists. This variable describes presence or absence of Mechanical DVT . | 0=No;1=Yes |
| DVTProphylaxisMechType | FALSE | TRUE | text | These variables aim to document specific information on the use of DVT prophylaxis. Little evidence exists on the use and timing of DVT prophylaxis after TBI, and considerable practice variation exists. This variable describes the Type of prophylaxis in case of Mechanical DVT. | Not Applicable |
| DVTProphylaxisPharm | FALSE | TRUE | integer | These variables aim to document specific information on the use of DVT prophylaxis. Little evidence exists on the use and timing of DVT prophylaxis after TBI, and considerable practice variation exists. This variable describes presence or absence of Pharmacologic DVT. | 0=No;1=Yes |
| EnteralNutrition | FALSE | TRUE | integer | These variables aim to document specific information on nutritional support, provided by parenteral and/or enteral routes. This variable describes the absence or presence of Enteral Nutrition. | 0=No;1=Yes |
| EnteralNutritionRoute | FALSE | TRUE | integer | These variables aim to document specific information on nutritional support, provided by parenteral and/or enteral routes. This variable describes the route of administration for Enteral Nutrition. | 1=Nasogastric tube;2=Transpyloric tube;3=Gastrostomy |
| HighestDailyDose | FALSE | TRUE | text | Details on medication captured information on Class, Agent, Reason, Highest daily dose, Route, start and stop date and whether or not medication has been ongoing after discharge. This variable describes the | Not Applicable |

| Name | Static | Intervention | Format | Description | Possible Values |
|---|---|---|---|---|---|
| | | | | Highest Daily Dose. These details should be entered for each agent. | |
| HospSecondInsultsNeuroWorseAction1 | FALSE | TRUE | integer | Action taken in case of Neuroworsening. The importance of Neuroworsening was first described by Morris and Marshall. The occurrence of neuroworsening is related to poorer outcome in subjects with moderate to severe TBI. Neuroworsening is defined as 1) a decrease in GCS motor score of 2 or more points; 2) a new loss of pupillary reactivity or development of pupillary assymmetry >= 2mm; 3) deterioration in neurological or CT status sufficient to warrant immediate medical or surgical intervention | 1=None; 2=Unscheduled CT scan; 3=Change in medical therapy' 4=Surgical intervention |
| HospSecondInsultsNeuroWorseAction2 | FALSE | TRUE | integer | Action taken in case of Neuroworsening. The importance of Neuroworsening was first described by Morris and Marshall. The occurrence of neuroworsening is related to poorer outcome in subjects with moderate to severe TBI. Neuroworsening is defined as 1) a decrease in GCS motor score of 2 or more points; 2) a new loss of pupillary reactivity or development of pupillary assymmetry >= 2mm; 3) deterioration in neurological or CT status sufficient to warrant immediate medical or surgical intervention | 1=None; 2=Unscheduled CT scan; 3=Change in medical therapy' 4=Surgical intervention |
| HospSecondInsultsNeuroWorseAction3 | FALSE | TRUE | integer | Action taken in case of Neuroworsening. The importance of Neuroworsening was first described by Morris and Marshall. The occurrence of neuroworsening is related to poorer outcome in subjects with moderate to severe TBI. Neuroworsening is defined as 1) a decrease in GCS motor score of 2 or more points; 2) a new loss of pupillary reactivity or development of pupillary assymmetry >= 2mm; 3) deterioration in neurological or CT status sufficient to warrant immediate medical or surgical intervention | 1=None; 2=Unscheduled CT scan; 3=Change in medical therapy' 4=Surgical intervention |
| HospSecondInsultsNeuroWorseAction4 | FALSE | TRUE | integer | Action taken in case of Neuroworsening. The importance of Neuroworsening was first described by Morris and Marshall. The occurrence of neuroworsening is related to poorer outcome in subjects with moderate to severe TBI. Neuroworsening is defined as 1) a decrease in GCS motor score of 2 or more points; 2) a new loss of pupillary reactivity or development of pupillary assymmetry >= 2mm; 3) deterioration in neurological or CT status sufficient to warrant immediate medical or surgical intervention | 1=None; 2=Unscheduled CT scan; 3=Change in medical therapy' 4=Surgical intervention |
| HVTIL | FALSE | TRUE | integer | Bihourly change in therapy intensity level (TIL) | 0=No change;1=Increasing intensity;2=Decreasing intensity |
| HVTILChangeReason | FALSE | FALSE | integer | Reason for bihourly change in therapy intensity level (TIL) | 1=Intensified: Clinical deterioration;2=Intensified: Suspicion of increased of ICP (not measured);3=Intensified: Increased ICP (documented);4=Intensified: Clinical decision to target other mechanism;5=Intensified: Change of doctor (different shift);6=Decreasing: Clinical improvement;7=Decreasing: Adequate control over ICP;8=Decreasing: Upper treatment limit reached/past;9=Decreasing: Further treatment considered futile;10=Decreasing: Change of doctor (different shift) |
| ICUAdmisStatusHaemoStable | TRUE | FALSE | integer | Reflects the status of the patient on admission to the ICU: Haemodynamically stable | 0=No;1=Yes;88=Unknown |
| ICUAdmisStatusIntubated | TRUE | FALSE | integer | Reflects the status of the patient on admission to the ICU: Intubated | 0=No;1=Yes;88=Unknown |
| ICUAdmisStatusMechVent | TRUE | FALSE | integer | Reflects the status of the patient on admission to the ICU: Mechanically ventilated | 0=No;1=Yes;88=Unknown |
| Intubation | FALSE | TRUE | integer | This variable describes the absence or presence of Intubation as Ventilation Management (only for ICU patients). | 0=No;1=Yes |
| MechVentilation | FALSE | TRUE | integer | This variable describes the absence or presence of Mechanical Ventilation (any respiratory mode except for CPAP). | 0=No;1=Yes |

| Name | Static | Intervention | Format | Description | Possible Values |
|---|---|---|---|---|---|
| Nasogastric | FALSE | TRUE | integer | Reflects absence or presence of a Nasogastric feeding tube. | 0=No;1=Yes |
| OxygenAdm | FALSE | TRUE | integer | Reflects presence or absence of Oxygen Administration. | 0=No;1=Yes;88=Unknown |
| ParenteralNutrition | FALSE | TRUE | integer | These variables aim to document specific information on nutritional support, provided by parenteral and/or enteral routes. This variable describes the absence or presence of Parenteral Nutrition. | 0=No;1=Yes |
| PEGTube | FALSE | TRUE | integer | This variable describes the absence or presence of a PEG tube (gastrostomy). | 0=No;1=Yes |
| Reason | FALSE | FALSE | integer | Details on medication captured information on Class, Agent, Reason, Highest daily dose, Route, start and stop date and whether or not medication has been ongoing after discharge. This variable describes the Reason for medication. These details should be entered for each agent. | 1=Sedatives/treatment of agitation: mechanical ventilation;2=Sedatives/treatment of agitation: metabolic suppression;3=Anti-epileptic: prophylaxis;4=Anti-epileptic: treatment of overt seizure;6=Anti-epileptic: treatment of (silent) seizure activity;7=Antibiotics: fever, no clear focus;8=Antibiotics: pneumonia;9=Antibiotics: urinary tract infection;10=Antibiotics: catheter related bloodstream infection;11=Antibiotics: intracranial abces/empyeme;12=Antibiotics: periprocedural prophylaxis;13=Antibiotics: meningitis;14=Anti-hypertensive: to lower blood pressure;15=Anti-hypertensive: treatment agitation;16=Calcium channel blockers: prevention of vasospasm;17=Calcium channel blockers: treatment of vasospasm;18=Calcium channel blockers: anti-hypertensive;19=Calcium channel blockers: cardiac indication;20=Steroids: traumatic brain injury;21=Steroids: ARDS;22=Steroids: hypopituitarism;23=Steroids: sepsis;24=Antacids: gastric protection;25=Antacids: reflux;26=H2 receptor antagonist: gastric protection;27=H2 receptor antagonist: treatment of ulcer;28=Proton pump inhibitors: gastric protection;29=Proton pump inhibitors: treatment of ulcer;30=Prokinetics: gastric retention;31=Prokinetics: vomiting;32=Prokinetics: constipation;33=Prokinetics: routine care;35=Analgesic: other;36=Neuromuscular blockade: other;37=Sedatives/treatment of agitation: Other;38=Anti-epileptic: Other;39=Antibiotics: Other;40=Anti-hypertensive: Other;41=Calcium channel blockers: Other;42=Steroids: Other;43=Antacids: Other;44=H2 receptor antagonist: Other;45=Proton pump inhibitors: Other;46=Prokinetics: Other;99=Other, specify in Agent Other: Other |
| ReasonOther | FALSE | FALSE | text | Details on medication captured information on Class, Agent, Reason, Highest daily dose, Route, start and stop date and whether or not medication has been ongoing after discharge. This variable describes the Reason for medication if this was "other than the predefined ones. These details should be entered for each agent. | Not Applicable |
| ReIntubation | FALSE | TRUE | integer | Reflects if there has been a need for re-intubation. | 0=No;1=Yes |
| ReMechVentilation | FALSE | TRUE | integer | Reflects the need for re-instituting mechanical ventilation. | 0=No;1=Yes |
| ReMechVentilationReason | FALSE | FALSE | integer | Reflects the reason for the need of re-instituting mechanical ventilation. | 1=Respiratory failure;2=Neurologic deterioration;3=Spontaneous hyperventilation;4=Sepsis;99=Other |
| ReMechVentilationReasonOther | FALSE | FALSE | text | Reflects the "other" reason for the need of re-instituting mechanical ventilation. | Not Applicable |
| Route | FALSE | TRUE | text | Details on medication captured information on Class, Agent, Reason, Highest daily dose, Route, start and stop date and whether or not medication has been ongoing after discharge. This variable describes the Route. These details should be entered for each agent. | ED=Epidural;Ih=Inhaled;Im=Intramuscular;IvCont=Continuous IV;IvInt=Intermittent IV;PO=Oral;Pv=Vaginal;Re=Rectal;Sc=Subcutaneous;To=Topical |
| TILCCSFDrainageVolume | FALSE | TRUE | integer | Specification of volume drained, only applicable if "DailyTIL.TILCSFDrainage" was "yes" | Not Applicable |
| TILCSFDrainage | FALSE | TRUE | integer | Daily TIL: reflects if CSF drainage occurred yes or no | 0=No;1=Yes |
| TILDobutamineDose | FALSE | TRUE | text | Indicates the total dose of dobutamine administered (in mg.) if applicable. Calculated over a 24-hour period | Not Applicable |
| TILDopamineDose | FALSE | TRUE | text | Indicates the total dose of dopamine administered (in mg.) if applicable. Calculated over a 24-hour period | Not Applicable |

| Name | Static | Intervention | Format | Description | Possible Values |
|---|---|---|---|---|---|
| TILFactorsCaloricIntakeEnteralKcal | FALSE | TRUE | integer | Daily TIL: reflects the caloric intake via Enteral route in Kcal | Not Applicable |
| TILFactorsCaloricIntakeParenKcal | FALSE | TRUE | integer | Daily TIL: reflects the caloric intake via Parenteral route in Kcal | Not Applicable |
| TILFactorsCaloricIntakeRouteEnteral | FALSE | TRUE | integer | The variables "parenteral" and "enteral route" are used to document if the patient received enteral feeding or not and if so, what the total number of Kcal was given by each route. | 1=Parenteral route;2=Enteral route |
| TILFactorsCaloricIntakeRouteParen | FALSE | TRUE | integer | The variables "parenteral" and "enteral route" are used to document if the patient received enteral feeding or not and if so, what the total number of Kcal was given by each route. | 0=No;1=Yes |
| TILFactorsCoagulation | FALSE | TRUE | integer | A maximum of 4 "types" of treatment (drop down box) can be selected and entered under treatment 1-4. Details on volume/dose of the products administered should correspond to treatment 1-4 and be entered in the variable volume 1-4. | 0=No;1=Yes, for clinical reasons;2=Yes, according to study protocol;88=Unknown |
| TILFactorsCoagulationHemoglobinAfter | FALSE | FALSE | decimal | Reflects the level of hemoglobin after transfusion in the standard unit (g/dL) | Not Applicable |
| TILFactorsCoagulationHemoglobinBefore | FALSE | FALSE | decimal | Reflects the level of hemoglobin before transfusion in the standard unit (g/dL) | Not Applicable |
| TILFactorsCoagulationType1 | FALSE | TRUE | integer | Reflects the details on volume/dose of the products administered | 1=Packed red blood cell concentrates (pRBCs);2=Fresh whole blood;3=Fresh frozen plasma (FFP);4=Freeze dried plasma / lypholized plasma;5=Platelet concentrates;6=PCC (prothrombin complex concentrates);7=Fibrinogen concentrate;8=Albumine;9=Recombinant factor FVIIa;10=Tranexamic acid (TXA);11=Cryoprecipitate;12=Desmopression (DDAVP);13=Factor XIII;14=Calcium |
| TILFactorsCoagulationType2 | FALSE | TRUE | integer | Reflects the details on volume/dose of the products administered | 1=Packed red blood cell concentrates (pRBCs);2=Fresh whole blood;3=Fresh frozen plasma (FFP);4=Freeze dried plasma / lypholized plasma;5=Platelet concentrates;6=PCC (prothrombin complex concentrates);7=Fibrinogen concentrate;8=Albumine;9=Recombinant factor FVIIa;10=Tranexamic acid (TXA);11=Cryoprecipitate;12=Desmopression (DDAVP);13=Factor XIII;14=Calcium |
| TILFactorsCoagulationType3 | FALSE | TRUE | integer | Reflects the details on volume/dose of the products administered | 1=Packed red blood cell concentrates (pRBCs);2=Fresh whole blood;3=Fresh frozen plasma (FFP);4=Freeze dried plasma / lypholized plasma;5=Platelet concentrates;6=PCC (prothrombin complex concentrates);7=Fibrinogen concentrate;8=Albumine;9=Recombinant factor FVIIa;10=Tranexamic acid (TXA);11=Cryoprecipitate;12=Desmopression (DDAVP);13=Factor XIII;14=Calcium |
| TILFactorsCoagulationType4 | FALSE | TRUE | integer | Reflects the details on volume/dose of the products administered | 1=Packed red blood cell concentrates (pRBCs);2=Fresh whole blood;3=Fresh frozen plasma (FFP);4=Freeze dried plasma / lypholized plasma;5=Platelet concentrates;6=PCC (prothrombin complex concentrates);7=Fibrinogen concentrate;8=Albumine;9=Recombinant factor FVIIa;10=Tranexamic acid (TXA);11=Cryoprecipitate;12=Desmopression (DDAVP);13=Factor XIII;14=Calcium |
| TILFactorsCoagulationVolume1 | FALSE | TRUE | text | Reflects the details on volume/dose of the products administered | Not Applicable |
| TILFactorsCoagulationVolume2 | FALSE | TRUE | text | Reflects the details on volume/dose of the products administered | Not Applicable |
| TILFactorsCoagulationVolume3 | FALSE | TRUE | text | Reflects the details on volume/dose of the products administered | Not Applicable |
| TILFactorsCoagulationVolume4 | FALSE | TRUE | text | Reflects the details on volume/dose of the products administered | Not Applicable |
| TILFactorsGlucoseManagement | FALSE | TRUE | integer | Indicates whether glucose management was applied and if so, which therapy used (prophylactic, insulin administration of tight glycemic control). | 0=No specific therapy;1=Prophylactic;2=Insulin administration to correct hyperglycemias;3=Tight glycemic control (targeting blood glucose levels of 80-110mg/dL [4.4-6.1mmol/L]) |
| TILFever | FALSE | TRUE | integer | Records for the Daily TIL whether there was treatment of fever (temperature <38¬∞C) or spontaneous temperature of 34.5¬∞C | 0=No;1=Yes |
| TILFeverHypothermia | FALSE | FALSE | integer | Records for the Daily TIL whether there was hypothermia below 35¬∞C | 0=No;1=Yes |

| Name | Static | Intervention | Format | Description | Possible Values |
|---|---|---|---|---|---|
| TILFeverMildHypothermia | FALSE | FALSE | integer | Records for the Daily TIL whether there was mild hypothermia for ICP control with a lower limit of 35¬∞C. | 0=No;1=Yes |
| TILFluidColloids | FALSE | TRUE | integer | Indicates whether colloids were administered with regard to Fluid Balance. | 0=No;1=Yes;88=Unknown |
| TILFluidColloidsTotalVolume | FALSE | TRUE | integer | Total volume of colloids administered (in ml) | Not Applicable |
| TILFluidColloidsType | FALSE | TRUE | integer | Type of colloids administered | 1=Albumin 5%;2=Albumin 20%;3=Dextran;4=Gelatin (e.g. gelofusion);5=HES (hydroxyethyl starches);6=Tetrastarches (e.g. HES 130/04) |
| TILFluidIn | FALSE | TRUE | integer | Recorded preferably over 24-hour period, exact details to be derived from start and stop date/time for calculation | Not Applicable |
| TILFluidLoading | FALSE | TRUE | integer | Records for the Daily TIL whether there was fluid loading for maintenance of cerebral perfusion. | 0=No;1=Yes |
| TILFluidLoadingVasopressor | FALSE | FALSE | integer | Records for the Daily TIL whether there was vasopressor therapy required for management of cerebral perfusion | 0=No;1=Yes |
| TILFluidOutCSFDrain | FALSE | TRUE | integer | Daily TIL - Number of fluid out: CSF drainage in ml | Not Applicable |
| TILFluidOutGastric | FALSE | FALSE | integer | Daily TIL - Number of fluid out: Gastic loss in ml | Not Applicable |
| TILFluidOutOther | FALSE | FALSE | integer | Daily TIL - Number of fluid out: other fluid (than Urine, Gastic loss or CSF drainage) in ml | Not Applicable |
| TILFluidOutUrine | FALSE | FALSE | integer | Daily TIL - Number of fluid out: Urine in ml | Not Applicable |
| TILFluidsRenalReplacement | FALSE | FALSE | integer | Indicates for the fluid balance whether there was a need for renal replacement therapy. | 0=No;1=Yes |
| TILHyperosmolarTherapy | FALSE | TRUE | integer | Records for the Daily TIL whether there was Hyperosmolar therapy with mannitol up to 2 g/kg/24 hours | 0=No;1=Yes |
| TILHyperosomolarTherapyHigher | FALSE | TRUE | integer | Records for the Daily TIL whether there was Hyperosmolar therapy with hypertonic saline > 0.3 g/kg/24 hours | 0=No;1=Yes |
| TILHyperosomolarTherapyHypertonicLow | FALSE | TRUE | integer | Records for the Daily TIL whether there was Hyperosmolar therapy with hypertonic saline up to 0.3 g/kg/24 hours | 0=No;1=Yes |
| TILHyperosomolarTherapyMannitolGreater2g | FALSE | TRUE | integer | Records for the Daily TIL whether there was Hyperosmolar therapy with mannitol > 2 g/kg/24 hours | 0=No;1=Yes |
| TILHypertonicSalineDose | FALSE | TRUE | text | Indicates the total dose of hypertonic saline administered (in g.) if applicable. Calculated over a 24-hour period. | Not Applicable |
| TILHyperventilation | FALSE | TRUE | integer | Records for the Daily TIL whether there was Mild hypocapnia for ICP control [PaCO2 4.6 - 5.3 kPa (35 - 40 mmHg)] | 0=No;1=Yes |
| TILHyperventilationIntensive | FALSE | TRUE | integer | Records for the Daily TIL whether there was Intensive hypocapnia for ICP control [PaCO2 < 4.0 kPa (30 mmHg)] | 0=No;1=Yes |
| TILHyperventilationModerate | FALSE | TRUE | integer | Records for the Daily TIL whether there was Moderate hypocapnia for ICP control [PaCO2 4.0 - 4.5 kPa (30 - 35 mmHg)] | 0=No;1=Yes |
| TILMannitolDose | FALSE | TRUE | text | Indicates the total dose of Mannitol administered (in g.) if applicable. Calculated over a 24-hour period | Not Applicable |
| TILNoradrenalineDose | FALSE | TRUE | text | Indicates the total dose of Noradrenaline administered (in mg.) if applicable. Calculated over a 24-hour period | Not Applicable |
| TILOtherDose | FALSE | TRUE | text | Indicates the dose (in mg.) of other vasopressors drugs administered (if applicable) | Not Applicable |
| TILOtherTxt | FALSE | TRUE | text | Indicates which other vasopressors drugs was administered (if applicable) | Not Applicable |
| TILOtherVaso | FALSE | TRUE | integer | Indicates whether any other Vasopressor drug was administered (other than Dobutamine, Dopamine, Noradrenaline or Phenylephrine) | 0=No;1=Yes |
| TILPhenylephrineDose | FALSE | TRUE | text | Indicates the total dose of phenylephrine administered (in mg.) if applicable. Calculated over a 24-hour period | Not Applicable |

| Name | Static | Intervention | Format | Description | Possible Values |
|---|---|---|---|---|---|
| TILPhysicianConcernsContusionpregression | FALSE | FALSE | integer | Reflects daily the physician concern and satisfaction with regard to contusion progression. Scales from 1 (not concerned) to 10 (very concerned) | 1=1;2=2;3=3;4=4;5=5;6=6;7=7;8=8;9=9;10=10 |
| TILPhysicianConcernsEpilepsy | FALSE | FALSE | integer | Reflects daily the physician concern and satisfaction with regard to epilepsy. Scales from 1 (not concerned) to 10 (very concerned) | 1=1;2=2;3=3;4=4;5=5;6=6;7=7;8=8;9=9;10=10 |
| TILPhysicianConcernsFocalSwelling | FALSE | FALSE | integer | Reflects daily the physician concern and satisfaction with regard to focal swelling. Scales from 1 (not concerned) to 10 (very concerned) | 1=1;2=2;3=3;4=4;5=5;6=6;7=7;8=8;9=9;10=10 |
| TILPhysicianConcernsHematomaProgression | FALSE | FALSE | integer | Reflects daily the physician concern and satisfaction with regard to hematoma progression. Scales from 1 (not concerned) to 10 (very concerned) | 1=1;2=2;3=3;4=4;5=5;6=6;7=7;8=8;9=9;10=10 |
| TILPhysicianConcernsHypoperfusion | FALSE | FALSE | integer | Reflects daily the physician concern and satisfaction with regard to suspected hypoperfusion. Scales from 1 (not concerned) to 10 (very concerned) | 1=1;2=2;3=3;4=4;5=5;6=6;7=7;8=8;9=9;10=10 |
| TILPhysicianConcernsIntracranialInfection | FALSE | FALSE | integer | Reflects daily the physician concern and satisfaction with regard to intracranial infections. Scales from 1 (not concerned) to 10 (very concerned) | 1=1;2=2;3=3;4=4;5=5;6=6;7=7;8=8;9=9;10=10 |
| TILPhysicianConcernsVasospasm | FALSE | FALSE | integer | Reflects daily the physician concern and satisfaction with regard to vasospasm. Scales from 1 (not concerned) to 10 (very concerned) | 1=1;2=2;3=3;4=4;5=5;6=6;7=7;8=8;9=9;10=10 |
| TILPhysicianOverallSatisfaction | FALSE | FALSE | integer | This variable aims to capture the overall satisfaction of the physician with the clinical course of this patient; "not at all satisfied" would indicate that the patient did much more poorly than expected; "very satisfied" would indicate that the patient did much better than expected. Physician satisfaction should be assessed on a daily basis, | 0=Not at all;1=Slightly;2=Moderately;3=Quite;4=Very |
| TILPhysicianOverallSatisfactionSurvival | FALSE | FALSE | integer | This variable aims to capture the opinion of the treating physician as to whether the short time survival change have chnged in comparision to the previous assessment | 1=Much worse;2=A little worse;3=Unchanged;4=A little better;5=Much better |
| TILPosition | FALSE | TRUE | integer | Records for the Daily TIL whether there was head elevation for ICP control | 0=No;1=Yes |
| TILPositionNursedFlat | FALSE | TRUE | integer | Records for the Daily TIL whether there was a patient position of Nursed flat (180¬∞C) for CPP management | 0=No;1=Yes |
| TILReasonForChange | FALSE | FALSE | integer | Reflects the reason for change in TIL therapy over the day. | 0=No change;1=Intensified: Clinical deterioration;2=Intensified:Suspicion of increased of ICP (not measured);3=Intensified:Increased ICP (documented);4=Intensified:Clinical decision to target other mechanism;5=Intensified:Change of doctor (different shift);6=Decreasing:Clinical improvement;7=Decreasing:Adequate control over ICP;8=Decreasing:Upper treatment limit reached/past;9=Decreasing:Further treatment considered futile;10=Decreasing:Change of doctor (different shift) |
| TILSedation | FALSE | TRUE | integer | Records for the Daily TIL whether there was sedation (low dose as required for mechanical ventilation) | 0=No;1=Yes |
| TILSedationHigher | FALSE | TRUE | integer | Records for the Daily TIL whether there was a higher dose sedation for ICP control (not aiming for burst supression) | 0=No;1=Yes |
| TILSedationMetabolic | FALSE | TRUE | integer | Records for the Daily TIL whether there was metabolic suppression for ICP control with high dose barbiturates or propofol | 0=No;1=Yes |
| TILSedationNeuromuscular | FALSE | TRUE | integer | Records for the Daily TIL whether there was neuromuscular blockade (paralysis) | 0=No;1=Yes |
| TILSedationScaleUsed | FALSE | TRUE | integer | Reflects with regard to sedation management whether a sedation scale (SAS, RASS, MASS, Ramsay, etc) was used to adjust sedatives? | 0=No;1=Yes;77=N/A |
| TILSedativesInterrupted | FALSE | TRUE | integer | Reflects with regard to sedation management whether infusions of sedatives (opioids) were interrupted during the day if the patient was receiving any. | 0=No;1=Yes;77=N/A |

| Name | Static | Intervention | Format | Description | Possible Values |
| --- | --- | --- | --- | --- | --- |
| TotalTIL | FALSE | TRUE | integer | Calculated centrally - 24 hour TILS as the worst sum TILs for each day for the ICU timepoints (day 1-7, 10, 14, 21 and 28) | Not Applicable |
| Tracheostomy | FALSE | TRUE | integer | Describes absence or presence of a Tracheostomy. | 0=No;1=Yes |
| Transfer | FALSE | FALSE | | | Not Applicable |
| TransReason | FALSE | FALSE | integer | WHY question: documents reason for transition of care | 1=Mechanical ventilation;2=Frequent neurological observations;3=Haemodynamic invasive monitoring;4=Extracranial injuries;5=Neurological operation;6=Clinical deterioration;7=CT abnormalities;8=Clinical observation for TBI;9=No ICU bed available;10=Could be discharged home, but no adequate supervision;11=Improvement;12=Neurological deterioration;13=Systemic compilation;14=CT progression;15=Planned surgery;16=Condition stable;17=(acute) Treatment goals accomplished;18=Need to free a bed;19=Further improvement;20=Clinical rehab completed;21=Lack of improvement;22=Late neurological deterioration;23=Problems unrelated to trauma;24=Post operative care;25=Neurological complication;99=Other |
| TransTiming | FALSE | FALSE | text | Transitions of care may be premature or delayed because of logistic problems. This variable aims to capture such information. | APP=Appropriate;DEL=Delayed;PRE=Premature |
| UrineCath | FALSE | TRUE | integer | Describes absence or presence of an Urinary catheter. | 0=No;1=Yes |

*ICU vitals and assessments*

| Name | Static | Intervention | Format | Description | Possible Values |
| --- | --- | --- | --- | --- | --- |
| DVAssmtConditions | FALSE | FALSE | integer | Reflects the assessment condition for the best GCS assessments as part of the daily vitals. Only applicable to ICU stratum. | 0=No sedation or paralysis;1=Sedated;2=Paralyzed;3=Temporary stop of sedation/paralysis;4=Reversal of sedation/paralysis;5=Active reversal (pharmacologic) of sedation/paralysis;99=Other |
| DVAssmtConditions Worst | FALSE | FALSE | integer | Reflects the assessment condition for the worst GCS assessments as part of the daily vitals. Only applicable to ICU stratum. | 0=No sedation or paralysis;1=Sedated;2=Paralyzed;3=Temporary stop of sedation/paralysis;4=Reversal of sedation/paralysis;5=Active reversal (pharmacologic) of sedation/paralysis;99=Other |
| DVBestPupilSymmetry | FALSE | FALSE | integer | Reflects if the pupil are of equal size (symmetry) during the day as part of the Daily vitals. | 1=Equal;2=Unequal R>L;3=Unequal L>R |
| DVBloodOxySatHigh | FALSE | FALSE | decimal | Reflects Highest oxygen saturation (in %) as determined from Arterial Blood Gas | Not Applicable |
| DVBloodOxySatLow | FALSE | FALSE | decimal | Reflects Lowest oxygen saturation (in %) as determined from Arterial Blood Gas | Not Applicable |
| DVChangeCauseWorst | FALSE | FALSE | integer | Reflects the cause of change for the worst GCS as part of the Daily Vitals. | 1=Mainly intracranial;2=Mainly extracranial;3=Both to an equal extent;88=Unknown |
| DVChangeInOneDay | FALSE | FALSE | integer | Reflects presence or absence of change over past 24 hours for the Best GCS as part of the Daily Vitals. | 0=No change;1=Improving;2=Episode of deterioration;3=Sustained deterioration;4=Fluctuating |
| DVChangeInOneDay Worst | FALSE | FALSE | integer | Reflects presence or absence of change over past 24 hours for the Worst GCS as part of the Daily Vitals. | 0=No change;1=Improving;2=Episode of deterioration;3=Sustained deterioration;4=Fluctuating |
| DVDBP | FALSE | FALSE | integer | The diastolic blood pressure at the timepoint of the highest systolic blood pressure. | Not Applicable |
| DVDBPLow | FALSE | FALSE | integer | The diastolic blood pressure at the timepoint of the lowest systolic blood pressure. | Not Applicable |
| DVFiO2AtHighPaO2 | FALSE | FALSE | decimal | FiO2 at time of sampling - Highest PaO2/FiO3 (in %) | Not Applicable |
| DVFiO2AtLowPaO2 | FALSE | FALSE | decimal | FiO2 at time of sampling - Lowest PaO2/FiO3 (in %) | Not Applicable |
| DVFourScoreBrainstem | FALSE | FALSE | integer | Four Score for the Brainstem reflexes | 0=Absent pupil, corneal;1=Pupil and corneal reflexes absent;2=Pupil or corneal reflexes absent;3=One pupil wide and fixed;4=Pupil and corneal reflexes present;88=Unknown |
| DVFourScoreEye | FALSE | FALSE | integer | Four Score for the Eye response | 0=Eyelids remain closed with pain;1=Eyelids closed but opens to pain;2=Eyelids closed but opens to loud voice;3=Eyelids open but not tracking;4=Eyelids open or opened tracking or blinking to command;88=Unknown |
| DVFourScoreMotor | FALSE | FALSE | integer | Four Score for the Motor response | 0=No response to pain or generalized myoclonus status epilepticus;1=Extensor posturing;2=Flexion response to pain;3=Localizing to pain;4=Thumbs up, fist;88=Unknown |
| DVFourScoreRespiration | FALSE | FALSE | integer | Four Score for the Respiration | 0=Breathes at ventilator rate or apnea;1=Breathes above ventilator rate;2=Not intubated, irregular breathing pattern;3=Not intubated, Cheyne- Stokes breathing pattern;4=Not intubated, regular breathing pattern;88=Unknown |
| DVFourScoreTotal | FALSE | FALSE | integer | DVFourScoreEye + DVFourScoreMotor + DVFourScoreBraintstem + | Not Applicable |

| Name | Static | Intervention | Format | Description | Possible Values |
|---|---|---|---|---|---|
| | | | | DVFourScoreRespiration. No score if any of the components =88 (unknown) or NULL | |
| DVGCSBest | FALSE | FALSE | integer | Reflects if the Worst GCS is the same as the best GCS as part of the Daily Vitals. | 0=No;1=Yes |
| DVGCSBestChangeCause | FALSE | FALSE | integer | Cause of change for the Best GCS as part of the Daily Vitals. | 1=Mainly intracranial;2=Mainly extracranial;3=Both to an equal extent;88=Unknown |
| DVGCSEyes | FALSE | FALSE | text | Best GCS eye opening as part of the Daily Vitals. | 1=1-None;2=2-To pain;3=3-To speech;4=4-Spontaneously;O=Untestable (Other);S=Untestable (swollen);UNK=Unknown |
| DVGCSMotor | FALSE | FALSE | text | Best GCS motor score as part of the Daily Vitals. | 1=1-None;2=2-Abnormal extension;3=3-Abnormal flexion;4=4-Normal flexion/withdrawal;5=5-Localizes to pain;6=6-Obeys command;O=Untestable (Other);P=Untestable (Deep sedation/paralyzed);UN=Unknown |
| DVGCSScore | FALSE | FALSE | text | Best GCS Score: DVGCSEyes + DVGCSMotor + DVGCSVerbal. If one or more of these is Untestable or unknown then = "No Sum" | 3=3;4=4;5=5;6=6;7=7;8=8;9=9;10=10;11=11;12=12;13=13;14=14;15=15;No Sum=Untestable/Unknown |
| DVGCSVerbal | FALSE | FALSE | text | Best GCS verbal score as part of the daily vitals. | 1=1-None;2=2- Incomprehensible sound;3=3-Inappropriate words;4=4-Confused;5=5-Oriented;O=Untestable (Other);T=Untestable (Tracheotomy/endotracheal tube);UN=Unknown |
| DVGCSWorstEyes | FALSE | FALSE | text | Worst GCS eye opening as part of the daily vitals. | 1=1-None;2=2-To pain;3=3-To speech;4=4-Spontaneously;O=Untestable (Other);S=Untestable (swollen);UNK=Unknown |
| DVGCSWorstMotor | FALSE | FALSE | text | Worst GCS motor score as part of the daily vitals. | 1=1-None;2=2-Abnormal extension;3=3-Abnormal flexion;4=4-Normal flexion/withdrawal;5=5-Localizes to pain;6=6-Obeys command;O=Untestable (Other);P=Untestable (Deep sedation/paralyzed);UN=Unknown |
| DVGCSWorstScore | FALSE | FALSE | text | Worst GCS: DVGCSWorstEyes + DVGCSWorstMotor + DVGCSWorstVerbal. If one or more of these is Untestable or unknown then = "No Sum" | 3=3;4=4;5=5;6=6;7=7;8=8;9=9;10=10;11=11;12=12;13=13;14=14;15=15;No Sum=Untestable/Unknown |
| DVGCSWorstVerbal | FALSE | FALSE | text | Worst GCS Verbal score as part of the daily vitals | 1=1-None;2=2- Incomprehensible sound;3=3-Inappropriate words;4=4-Confused;5=5-Oriented;O=Untestable (Other);T=Untestable (Tracheotomy/endotracheal tube);UN=Unknown |
| DVHighestPaCO2 | FALSE | FALSE | decimal | Highest PaCO2 on a day, in mmHg. | Not Applicable |
| DVHighestPaO2 | FALSE | FALSE | decimal | Highest PaO2 on a day, in mmHg | Not Applicable |
| DVHighestPaO2OverHighestFiO2 | FALSE | FALSE | decimal | FiO2 at time of sampling - Highest PaO2/FiO4 | Not Applicable |
| DVHighestpH | FALSE | FALSE | decimal | Highest pH as part of the Daily vitals Blood Gas Analysis | Not Applicable |
| DVHR | FALSE | FALSE | integer | The highest heart rate during the day. | Not Applicable |
| DVHRLow | FALSE | FALSE | integer | The lowest heart rate of a day. | Not Applicable |
| DVLowestPaCO2 | FALSE | FALSE | decimal | The lowest value of PaCO2 measured on each day, in mmHg. | Not Applicable |
| DVLowestPaO2 | FALSE | FALSE | decimal | Lowest PaO2 in mmHg as part of the Daily Vitals Blood Gas Analysis | Not Applicable |
| DVLowestPaO2OverLowestFiO2 | FALSE | FALSE | decimal | FiO2 at time of sampling - Lowest PaO2/FiO4 as part of the Daily Vitals | Not Applicable |
| DVLowestpH | FALSE | FALSE | decimal | Lowest pH as part of the Daily Vitals Blood Gas Analysis | Not Applicable |
| DVPupilLftEyeMeasr | FALSE | FALSE | integer | Best Pupil Left eye size | Not Applicable |
| DVPupilReactivityLghtLftEyeReslt | FALSE | FALSE | integer | The best pupillary reactivity of the left eye during the day | 1=+ (Brisk);2=+ (Sluggish);3=- (Negative) |
| DVPupilReactivityLghtRtEyeReslt | FALSE | FALSE | integer | Best pupillary reactivity of the right eye during the day | 1=+ (Brisk);2=+ (Sluggish);3=- (Negative) |
| DVPupilRtEyeMeasr | FALSE | FALSE | integer | Best Pupil Right eye Size | Not Applicable |
| DVSBP | FALSE | FALSE | integer | The highest systolic blood pressure for each day where vitals are recorded. | Not Applicable |
| DVSBPLow | FALSE | FALSE | integer | The lowest systolic blood pressure for each day. | Not Applicable |
| DVSpO2 | FALSE | FALSE | integer | Highest Oxygen saturation (pulse oximetry) in % for each day where vitals are recorded. | Not Applicable |
| DVSpO2Low | FALSE | FALSE | integer | Lowest Oxygen saturation (pulse oximetry) in % for each day where vitals are recorded. | Not Applicable |

| Name | Static | Intervention | Format | Description | Possible Values |
|---|---|---|---|---|---|
| DVTempHighC | FALSE | FALSE | decimal | The highest body temperature of the day in degrees Celsius. For the associated time point, see Vitals.DVTempHighDateTime. For location of where the temperature was measured, see Vitals.DVTempLocation. | Not Applicable |
| DVTempLocation | FALSE | FALSE | integer | Location of thermometer probe for temperature measurement. | 1=External-axillary;2=External-skin;3=Core- rectal;4=Core- bladder;5=Core- oesophageal;6=Core- tympanic;7=Core- nasopharynx |
| DVTempLowC | FALSE | FALSE | decimal | The lowest temperature of a day in degrees Celsius. For the timepoint, see Vitals.DVTempLowDateTime. For location of where the temperature was measured, see Vitals.DVTempLocation. | Not Applicable |
| DVWorstPupilLftEyeMeasr | FALSE | FALSE | integer | Worst Pupils Left eye Size | Not Applicable |
| DVWorstPupilReactivityLghtLftEyeResult | FALSE | FALSE | integer | Worst Pupils Left eye Reactivity | 1=+ (Brisk);2=+ (Sluggish);3=- (Negative) |
| DVWorstPupilReactivityLghtRghtEyeResult | FALSE | FALSE | integer | Worst Pupils Right eye Reactivity | 1=+ (Brisk);2=+ (Sluggish);3=- (Negative) |
| DVWorstPupilRghtEyeMeasr | FALSE | FALSE | integer | Worst Pupils Right eye Size | Not Applicable |
| DVWorstPupilSymmetry | FALSE | FALSE | integer | Worst Pupil symmetry | 1=Equal;2=Unequal R>L;3=Unequal L>R |
| HosComplEventHypocapnia | FALSE | FALSE | integer | Reflects Inadvertent hypocapnia. Second Insults reported in Vitals relate to the the hospital phase (both ward and ICU). Inadvertent hypocapnia is defined as a PaCO2 <=3.3 kPa (25 mmHg) | 0=No;1=Single episode, short duration;2=Multiple episodes or prolonged duration;88=Unknown |
| HosComplEventHypotension | FALSE | FALSE | integer | Reflects Hypotensive Episodes. Second Insults reported in Vitals relate to the the hospital phase (both ward and ICU). Pre-hospital hypoxia is documented at: InjuryHx.EDComplEventHypotension. Definite hypotension is defined as a documented systolic BP < 90 mm Hg (adults) | 0=No;1=Single episode, short duration;2=Multiple episodes or prolonged duration;88=Unknown |
| HosComplEventHypoxia | FALSE | FALSE | integer | Reflects Hypoxic Episodes. Second Insults reported in Vitals relate to the hospital phase (both ward and ICU). Pre-hospital hypoxia is documented at: InjuryHx.EDComplEventHypoxia. Definite hypoxia is defined as a documented PaO2 <8 kPa (60 mm Hg) and/or SaO2<90% | 0=No;1=Single episode, short duration;2=Multiple episodes or prolonged duration;88=Unknown |
| HosComplEventSeizures | FALSE | FALSE | integer | Reflects absence or presence of seizures as Intracranial Second Insult. Second Insults reported in Vitals relate to the the hospital phase (both ward and ICU). | 0=No;1=Single episode, short duration;2=Partial/Focal;3=Generalized;4=Status epilepticus;5=Silent seizure activity (only electrical, no clinical manifestation);88=Unknown |
| HospSecondInsultsNeuroWorse | FALSE | FALSE | integer | The importance of Neuroworsening was first described by Morris and Marshall. The occurrence of neuroworsening is related to poorer outcome in subjects with moderate to severe TBI. Neuroworsening is defined as 1) a decrease in GCS motor score of 2 or more points; 2) a new loss of pupillary reactivity or development of pupillary assymmetry >= 2mm; 3) deterioration in neurological or CT status sufficient to warrant immediate medical or surgical intervention | 0=No;1=Yes;88=Unknown |
| HospSecondInsultsNeuroWorseYes1 | FALSE | FALSE | integer | This variable provides a specification of the type of neuroworsening if it occurs. | 1=Decrease in motor score >= 2 points; 2=Development of pupillary abnormalities; 3=Other neurological and/or CT deterioration |
| HospSecondInsultsNeuroWorseYes2 | FALSE | FALSE | integer | This variable provides a specification of the type of neuroworsening if it occurs. | 1=Decrease in motor score >= 2 points; 2=Development of pupillary abnormalities; 3=Other neurological and/or CT deterioration |
| HospSecondInsultsNeuroWorseYes3 | FALSE | FALSE | integer | This variable provides a specification of the type of neuroworsening if it occurs. | 1=Decrease in motor score >= 2 points; 2=Development of pupillary abnormalities; 3=Other neurological and/or CT deterioration |
| HourlyValueAccurate | FALSE | FALSE | integer | Are the measured values for ICP over this day considered to be accurate? An explanation can be found in HourlyValues.HourlyValueNotAccurateProblems. Associated with the date in HourlyValues.HVDate. | 0=No;1=Yes;2=Doubtful |
| HourlyValueLevelABP | FALSE | FALSE | integer | Differences in level of zeroing the arterial blood pressure transducer may affect | 1=Right atrium;2=Level of arterial catheter;99=Other |

| Name | Static | Intervention | Format | Description | Possible Values |
|---|---|---|---|---|---|
| | | | | calculations of CPP and comparisons between centres. | |
| HourlyValueLevelABPOther | FALSE | FALSE | text | A free text description of the level of ABP transducer if HourlyValues.HourlyValueLevelABP is "other". | Not Applicable |
| HourlyValueNotAccurateProblems | FALSE | FALSE | text | Explanation in free text of the variable HourlyValues.HourlyValueAccurate. Associated with the date in HourlyValues.HVDate. | Not Applicable |
| HVDBP | FALSE | FALSE | integer | Bihourly diastolic blood pressure | Not Applicable |
| HVSBP | FALSE | FALSE | integer | Bihourly systolic blood pressure | Not Applicable |

*Surgery and neuromonitoring*

| Name | Static | Intervention | Format | Description | Possible Values |
|---|---|---|---|---|---|
| DecompressiveCran | TRUE | FALSE | integer | These variables specifically focus on decompressive craniectomy (DC). Considerable uncertainty exists on which patients may benefit from DC, as well as on the timing. Previous studies have shown that in approximately one third of cases the DC was too small in size. | 0=No;1=Yes |
| DecompressiveCranLocation | TRUE | FALSE | integer | Documents type of decompressive craniectomy. | 1=Bifrontal;2=Hemicraniectomy- left side;3=Hemicraniectomy- right side;4=Posterior fossa |
| DecompressiveCranReason | TRUE | FALSE | integer | WHY question: documents main reason for performing DC | 1=Pre-emptive approach to treatment of (suspected) raised ICP (not last resort);2=Raised ICP, refractory to medical management (last resort);3=ICP not monitored, but CT evidence of raised ICP;4=Not directly planned, but decided on because of intra-operative brain swelling;5=Routinely performed with every ASDH or Contusion evacuation;6=Development of cerebral infarction |
| DecompressiveCranType | TRUE | FALSE | integer | The indications for the decompressive craniectomy can be documented here. | 1=Isolated procedure;2=In association with ASDH removal;3=In association with contusion/ICH removal;4=In association with ASDH and contusion/ICH removal |
| DecompressiveSize | TRUE | FALSE | integer | In some sites, the size of DC was recorded by Investigators. No generable way of calculating was proposed. | Not Applicable |
| DVBrainTempCLow | FALSE | FALSE | decimal | Brain temperature at the timepoint of the lowest body temperature of the day, degrees Celsius. | Not Applicable |
| DVTempBrainC | FALSE | FALSE | decimal | Brain temperature at the timepoint of the highest body temperature of the day, degrees Celsius. | Not Applicable |
| HourlyValueICPDiscontinued | FALSE | TRUE | integer | Answer to the question: "Was active ICP treatment discontinued (due to poor prognosis)?" | 0=No;1=Yes |
| HourlyValueLevelICP | FALSE | FALSE | integer | Only applicable in case ICP monitored; Differences in level of zeroing ICP may affect calculation of CPP and comparisons between centres. | 1=Foramen of Monro;2=Same level as ABP;3=Meatus externus (ear) |
| HVCPP | FALSE | FALSE | integer | Bihourly cerebral perfusion pressure | Not Applicable |
| HVICP | FALSE | FALSE | integer | Bihourly intracranial pressure | Not Applicable |
| ICPMonitorStop | FALSE | TRUE | integer | If ICP monitoring has stopped or not. Time and date is found in Hospital.ICPRemTime and Hospital.ICPRemDate. Reason for stopping ICP monitoring is found in Hospital.ICPStopReason. | 0=No;1=Yes |
| ICPMonitorStopReasonOther | FALSE | FALSE | text | Specifies the "other" reason if the reason for stopping ICP was not on the pre-defined list. Check also "Hospital.ICPStopReason" | Not Applicable |
| ICUCatheterICP | FALSE | TRUE | integer | Reflects if the ICP catheter has been revised in case of ICP monitoring | 0=No;1=Yes |
| ICUProblemsICP | FALSE | FALSE | integer | In case of brain specific monitoring in the ICU, all details on the ICP monitoring were recorded. This variable reflects Problems in ICP monitoring. | 0=No;1=Yes |
| ICUProblemsICPYes | FALSE | FALSE | integer | In case of brain specific monitoring in the ICU, all details on the ICP monitoring were | 1=Accidental catheter removal;2=Catheter obstruction/failure;3=Suspicion of inaccurate measurement |

| Name | Static | Intervention | Format | Description | Possible Values |
|---|---|---|---|---|---|
| | | | | recorded. This variable explains the Problems in ICP monitoring if applicable. | |
| ICURaisedICP | FALSE | FALSE | integer | In case of brain specific monitoring in the ICU, all details on the ICP monitoring were recorded. This variable reflects if there was Raised ICP (sustained). | 0=No;1=Yes, controlled;2=Yes, refractory |
| ICUReasonForTypeICPMont | FALSE | FALSE | integer | In case of brain specific monitoring in the ICU, all details on the ICP monitoring were recorded. This variable reflects WHY Question: reason for choosing ventricular monitor | 1=Routine in our department;2=Not routine, but enlarged ventricles;3=No parenchymal device available;99=Other |
| ICUReasonForTypeICPMontPare | FALSE | FALSE | integer | In case of brain specific monitoring in the ICU, all details on the ICP monitoring were recorded. This variable reflects WHY Question: reason for choosing parenchymal monitor. | 1=Routine in our department;2=Not routine, but small ventricles;3=Mainly motivated by time of day;4=No OR available for placement ventr. catheter;5=Failed implantation ventricular catheter;99=Other |
| ICUReasonForTypeICUMontOther | FALSE | FALSE | text | In case of brain specific monitoring in the ICU, all details on the ICP monitoring were recorded. This variable reflects "Other Reason of choice for ventricular/ventricular+sensor monitoring" | Not Applicable |
| ICUReasonForTypeICUMontParOther | FALSE | FALSE | text | In case of brain specific monitoring in the ICU, all details on the ICP monitoring were recorded. This variable reflects 'Other Reason of choice for parenchymal sensor'. | Not Applicable |
| ICUReasonICP | FALSE | FALSE | integer | In case of brain specific monitoring in the ICU, all details on the ICP monitoring were recorded. This variable reflects WHY Question: documents reason for monitoring ICP in patient admitted to ICU | 1=Guideline criteria;2=Radiological signs raised ICP;3=Clinical suspicion raised ICP;4=Anaesthesia or mechanical ventilation required for extracranial injuries;5=To inform surgical indication for mass lesion;99=Other |
| ICUReasonICPOther | FALSE | FALSE | text | In case of brain specific monitoring in the ICU, all details on the ICP monitoring were recorded. This variable is a free text field for description of why the patient has ICP monitoring if there is another reason than pre-specified in Hospital.ICUReasonICP. | Not Applicable |
| ShortTermSurvivalNoSurg | FALSE | FALSE | integer | Aims to capture information on the surgeon's expectations, eg if the surgeon considers a realistic expectation of benefit, or performs the surgery as a "last resort" in a likely hopeless case --> The short term survival chances of the patient if I DO NOT operate (1-100) | Not Applicable |
| ShortTermSurvivalYesSurg | FALSE | FALSE | integer | Aims to capture information on the surgeon's expectations, eg if the surgeon considers a realistic expectation of benefit, or performs the surgery as a "last resort" in a likely hopeless case --> The short term survival chances of the patient if I DO operate (1-100) | Not Applicable |
| SurgeryCranialDelay | FALSE | FALSE | integer | Reflects reason for delay of cranial surgery (if any). | 1=Transferral from other hospital;2=Haemodynamic instability;3=No OR available;4=Surgeon delayed;5=No delay;99=Other |
| SurgeryCranialReason | FALSE | FALSE | integer | WHY Question: aims to document the reason for intracranial surgery | 1=Emergency/Life saving;2=Clinical deterioration;3=Mass effect on CT;4=Radiological progression;5=suspicion of) raised ICP;6=Guideline adherence;7=To prevent deterioration |
| SurgeryDescCranial | FALSE | TRUE | integer | The cranial surgeries information was to be entered in the e-CRF in tables for which you could add as many rows as you wish. The tables consisted of - Surgery start date - Surgery start time - Surgery end date - Surgery end time - Cranial surgery code - Reason - Delay - Short time survival if you do not operate - Short time survival if you do operate For the ,Äúcode,Äù there was a choice of 23 values in a drop-down box. For each row you entered in the table, you could select only 1 code in the drop-down. But, if a patient had several surgical options that were applicable, you could enter several rows with each the same start and end date, but marking the different applicable surgical codes in each row separately. | 1=Aneurysm (non trauma);2=Acute subdural hematoma;3=Contusion;4=Craniofacial surgery;5=CSF shunt;6=Chronic subdural hematoma;7=Decompressive craniectomy-hemicraniectomy;8=Depressed skull fracture;9=Epidural hematoma;10=Intracerebral hematoma;11=Infection;12=Optic nerve decompression;13=Posterior fossa surgery;14=Skull base fracture;15=Ventriculostomy for CSF drainage;16=Debridement minimal for penetrating injuries;17=Debridement extensive for penetrating injuries;18=Foreign body removal;19=Bone flap replacement;20=Cranioplasty;21=Other;71=Decompressive craniectomy - bifrontal;72=Decompressive craniectomy - removal previous bone flap |
| SurgeryDescExtraCranial | FALSE | TRUE | integer | The extra-cranial surgeries information was to be entered in the e-CRF in tables for | 22=Maxillofacial;23=Extremity fracture lower limb (internal fixation);24=Extremity fracture lower limb (external |

| Name | Static | Intervention | Format | Description | Possible Values |
|---|---|---|---|---|---|
| | | | | which you could add as many rows as you wish. The tables consisted of - Surgery start date - Surgery start time - Surgery end date - Surgery end time - Extracranial surgery code - Reason - Delay - Short time survival if you do not operate - Short time survival if you do operate For the ‚Äúcode‚Äù there was a choice of 18 values in a drop-down box. For each row you entered in the table, you could select only 1 code in the drop-down. But, if a patient had several surgical options that were applicable, you could enter several rows with each the same start and end date, but marking the different applicable surgical codes in each row separately. | fixation);25=Extremity fracture upper limb (internal fixation);26=Extremity fracture upper limb (external fixation);27=Fasciotomy;28=Laparotomy (abdomen);29=Pelvic fracture (internal fixation);30=Pelvic fracture (external fixation);31=Spinal stabilisation/cervical;32=Spinal stabilisation/thoracic;33=Spinal stabilisation/lumbar;34=Thoracotomy;35=Tracheostomy;36=Vascular (operative);37=Vascular (endovascular treatment);38=Wound closure/graft;39=Other |
| SurgeryExtraCranialDelay | FALSE | FALSE | integer | The extra-cranial surgeries information was to be entered in the e-CRF in tables for which you could add as many rows as you wish. The tables consisted of - Surgery start date - Surgery start time - Surgery end date - Surgery end time - Extracranial surgery code - Reason - Delay - Short time survival if you do not operate - Short time survival if you do operate | 1=Transferral from other hospital;2=Haemodynamic instability;3=No OR available;4=Surgeon delayed;5=No Delay;99=Other |
| SurgeryExtraCranialReason | FALSE | FALSE | integer | The extra-cranial surgeries information was to be entered in the e-CRF in tables for which you could add as many rows as you wish. The tables consisted of - Surgery start date - Surgery start time - Surgery end date - Surgery end time - Extracranial surgery code - Reason - Delay - Short time survival if you do not operate - Short time survival if you do operate | 1=Emergency/Lifesaving;2=Elective;3=Treatment of complication;4=Airway management;99=Other |
| SurgIntervenAppro | TRUE | FALSE | integer | WHY question: How strongly does the surgeon feels that this surgical intervention is appropriate in terms of the expected benefit to final clinical outcome? | 0=0;1=1;2=2;3=3;4=4;5=5;6=6;7=7;8=8;9=9;10=10 |
| TILICPSurgery | FALSE | TRUE | integer | Records for the Daily TIL whether there was an intracranial operation for progressive mass lesion, not scheduled on admission | 0=No;1=Yes |
| TILICPSurgeryDecomCranectomy | FALSE | TRUE | integer | Records for the Daily TIL whether there was a Decompressive Craniectomy | 0=No;1=Yes |
| TILPhysicianConcernsCPP | FALSE | FALSE | integer | Reflects daily the physician concern and satisfaction with regard to CPP. Scales from 1 (not concerned) to 10 (very concerned) | 1=1;2=2;3=3;4=4;5=5;6=6;7=7;8=8;9=9;10=10 |
| TILPhysicianConcernsICP | FALSE | FALSE | integer | Reflects daily the physician concern and satisfaction with regard to ICP. Scales from 1 (not concerned) to 10 (very concerned) | 1=1;2=2;3=3;4=4;5=5;6=6;7=7;8=8;9=9;10=10 |
| TILPhysicianSatICP | FALSE | FALSE | integer | Documents physician satisfaction with ICP control | 1=Not at all;2=Slightly;3=Moderate;4=Quite;5=Very;77=N/A (no ICP monitoring) |

## Supplementary Note 2. Physician-based impression variables.

| Name | Static | Format | Description | Possible Values |
|---|---|---|---|---|
| InjAIS | TRUE | integer | In the original AIS classification of injury severity, the grading is from 1 (minor) to 6 (unsurvivable). We added a score of 0 to designate absence of injuries. This is the AIS score for body regions as specified by AIS.InjBodyRegion. | 0=None;1=Minor: no treatment needed;2=Moderate: requires only outpatient treatment;3=Serious: requires non-ICU hospital admission;4=Severe: requires ICU observation and/or basic treatment;5=Critical: requires intubation, mechanical ventilation or vasopressors for blood pressure support;6=Unsurvivable: not survivable |
| HVTILChangeReason | FALSE | integer | Bihourly reason for change in therapy intensity level | 1=Intensified: Clinical deterioration;2=Intensified: Suspicion of increased of ICP (not measured);3=Intensified: Increased ICP (documented);4=Intensified: Clinical decision to target other mechanism;5=Intensified: Change of doctor (different shift);6=Decreasing: Clinical improvement;7=Decreasing: Adequate control over ICP;8=Decreasing: Upper treatment limit reached/past;9=Decreasing: Further treatment considered futile;10=Decreasing: Change of doctor (different shift) |
| TILPhysicianOverallSatisfaction | FALSE | integer | This variable aims to capture the overall satisfaction of the physician with the clinical course of this patient; "not at all satisfied" would indicate that the patient did much more poorly than expected; "very satisfied" would indicate that the patient did much better than expected. Physician satisfaction should be assessed on a daily basis, | 0=Not at all;1=Slightly;2=Moderately;3=Quite;4=Very |
| TILPhysicianOverallSatisfactionSurvival | FALSE | integer | This variable aims to capture the opinion of the treating physician as to whether the short time survival change have chnged in comparision to the previous assessment | 1=Much worse;2=A little worse;3=Unchanged;4=A little better;5=Much better |
| TILReasonForChange | FALSE | integer | Reflects the reason for change in TIL therapy over the day. | 0=No change;1=Intensified: Clinical deterioration;2=Intensified:Suspicion of increased of ICP (not measured);3=Intensified:Increased ICP (documented);4=Intensified:Clinical decision to target other mechanism;5=Intensified:Change of doctor (different shift);6=Decreasing:Clinical improvement;7=Decreasing:Adequate control over ICP;8=Decreasing:Upper treatment limit reached/past;9=Decreasing:Further treatment considered futile;10=Decreasing:Change of doctor (different shift) |
| TotalTIL | FALSE | integer | Calculated centrally - 24 hour TILS as the worst sum TILs for each day for the ICU timepoints (day 1-7, 10, 14, 21 and 28) | Not Applicable |
| AbdomenPelvicContentsAIS | TRUE | integer | AIS score for the Abdomen/Pelvic Contents In the original AIS classification of injury severity, the grading is from 1 (minor) to 6 (unsurvivable). We added a score of 0 to designate absence of injuries. | 0=None;1=Minor: no treatment needed;2=Moderate: requires only outpatient treatment;3=Serious: requires non-ICU hospital admission;4=Severe: requires ICU observation and/or basic treatment;5=Critical: requires intubation, mechanical ventilation or vasopressors for blood pressure support;6=Unsurvivable: not survivable |
| BaselineGOS6MoExpectedDeathRisk | TRUE | text | At ER discharge, physician estimate of six month outcome was recorded as a baseline risk assessment: "Given all current available information, what is, in your subjective opinion, the most likely 6-month outcome of this patient? To be based upon information on discharge ER or admission to hospital/ICU". This reflects the Risk of death in % | Not Applicable |
| BaselineGOS6MoExpectedOutcome | TRUE | text | At ER discharge, physician estimate of six month outcome was recorded as a baseline risk assessment: "Given all current available information, what is, in your subjective opinion, the most likely 6-month outcome of this patient? To be based upon information on discharge ER or admission to hospital/ICU". This reflects the Expected outcome (GOS) | D=D - Death;GR=GR - Good Recovery;MD=MD - Moderate Disability;SD=SD - Severe Disability;V=V - Vegetative State |
| BaselineGOS6MoUnfavourableOutcomeRisk | TRUE | text | At ER discharge, physician estimate of six month outcome was recorded as a baseline risk assessment: "Given all current available information, what is, in your subjective opinion, the most likely 6-month outcome of this patient? To be based upon information on discharge ER or admission to hospital/ICU". This reflects the Risk of unfavorable outcome (D, VS, SD) in % | Not Applicable |
| BaselinePhysEstOf6MoOutcomePhysicianQual | TRUE | integer | At ER discharge, physician estimate of six month outcome was recorded as a baseline risk assessment: "Given all current available information, what is, in your subjective opinion, the most likely 6-month outcome of this patient? To be based upon information on discharge ER or admission to hospital/ICU". This reflects the qualification of the physician who provided prognostic estimate on ER discharge/admission to hospital/ICU | 1=Resident;2=Junior staff (< 5 years);3=Senior staff (>= 5 years);4=Head of department |
| BaselinePhysEstOf6MoOutcomePhysicianType | TRUE | integer | At ER discharge, physician estimate of six month outcome was recorded as a baseline risk assessment: "Given all current available information, what is, in your subjective opinion, the most likely 6-month outcome of this patient? To be based upon information on discharge ER or admission to | 1=ER Physician;2=Intensive Care;3=Neurology;4=Neurosurgery;5=Traumatology;88=Unknown |
| | | | hospital/ICU". This reflects the type of the physician who provided prognostic estimate on ER discharge/admission to hospital/ICU | |
| BestOfAbdomenPelvicLumbarISS | TRUE | integer | AbdomenPelvicLumbar region (Highest AIS of the region)^2 compare AbdomenPelvicContentsAIS, LumbarSpineAIS. This score is taken forward for ISS calculation | Not Applicable |
| BestOfChestSpineISS | TRUE | integer | (highest AIS of the region)^2 Compare ThoraxChestAIS, ThoracicSpineAIS and select the highest for ISS calculation | Not Applicable |
| BestOfExternalISS | TRUE | integer | External region (ExternaAIS)^2 select the highest external AIS severity code for ISS calculation. | Not Applicable |
| BestOfExtremitiesISS | TRUE | integer | Extremities region (Highest AIS of the region)^2 compare UpperExtremitiesAIS, LowerExtremitiesAIS, PelvicGirdleAIS select the highest for ISS calculation | Not Applicable |
| BestOfFaceISS | TRUE | integer | Face region (FaceAIS)^2 select the highest facial injury for ISS calculation | Not Applicable |
| BestOfHeadBrainCervicalISS | TRUE | integer | HeadBrainCervical region (Highest AIS of the region)^2 Compare HeadNeckAIS, InjuryHx.BrainInjuryAIS, CervicalSpineAIS select the highest scoring injury in any of these 3 areas for ISS calculation | Not Applicable |
| BrainInjuryAIS | TRUE | integer | AIS score for the Brain Injury In the original AIS classification of injury severity, the grading is from 1 (minor) to 6 (unsurvivable). We added a score of 0 to designate absence of injuries. | 0=None;1=Minor: no treatment needed;2=Moderate: requires only outpatient treatment;3=Serious: requires non-ICU hospital admission;4=Severe: requires ICU observation and/or basic treatment;5=Critical: requires intubation, mechanical ventilation or vasopressors for blood pressure support;6=Unsurvivable: not survivable |
| CervicalSpineAIS | TRUE | integer | AIS score for the Cervical Spine region. In the original AIS classification of injury severity, the grading is from 1 (minor) to 6 (unsurvivable). We added a score of 0 to designate absence of injuries. | 0=None;1=Minor: no treatment needed;2=Moderate: requires only outpatient treatment;3=Serious: requires non-ICU hospital admission;4=Severe: requires ICU observation and/or basic treatment;5=Critical: requires intubation, mechanical ventilation or vasopressors for blood pressure support;6=Unsurvivable: not survivable |
| DispER | TRUE | integer | Destination of the patient at ER discharge. | 1=Discharge home;2=Discharge other facility;3=Hospital admission--Ward;4=Hospital admission--Intermediate/high care unit;5=Hospital admission--ICU;6=Hospital admission--OR for immediate surgical procedure;7=Death;8=Hospital admission--Other (e.g. observation unit);88=Unknown |
| EmerSurgIntraCranSurviveNoSurg | TRUE | integer | "InjuryHx.EmerSurgIntraCranSurviveNoSurg" and "InjuryHx.EmerSurgIntraCranSurviveYesSurg" These 2 variables aim to capture information on the surgeon's expectations, eg if the surgeon considers a realistic expectation of benefit, or performs the surgery as a "last resort" in a likely hopeless case. 'The short term survival chances of the patients if I DO NOT operate will be (in %)' | Not Applicable |
| EmerSurgIntraCranSurviveYesSurg | TRUE | integer | "InjuryHx.EmerSurgIntraCranSurviveNoSurg" and "InjuryHx.EmerSurgIntraCranSurviveYesSurg" These 2 variables aim to capture information on the surgeon's expectations, eg if the surgeon considers a realistic expectation of benefit, or performs the surgery as a "last resort" in a likely hopeless case. 'The short term survival chances of the patients if I DO operate will be (in %)' | Not Applicable |
| ExternaAIS | TRUE | integer | AIS score for the External skin In the original AIS classification of injury severity, the grading is from 1 (minor) to 6 (unsurvivable). We added a score of 0 to designate absence of injuries. | 0=None;1=Minor: no treatment needed;2=Moderate: requires only outpatient treatment;3=Serious: requires non-ICU hospital admission;4=Severe: requires ICU observation and/or basic treatment;5=Critical: requires intubation, mechanical ventilation or vasopressors for blood pressure support;6=Unsurvivable: not survivable |
| FaceAIS | TRUE | integer | AIS score for Face (incl.maxillofacial) In the original AIS classification of injury severity, the grading is from 1 (minor) to 6 (unsurvivable). We added a score of 0 to designate absence of injuries. | 0=None;1=Minor: no treatment needed;2=Moderate: requires only outpatient treatment;3=Serious: requires non-ICU hospital admission;4=Severe: requires ICU observation and/or basic treatment;5=Critical: requires intubation, mechanical ventilation or vasopressors for blood pressure support;6=Unsurvivable: not survivable |
| HeadNeckAIS | TRUE | integer | AIS score for the Head Neck region In the original AIS classification of injury severity, the grading is from 1 (minor) to 6 (unsurvivable). We added a score of 0 to designate absence of injuries. | 0=None;1=Minor: no treatment needed;2=Moderate: requires only outpatient treatment;3=Serious: requires non-ICU hospital admission;4=Severe: requires ICU observation and/or basic treatment;5=Critical: requires intubation, mechanical ventilation or vasopressors for blood pressure support;6=Unsurvivable: not survivable |
| LowerExtremitiesAIS | TRUE | integer | AIS score for the Lower extremities. In the original AIS classification of injury severity, the grading is from 1 (minor) to 6 (unsurvivable). We added a score of 0 to designate absence of injuries. | 0=None;1=Minor: no treatment needed;2=Moderate: requires only outpatient treatment;3=Serious: requires non-ICU hospital admission;4=Severe: requires ICU observation and/or basic treatment;5=Critical: requires intubation, mechanical ventilation or vasopressors for blood pressure support;6=Unsurvivable: not survivable |
| LumbarSpineAIS | TRUE | integer | AIS score for the Lumbar Spine region. In the original AIS classification of injury severity, the grading is from 1 (minor) | 0=None;1=Minor: no treatment needed;2=Moderate: requires only outpatient treatment;3=Serious: requires non-ICU |

| | | | |
|---|---|---|---|
| | | to 6 (unsurvivable). We added a score of 0 to designate absence of injuries. | hospital admission;4=Severe: requires ICU observation and/or basic treatment;5=Critical: requires intubation, mechanical ventilation or vasopressors for blood pressure support;6=Unsurvivable: not survivable |
| PelvicGirdleAIS | TRUE | integer | AIS score for the Pelvic Girdle region. In the original AIS classification of injury severity, the grading is from 1 (minor) to 6 (unsurvivable). We added a score of 0 to designate absence of injuries. | 0=None;1=Minor: no treatment needed;2=Moderate: requires only outpatient treatment;3=Serious: requires non-ICU hospital admission;4=Severe: requires ICU observation and/or basic treatment;5=Critical: requires intubation, mechanical ventilation or vasopressors for blood pressure support;6=Unsurvivable: not survivable |
| SurgIntervenAppro | TRUE | integer | WHY question: How strongly does the surgeon feels that this surgical intervention is appropriate in terms of the expected benefit to final clinical outcome? | 0=0;1=1;2=2;3=3;4=4;5=5;6=6;7=7;8=8;9=9;10=10 |
| ThoracicSpineAIS | TRUE | integer | AIS score for the Thoracic spine Region In the original AIS classification of injury severity, the grading is from 1 (minor) to 6 (unsurvivable). We added a score of 0 to designate absence of injuries. | 0=None;1=Minor: no treatment needed;2=Moderate: requires only outpatient treatment;3=Serious: requires non-ICU hospital admission;4=Severe: requires ICU observation and/or basic treatment;5=Critical: requires intubation, mechanical ventilation or vasopressors for blood pressure support;6=Unsurvivable: not survivable |
| ThoraxChestAIS | TRUE | integer | AIS score for the Thorax Chest region. In the original AIS classification of injury severity, the grading is from 1 (minor) to 6 (unsurvivable). We added a score of 0 to designate absence of injuries. | 0=None;1=Minor: no treatment needed;2=Moderate: requires only outpatient treatment;3=Serious: requires non-ICU hospital admission;4=Severe: requires ICU observation and/or basic treatment;5=Critical: requires intubation, mechanical ventilation or vasopressors for blood pressure support;6=Unsurvivable: not survivable |
| TotalISS | TRUE | integer | The Injury Severity Score is calculated as the sum of the squares of the the 3 body regions with the highest AIS score. The max score for the ISS = 75. If any body region AIS is assigned a score of "6", the ISS is automatically set to 75 (highest score). In the calculation of the ISS, only the 6 main body regions are taken into consideration. | Not Applicable |
| UpperExtremitiesAIS | TRUE | integer | AIS score for the Upper extremities. In the original AIS classification of injury severity, the grading is from 1 (minor) to 6 (unsurvivable). We added a score of 0 to designate absence of injuries. | 0=None;1=Minor: no treatment needed;2=Moderate: requires only outpatient treatment;3=Serious: requires non-ICU hospital admission;4=Severe: requires ICU observation and/or basic treatment;5=Critical: requires intubation, mechanical ventilation or vasopressors for blood pressure support;6=Unsurvivable: not survivable |
| CRFCTReason | FALSE | text | This variable contains the main reason why a CT-scan, during hospital stay, was performed. One of following options: standard follow-up, post-operative control, clinical deterioration, (suspicion of) increasing ICP, lack of improvement, unknown, other (specified in CTMRI.CTReasonOther) The reason for making an early CT-scan/ER scan can be found in: CTMRI.CTERReason | CD=Clinical deterioration;ICUADM88=Unknown;ICUADM99=Other;IICP=(Suspicion of) Increasing ICP;LOP=Lack of improvement;POC=Post-operative control;SFU=Standard follow-up |
| CTNoOpMotiv | FALSE | integer | WHY question: documents reason for not having an indication for (intra)cranial surgery. | 0=No surgical lesion;1=Lesion present, but Acceptable/good neurologic condition;2=Lesion present, but Guideline adherence;3=Lesion present, but Little/no mass effect;4=Lesion present, but Not hospital policy;5=Lesion present, but Extremely poor prognosis;6=Lesion present, but Brain Death;7=Lesion present, but Old age;8=Lesion present, but Wish family;88=Unknown;99=Lesion present, but Other |
| CTNoOpMotivOther | FALSE | text | Specification, only applicable if "CTMRI.CTNoOpMotiv" was "other" | Not Applicable |
| CTYesOpMotiv | FALSE | integer | WHY question: documents reason for having an indication for (intra)cranial surgery. | 1=Emergency/life saving;2=Clinical deterioration;3=Mass effect on CT;4=Radiological progression;5=(Suspicion of) raised ICP;6=Guideline adherence;7=To prevent deterioration;8=Depressed skull fracture;99=Other |
| CTYesOpMotivOther | FALSE | text | Free text if "CTMRI.CTYesOpMotiv" was marked as 'Other'. Relates to the WHY question: documents reason for having an indication for (intra)cranial surgery. | Not Applicable |
| ERCTNoOpMotiv | TRUE | integer | In emergency room: WHY question: documents reason for not having an indication for (intra)cranial surgery. | 0=No surgical lesion;1=Lesion present, but Acceptable/good neurologic condition;2=Lesion present, but Guideline adherence;3=Lesion present, but Little/no mass effect;4=Lesion present, but Not hospital policy;5=Lesion present, but Extremely poor prognosis;6=Lesion present, but Brain Death;7=Lesion present, but Old age;8=Lesion present, but Wish family;88=Unknown;99=Lesion present, but Other |
| ERCTNoOpMotivOther | TRUE | text | In emergency room: Specification, only applicable if "CTMRI.CTNoOpMotiv" was "other" | Not Applicable |
| ERCTYesOpMotiv | TRUE | integer | In emergency room: WHY question: documents reason for having an indication for (intra)cranial surgery. | 1=Emergency/life saving;2=Clinical deterioration;3=Mass effect on CT;4=Radiological progression;5=(Suspicion of) raised ICP;6=Guideline adherence;7=To prevent deterioration;8=Depressed skull fracture;99=Other |
| ERCTYesOpMotivOther | TRUE | text | In emergency room: Free text if "CTMRI.CTYesOpMotiv" was marked as 'Other'. Relates to the WHY question: | Not Applicable |
| | | | documents reason for having an indication for (intra)cranial surgery. | |
| ShortTermSurvivalNoSurg | FALSE | integer | Aims to capture information on the surgeon's expectations, eg if the surgeon considers a realistic expectation of benefit, or performs the surgery as a "last resort" in a likely hopeless case --> The short term survival chances of the patient if I DO NOT operate (1-100) | Not Applicable |
| ShortTermSurvivalYesSurg | FALSE | integer | Aims to capture information on the surgeon's expectations, eg if the surgeon considers a realistic expectation of benefit, or performs the surgery as a "last resort" in a likely hopeless case --> The short term survival chances of the patient if I DO operate (1-100) | Not Applicable |
| SurgeryCranialReason | FALSE | integer | WHY Question: aims to document the reason for intracranial surgery | 1=Emergency/Life saving;2=Clinical deterioration;3=Mass effect on CT;4=Radiological progression;5=suspicion of) raised ICP;6=Guideline adherence;7=To prevent deterioration |
| SurgeryExtraCranialReason | FALSE | integer | The extra-cranial surgeries information was to be entered in the e-CRF in tables for which you could add as many rows as you wish. The tables consisted of - Surgery start date - Surgery start time - Surgery end date - Surgery end time - Extracranial surgery code - Reason - Delay - Short time survival if you do not operate - Short time survival if you do operate | 1=Emergency/Lifesaving;2=Elective;3=Treatment of complication;4=Airway management;99=Other |
| TransReason | FALSE | integer | WHY question: documents reason for transition of care | 1=Mechanical ventilation;2=Frequent neurological observations;3=Haemodynamic invasive monitoring;4=Extracranial injuries;5=Neurological operation;6=Clinical deterioration;7=CT abnormalities;8=Clinical observation for TBI;9=No ICU bed available;10=Could be discharged home, but no adequate supervision;11=Improvement;12=Neurological deterioration;13=Systemic compilation;14=CT progression;15=Planned surgery;16=Condition stable;17=(acute) Treatment goals accomplished;18=Need to free a bed;19=Further improvement;20=Clinical rehab completed;21=Lack of improvement;22=Late neurological deterioration;23=Problems unrelated to trauma;24=Post operative care;25=Neurological complication;99=Other |

## Supplementary Note 3. Manually excluded variables indicating death or withdrawal of life-sustaining treatment.

| Name | Format | Description | Possible Values |
|---|---|---|---|
| BrainDeathDate | date | Reflects the date of Brain death in case of Withdrawal of life-sustaining measures | Not Applicable |
| BrainDeathTime | text | Reflects the Time of Brain death in case of Withdrawal of life-sustaining measures Check also "Hospital.BrainDeathDate" for the date. | Not Applicable |
| CTPatientLocation | text | This variable describes the in-hospital location of the patient when the CT-scan was performed and was not meant to describe the location of the CT-scanner. Three options: ER, Ward/Admission, ICU | ADMIS=Ward/Admission;ED=ER;ICU=ICU |
| DeadAge | integer | Reflects if the reason for Withdrawal of life-sustaining measures was age. | 0=No;1=Yes;88=Unknown |
| DeadCoMorbidities | integer | Reflects if the reason for Withdrawal of life-sustaining measures was co-morbidities. | 0=No;1=Yes;88=Unknown |
| DeadDeterminationOfBrainDeath | integer | Reflects if the reason for Withdrawal of life-sustaining measures was Determination of brain death (according to national law). | 0=No;1=Yes;88=Unknown |
| DeadOrganDonation | integer | Reflects if Withdrawal of life-sustaining measures was followed by organ donation. | 0=No;1=Yes;88=Unknown |
| DeadPatWill | integer | Reflects if the reason for Withdrawal of life-sustaining measures was Following living will of patient. | 0=No;1=Yes;88=Unknown |
| DeadRequestRelatives | integer | Reflects if the reason for Withdrawal of life-sustaining measures was On request of relatives. | 0=No;1=Yes;88=Unknown |
| DeadSeverityofTBI | integer | Reflects if the reason for Withdrawal of life-sustaining measures was Severity of TBI. | 0=No;1=Yes;88=Unknown |
| DeathAutopsy | integer | Reflects if an autopsy was performed after the death of the patient. | 0=No;1=Yes, forensic;2=Yes, clinical;88=Unknown |
| DeathCause | integer | Cause of death in or outside the hospital | 1=Head injury/initial injury;2=Head injury/secondary intracranial damage;3=Systemic trauma;4=Medical complications;88=Unknown;99=Other |
| DeathCauseOther | text | "Other" cause of death in or outside the hospital (than predefined list). | Not Applicable |
| DeathDate | date | Date of death also recorded on hospital discharge and at followup: FollowUp.FUPrincipalDeathCause;Death may also have been recorded in the ER forms: Subject.DeathDate | Not Applicable |
| DeathERDeclaredBrainDeadFollowingNationalCriteria | integer | Reflects if patient was declared brain dead following national criteria. Only applicable if patient declared "dead" on the ER | 0=No;1=Yes;88=Unknown |
| DeathERDOA | integer | Reflects if patient is declared dead on the ER --> Dead on arrival (DOA). Only applicable if patient declared "dead" on the ER. | 0=No;1=Yes;88=Unknown |
| DeathERUnsuccResusForExtraCranInj | integer | Reflects unsuccessful resuscitation for extra cranial injuries if patient is declared dead on the ER. Only applicable if patient declared "dead" on the ER | 0=No;1=Yes;88=Unknown |
| DeathERWithdrawalLifeSuppForSeverityOfTBI | integer | Reflects Withdrawal of life¬≠sustaining measures for severity of TBI if patient is declared dead on the ER. Only applicable if patient declared "dead" on the ER | 0=No;1=Yes;88=Unknown |
| DeathTime | text | Time of death | Not Applicable |
| DischargeStatus | integer | Assessment by investigator Reflects if the patient was dead or alive at discharge. | 0=Dead;1=Alive;88=Unknown |
| EOSReason | integer | Reason for end of study participation | 1=Completion of study;2=Inability to obtain follow-up;3=Withdrawal from study (by patient or representative);4=Adverse event(s);5=Decision for DNR*:;6=Withdrawal of support;7=Death;99=Other |
| EOSReasonOtherTxt | text | "Other" reason for end of study participation than the predefined list. | Not Applicable |
| ICDCode1 | text | Up to 16 fields available to enter diagnosis as recorded by hospital administration according to ICD codes; applicable to patients admitted/discharged from hospital. For patients discharged directly from the ER, ICD codes are documented in: InjuryHx.ERDestICDCodes1 | Not Applicable |
| ICDCode2 | text | Up to 16 fields available to enter diagnosis as recorded by hospital administration according to ICD codes; applicable to patients admitted/discharged from hospital. For patients discharged directly from the ER, ICD codes are documented in: InjuryHx.ERDestICDCodes1 | Not Applicable |
| ICDCode3 | text | Up to 16 fields available to enter diagnosis as recorded by hospital administration according to ICD codes; applicable to patients admitted/discharged from hospital. For patients discharged directly from the ER, ICD codes are documented in: InjuryHx.ERDestICDCodes1 | Not Applicable |
| ICDCode4 | text | Up to 16 fields available to enter diagnosis as recorded by hospital administration according to ICD codes; applicable to patients admitted/discharged from hospital. | Not Applicable |

| Name | Format | Description | Possible Values |
|---|---|---|---|
| | | For patients discharged directly from the ER, ICD codes are documented in: InjuryHx.ERDestICDCodes1 | |
| ICDCode5 | text | Up to 16 fields available to enter diagnosis as recorded by hospital administration according to ICD codes; applicable to patients admitted/discharged from hospital.<br>For patients discharged directly from the ER, ICD codes are documented in: InjuryHx.ERDestICDCodes1 | Not Applicable |
| ICDCode6 | text | Up to 16 fields available to enter diagnosis as recorded by hospital administration according to ICD codes; applicable to patients admitted/discharged from hospital.<br>For patients discharged directly from the ER, ICD codes are documented in: InjuryHx.ERDestICDCodes1 | Not Applicable |
| ICDCodeVersion | integer | This variable reflects if the ICD code version 9 or version 10 was used.<br>Up to 16 fields are available to enter diagnosis as recorded by hospital administration according to ICD codes. | 9=ICD-9;10=ICD-10 |
| ICPStopReason | integer | Reason for stopping ICP. Also check Hospital.ICPMonitorStop. | 1=Clinically improved;2=ICP stable and < 20 mmHg;3=Monitor/catheter failure;4=Patient considered unsalvageable;5=Patient died;99=Other |
| ICPStopReason | integer | Reason for stopping ICP. Also check Hospital.ICPMonitorStop. | 1=Clinically improved;2=ICP stable and < 20 mmHg;3=Monitor/catheter failure;4=Patient considered unsalvageable;5=Patient died;99=Other |
| ICUDischargeICDCode1 | text | The intent here is to register ICD code as recorded in hospital administrative files for patients directly discharged from the ICU. Up to 16 codes can be entered, ICD codes are further captured at ER discharge and at hospital discharge. | Not Applicable |
| ICUDischargeICDCode2 | text | The intent here is to register ICD code as recorded in hospital administrative files for patients directly discharged from the ICU. Up to 16 codes can be entered, ICD codes are further captured at ER discharge and at hospital discharge. | Not Applicable |
| ICUDischargeICDCode3 | text | The intent here is to register ICD code as recorded in hospital administrative files for patients directly discharged from the ICU. Up to 16 codes can be entered, ICD codes are further captured at ER discharge and at hospital discharge. | Not Applicable |
| ICUDischargeICDCode4 | text | The intent here is to register ICD code as recorded in hospital administrative files for patients directly discharged from the ICU. Up to 16 codes can be entered, ICD codes are further captured at ER discharge and at hospital discharge. | Not Applicable |
| ICUDischargeICDCode5 | text | The intent here is to register ICD code as recorded in hospital administrative files for patients directly discharged from the ICU. Up to 16 codes can be entered, ICD codes are further captured at ER discharge and at hospital discharge. | Not Applicable |
| ICUDischargeICDCode6 | text | The intent here is to register ICD code as recorded in hospital administrative files for patients directly discharged from the ICU. Up to 16 codes can be entered, ICD codes are further captured at ER discharge and at hospital discharge. | Not Applicable |
| ICUDischargeICDCodeVersion | integer | This variable reflects if the ICD code version 9 or version 10 was used.<br>Up to 16 fields are available to enter diagnosis as recorded by hospital administration according to ICD codes. | 9=9;10=10 |
| ICUDischargeStatus | integer | Reflects if patient was alive or dead on discharge from ICU | 1=Alive;2=Dead;88=Unknown |
| ICUDischargeTo | integer | Reflects location to which the patient was discharged from ICU | 1=General ward;2=Other ICU;3=Other hospital;4=Rehab unit;5=Home;6=Nursing home;7=Step down/high care unit;88=Unknown;99=Other |
| ICUDischargeToOther | text | Specifies the "other" location to which the patient was discharged from ICU Check also "Hospital.ICUDischargeTo" | Not Applicable |
| ICUDisPatDeadAtICU | integer | Reflects if the patient was declared dead on the ICU.<br>Intended as an introductory question for the details on withdrawal of treatment, brain death and organ donation | 0=No;1=Yes |
| ICUDisSupportWithdrawnDate | date | This variable documents date and time at which life prolonging therapy was withdrawn<br>(together with "Hospital.ICUDisSupportWithdrawnTime") | Not Applicable |
| ICUDisSupportWithdrawnTime | text | This variable documents date and time at which life prolonging therapy was withdrawn<br>(together with "Hospital.ICUDisSupportWithdrawnDate") | Not Applicable |
| ICUDisWithdrawalTreatmentDecisionDate | date | Investigators were requested to record the details of Withdrawal of Treatment or Life support if applicable. | Not Applicable |
| ICUDisWithdrawalTreatmentDecisionTime | text | Investigators were requested to record the details of Withdrawal of Treatment or Life support if applicable. | Not Applicable |
| ICUDisWithdrawlTreatmentDecision | integer | Investigators were requested to record the details of Withdrawal of Treatment or Life support if applicable. | 1=Multi disciplinary;2=By a single physician;3=With relatives |
| IntubationStopReason | integer | This variable describes the Stop Reason of Extubation in case of Ventilation Management (only for ICU patients). | 1=Respiratory stable;2=Accidental;3=Withdrawal of care |

| Name | Format | Description | Possible Values |
|---|---|---|---|
| LengthOfStay | decimal | This variable reflects the length of stay of the patient at the study hospital. It has been derived using the information of the date and time of arrival at the study hospital and date and time of (study) hospital discharge. | Not Applicable |
| MonJugularSatStopReason | integer | Beside brain specific ICP monitoring in the ICU, details were recorded on other types of monitoring in the ICU. This variable reflects the reason for stopping if there was Jugular oximetry. | 1=Monitor/catheter failure;2=Patient considered unsalvageable;3=Patient died;4=Clinically no longer required |
| MonLicoxStopReason | integer | Beside brain specific ICP monitoring in the ICU, details were recorded on other types of monitoring in the ICU. This variable reflects the reason for stopping if there was Brain tissue PO2 monitoring. | 1=Monitor/catheter failure;2=Patient considered unsalvageable;3=Patient died;4=Clinically no longer required |
| MonLicoxStopReason | integer | Beside brain specific ICP monitoring in the ICU, details were recorded on other types of monitoring in the ICU. This variable reflects the reason for stopping if there was Brain tissue PO2 monitoring. | 1=Monitor/catheter failure;2=Patient considered unsalvageable;3=Patient died;4=Clinically no longer required |
| MonMicrodialysisStopReason | integer | Beside brain specific ICP monitoring in the ICU, details were recorded on other types of monitoring in the ICU. This variable reflects the reason for stopping if there was Microdialysis. | 1=Monitor/catheter failure;2=Patient considered unsalvageable;3=Patient died;4=Clinically no longer required |
| OrganDonationDate | date | Reflects the date of organ donation in case of Withdrawal of life-sustaining measures, if applicable. | Not Applicable |
| OrganDonationTime | text | Reflects the time of organ donation in case of Withdrawal of life-sustaining measures, if applicable. | Not Applicable |
| SupportWithdrawnDate | date | Investigators were requested to record the details of Withdrawal of Treatment or Life support if applicable. | Not Applicable |
| SupportWithdrawnTime | text | Investigators were requested to record the details of Withdrawal of Treatment or Life support if applicable. | Not Applicable |
| TimeSinceICUAdmisDeath | text | Investigators were requested to record the details of Withdrawal of Treatment or Life support if applicable. This reflects the time between admission in the ICU and death | Not Applicable |
| WithdrawalOption | integer | In case of complete withdrawal, all data have been deleted from the database | 1=Complete Withdrawal (no further contact, destruction of all data and samples collected up to that point);2=No further study related activities, but consent to access of clinical notes and use of existing data |
| WithdrawalTreatmentDecision | integer | Intended only to be scored if a medical decision was made to withdraw active treatment because of anticipated poor prognosis. However, some investigators may have scored this when patients had recovered to an extent that active treatment was no longer necessary. | 1=Multi disciplinary;2=By a single physician;3=With relatives |
| WithdrawalTreatmentDecisionDate | date | Investigators were requested to record the details of Withdrawal of Treatment or Life support if applicable. Intended only to be scored if a medical decision was made to withdraw active treatment because of anticipated poor prognosis. However, some investigators may have scored this when patients had recovered to an extent that active treatment was no longer necessary. | Not Applicable |
| WithdrawalTreatmentDecisionTime | text | Investigators were requested to record the details of Withdrawal of Treatment or Life support if applicable. Intended only to be scored if a medical decision was made to withdraw active treatment because of anticipated poor prognosis. However, some investigators may have scored this when patients had recovered to an extent that active treatment was no longer necessary. | Not Applicable |
| WithdrawSuppDateTime | datetime | Withdrawal of Support Date & Time if Withdrawal of life-sustaining support was the reason for end of study participation. | Not Applicable |

**Supplementary Note 4. The CENTER-TBI investigators and participants.**

The co-lead investigators of CENTER-TBI are designated with an asterisk (*), and their contact email addresses are listed below.

Cecilia Åkerlund[1], Krisztina Amrein[2], Nada Andelic[3], Lasse Andreassen[4], Audny Anke[5], Anna Antoni[6], Gérard Audibert[7], Philippe Azouvi[8], Maria Luisa Azzolini[9], Ronald Bartels[10], Pál Barzó[11], Romuald Beauvais[12], Ronny Beer[13], Bo-Michael Bellander[14], Antonio Belli[15], Habib Benali[16], Maurizio Berardino[17], Luigi Beretta[9], Morten Blaabjerg[18], Peter Bragge[19], Alexandra Brazinova[20], Vibeke Brinck[21], Joanne Brooker[22], Camilla Brorsson[23], Andras Buki[24], Monika Bullinger[25], Manuel Cabeleira[26], Alessio Caccioppola[27], Emiliana Calappi[27], Maria Rosa Calvi[9], Peter Cameron[28], Guillermo Carbayo Lozano[29], Marco Carbonara[27], Simona Cavallo[17], Giorgio Chevallard[30], Arturo Chieregato[30], Giuseppe Citerio[31,32], Hans Clusmann[33], Mark Coburn[34], Jonathan Coles[35], Jamie D. Cooper[36], Marta Correia[37], Amra Čović [38], Nicola Curry[39], Endre Czeiter[24], Marek Czosnyka[26], Claire Dahyot-Fizelier[40], Paul Dark[41], Helen Dawes[42], Véronique De Keyser[43], Vincent Degos[16], Francesco Della Corte[44], Hugo den Boogert[10], Bart Depreitere[45], Đula Đilvesi[46], Abhishek Dixit[47], Emma Donoghue[22], Jens Dreier[48], Guy-Loup Dulière[49], Ari Ercole[47], Patrick Esser[42], Erzsébet Ezer[50], Martin Fabricius[51], Valery L. Feigin[52], Kelly Foks[53], Shirin Frisvold[54], Alex Furmanov[55], Pablo Gagliardo[56], Damien Galanaud[16], Dashiell Gantner[28], Guoyi Gao[57], Pradeep George[58], Alexandre Ghuysen[59], Lelde Giga[60], Ben Glocker[61], Jagoš Golubovic[46], Pedro A. Gomez[62], Johannes Gratz[63], Benjamin Gravesteijn[64], Francesca Grossi[44], Russell L. Gruen[65], Deepak Gupta[66], Juanita A. Haagsma[64], Iain Haitsma[67], Raimund Helbok[13], Eirik Helseth[68], Lindsay Horton[69], Jilske Huijben[64], Peter J. Hutchinson[70], Bram Jacobs[71], Stefan Jankowski[72], Mike Jarrett[21], Ji-yao Jiang[58], Faye Johnson[73], Kelly Jones[52], Mladen Karan[46], Angelos G. Kolias[70], Erwin Kompanje[74], Daniel Kondziella[51], Evgenios Kornaropoulos[47], Lars-Owe Koskinen[75], Noémi Kovács[76], Ana Kowark[77], Alfonso Lagares[62], Linda Lanyon[58], Steven Laureys[78], Fiona Lecky[79,80], Didier Ledoux[78], Rolf Lefering[81], Valerie Legrand[82], Aurelie Lejeune[83], Leon Levi[84], Roger Lightfoot[85], Hester Lingsma[64], Andrew I.R. Maas[43,*], Ana M. Castaño-León[62], Marc Maegele[86], Marek Majdan[20], Alex Manara[87], Geoffrey Manley[88], Costanza Martino[89], Hugues Maréchal[49], Julia Mattern[90], Catherine McMahon[91], Béla Melegh[92], David Menon[47,*], Tomas Menovsky[43], Ana Mikolic[64], Benoit Misset[78], Visakh Muraleedharan[58], Lynnette Murray[28], Ancuta Negru[93], David Nelson[1], Virginia Newcombe[47], Daan Nieboer[64], József Nyirádi[2], Otesile Olubukola[79], Matej Oresic[94], Fabrizio Ortolano[27], Aarno Palotie[95,96,97], Paul M. Parizel[98], Jean-François Payen[99], Natascha Perera[12], Vincent Perlbarg[16], Paolo Persona[100], Wilco Peul[101], Anna Piippo-Karjalainen[102], Matti Pirinen[95], Dana Pisica[64], Horia Ples[93], Suzanne Polinder[64], Inigo Pomposo[29], Jussi P. Posti[103], Louis Puybasset[104], Andreea Radoi[105], Arminas Ragauskas[106], Rahul Raj[102], Malinka Rambadagalla[107], Isabel Retel Helmrich[64], Jonathan Rhodes[108], Sylvia Richardson[109], Sophie Richter[47], Samuli Ripatti[95], Saulius Rocka[106], Cecilie Roe[110], Olav Roise[111,112], Jonathan Rosand[113], Jeffrey V. Rosenfeld[114], Christina Rosenlund[115], Guy Rosenthal[55], Rolf Rossaint[77], Sandra Rossi[100], Daniel Rueckert[61] Martin Rusnák[116], Juan Sahuquillo[105], Oliver Sakowitz[90,117], Renan Sanchez-Porras[117], Janos Sandor[118], Nadine Schäfer[81], Silke Schmidt[119], Herbert Schoechl[120], Guus Schoonman[121], Rico Frederik Schou[122], Elisabeth Schwendenwein[6], Charlie Sewalt[64], Ranjit D. Singh[101], Toril Skandsen[123,124,] Peter Smielewski[26], Abayomi Sorinola[125], Emmanuel Stamatakis[47], Simon Stanworth[39], Robert Stevens[126], William Stewart[127], Ewout W. Steyerberg[64,128], Nino Stocchetti[129], Nina Sundström[130], Riikka Takala[131], Viktória Tamás[125], Tomas Tamosuitis[132], Mark Steven Taylor[20], Braden Te Ao[52], Olli Tenovuo[103], Alice Theadom[52], Matt Thomas[87], Dick Tibboel[133], Marjolein Timmers[74], Christos Tolias[134], Tony Trapani[28], Cristina Maria Tudora[93], Andreas Unterberg[90], Peter Vajkoczy [135], Shirley Vallance[28], Egils Valeinis[60], Zoltán Vámos[50], Mathieu van der Jagt[136], Gregory Van der Steen[43], Joukje van der Naalt[71], Jeroen T.J.M. van Dijck[101], Inge A. M. van Erp[101], Thomas A. van Essen[101], Wim Van Hecke[137], Caroline van Heugten[138], Dominique Van Praag[139], Ernest van Veen[64], Thijs Vande Vyvere[137], Roel P. J. van Wijk[101], Alessia Vargiolu[32], Emmanuel Vega[83], Kimberley Velt[64], Jan Verheyden[137], Paul M. Vespa[140], Anne Vik[123,141], Rimantas Vilcinis[132], Victor Volovici[67], Nicole von Steinbüchel[38], Daphne Voormolen[64], Petar Vulekovic[46], Kevin K.W. Wang[142], Daniel Whitehouse[47], Eveline Wiegers[64], Guy Williams[47], Lindsay Wilson[69], Stefan Winzeck[47], Stefan Wolf[143], Zhihui Yang[113], Peter Ylén[144], Alexander Younsi[90], Frederick A. Zeiler[47,145], Veronika Zelinkova[20], Agate Ziverte[60,] Tommaso Zoerle[27]

[1]Department of Physiology and Pharmacology, Section of Perioperative Medicine and Intensive Care, Karolinska Institutet, Stockholm, Sweden
[2]János Szentágothai Research Centre, University of Pécs, Pécs, Hungary
[3]Division of Surgery and Clinical Neuroscience, Department of Physical Medicine and Rehabilitation, Oslo University Hospital and University of Oslo, Oslo, Norway
[4]Department of Neurosurgery, University Hospital Northern Norway, Tromso, Norway
[5]Department of Physical Medicine and Rehabilitation, University Hospital Northern Norway, Tromso, Norway
[6]Trauma Surgery, Medical University Vienna, Vienna, Austria
[7]Department of Anesthesiology & Intensive Care, University Hospital Nancy, Nancy, France
[8]Raymond Poincare hospital, Assistance Publique – Hopitaux de Paris, Paris, France
[9]Department of Anesthesiology & Intensive Care, S Raffaele University Hospital, Milan, Italy
[10]Department of Neurosurgery, Radboud University Medical Center, Nijmegen, The Netherlands
[11]Department of Neurosurgery, University of Szeged, Szeged, Hungary
[12]International Projects Management, ARTTIC, Munchen, Germany
[13]Department of Neurology, Neurological Intensive Care Unit, Medical University of Innsbruck, Innsbruck, Austria
[14]Department of Neurosurgery & Anesthesia & intensive care medicine, Karolinska University Hospital, Stockholm, Sweden
[15]NIHR Surgical Reconstruction and Microbiology Research Centre, Birmingham, UK
[16]Anesthesie-Réanimation, Assistance Publique – Hopitaux de Paris, Paris, France
[17]Department of Anesthesia & ICU, AOU Città della Salute e della Scienza di Torino - Orthopedic and Trauma Center, Torino, Italy

[18]Department of Neurology, Odense University Hospital, Odense, Denmark
[19]BehaviourWorks Australia, Monash Sustainability Institute, Monash University, Victoria, Australia
[20]Department of Public Health, Faculty of Health Sciences and Social Work, Trnava University, Trnava, Slovakia
[21]Quesgen Systems Inc., Burlingame, California, USA
[22]Australian & New Zealand Intensive Care Research Centre, Department of Epidemiology and Preventive Medicine, School of Public Health and Preventive Medicine, Monash University, Melbourne, Australia
[23]Department of Surgery and Perioperative Science, Umeå University, Umeå, Sweden
[24]Department of Neurosurgery, Medical School, University of Pécs, Hungary and Neurotrauma Research Group, János Szentágothai Research Centre, University of Pécs, Hungary
[25]Department of Medical Psychology, Universitätsklinikum Hamburg-Eppendorf, Hamburg, Germany
[26]Brain Physics Lab, Division of Neurosurgery, Dept of Clinical Neurosciences, University of Cambridge, Addenbrooke's Hospital, Cambridge, UK
[27]Neuro ICU, Fondazione IRCCS Cà Granda Ospedale Maggiore Policlinico, Milan, Italy
[28]ANZIC Research Centre, Monash University, Department of Epidemiology and Preventive Medicine, Melbourne, Victoria, Australia
[29]Department of Neurosurgery, Hospital of Cruces, Bilbao, Spain
[30]NeuroIntensive Care, Niguarda Hospital, Milan, Italy
[31]School of Medicine and Surgery, Università Milano Bicocca, Milano, Italy
[32]NeuroIntensive Care, ASST di Monza, Monza, Italy
[33]Department of Neurosurgery, Medical Faculty RWTH Aachen University, Aachen, Germany
[34]Department of Anesthesiology and Intensive Care Medicine, University Hospital Bonn, Bonn, Germany
[35]Department of Anesthesia & Neurointensive Care, Cambridge University Hospital NHS Foundation Trust, Cambridge, UK
[36]School of Public Health & PM, Monash University and The Alfred Hospital, Melbourne, Victoria, Australia
[37]Radiology/MRI department, MRC Cognition and Brain Sciences Unit, Cambridge, UK
[38]Institute of Medical Psychology and Medical Sociology, Universitätsmedizin Göttingen, Göttingen, Germany
[39]Oxford University Hospitals NHS Trust, Oxford, UK
[40]Intensive Care Unit, CHU Poitiers, Potiers, France
[41]University of Manchester NIHR Biomedical Research Centre, Critical Care Directorate, Salford Royal Hospital NHS Foundation Trust, Salford, UK
[42]Movement Science Group, Faculty of Health and Life Sciences, Oxford Brookes University, Oxford, UK
[43]Department of Neurosurgery, Antwerp University Hospital and University of Antwerp, Edegem, Belgium
[44]Department of Anesthesia & Intensive Care, Maggiore Della Carità Hospital, Novara, Italy
[45]Department of Neurosurgery, University Hospitals Leuven, Leuven, Belgium
[46]Department of Neurosurgery, Clinical centre of Vojvodina, Faculty of Medicine, University of Novi Sad, Novi Sad, Serbia
[47]Division of Anaesthesia, University of Cambridge, Addenbrooke's Hospital, Cambridge, UK
[48]Center for Stroke Research Berlin, Charité – Universitätsmedizin Berlin, corporate member of Freie Universität Berlin, Humboldt-Universität zu Berlin, and Berlin Institute of Health, Berlin, Germany
[49]Intensive Care Unit, CHR Citadelle, Liège, Belgium
[50]Department of Anaesthesiology and Intensive Therapy, University of Pécs, Pécs, Hungary
[51]Departments of Neurology, Clinical Neurophysiology and Neuroanesthesiology, Region Hovedstaden Rigshospitalet, Copenhagen, Denmark
[52]National Institute for Stroke and Applied Neurosciences, Faculty of Health and Environmental Studies, Auckland University of Technology, Auckland, New Zealand
[53]Department of Neurology, Erasmus MC, Rotterdam, the Netherlands
[54]Department of Anesthesiology and Intensive care, University Hospital Northern Norway, Tromso, Norway
[55]Department of Neurosurgery, Hadassah-hebrew University Medical center, Jerusalem, Israel
[56]Fundación Instituto Valenciano de Neurorrehabilitación (FIVAN), Valencia, Spain
[57]Department of Neurosurgery, Shanghai Renji hospital, Shanghai Jiaotong University/school of medicine, Shanghai, China
[58]Karolinska Institutet, INCF International Neuroinformatics Coordinating Facility, Stockholm, Sweden
[59]Emergency Department, CHU, Liège, Belgium
[60]Neurosurgery clinic, Pauls Stradins Clinical University Hospital, Riga, Latvia
[61]Department of Computing, Imperial College London, London, UK
[62]Department of Neurosurgery, Hospital Universitario 12 de Octubre, Madrid, Spain
[63]Department of Anesthesia, Critical Care and Pain Medicine, Medical University of Vienna, Austria
[64]Department of Public Health, Erasmus Medical Center-University Medical Center, Rotterdam, The Netherlands
[65]College of Health and Medicine, Australian National University, Canberra, Australia
[66]Department of Neurosurgery, Neurosciences Centre & JPN Apex trauma centre, All India Institute of Medical Sciences, New Delhi-110029, India
[67]Department of Neurosurgery, Erasmus MC, Rotterdam, the Netherlands
[68]Department of Neurosurgery, Oslo University Hospital, Oslo, Norway
[69]Division of Psychology, University of Stirling, Stirling, UK

[70]Division of Neurosurgery, Department of Clinical Neurosciences, Addenbrooke's Hospital & University of Cambridge, Cambridge, UK
[71]Department of Neurology, University of Groningen, University Medical Center Groningen, Groningen, Netherlands
[72]Neurointensive Care, Sheffield Teaching Hospitals NHS Foundation Trust, Sheffield, UK
[73]Salford Royal Hospital NHS Foundation Trust Acute Research Delivery Team, Salford, UK
[74]Department of Intensive Care and Department of Ethics and Philosophy of Medicine, Erasmus Medical Center, Rotterdam, The Netherlands
[75]Department of Clinical Neuroscience, Neurosurgery, Umeå University, Umeå, Sweden
[76]Hungarian Brain Research Program - Grant No. KTIA_13_Not ApplicableP-A-II/8, University of Pécs, Pécs, Hungary
[77]Department of Anaesthesiology, University Hospital of Aachen, Aachen, Germany
[78]Cyclotron Research Center, University of Liège, Liège, Belgium
[79]Centre for Urgent and Emergency Care Research (CURE), Health Services Research Section, School of Health and Related Research (ScHARR), University of Sheffield, Sheffield, UK
[80]Emergency Department, Salford Royal Hospital, Salford UK
[81]Institute of Research in Operative Medicine (IFOM), Witten/Herdecke University, Cologne, Germany
[82]VP Global Project Management CNS, ICON, Paris, France
[83]Department of Anesthesiology-Intensive Care, Lille University Hospital, Lille, France
[84]Department of Neurosurgery, Rambam Medical Center, Haifa, Israel
[85]Department of Anesthesiology & Intensive Care, University Hospitals Southhampton NHS Trust, Southhampton, UK
[86]Cologne-Merheim Medical Center (CMMC), Department of Traumatology, Orthopedic Surgery and Sportmedicine, Witten/Herdecke University, Cologne, Germany
[87]Intensive Care Unit, Southmead Hospital, Bristol, Bristol, UK
[88]Department of Neurological Surgery, University of California, San Francisco, California, USA
[89]Department of Anesthesia & Intensive Care,M. Bufalini Hospital, Cesena, Italy
[90]Department of Neurosurgery, University Hospital Heidelberg, Heidelberg, Germany
[91]Department of Neurosurgery, The Walton centre NHS Foundation Trust, Liverpool, UK
[92]Department of Medical Genetics, University of Pécs, Pécs, Hungary
[93]Department of Neurosurgery, Emergency County Hospital Timisoara, Timisoara, Romania
[94]School of Medical Sciences, Örebro University, Örebro, Sweden
[95]Institute for Molecular Medicine Finland, University of Helsinki, Helsinki, Finland
[96]Analytic and Translational Genetics Unit, Department of Medicine; Psychiatric & Neurodevelopmental Genetics Unit, Department of Psychiatry; Department of Neurology, Massachusetts General Hospital, Boston, MA, USA
[97]Program in Medical and Population Genetics; The Stanley Center for Psychiatric Research, The Broad Institute of MIT and Harvard, Cambridge, MA, USA
[98]Department of Radiology, University of Antwerp, Edegem, Belgium
[99]Department of Anesthesiology & Intensive Care, University Hospital of Grenoble, Grenoble, France
[100]Department of Anesthesia & Intensive Care, Azienda Ospedaliera Università di Padova, Padova, Italy
[101]Dept. of Neurosurgery, Leiden University Medical Center, Leiden, The Netherlands and Dept. of Neurosurgery, Medical Center Haaglanden, The Hague, The Netherlands
[102]Department of Neurosurgery, Helsinki University Central Hospital
[103]Division of Clinical Neurosciences, Department of Neurosurgery and Turku Brain Injury Centre, Turku University Hospital and University of Turku, Turku, Finland
[104]Department of Anesthesiology and Critical Care, Pitié -Salpêtrière Teaching Hospital, Assistance Publique, Hôpitaux de Paris and University Pierre et Marie Curie, Paris, France
[105]Neurotraumatology and Neurosurgery Research Unit (UNINN), Vall d'Hebron Research Institute, Barcelona, Spain
[106]Department of Neurosurgery, Kaunas University of technology and Vilnius University, Vilnius, Lithuania
[107]Department of Neurosurgery, Rezekne Hospital, Latvia
[108]Department of Anaesthesia, Critical Care & Pain Medicine NHS Lothian & University of Edinburg, Edinburgh, UK
[109]Director, MRC Biostatistics Unit, Cambridge Institute of Public Health, Cambridge, UK
[110]Department of Physical Medicine and Rehabilitation, Oslo University Hospital/University of Oslo, Oslo, Norway
[111]Division of Orthopedics, Oslo University Hospital, Oslo, Norway
[112]Institue of Clinical Medicine, Faculty of Medicine, University of Oslo, Oslo, Norway
[113]Broad Institute, Cambridge MA Harvard Medical School, Boston MA, Massachusetts General Hospital, Boston MA, USA
[114]National Trauma Research Institute, The Alfred Hospital, Monash University, Melbourne, Victoria, Australia
[115]Department of Neurosurgery, Odense University Hospital, Odense, Denmark
[116]International Neurotrauma Research Organisation, Vienna, Austria
[117]Klinik für Neurochirurgie, Klinikum Ludwigsburg, Ludwigsburg, Germany
[118]Division of Biostatistics and Epidemiology, Department of Preventive Medicine, University of Debrecen, Debrecen, Hungary
[119]Department Health and Prevention, University Greifswald, Greifswald, Germany
[120]Department of Anaesthesiology and Intensive Care, AUVA Trauma Hospital, Salzburg, Austria
[121]Department of Neurology, Elisabeth-TweeSteden Ziekenhuis, Tilburg, the Netherlands

[122]Department of Neuroanesthesia and Neurointensive Care, Odense University Hospital, Odense, Denmark
[123]Department of Neuromedicine and Movement Science, Norwegian University of Science and Technology, NTNU, Trondheim, Norway
[124]Department of Physical Medicine and Rehabilitation, St.Olavs Hospital, Trondheim University Hospital, Trondheim, Norway
[125]Department of Neurosurgery, University of Pécs, Pécs, Hungary
[126]Division of Neuroscience Critical Care, Johns Hopkins University School of Medicine, Baltimore, USA
[127]Department of Neuropathology, Queen Elizabeth University Hospital and University of Glasgow, Glasgow, UK
[128]Dept. of Department of Biomedical Data Sciences, Leiden University Medical Center, Leiden, The Netherlands
[129]Department of Pathophysiology and Transplantation, Milan University, and Neuroscience ICU, Fondazione IRCCS Cà Granda Ospedale Maggiore Policlinico, Milano, Italy
[130]Department of Radiation Sciences, Biomedical Engineering, Umeå University, Umeå, Sweden
[131]Perioperative Services, Intensive Care Medicine and Pain Management, Turku University Hospital and University of Turku, Turku, Finland
[132]Department of Neurosurgery, Kaunas University of Health Sciences, Kaunas, Lithuania
[133]Intensive Care and Department of Pediatric Surgery, Erasmus Medical Center, Sophia Children's Hospital, Rotterdam, The Netherlands
[134]Department of Neurosurgery, Kings college London, London, UK
[135]Neurologie, Neurochirurgie und Psychiatrie, Charité – Universitätsmedizin Berlin, Berlin, Germany
[136]Department of Intensive Care Adults, Erasmus MC– University Medical Center Rotterdam, Rotterdam, the Netherlands
[137]icoMetrix NV, Leuven, Belgium
[138]Movement Science Group, Faculty of Health and Life Sciences, Oxford Brookes University, Oxford, UK
[139]Psychology Department, Antwerp University Hospital, Edegem, Belgium
[140]Director of Neurocritical Care, University of California, Los Angeles, USA
[141]Department of Neurosurgery, St.Olavs Hospital, Trondheim University Hospital, Trondheim, Norway
[142]Department of Emergency Medicine, University of Florida, Gainesville, Florida, USA
[143]Department of Neurosurgery, Charité – Universitätsmedizin Berlin, corporate member of Freie Universität Berlin, Humboldt-Universität zu Berlin, and Berlin Institute of Health, Berlin, Germany
[144]VTT Technical Research Centre, Tampere, Finland
[145]Section of Neurosurgery, Department of Surgery, Rady Faculty of Health Sciences, University of Manitoba, Winnipeg, MB, Canada

*Co-lead investigators: andrew.maas@uza.be (AIRM) and dkm13@cam.ac.uk (DM)

**SUPPLEMENTARY METHODS**

**Hyperparameter optimisation report**

*Summary*

Combinations of the listed hyperparameters were tested on the validation sets of our repeated *k*-fold cross-validation (20 repeats, 5 folds) in successive model versions. A single combination of model hyperparameters is known as a configuration. Configurations which significantly ($\alpha$=0.01) underperformed in calibration and discrimination on the validation set were dropped out after each repeat using the Bootstrap Bias Corrected with Dropping Cross-Validation (BBCD-CV) method[A3]. For greater detail regarding the role of each hyperparameter in model function, please see the model code in https://github.com/sbhattacharyay/dynamic_GOSE_model/blob/main/scripts/models/dynamic_APM.py.

*Overview of tested hyperparameters*

- Time window length: length, in hours, of the non-overlapping windows into which ICU stays were partitioned.
  - Tested values: 2, 8, 12, 24
  - Optimal value: 2
- Quantile bin count: number of quantile bins into which numerical variables were discretised for tokenisation.
  - Tested values: 3, 4, 5, 7, 10, 20
  - Optimal value: 20
- Embedding vector dimension: length of vectors learned for each token in the embedding layer.
  - Tested values: 16, 32, 64, 128
  - Optimal value: 128
- Recurrent neural network (RNN) architecture: type of RNN structure.
  - Tested values: long short-term memory (LSTM), gated recurrent unit (GRU)
  - Optimal value: GRU
- Output layer: Encoding of six-month Glasgow Outcome Scale – Extended (GOSE) in model. See the following reference for greater detail[A4].
  - Tested values: softmax (i.e., multinomial encoding), sigmoid (i.e., ordinal encoding)
  - Optimal value: softmax
- Inclusion of temporal tokens: whether to include tokens representing the time of day or time elapsed since ICU admission in patient time windows.
  - Tested values: only time of day tokens, only time from admission tokens, neither, both
  - Optimal value: neither
- Token presentation: how tokens are presented (i.e., transformed or not) to models. See the following reference for greater detail on how "dt-patient-matrix" patient windows are formed[A5].
  - Tested values: "patient-matrix" (i.e., no transformation of tokens), "dt-patient-matrix" (i.e., each patient window contains the difference of tokens from the previous window)
  - Optimal value: "patient-matrix"
- RNN hidden state dimension: dimension of the RNN hidden state.
  - Tested values: 16, 32, 64, 128, 256
  - Optimal value: 128
- Number of RNN layers: number of hidden layers in the RNN architecture.
  - Tested values: 1, 2
  - Optimal value: 1
- Window limit during training: limit to the number of time windows per training set patient considered during training.
  - Tested values: None, 12, 24, 36, 84
  - Optimal value: 84
- Dropout: proportion of embedding layer weights randomly set to zero during training to regularise against overfitting.
  - Tested values: 0, 0.2
  - Optimal value: 0.2
- Learning rate: initial learning rate of Adam stochastic gradient descent algorithm for parameter optimisation.
  - Tested values: 0.001, 0.0003
  - Optimal value: 0.001
- Batch size: number of training set patients used during each iteration of parameter updating during training.
  - Tested values: 1, 4, 8, 32, 64, 128
  - Optimal value: 1
- Maximum training epochs: maximum number of rounds updating model parameters across the full training set of patients.
  - Tested values: 4, 8, 20, 30
  - Optimal value: 30
- Early stopping patience: number of epochs of no improvement in validation set discrimination performance after which training is stopped.
  - Tested values: 10 (if maximum training epochs was 20 or 30)

- Optimal value: 10

*Tested hyperparameters per model version*
In successive model versions, we tested different combinations of model hyperparameters, using top performing combinations of untested hyperparameters from previous versions.

Version 0-0: The purpose of this version was to initialise hyperparameter optimisation and test a wide range of configurations.
Tested configurations:

- Quantile bin count:
  - Tested values: 3, 4, 5, 7, 10, 20
  - Optimal value: 20
- Embedding vector dimension:
  - Tested values: 16, 32, 64
  - Optimal value: 64
- RNN architecture:
  - Tested values: LSTM, GRU
  - Optimal value: GRU
- Output layer:
  - Tested values: softmax (i.e., multinomial encoding), sigmoid (i.e., ordinal encoding)
  - Optimal value: softmax
- RNN hidden state dimension:
  - Tested values: 16, 32, 64, 128, 256
  - Optimal value: 128
- Number of RNN layers:
  - Tested values: 1, 2
  - Optimal value: 1
- Dropout: proportion of embedding layer weights randomly set to zero during training to regularise against overfitting.
  - Tested values: 0, 0.2
  - Optimal value: 0.2

Fixed configurations:

- Time window length: 2
- Inclusion of temporal tokens: neither
- Token presentation: "patient-matrix"
- Window limit during training: None
- Learning rate: 0.001
- Batch size: 32
- Maximum training epochs: 30
- Early stopping patience: 10

Version 0-1: The purpose of this version was to test the optimal configurations of the prior version with minor adjustments to the tokenisation algorithm.
Fixed configurations:

- Time window length: 2
- Quantile bin count: 20
- Embedding vector dimension: 128
- RNN architecture: GRU
- Output layer: softmax
- Inclusion of temporal tokens: neither
- Token presentation: "patient-matrix"
- RNN hidden state dimension: 128
- Number of RNN layers: 1
- Window limit during training: None
- Dropout: 0
- Learning rate: 0.001
- Batch size: 32
- Maximum training epochs: 30
- Early stopping patience: 10

Version 0-2: The purpose of this version was to test different optimisation parameters on the optimal configurations of the prior version for each output encoding type.
Tested configurations:

- Quantile bin count:
  - Tested values: 4, 20
  - Optimal value: 20
- Output layer:
  - Tested values: softmax (i.e., multinomial encoding), sigmoid (i.e., ordinal encoding)
  - Optimal value: softmax
- RNN hidden state dimension:
  - Tested values: 16, 128
  - Optimal value: 128
- Learning rate:
  - Tested values: 0.001, 0.0003
  - Optimal value: 0.001
- Batch size:
  - Tested values: 1, 8, 32, 64
  - Optimal value: 1

Fixed configurations:

- Time window length: 2
- Embedding vector dimension: 64
- RNN architecture: GRU
- Inclusion of temporal tokens: neither
- Token presentation: "patient-matrix"
- Number of RNN layers: 1
- Window limit during training: None
- Dropout: 0
- Maximum training epochs: 30
- Early stopping patience: 10

Version 0-3: The purpose of this version was to test different optimisation parameters on the optimal configurations of the prior version for each output encoding type.
Tested configurations:

- Quantile bin count:
  - Tested values: 4, 20
  - Optimal value: 20
- Output layer:
  - Tested values: softmax (i.e., multinomial encoding), sigmoid (i.e., ordinal encoding)
  - Optimal value: softmax
- RNN hidden state dimension:
  - Tested values: 16, 128
  - Optimal value: 128
- Maximum training epochs:
  - Tested values: 4, 8
  - Optimal value: 8

Fixed configurations:

- Time window length: 2
- Embedding vector dimension: 64
- RNN architecture: GRU
- Inclusion of temporal tokens: neither
- Token presentation: "patient-matrix"
- Number of RNN layers: 1
- Window limit during training: None
- Dropout: 0
- Learning rate: 0.001
- Batch size: 1
- Early stopping patience: 10

Version 1-0: The purpose of this version was to test the optimal configurations of the prior version with repeated cross-validation
Tested configurations:

- Quantile bin count:
  - Tested values: 4, 20
  - Optimal value: 20
- Output layer:
  - Tested values: softmax (i.e., multinomial encoding), sigmoid (i.e., ordinal encoding)
  - Optimal value: softmax
- RNN hidden state dimension:
  - Tested values: 16, 128
  - Optimal value: 128
- Maximum training epochs:
  - Tested values: 4, 8
  - Optimal value: 8

Fixed configurations:

- Time window length: 2
- Embedding vector dimension: 64
- RNN architecture: GRU
- Inclusion of temporal tokens: neither
- Token presentation: "patient-matrix"
- Number of RNN layers: 1
- Window limit during training: None
- Dropout: 0
- Learning rate: 0.001
- Batch size: 1
- Early stopping patience: 10

Version 2-0: The purpose of this version was to test the effect of modifying the time window length on model information and calibration.
Tested configurations:

- Time window length:
  - Tested values: 2, 8, 12, 24
  - Optimal value: 2

Fixed configurations:

- Quantile bin count: 20
- Embedding vector dimension: 64
- RNN architecture: GRU
- Output layer: softmax
- Inclusion of temporal tokens: neither
- Token presentation: "patient-matrix"
- RNN hidden state dimension: 128
- Number of RNN layers: 1
- Window limit during training: None
- Dropout: 0.2
- Learning rate: 0.001
- Batch size: 4
- Maximum training epochs: 20
- Early stopping patience: 10

Version 3-0:
Fixed configurations:

- Time window length: 2
- Quantile bin count: 20
- Embedding vector dimension: 64
- RNN architecture: GRU
- Output layer: softmax

- Inclusion of temporal tokens: neither
- Token presentation: "patient-matrix"
- RNN hidden state dimension: 128
- Number of RNN layers: 1
- Window limit during training: None
- Dropout: 0.2
- Learning rate: 0.001
- Batch size: 4
- Maximum training epochs: 20
- Early stopping patience: 10

Version 4-0: The purpose of this version was to test the effect of limiting the number of time windows per patient read from the training set during model training.
Tested configurations:

- Window limit during training: limit to the number of time windows per training set patient considered during training.
    - Tested values: None, 84
    - Optimal value: 84
- Batch size:
    - Tested values: 1, 4, 32, 128
    - Optimal value: 1

Fixed configurations:

- Time window length: 2
- Quantile bin count: 20
- Embedding vector dimension: 64
- RNN architecture: GRU
- Output layer: softmax
- Inclusion of temporal tokens: neither
- Token presentation: "patient-matrix"
- RNN hidden state dimension: 128
- Number of RNN layers: 1
- Dropout: 0.2
- Learning rate: 0.001
- Maximum training epochs: 30
- Early stopping patience: 10

Version 5-0: The purpose of this version was to test both the effect of limiting the number of time windows per patient read from the training set during model training and the pre-training transformation of patient windows to only focus on changes in clinical events.
Tested configurations:

- Token presentation:
    - Tested values: "patient-matrix" (i.e., no transformation of tokens), "dt-patient-matrix" (i.e., each patient window contains the difference of tokens from the previous window)
    - Optimal value: "patient-matrix"
- Window limit during training: limit to the number of time windows per training set patient considered during training.
    - Tested values: None, 12, 24, 36, 84
    - Optimal value: 84

Fixed configurations:

- Time window length: 2
- Quantile bin count: 20
- Embedding vector dimension: 64
- RNN architecture: GRU
- Output layer: softmax
- Inclusion of temporal tokens: neither
- RNN hidden state dimension: 128
- Number of RNN layers: 1
- Dropout: 0.2
- Learning rate: 0.001
- Batch size: 1

- Maximum training epochs: 30
- Early stopping patience: 10

Version 6-0: This version resulted in the best performing models, overall. We also tested calibration techniques on the validation set to improve calibration performance. These techniques included matrix scaling, vector scaling, and temperature scaling on validation set predictions[A6].

Tested configurations:

- Embedding vector dimension:
  - Tested values: 32, 64, 128
  - Optimal value: 128
- Inclusion of temporal tokens:
  - Tested values: only time of day tokens, only time from admission tokens, neither, both
  - Optimal value: neither
- RNN hidden state dimension:
  - Tested values: 32, 64, 128
  - Optimal value: 128
- Token presentation:
  - Tested values: "patient-matrix" (i.e., no transformation of tokens), "dt-patient-matrix" (i.e., each patient window contains the difference of tokens from the previous window)
  - Optimal value: "patient-matrix"
- Window limit during training: limit to the number of time windows per training set patient considered during training.
  - Tested values: None, 12, 24, 84
  - Optimal value: 84

Fixed configurations:

- Time window length: 2
- Quantile bin count: 20
- RNN architecture: GRU
- Output layer: softmax
- Number of RNN layers: 1
- Dropout: 0.2
- Learning rate: 0.001
- Batch size: 1
- Maximum training epochs: 30
- Early stopping patience: 10

Version 7-0: This purpose of this version was to test model performance after excluding explicit physician-based impressions (see Supplementary Note 2).

Tested configurations:

- RNN architecture:
  - Tested values: LSTM, GRU
  - Optimal value: LSTM
- Embedding vector dimension:
  - Tested values: 32, 64, 128
  - Optimal value: 128
- Inclusion of temporal tokens:
  - Tested values: only time of day tokens, only time from admission tokens, neither, both
  - Optimal value: neither
- RNN hidden state dimension:
  - Tested values: 32, 64, 128
  - Optimal value: 64
- Token presentation:
  - Tested values: "patient-matrix" (i.e., no transformation of tokens), "dt-patient-matrix" (i.e., each patient window contains the difference of tokens from the previous window)
  - Optimal value: "patient-matrix"
- Window limit during training: limit to the number of time windows per training set patient considered during training.
  - Tested values: None, 12, 24, 84
  - Optimal value: 84

Fixed configurations:

- Time window length: 2
- Quantile bin count: 20
- Output layer: softmax
- Number of RNN layers: 1
- Dropout: 0.2
- Learning rate: 0.001
- Batch size: 1
- Maximum training epochs: 30
- Early stopping patience: 10

**Calculation of Somers' $D_{xy}$**

The calculation of Somers' $D_{xy}$ begins by converting the estimated probabilities of threshold-level six-month GOSE into individual GOSE probability scores ($\hat{p}_i$):

$$\hat{p}_i \triangleq \widehat{\Pr}(\text{GOSE} = \text{GOSE}_i) = \begin{cases} 1 - \widehat{\Pr}(\text{GOSE} > \text{GOSE}_i), \text{ for } i = 0 \\ \widehat{\Pr}(\text{GOSE} > \text{GOSE}_{i-1}) - \widehat{\Pr}(\text{GOSE} > \text{GOSE}_i), \text{ for } i \in [1,2,3,4,5] \\ \widehat{\Pr}(\text{GOSE} > \text{GOSE}_{i-1}), \text{ for } i = 6 \end{cases} \quad (1)$$

where $i \in [0,1,2,3,4,5,6]$ represents the index of possible GOSE ($\text{GOSE}_0 = 1$, $\text{GOSE}_1 = 2$ or 3, $\text{GOSE}_2 = 4$, $\text{GOSE}_3 = 5$, $\text{GOSE}_4 = 6$, $\text{GOSE}_5 = 7$, $\text{GOSE}_6 = 8$). Then, the expected six-month GOSE index is estimated by $\widehat{\mathbb{E}}[i] = \sum_{i=0}^{6} i \cdot \hat{p}_i$. A comparable pair is defined as a pair of patients with different six-month GOSE scores, and the total number of unique comparable pairs is denoted as $N^{\text{comp}}$. A concordant pair is a comparable pair in which the patient with the higher six-month GOSE also has a higher model-estimated $\widehat{\mathbb{E}}[i]$, and the total number of unique concordant pairs is denoted as $N^{\text{conc}}$. Since $N^{\text{conc}} = 0.5N^{\text{comp}}$ if there is no association between $\widehat{\mathbb{E}}[i]$ and six-month GOSE, Somers' $D_{xy}$ ($S$) is defined as:

$$S \triangleq \frac{N^{\text{conc}} - \frac{1}{2}N^{\text{comp}}}{\frac{1}{2}N^{\text{comp}}} \quad (2).$$

## SUPPLEMENTARY REFERENCES